%% file: main.tex
\documentclass[runningheads]{llncs}

\usepackage{eccv}

\usepackage{eccvabbrv}

\usepackage{graphicx}
\usepackage{booktabs}

\input{preamble}

\usepackage[accsupp]{axessibility}

\usepackage{hyperref}

\usepackage{orcidlink}

\begin{document}

% ---------------------------------------------------------------
\title{MUSES: The Multi-Sensor Semantic Perception Dataset for Driving under Uncertainty} 

\titlerunning{MUSES: The Multi-Sensor Semantic Perception Dataset}

\author{Tim Br{\"o}dermann$^{*}$\inst{1}\orcidlink{0009-0009-6428-3066} \and
David Bruggemann$^{*}$\inst{1}\orcidlink{0000-0002-2409-5548} \and
Christos Sakaridis\inst{1}\orcidlink{0000-0003-1127-8887},\\
Kevin Ta\inst{1}\orcidlink{0009-0005-5353-3630} \and
Odysseas Liagouris\inst{1}\orcidlink{0009-0008-3706-993X} \and
Jason Corkill\inst{1}\orcidlink{ 0009-0009-3855-1266} \and
Luc Van Gool\inst{1,2}\orcidlink{0000-0002-3445-5711}
}

\authorrunning{T.~Br{\"o}dermann et al.}

\institute{$^{*}$ Equal contribution $^{1}$ ETH Z\"urich, Switzerland $^{2}$ INSAIT, Sofia Un., Bulgaria
}

\maketitle
\input{sec/0_abstract}   

\input{sec/1_intro}
\input{sec/2_related}
\input{sec/3_dataset}

\input{sec/4_upq}

\input{sec/5_analysis}

\input{sec/6_conclusion}

\section*{Acknowledgment}

This work was supported by the ETH Future Computing Laboratory (EFCL), financed by a donation from Huawei Technologies.

% ---- Bibliography ----
\bibliographystyle{splncs04}
\bibliography{main}

\input{sec/X_suppl}
\input{sec/X_supplement_qual_figs}

\end{document}

%% file: preamble.tex
\usepackage[dvipsnames]{xcolor}

\usepackage{tikz}
\usetikzlibrary{patterns}
\usepackage{pgfplots}
\pgfplotsset{compat=1.18}
\tikzset{
  font={\fontsize{7}{8}\selectfont}}
\usetikzlibrary{shapes.geometric, arrows}
\tikzstyle{arrow} = [thick,->,>=stealth]

\usepackage{enumitem}

\usepackage{wrapfig}

\usepackage{multirow}

\newcommand{\tablefontsize}{\fontsize{8}{9}\selectfont}

\usepackage{colortbl}

\newcommand{\PAR}[1]{\vskip2pt \noindent{\bf #1}}

\newcommand{\Ours}{MUSES}

\newcommand{\yes}{\checkmark}
\newcommand{\no}{$\times$}

\definecolor{road}                {RGB}{128, 64,128}
\definecolor{sidewalk}            {RGB}{244, 35,232}
\definecolor{building}            {RGB}{ 70, 70, 70}
\definecolor{wall}                {RGB}{102,102,156}
\definecolor{fence}               {RGB}{190,153,153}
\definecolor{pole}                {RGB}{153,153,153}
\definecolor{traffic light}       {RGB}{250,170, 30}
\definecolor{traffic sign}        {RGB}{220,220,  0}
\definecolor{vegetation}          {RGB}{107,142, 35}
\definecolor{terrain}             {RGB}{152,251,152}
\definecolor{sky}                 {RGB}{ 70,130,180}
\definecolor{person}              {RGB}{220, 20, 60}
\definecolor{rider}               {RGB}{255,  0,  0}
\definecolor{car}                 {RGB}{  0,  0,142}
\definecolor{truck}               {RGB}{  0,  0, 70}
\definecolor{bus}                 {RGB}{  0, 60,100}
\definecolor{train}               {RGB}{  0, 80,100}
\definecolor{motorcycle}          {RGB}{  0,  0,230}
\definecolor{bicycle}             {RGB}{119, 11, 32}
\definecolor{void}                {RGB}{  0,  0,  0}

\definecolor{color1}{HTML}{FFA14C}
\definecolor{color2}{rgb}{0.6,0.8,0.1}
\definecolor{color3}{rgb}{0.552941176470588,0.627450980392157,0.796078431372549}
\definecolor{color4}{rgb}{0.905882352941176,0.541176470588235,0.764705882352941}

%% file: sec/0_abstract.tex
\begin{abstract}

Achieving level-5 driving automation in autonomous vehicles necessitates a robust semantic visual perception system capable of parsing data from different sensors across diverse conditions.
However, existing semantic perception datasets often lack important non-camera modalities typically used in autonomous vehicles, or they do not exploit such modalities to aid and improve semantic annotations in challenging conditions.
To address this, we introduce \Ours{}, the MUlti-SEnsor Semantic perception dataset for driving in adverse conditions under increased uncertainty. \Ours{} includes synchronized multimodal recordings with 2D panoptic annotations for 2500 images captured under diverse weather and illumination. The dataset integrates a frame camera, a lidar, a radar, an event camera, and an IMU/GNSS sensor. Our new two-stage panoptic annotation protocol captures both class-level and instance-level uncertainty in the ground truth and enables the novel task of uncertainty-aware panoptic segmentation, along with standard semantic and panoptic segmentation. \Ours{} proves both effective for training and challenging for evaluating models under diverse visual conditions, and it opens new avenues for research in multimodal and uncertainty-aware dense semantic perception. 
Our dataset and benchmark are publicly available at \url{https://muses.vision.ee.ethz.ch/}.

\end{abstract}

%% file: sec/1_intro.tex
\section{Introduction}
\label{sec:intro}

Experimental evidence~\cite{does:vision:matter:for:action} suggests that dense pixel-level semantic perception is one of the most central tasks for embodied intelligent agents such as autonomous cars. It densely parses the scene that surrounds the car into segments belonging to different semantic classes and/or distinct instances of these classes. The resulting high-level visual representation is an essential input to the downstream driving modules involving planning and control.

Due to their excellent spatial resolution, frame cameras are the standard sensor of choice in datasets for 2D pixel-level semantic perception of driving scenes~\cite{Cityscapes,Mapillary,ApolloScape,ACDC,wilddash,IDD,Wilddash2}.
However, their signal deteriorates severely in adverse visual conditions, such as nighttime, fog, rain, snow, or sun glare.
This implies that the semantic perception module of an autonomous car can benefit from additional sensor inputs such as lidars, radars, and event cameras.
These can complement frame cameras in different ways, such as via the robustness of lidars to ambient illumination, the robustness of radars to weather, and the high dynamic range and low latency of event cameras.
Through sensor fusion, complementary information from the additional sensors can be exploited in parts of the scene where the camera signal is unintelligible.

\begin{figure}[tb]
  \centering
  \begin{subfigure}[t]{0.32\linewidth}
    \includegraphics[width=\linewidth]{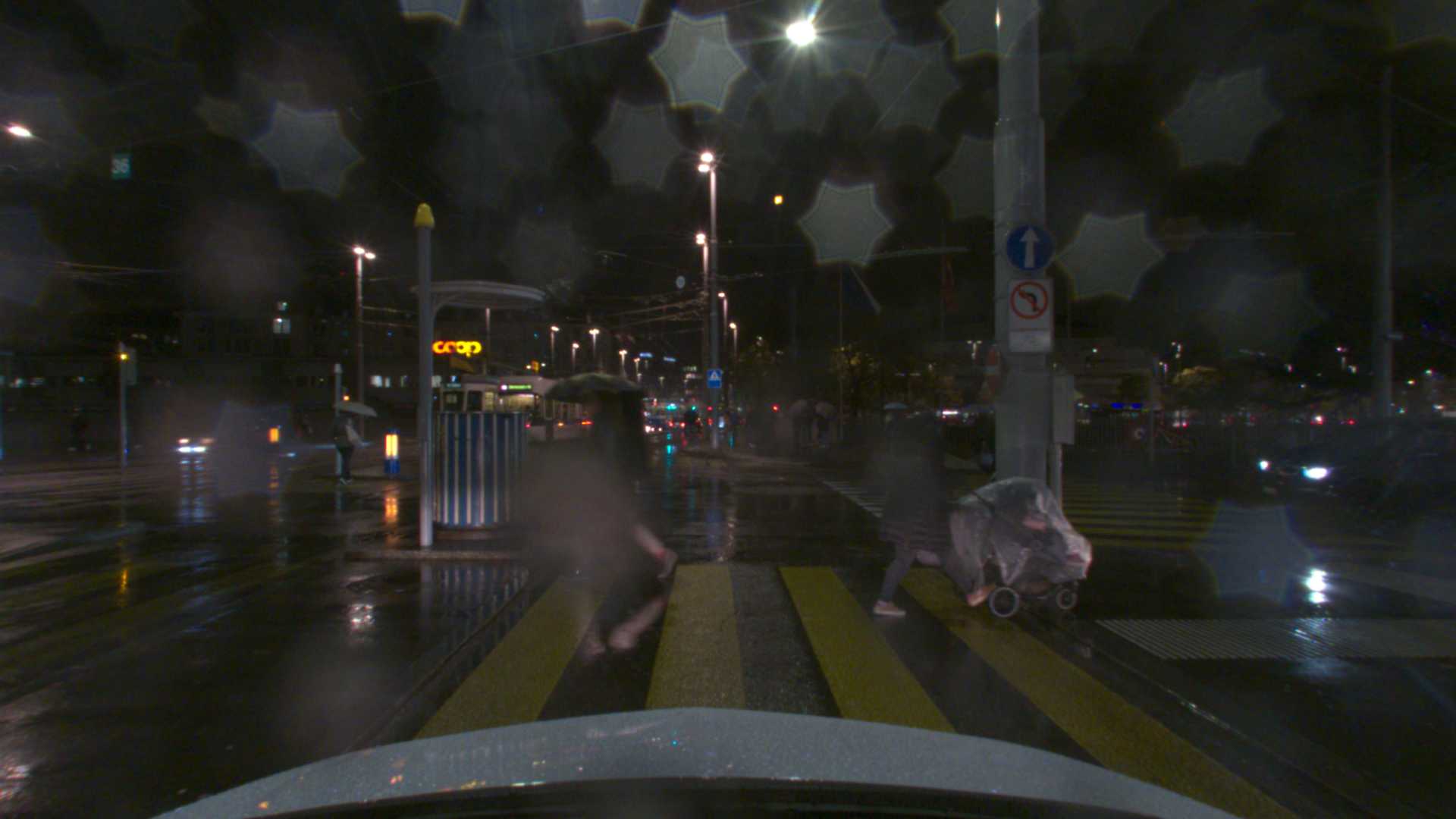}
  \end{subfigure}
  \begin{subfigure}[t]{0.32\linewidth}
    \includegraphics[width=\linewidth]{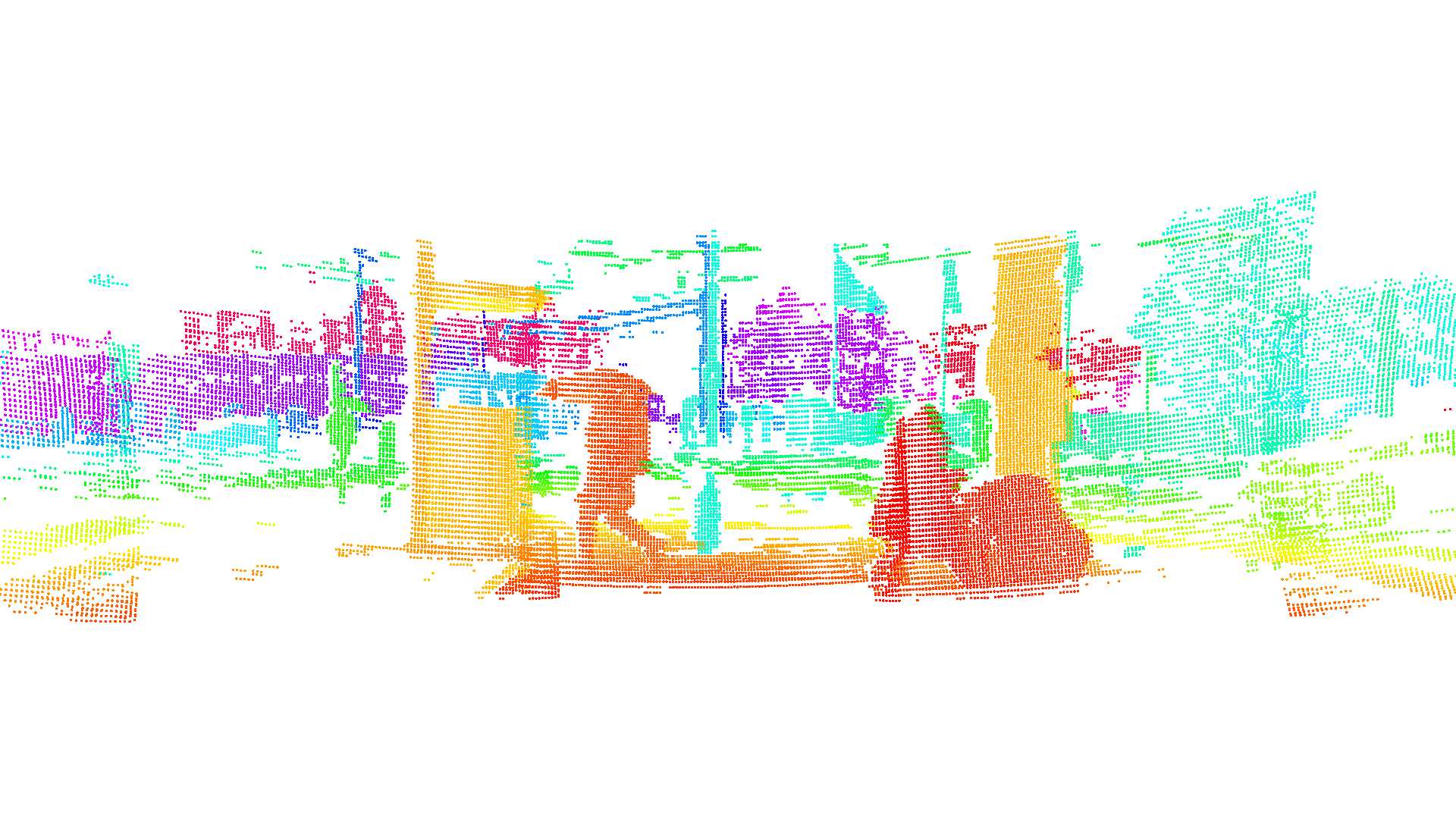}
  \end{subfigure}    
  \begin{subfigure}[t]{0.32\linewidth}
    \centering
    \includegraphics[width=0.5\linewidth]{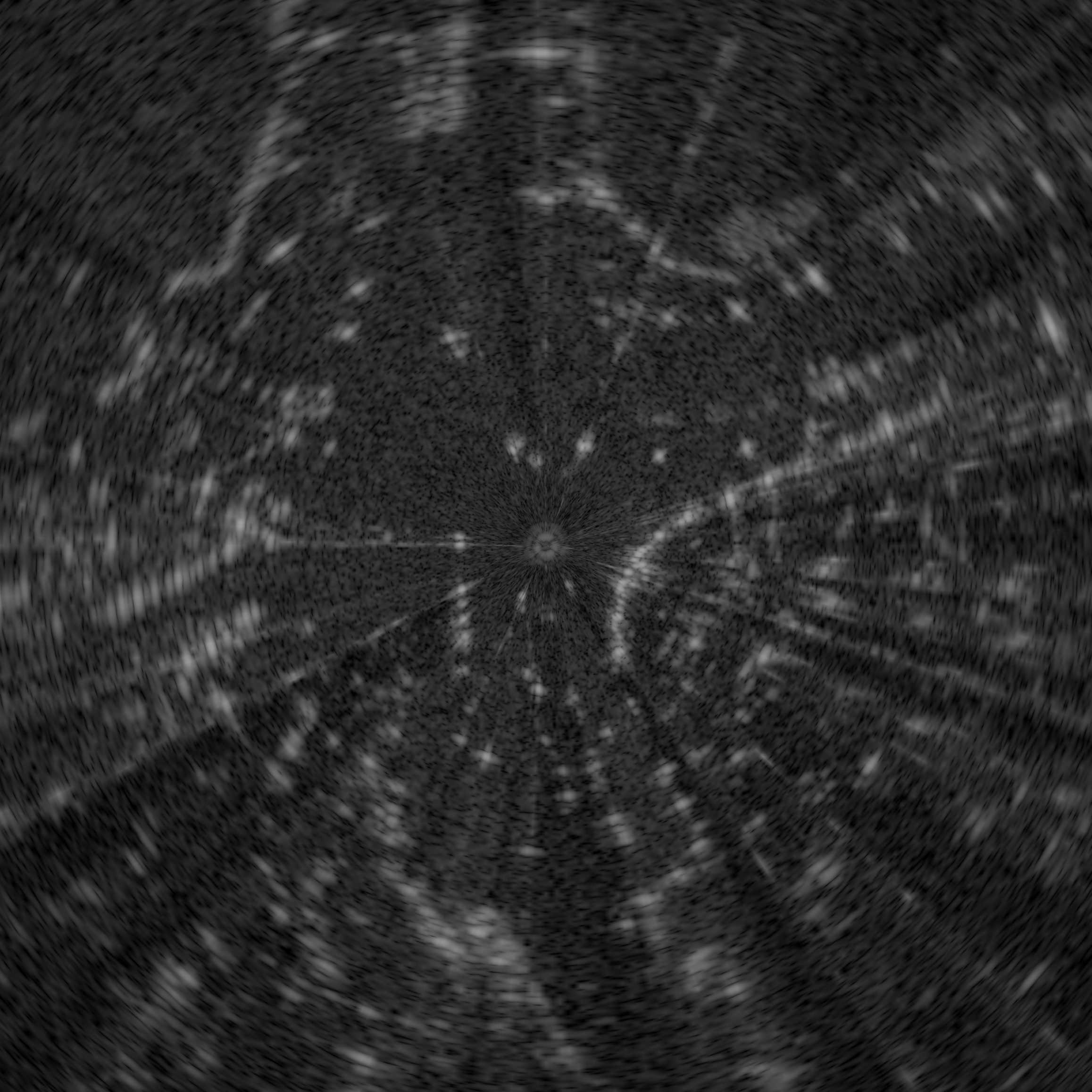}
  \end{subfigure}\\ 
  
  \begin{subfigure}[t]{0.32\linewidth}
    \includegraphics[width=\linewidth]{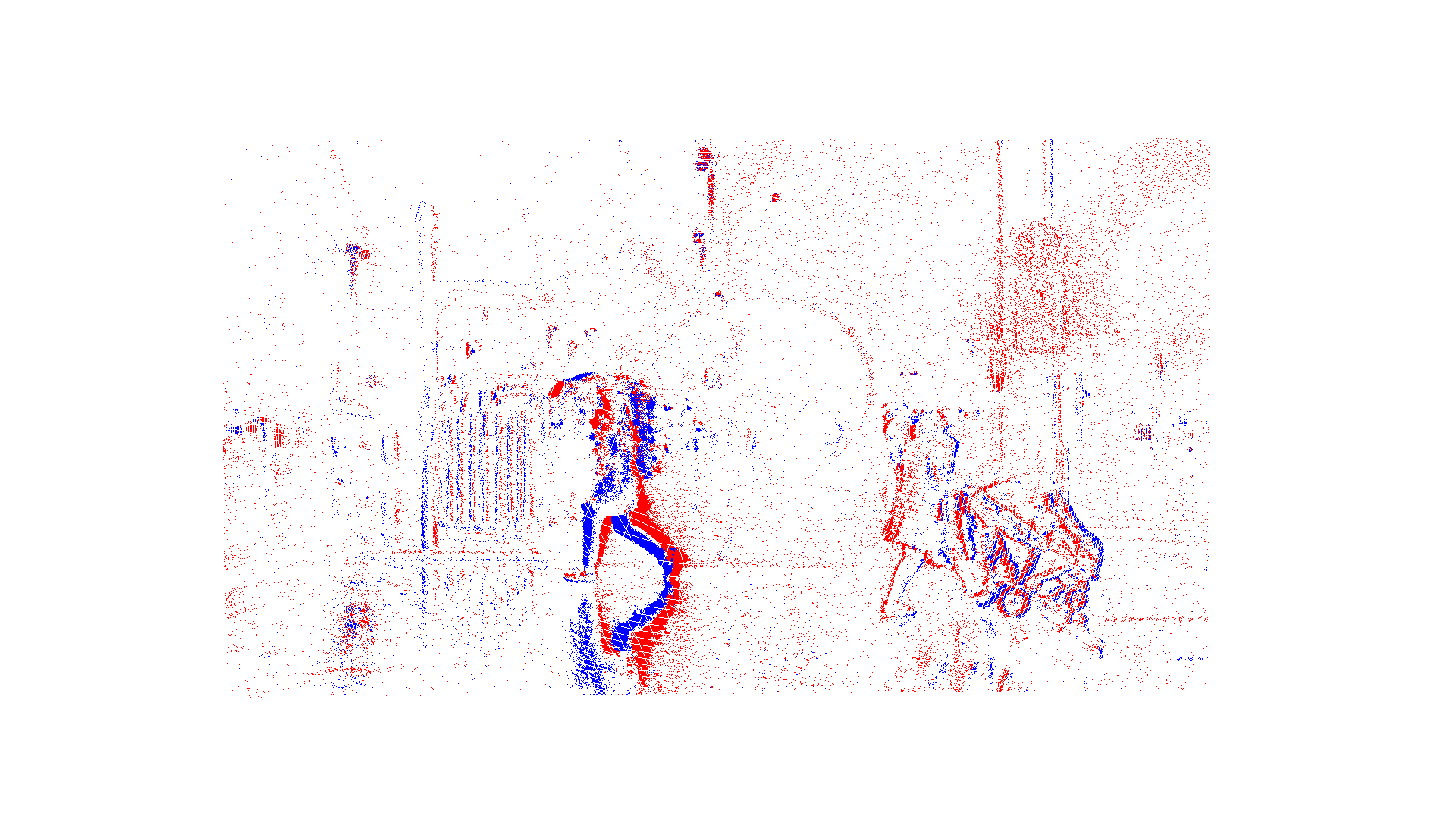}
  \end{subfigure}
  \begin{subfigure}[t]{0.32\linewidth}
    \includegraphics[width=\linewidth]{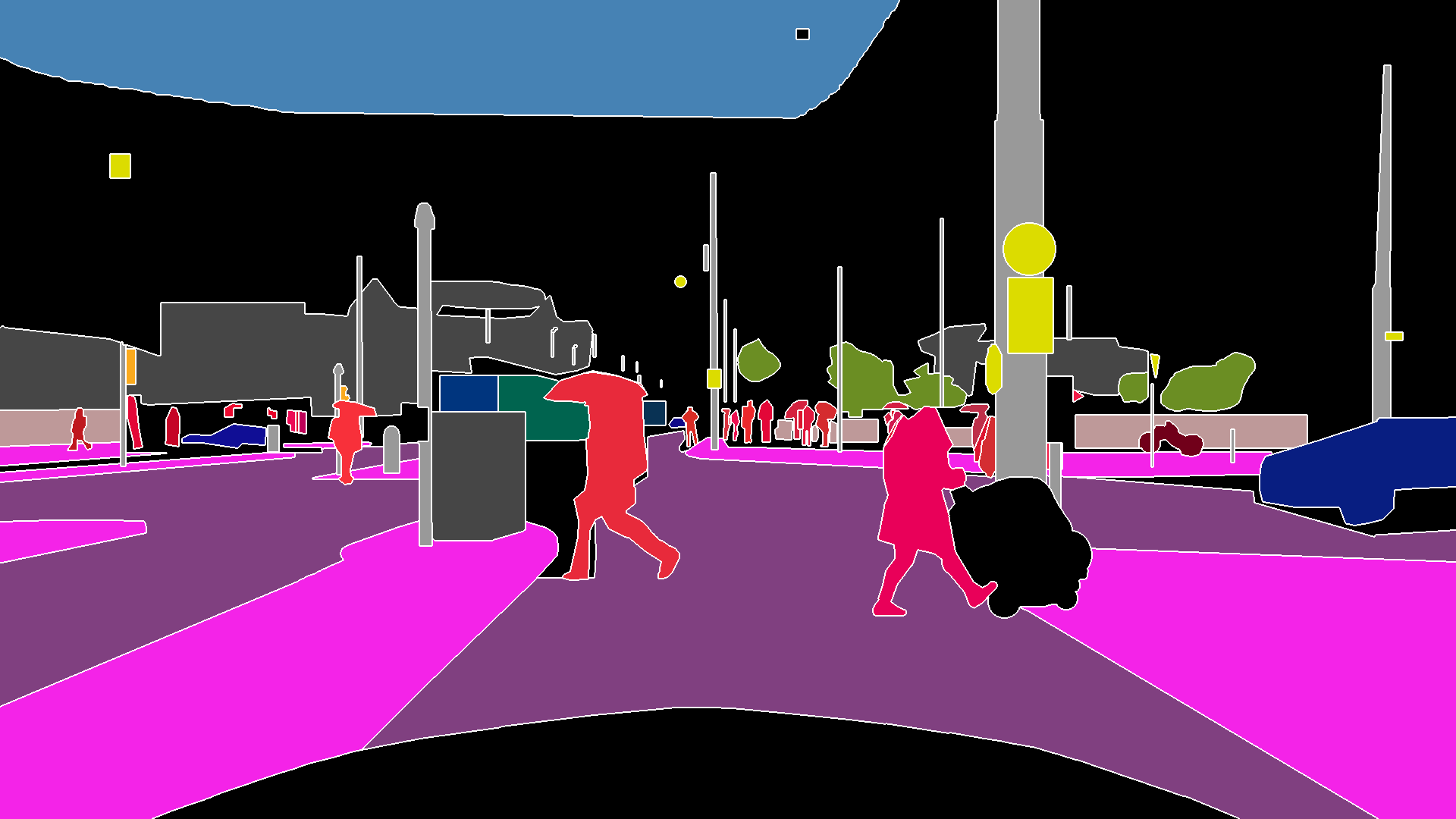}
  \end{subfigure}
  \begin{subfigure}[t]{0.32\linewidth}
    \includegraphics[width=\linewidth]{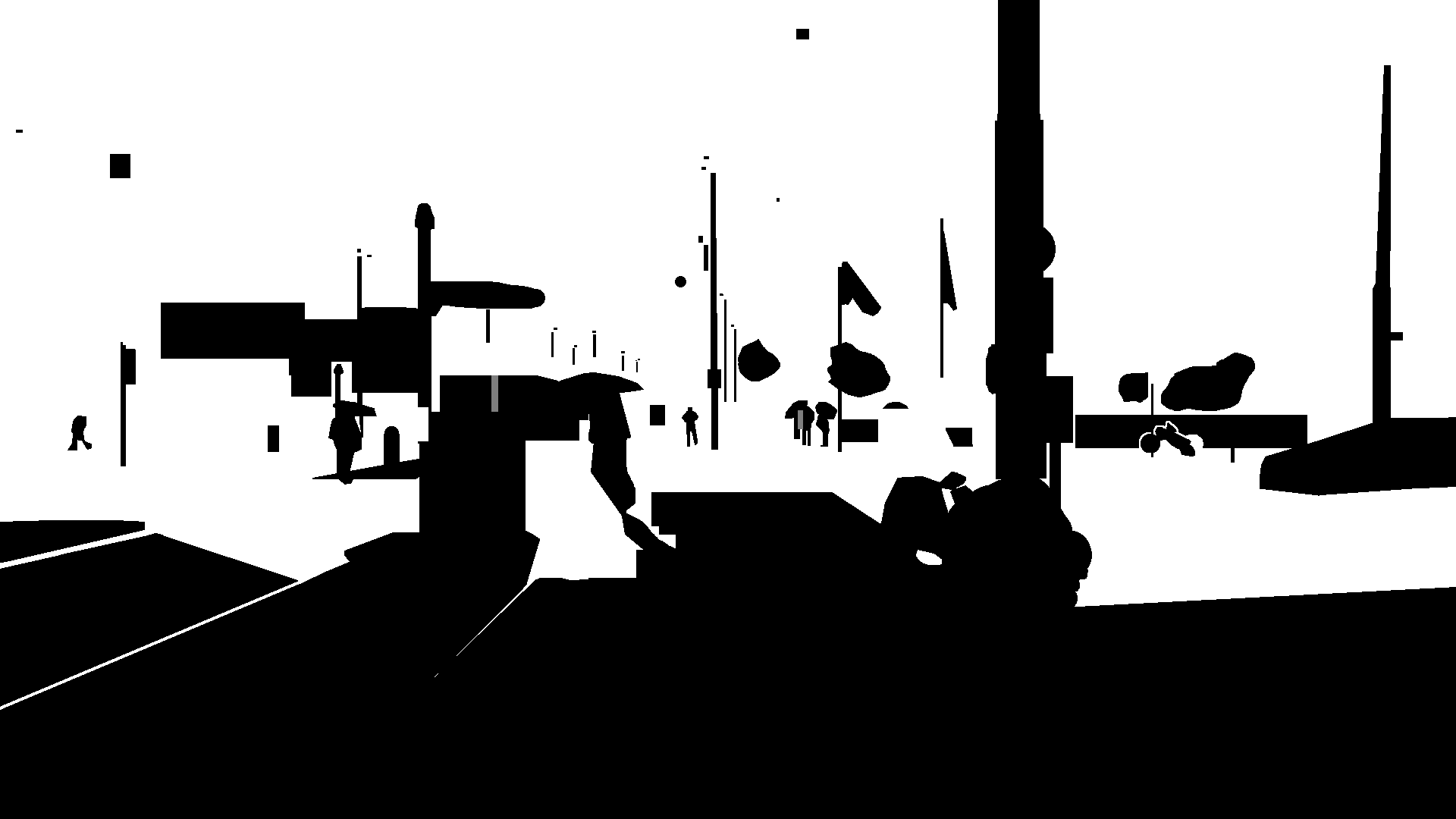}
  \end{subfigure}\\

  \caption{Annotated scene of \Ours{}. First row left to right: frame camera, lidar, radar; second row left to right: event camera, panoptic annotation, and difficulty annotation. 
  }
  \label{fig:teaser}
\end{figure}

To develop sensor fusion models for dense semantic perception in driving scenes, appropriate datasets are required.
The central features of such datasets are (i) the inclusion of multiple sensor modalities that are relevant for parsing driving scenes, (ii) the annotation of the input images of the dataset with fine pixel-level labels that facilitate panoptic segmentation, (iii) proper annotation of the aleatoric uncertainty in the semantic content of the input owing to the potentially low signal-to-noise ratio of the latter, and (iv) the representation of a diverse set of visual conditions, including adverse ones.
Existing driving datasets, however, feature only some of these attributes while missing others~\cite{nuScenes,choi2018kaist,patil2019h3d,chang2019argoverse,geyer2020a2d2,xiao2021pandaset}. 
In particular, multimodal datasets targeting adverse weather only feature coarse bounding boxes for traffic-related objects~\cite{seeing:through:fog,CADC,sheeny2021radiate,alibeigi2023zenseact}, pixel-level annotations just for the road~\cite{alibeigi2023zenseact}, or limited object classes~\cite{diaz2022ithaca365}.
These annotations preclude the training and evaluation of general pixel-level segmentation methods that parse the entire driving scene (e.g. sidewalk, pole, and other important classes for driving).
On the other hand, adverse-condition datasets that feature dense semantic labels are based solely on camera inputs~\cite{ACDC,Wilddash2}, which limits the number and extent of pixels that can reliably receive manual semantic labels.
There is a need for a multimodal driving dataset enabling 
dense semantic perception where 
adverse conditions are sufficiently represented.

% Millimeter-wave (MMW) radar
To respond to this need, we construct MUSES, the MUlti-SEnsor Semantic perception dataset for driving under uncertainty. MUSES consists of driving sequences that include synchronized, calibrated, multimodal recordings with a normal frame-based camera, a new-generation microelectromechanical-system (MEMS) mirror-based lidar, a frequency-modulated continuous wave (FMCW) scanning radar, a high-definition (HD) event camera, and an IMU/GNSS sensor (see \cref{fig:teaser}). 
These
sequences are recorded 
under a diverse set of weather and illumination conditions, 
covering various combinations of time of day and precipitation.
To the best of our knowledge, \Ours{} is the first adverse conditions-focused dataset that includes event camera or MEMS lidar data.

We annotate 2500 selected camera images
of MUSES with high-quality fine 2D pixel-level panoptic annotations, which directly afford semantic and instance annotations. The annotated images are evenly distributed over different combinations of time of day, visibility, and precipitation, in each of the training, validation, and test sets. Each of the panoptically annotated camera images is accompanied by the respective readings of the other four sensors, which render MUSES the first large-scale multimodal diverse-condition driving dataset for dense semantic perception tasks, including semantic, instance, and panoptic segmentation. Moreover, our specialized image annotation protocol leverages the respective readings of the additional modalities and a corresponding normal-condition sequence, allowing the annotators to also reliably label degraded image regions that are still discernible in other modalities but would otherwise be impossible to label only from the image itself. This results in better pixel coverage in the annotations and creates a more challenging evaluation setup.

As uncertainty quantification of the perception module 
is crucial for certifiable downstream control~\cite{bonzanini2021perception,yang2023safe,mesbah2016stochastic,bemporad2007robust},
our specialized annotation protocol accounts for high aleatoric uncertainty in the semantic content of images in MUSES.
In particular, we extend the uncertainty-aware annotation protocol of ACDC~\cite{ACDC} to model \emph{instance-level uncertainty}, 
besides ACDC's class-level uncertainty. 
This allows MUSES to support the novel task of \emph{uncertainty-aware panoptic segmentation}, which constitutes a generalization of both panoptic segmentation~\cite{kirillov2019panoptic} and uncertainty-aware semantic segmentation~\cite{ACDC}. Ground-truth class-level and instance-level difficulty maps are used in uncertainty-aware panoptic segmentation to evaluate respective class- and instance-level confidence maps jointly with the standard panoptic predictions via the novel average uncertainty-aware panoptic quality (AUPQ) metric, which rewards labels and confidence values that are consistent with their human counterparts.

We present a detailed analysis of \Ours{}, highlighting the challenges for annotation in adverse conditions, notably fog and nighttime. We further show that our two-stage annotation indeed enhances label coverage and difficulty. The investigation of sensor impact on semantic perception highlights the intricacies of multimodal fusion, emphasizing the need for specialized fusion methods. 
Also, we show that models trained on \Ours{} excel in cross-domain evaluation, which is attributable to the high diversity and annotation quality of our dataset. 
Overall, our experimental analysis establishes \Ours{} as a challenging benchmark for uni- and multimodal dense semantic perception tasks, including standard and uncertainty-aware semantic/panoptic segmentation.

\Ours{} opens up the following new research directions: 
(1) Sensor fusion for pixel-level semantic perception in adverse conditions, (2) exploring event camera utility in adverse weather for automated driving, 
(3) examining challenges and opportunities of new-generation automotive MEMS lidars for semantic perception, 
and (4)  researching the novel uncertainty-aware panoptic segmentation task.
MUSES promises to advance the field by enabling more robust and accurate perception systems, particularly in challenging conditions, thus enhancing the safety and efficiency of autonomous driving technology.

%% file: sec/2_related.tex
\section{Related Work}
\label{sec:related}

Two of the earliest and most influential datasets for autonomous driving are KITTI~\cite{geiger2012we}, which provides high-resolution stereo images, lidar point clouds, and IMU/GNSS data, and Cityscapes~\cite{Cityscapes}, which offers high-quality pixel-level semantic annotations for RGB images.
However, neither KITTI nor Cityscapes include any samples for adverse weather or illumination.
To address this limitation, several datasets have been proposed that explicitly incorporate visual hazards~\cite{zendel2017good} in their data collection.
These hazards can be caused by natural phenomena, such as rain, snow, fog, or night, or by artificial factors, such as lens flare, motion blur, or sensor noise.
Some of these datasets are Oxford RobotCar~\cite{Oxford}, Oxford Radar RobotCar~\cite{Oxford:Radar}, DENSE~\cite{seeing:through:fog}, CADC~\cite{CADC}, AWARE~\cite{pinchon2019all} and Boreas~\cite{burnett2022boreas}.
These datasets provide multi-sensor recordings from different modalities, such as RGB cameras, lidar scanners, radar sensors, or thermal cameras. 
However, most of them do not provide \emph{fine}, pixel-level \emph{semantic} annotations, which are essential for training semantic segmentation models that are crucial for downstream action tasks of embodied agents~\cite{does:vision:matter:for:action}.
Another set of datasets provides dense semantic annotations for RGB images under adverse conditions.
These include the small-scale Foggy Driving\cite{sakaridis2018semantic}, Foggy Zurich~\cite{sakaridis2018model}, Dark Zurich~\cite{MGCDA_UIoU} and Raincouver~\cite{tung2017raincouver} and the large-scale ACDC~\cite{ACDC}, Wilddash2~\cite{Wilddash2} and BDD100K~\cite{yu2020bdd100k}.
Although these datasets focus on diverse weather and illumination conditions, they are limited to vision-only data and are not usually sufficient for reliable perception under adverse conditions.
Some recent datasets like InfraParis~\cite{franchi2024infraparis} provide multimodal data with dense semantic annotations, some include adverse lighting conditions like~\cite{vertens2020heatnet} and the panoptic extension~\cite{mei2022waymo} of Waymo Open~\cite{Waymo:Open:Dataset}, or even adverse weather conditions~\cite{alibeigi2023zenseact,diaz2022ithaca365}.
They combine RGB frames mainly with lidar point clouds to provide a more comprehensive representation of the scene but do not include further sensors or specialized labeling for adverse conditions.
Another relevant dataset is DSEC-Semantic~\cite{sun2022ess,gehrig2021dsec}, which adds event cameras to the frame-lidar combination,
but it only provides semantic pseudo-labels.

Our dataset, \Ours{}, aims to fill the gap between the existing datasets for semantic perception under adverse conditions (see \cref{table:comparison:datasets}).
\Ours{} provides diverse-condition multimodal data from a normal frame camera, a lidar, a radar, and an event camera, and corresponding high-quality 2D pixel-level panoptic annotations.
We extend the uncertainty-aware semantic segmentation task introduced in~\cite{ACDC} to panoptic segmentation, by accounting for both class- and instance-level aleatoric uncertainty in predictions and ground truth.
The resulting novel task of uncertainty-aware panoptic segmentation is enabled by a specialized annotation protocol, and it is fundamentally different from previous work~\cite{sirohi2023uncertainty} that investigates epistemic uncertainty for panoptic segmentation.

\begin{table}[tb]
    \caption{Comparison of \Ours{} to densely labeled semantic adverse-condition datasets. ``Sem./Pan.\ seg.'': semantic/panoptic segmentation, ``Corr.\ normal'': includes image-level correspondences, ``Class/Inst.\ uncert.'': captures class/instance uncertainty, *: segmentation only for road/lanes, **: pseudo-labels.}
    \label{table:comparison:datasets}
    \centering
    \tablefontsize
    \setlength\tabcolsep{2pt}
    \begin{tabular*}{\linewidth}{@{\extracolsep{\fill}}lccccccccc@{}}
    \toprule        
    Dataset & \rotatebox[origin=c]{90}{\begin{tabular}{c}Frame\\camera\end{tabular}} & \rotatebox[origin=c]{90}{Lidar} & \rotatebox[origin=c]{90}{\begin{tabular}{c}Event\\camera\end{tabular}} & \rotatebox[origin=c]{90}{Radar} & \rotatebox[origin=c]{90}{\begin{tabular}{c}Sem.\\seg.\end{tabular}} & \rotatebox[origin=c]{90}{\begin{tabular}{c}Pan.\\seg.\end{tabular}} & \rotatebox[origin=c]{90}{\begin{tabular}{c}Corr.\\normal\end{tabular}} & \rotatebox[origin=c]{90}{\begin{tabular}{c}Class\\uncert.\end{tabular}} & \rotatebox[origin=c]{90}{\begin{tabular}{c}Inst.\\uncert.\end{tabular}}\\ % & [\# Adverse] 
    \midrule
    Foggy Zurich~\cite{CMAda:IJCV2020} & \yes & \no & \no & \no   & \yes & \no  & \no & \no & \no \\ % & 40 
    Dark Zurich~\cite{MGCDA_UIoU}   & \yes & \no & \no & \no   & \yes & \no  & \yes & \yes & \no \\ %& 201
    Wilddash2~\cite{Wilddash2}   & \yes & \no & \no & \no & \yes & \yes  & \no  & \no & \no \\ %& ? (\>315)
    BDD100K~\cite{yu2020bdd100k}  & \yes & \no & \no & \no   & \yes & \yes & \no  & \no & \no \\ %& 1346 
    ACDC~\cite{ACDC}   & \yes & \no & \no & \no  & \yes & \yes   & \yes & \yes & \no \\ %& 4006
    Zenseact Open Dataset~\cite{alibeigi2023zenseact} &   \yes & \yes  & \no & \no & * & \no & \no  & \no & \no \\ %& ? 
    WOD: PVPS~\cite{Waymo:Open:Dataset,mei2022waymo} & \yes & \yes & \no  & \no  & \yes & \yes & \no  & \no & \no \\ %& ? 
    DSEC-Semantic~\cite{gehrig2021dsec,sun2022ess}  & \yes & \yes & \yes  & \no & ** & \no & \no& \no& \no \\ %& ?
    \Ours{} (Ours)  & \yes  & \yes & \yes & \yes & \yes & \yes  & \yes & \yes & \yes\\ %& 2000
    \bottomrule
    \end{tabular*}
\end{table}

%% file: sec/3_dataset.tex
\section{The \Ours{} Dataset}
\label{sec:method}

\Ours{} is a multi-sensor dataset with 2D panoptic annotations for autonomous driving that emphasizes adverse conditions.
The dataset was recorded in Switzerland from November 2022 to July 2023. 
We recorded multiple driving sequences under adverse conditions (night, rain, snow, fog, or a combination thereof) and recorded each adverse-condition route again in clear-weather daytime conditions.
The multi-sensor readings serve two purposes: 1) they provide corroborative evidence during annotation to increase label coverage; 2) they support multimodal methods for dense semantic perception in adverse visual conditions. This section provides a detailed overview of the dataset. We detail the sensors in \cref{subsec:sensors}, the annotation protocol in \cref{subsec:annotation}, and the splits and benchmarks in \cref{subsec:overview}.

\subsection{Sensor Suite}
\label{subsec:sensors}

Our sensor suite consists of a high-resolution frame camera, an event camera, a MEMS lidar, an FMCW scanning radar, and an IMU/GNSS module, all mounted on the front part of the roof of our recording car (see~\cref{fig:car_pic}).
Since we recorded highly adverse scenes, the two cameras and the GNSS were placed inside a waterproof box with a transparent acrylic glass window.
To manage lens occlusion due to water droplets or snow, the acrylic glass window and lidar cover glass were treated with a hydrophobic coating and regularly wiped clean. 
The geometric calibration and synchronization of all the sensors are described in~\cref{subsec:calibration}, and the sensor specifications are summarized in \cref{tab:sensors:perception}.

\begin{table}
  \caption{Overview of sensor specifications for \Ours{}.}
  \label{tab:sensors:perception}
  \centering
  \tablefontsize
  \setlength\tabcolsep{3pt}
  \resizebox{\textwidth}{!}{%
  \begin{tabular}{lll}%{\linewidth}{@{\extracolsep{\fill}} ll@{}}
  \toprule
  Modality & Name & Specifications \\
  \midrule
  Frame camera & TRI023S-CC & \begin{tabular}{@{}l@{}} 8-bit RGB, 30 Hz, 1920\texttimes1080, HFOV: 77\textdegree, VFOV: 43\textdegree \end{tabular} \\
  \addlinespace[0.6ex] \rowcolor{gray!20} Event camera & Prophesee GEN4.1 & \begin{tabular}{@{}l@{}} 1280\texttimes720, 15M events/s, HFOV: 64\textdegree, VFOV: 39\textdegree \end{tabular} \\
  \addlinespace[0.6ex] Lidar & RS-LiDAR-M1	& \begin{tabular}{@{}l@{}} 10 Hz, avg.\ angular resolution: 0.2\textdegree, range: 200 m, \\HFOV: 120\textdegree, VFOV: 25\textdegree, 75K points/scan \end{tabular} \\
  \addlinespace[0.5ex] \rowcolor{gray!20} Radar & Navtech CIR-DEV & \begin{tabular}{@{}l@{}} 4 Hz, range resolution: 43.8 mm, horizontal angular \\resolution: 0.9\textdegree, range: 330 m\end{tabular} \\
  \addlinespace[0.6ex] IMU/GNSS & simpleRTK2B Fusion & RTK accuracy: \textless 10cm, 30 Hz \\
  \bottomrule
  \end{tabular}
  }
\end{table}

\PAR{Frame camera.}
We use a TRI023S-CC camera from Lucid Vision Labs with a global shutter and automatic exposure time.
The RGB images are undistorted and provided as 8-bit PNG files.
We use the frame camera as our frame-of-reference as our panoptic annotations pertain to the undistorted RGB images. To protect the privacy of all persons identifiable from our dataset, we used a semi-automatic anonymization pipeline based on~\cite{wang2019fast} to segment and blur faces and license plates. More details on anonymization are in~\cref{sec:anonymization}.

\PAR{Event camera.}
Event cameras exhibit high dynamic range and low latency, but the potential of event cameras in adverse conditions remains unexplored due to a lack of available datasets.
To foster research in this direction, we use a Prophesee Gen4.1 event camera with HD resolution. 
The contrast sensitivity is manually adjusted depending on the lighting conditions, to record more details in brightly lit scenes and limit noise under low illumination.
To further avoid peaks of events, we used an event rate controller.
The raw events\textemdash \ie, the 4-tuples (x, y, timestamp, polarity)\textemdash are undistorted, truncated to 3-second chunks, and stored in compressed HDF5 files.

\PAR{Lidar.}
Lidars are a core sensing modality for autonomous cars, providing high-resolution 3D maps irrespective of lighting conditions. % Comment out? C: No.
New-generation MEMS lidars offer advantages such as a compact form, minimal moving parts, and cost-effectiveness, but their irregular scanning pattern poses compatibility issues with standard motorized scanning lidar frameworks.
To facilitate further research on such sensors, we use the new-generation automotive RS-LiDAR-M1 lidar from RoboSense. The raw lidar point cloud is stored as a binary file with six entries (x, y, z, intensity, mirror \#, timestamp) for each point.

\PAR{Radar.}
Due to its robustness to any type of weather condition, the radar is an important component of reliable automotive perception stacks. % Comment out?
We use a Navtech CIR-DEV FMCW radar with a cosec dish that directs part of the emitted power below the horizontal plane for better short-range detection.
Inspired by~\cite{Oxford:Radar}, we store the radar data as a range-azimuth PNG file, containing per azimuth the timestamp, sweep counter, and raw received power readings in dB.

\PAR{IMU/GNSS.}
We use the simpleRTK2B board from ArduSimple as our IMU/
GNSS device. It offers dead reckoning and real-time kinematics (RTK). An extended Kalman filter (EKF) fuses GNSS and IMU, resulting in a global 6 DoF pose with respective velocities.

\subsection{Annotation}
\label{subsec:annotation}

\begin{wrapfigure}{r}{5.5cm}
\centering
\resizebox{\linewidth}{!}{%
\begin{tabular}{@{}c@{\hspace{0.05cm}}c@{}}
\includegraphics[width=0.49\linewidth, trim=430 450 1100 450, clip]{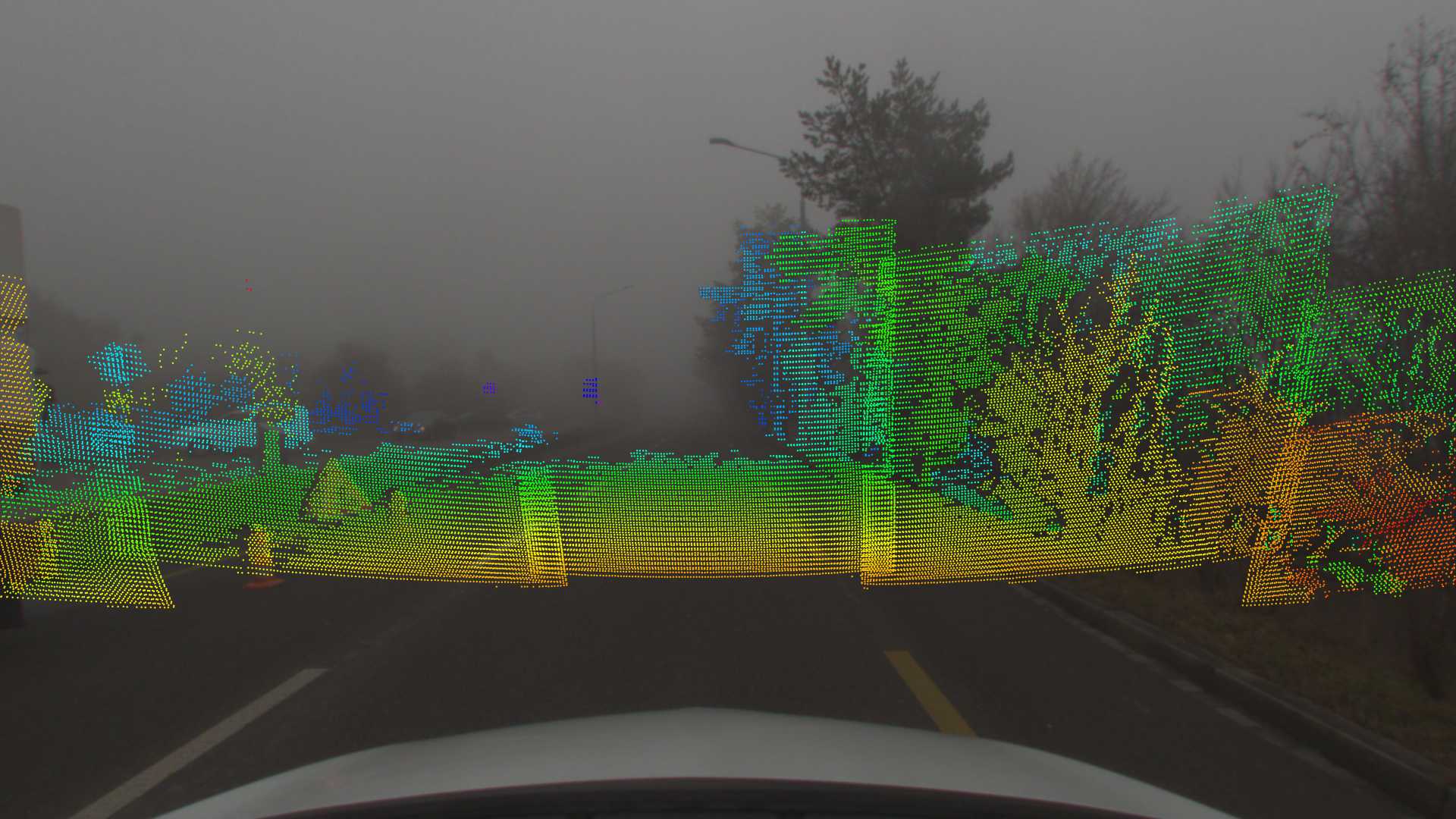} 
&
\includegraphics[width=0.49\linewidth, trim=430 450 1100 450, clip]{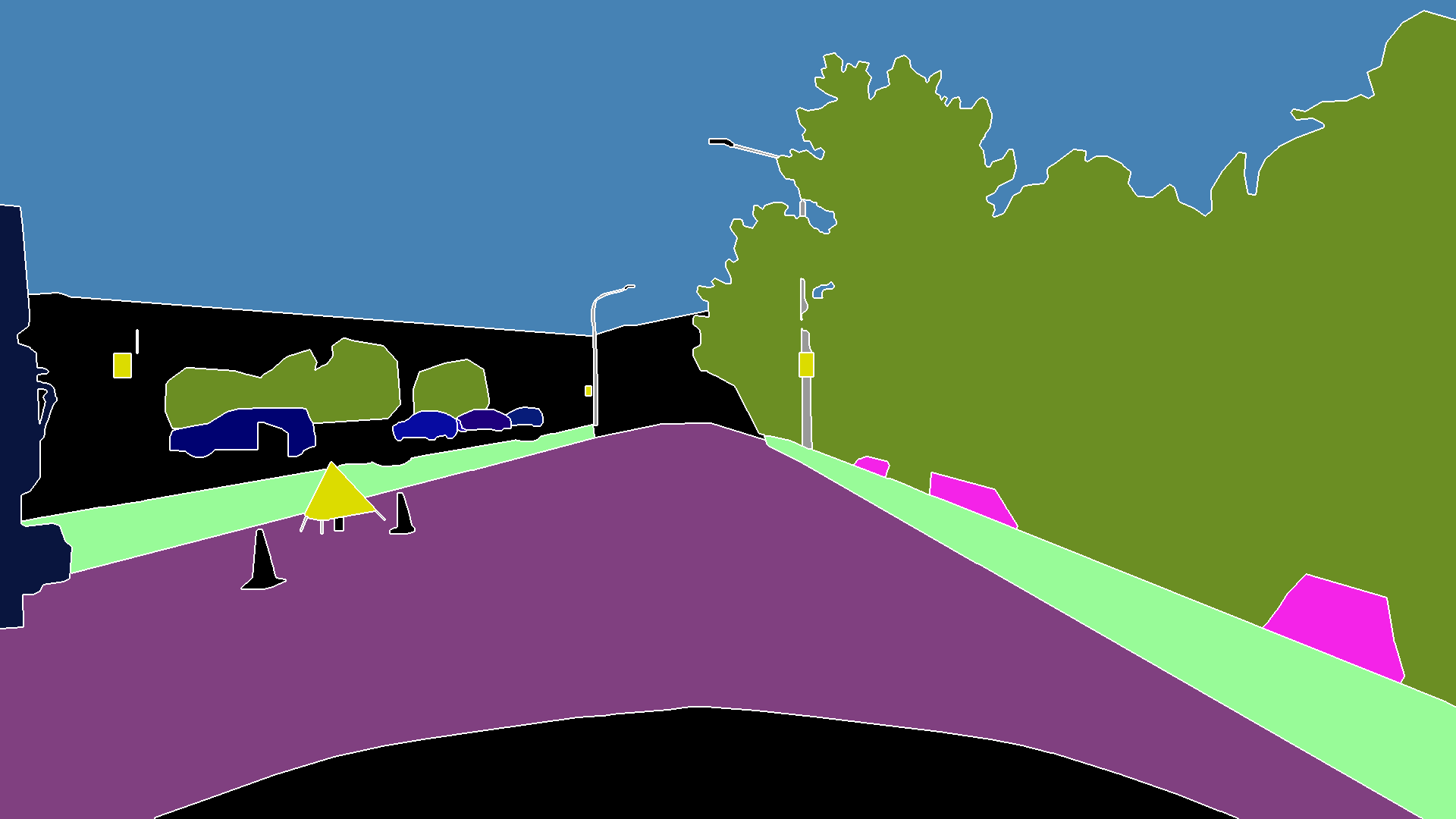}\\

\end{tabular}}%
\caption{3D vs.\ 2D. This example shows that the lidar point cloud (left, projected onto RGB image) can deteriorate in adverse weather, yielding insufficient information for 3D annotation. By contrast, the distant cars are captured with 2D annotations (right).}
\label{fig:annotation:why:2d}
\end{wrapfigure}

Along with the processed multi-sensor data, \Ours{} contains high-quality panoptic labels.
\Cref{fig:annotation:why:2d} illustrates that lidar point clouds provide insufficient information for accurate 3D annotation of objects in adverse conditions.
We thus label the 2D frame camera images, due to their high resolution and ease of interpretability for annotators.
The RGB images are labeled for panoptic segmentation with 19 classes, following the taxonomy of Cityscapes evaluation classes~\cite{Cityscapes}.
Even 2D panoptic labeling is highly challenging in adverse conditions because these images contain regions with indiscernible semantic content:
the image may be too dark or blurry to recognize an object, or a rain droplet on the lens may partially occlude the field-of-view.
To account for such cases, we asked annotators to follow the decision hierarchy shown in \cref{fig:ann_decision_hierarchy} for panoptic labeling.
If the object class is indiscernible, the pixel is labeled as \texttt{unknown\_class} (and by extension \texttt{unknown\_instance}), otherwise it belongs to either a \texttt{thing} class or a \texttt{stuff} class.
Pixels outside the 19 defined classes (\eg, a wheelchair) are given the fallback label \texttt{other\_class} and treated similarly to \texttt{stuff}.
If the pixel belongs to a \texttt{thing} class, the annotator can either assign it to an instance or label it as \texttt{unknown\_instance}.
This decision hierarchy results in two levels of uncertainty: class and instance uncertainty.
Instance uncertainty occurs when we know the class but not its instance membership, \eg, for pixels at the blurry boundary between two adjacent cars.

\begin{figure}[b]
\centering
\begin{tabular}{@{}c@{\hspace{0.05cm}}c@{\hspace{0.05cm}}c@{\hspace{0.05cm}}c@{}}
\subfloat{\fontsize{7}{8}\selectfont RGB Image} &
\subfloat{\fontsize{7}{8}\selectfont H1} &
\subfloat{\fontsize{7}{8}\selectfont H2}&
\subfloat{\fontsize{7}{8}\selectfont Difficulty Map} \\
\includegraphics[width=0.245
\textwidth, trim=250 350 1000 500, clip]{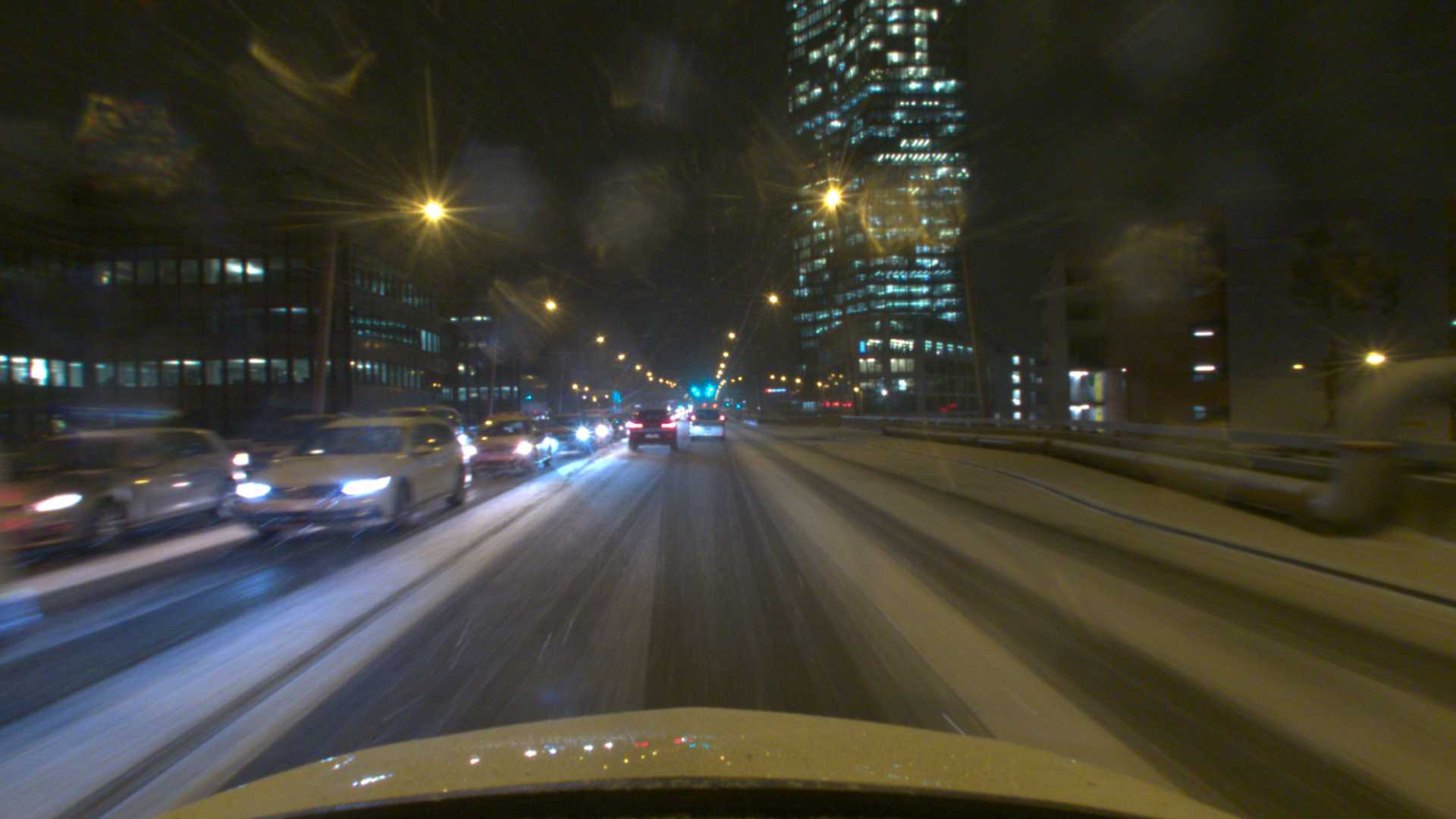} &
\includegraphics[width=0.245\textwidth, trim=250 350 1000 500, clip]{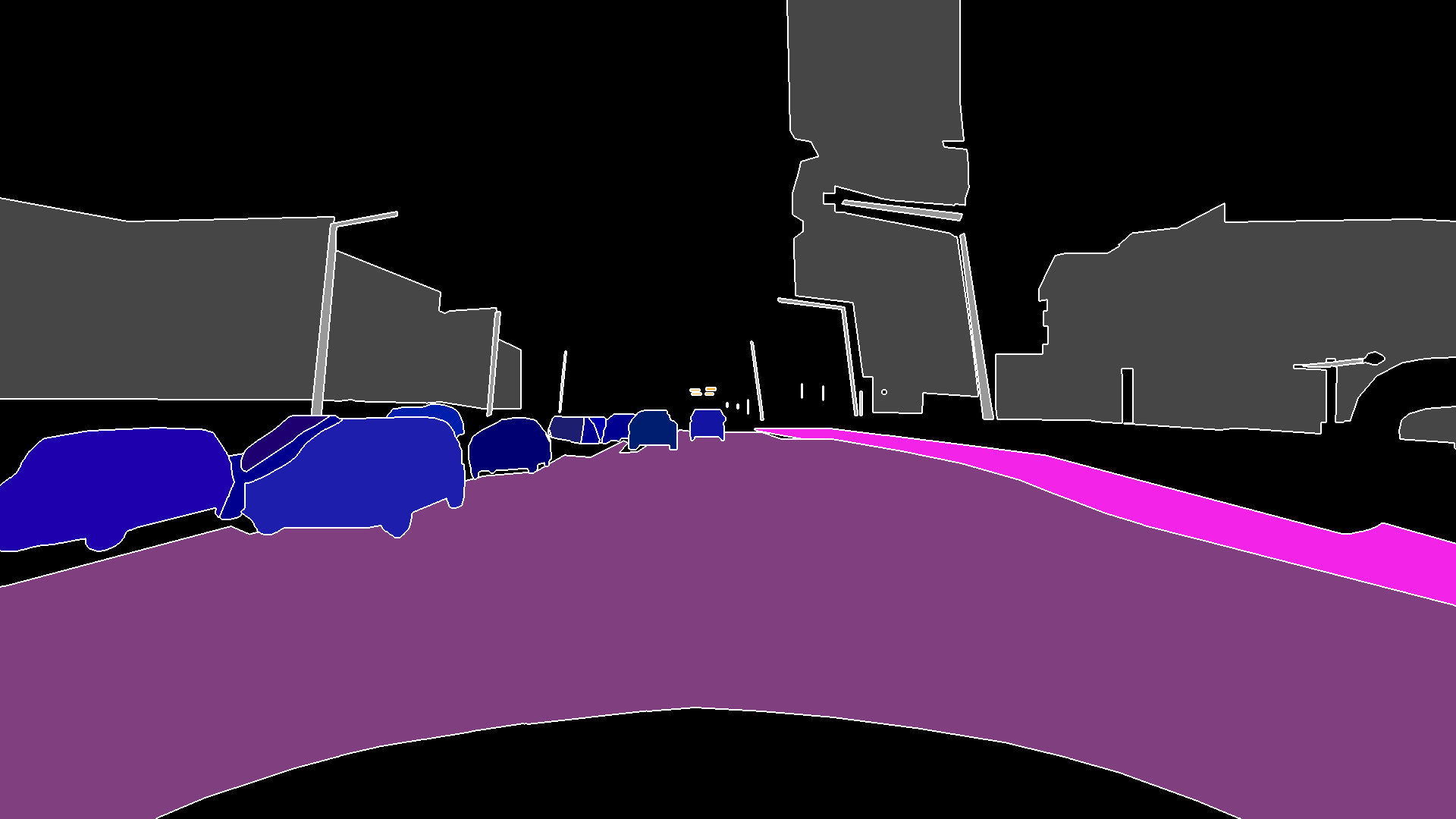} &
\includegraphics[width=0.245\textwidth, trim=250 350 1000 500, clip]{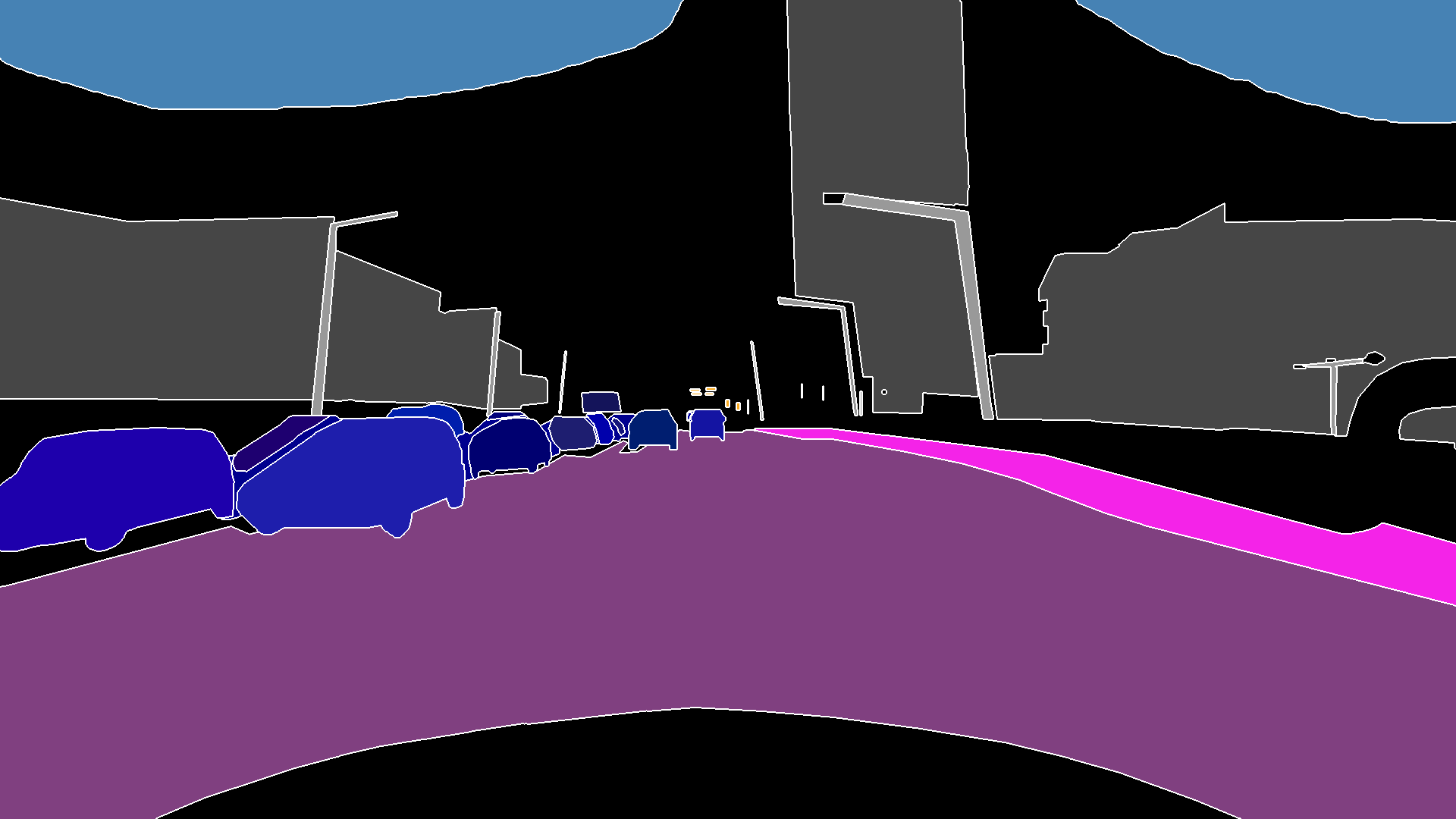} &
\includegraphics[width=0.245\textwidth, trim=250 350 1000 500, clip]{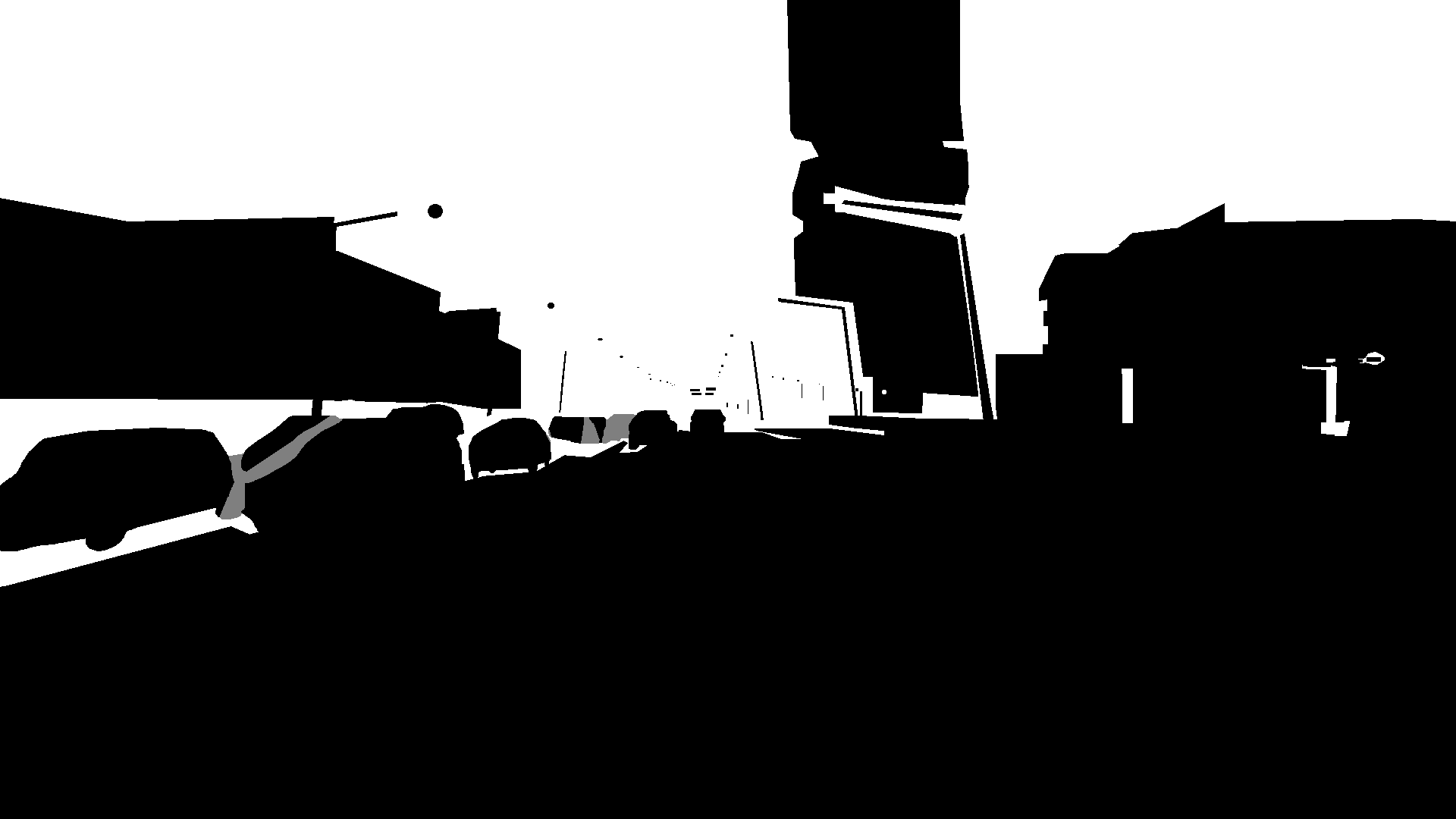}\\

\end{tabular}
\caption{Example of stage 1 and stage 2 panoptic annotations H1 and H2. The auxiliary data available in stage 2 allows better separation between the three car instances on the left, reducing the \texttt{unknown\_instance} area from H1 to H2, but keeping a \texttt{difficult\_instance} label (grey) in the difficulty map. Notice the additional class labels added to H2 for distant cars; the corresponding regions keep the \texttt{difficult\_class} (white) label in the difficulty map.}
\label{fig:annotation:output}
\end{figure}

\tikzstyle{process} = [rectangle, 
minimum width=1cm, 
minimum height=1cm, 
text centered, 
text width=2.2cm, 
draw=black, 
fill=color1!20]

\tikzstyle{decision} = [diamond, 
minimum width=0.8cm, 
minimum height=0.8cm, 
text centered, 
text width=1.4cm, 
draw=black, 
fill=color2!40]

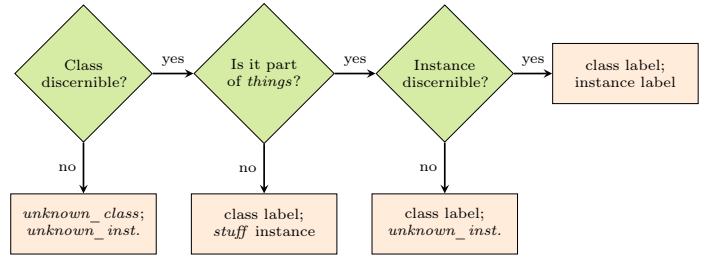
\begin{figure}
    \centering
    \resizebox{0.75\linewidth}{!}{%
    \begin{tikzpicture}[node distance=2cm]
    
    % \node (pixel) [startstop] {Pixel};
    \node (dec1) [decision] {Class\\discernible?};
    
    \node (pro1) [process, below of=dec1, yshift=-0.5cm] {\emph{unknown\_class};\\\emph{unknown\_inst.}};
    \node (dec2) [decision, right of=dec1, xshift=1cm] {Is it part\\of \emph{things}?};
    
    \node (pro2) [process, below of=dec2, yshift=-0.5cm] {class label;\\\emph{stuff} instance};
    \node (dec3) [decision, right of=dec2, xshift=1cm] {Instance\\discernible?};
    
    \node (pro3) [process, below of=dec3, yshift=-0.5cm] {class label;\\\emph{unknown\_inst.}};
    \node (pro4) [process, right of=dec3, xshift=1cm] {class label;\\instance label};
    
    % \draw [arrow] (pixel) -- (dec1);
    \draw [arrow] (dec1) -- node[anchor=east] {no} (pro1);
    \draw [arrow] (dec1) -- node[anchor=south] {yes} (dec2);
    \draw [arrow] (dec2) -- node[anchor=east] {no} (pro2);
    \draw [arrow] (dec2) -- node[anchor=south] {yes} (dec3);
    \draw [arrow] (dec3) -- node[anchor=east] {no} (pro3);
    \draw [arrow] (dec3) -- node[anchor=south] {yes} (pro4);
    
    \end{tikzpicture}%
    }
    \caption{Flowchart of uncertainty-aware panoptic annotation.}
    \label{fig:ann_decision_hierarchy}
\end{figure}

We design a two-stage annotation protocol.
In stage 1, the RGB image is labeled with an initial ground-truth panoptic label H1 following the hierarchical protocol in \cref{fig:ann_decision_hierarchy}.
In this stage, the annotator only has access to the RGB image under annotation.
In stage 2, H1 is refined with the same decision hierarchy to obtain the final ground truth panoptic label H2 (see \cref{fig:annotation:output}).
The refinement is made possible by additionally inspecting auxiliary data, which includes enhanced versions of the original image~\cite{shi2018nighttime}, event and lidar data, temporally adjacent frames of the sequence, and a corresponding sequence from the same location recorded under normal conditions.
This auxiliary data enables the annotators to fill in previously uncertain regions (\ie, droplets on RGB image in \cref{fig:annotation:aux:data}) and correct possible mistakes from stage 1.
In all stages, we asked annotators to prioritize traffic-related classes, \ie, \texttt{things} classes plus road, traffic light, and traffic sign.
The total annotation time was 11,827 hours, averaging 4.6 hours per image. It was divided equally between stage 1 and stage 2.

The two-stage annotation process allows us to assign ternary \emph{difficulty} levels to our labels (see last column in \cref{fig:annotation:output}).
Pixels with valid labels that are fully consistent in H1 and H2 are labeled \texttt{not\_difficult}.
Pixels with a consistent class label in H1 and H2, but an inconsistent instance label or the \texttt{unknown\_instance} label, receive a \texttt{difficult\_instance} label.
All other pixels, \ie, pixels with an inconsistent class label in H1 and H2 or pixels that are \texttt{unknown\_class}, receive a \texttt{difficult\_class} label.

\subsection{Splits and Benchmarks}
\label{subsec:overview}

\begin{figure}[b]
\centering
\begin{tabular}{@{}c@{\hspace{0.03cm}}c@{\hspace{0.03cm}}c@{\hspace{0.03cm}}c@{\hspace{0.03cm}}c@{\hspace{0.03cm}}c@{\hspace{0.03cm}}c@{}}
\subfloat{\tiny RGB Image} &
\subfloat{\tiny Lidar} &
\subfloat{\tiny Events}&
\subfloat{\tiny Radar}&
\subfloat{\tiny Corr. Image}&
\subfloat{\tiny Panoptic GT}&
\subfloat{\tiny Difficulty Map}  \\
\includegraphics[width=0.14\textwidth]{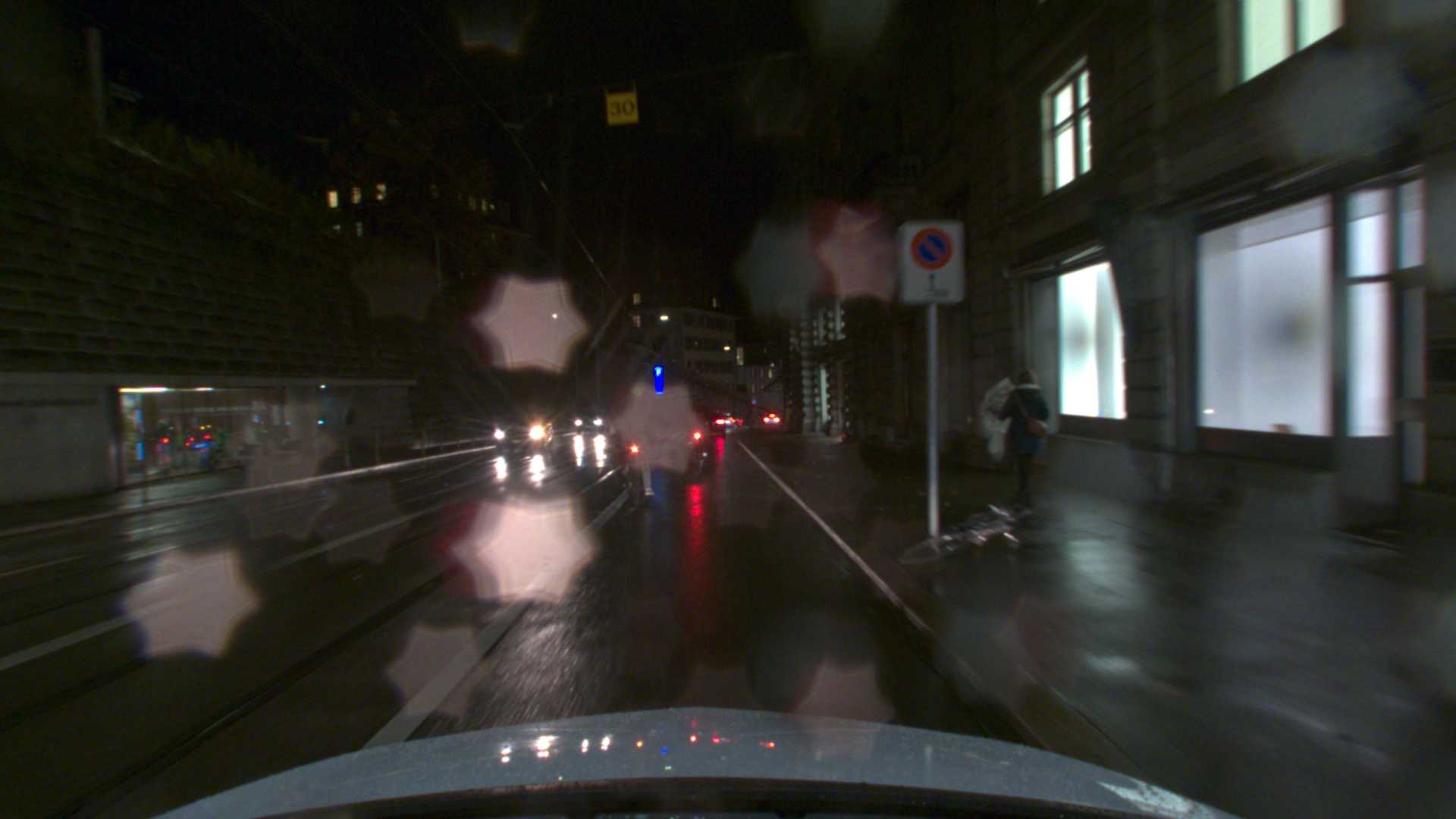} &
\includegraphics[width=0.14\textwidth]{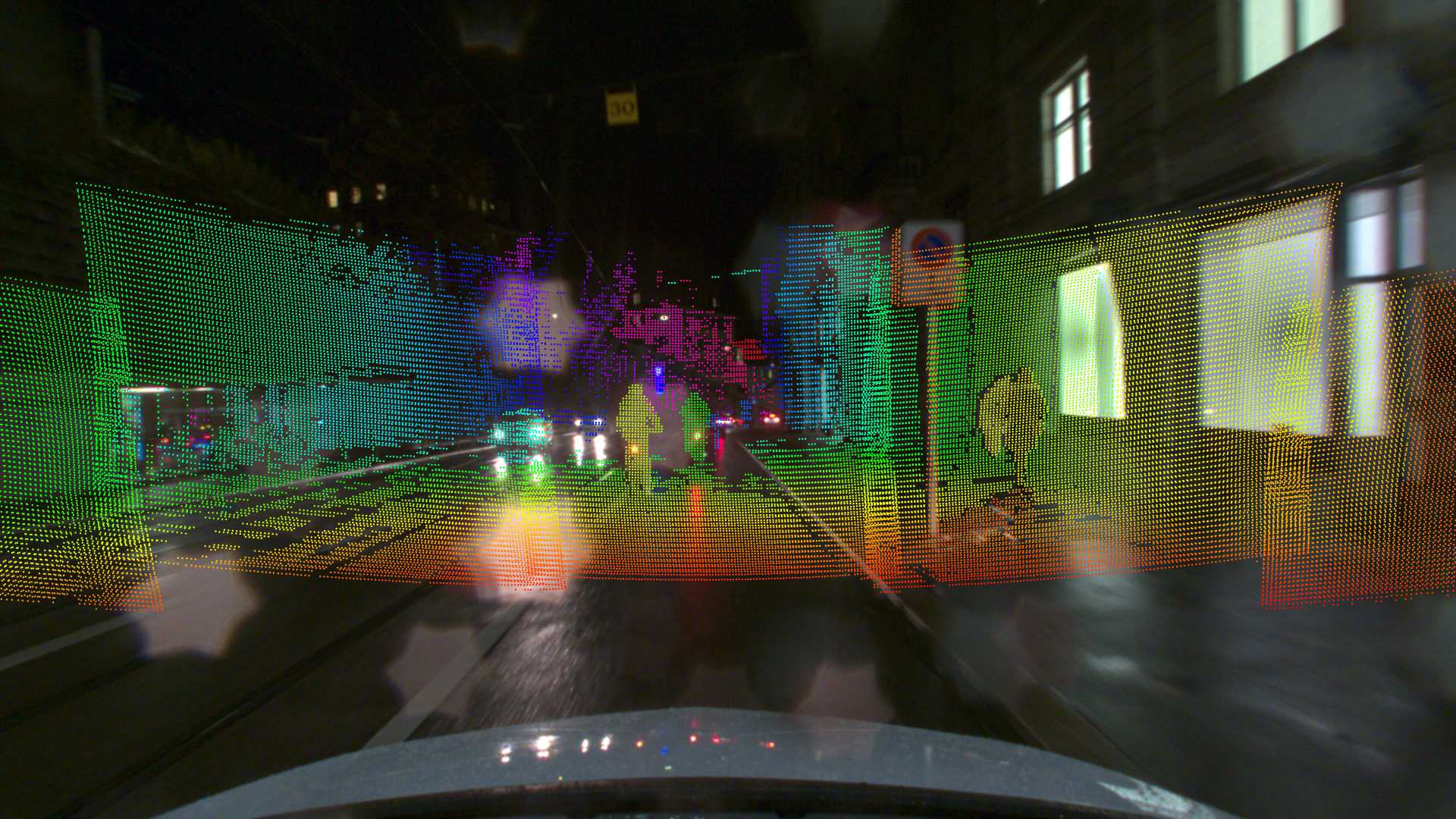} &
\includegraphics[width=0.14\textwidth]{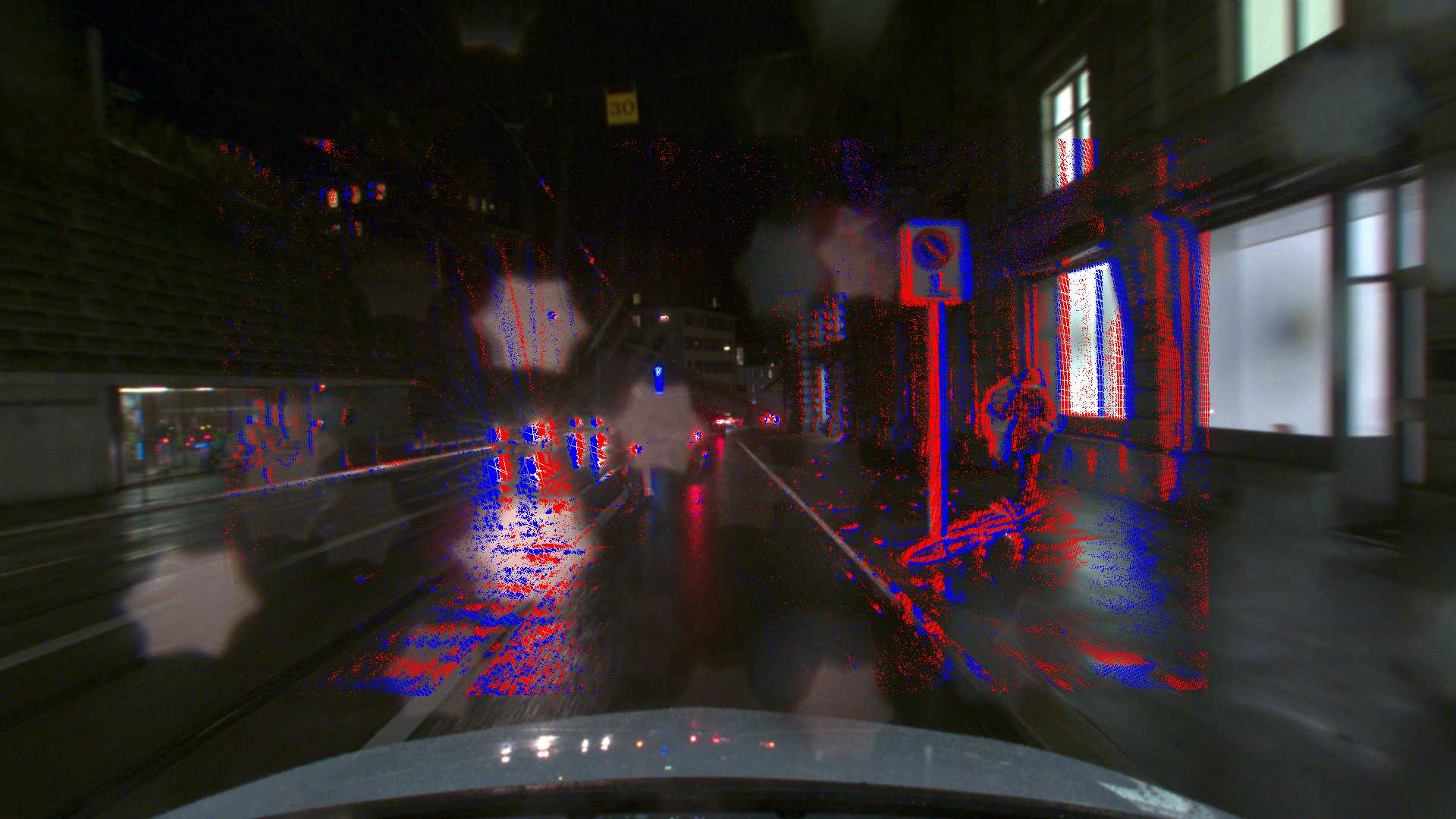} &
\includegraphics[angle=90, trim=0 6835 0 0, clip, width=0.14\textwidth]{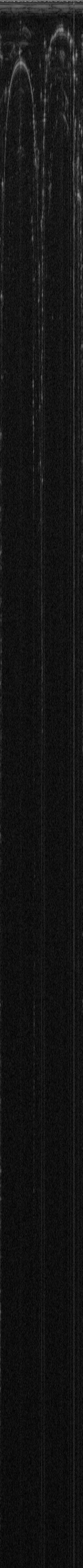}&
\includegraphics[width=0.14\textwidth]{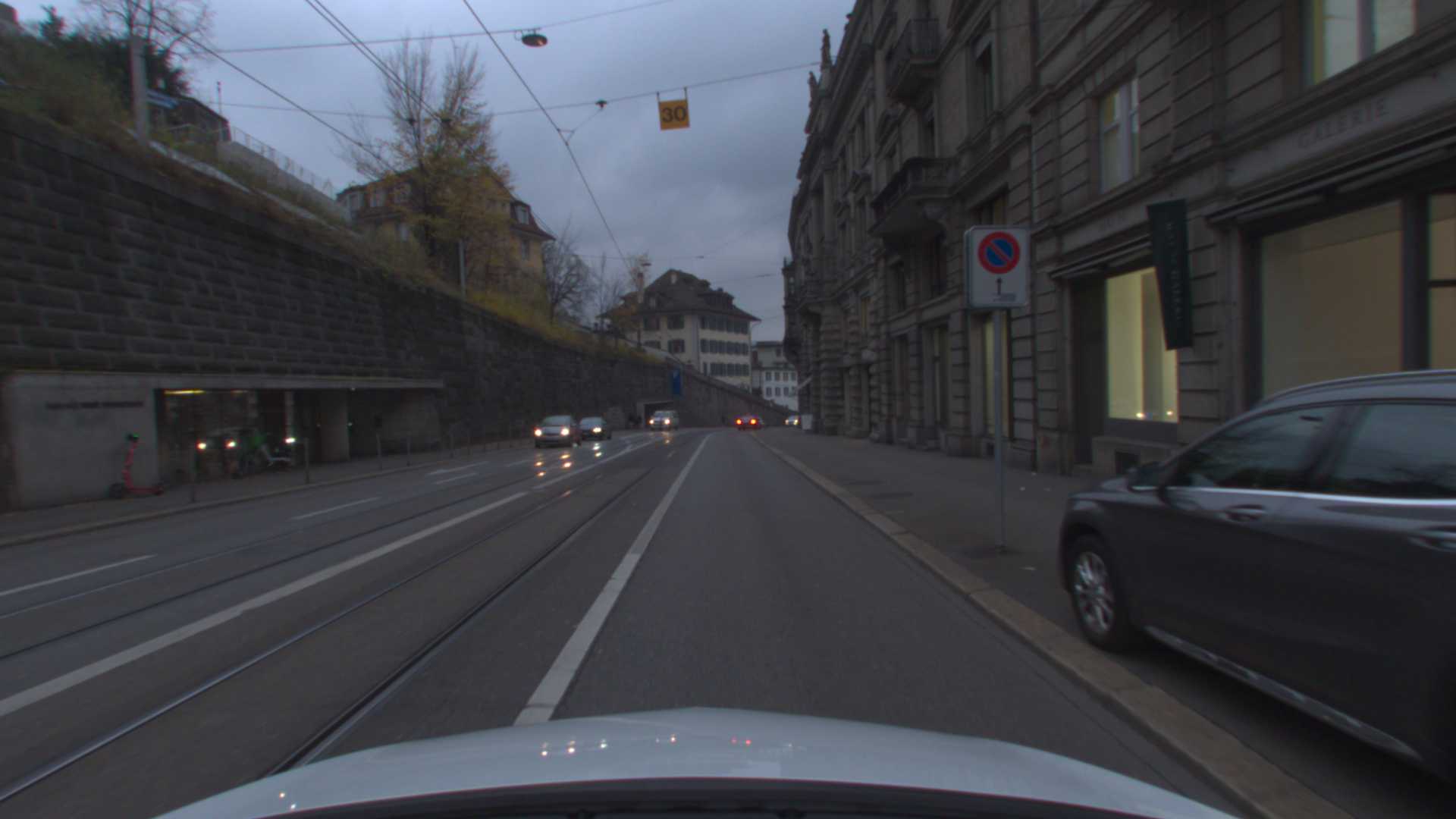} &
\includegraphics[width=0.14\textwidth]{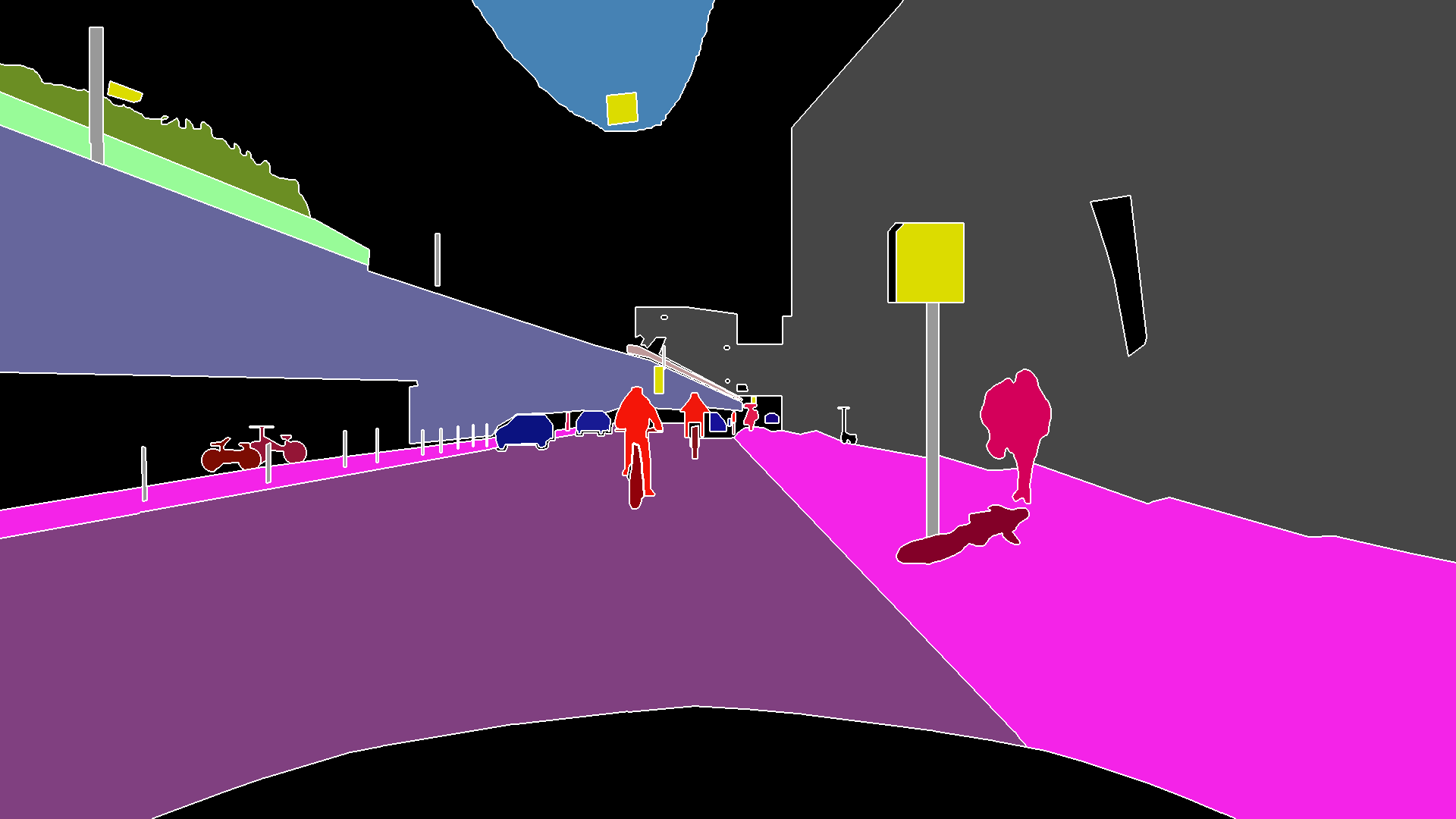} &
\includegraphics[width=0.14\textwidth]{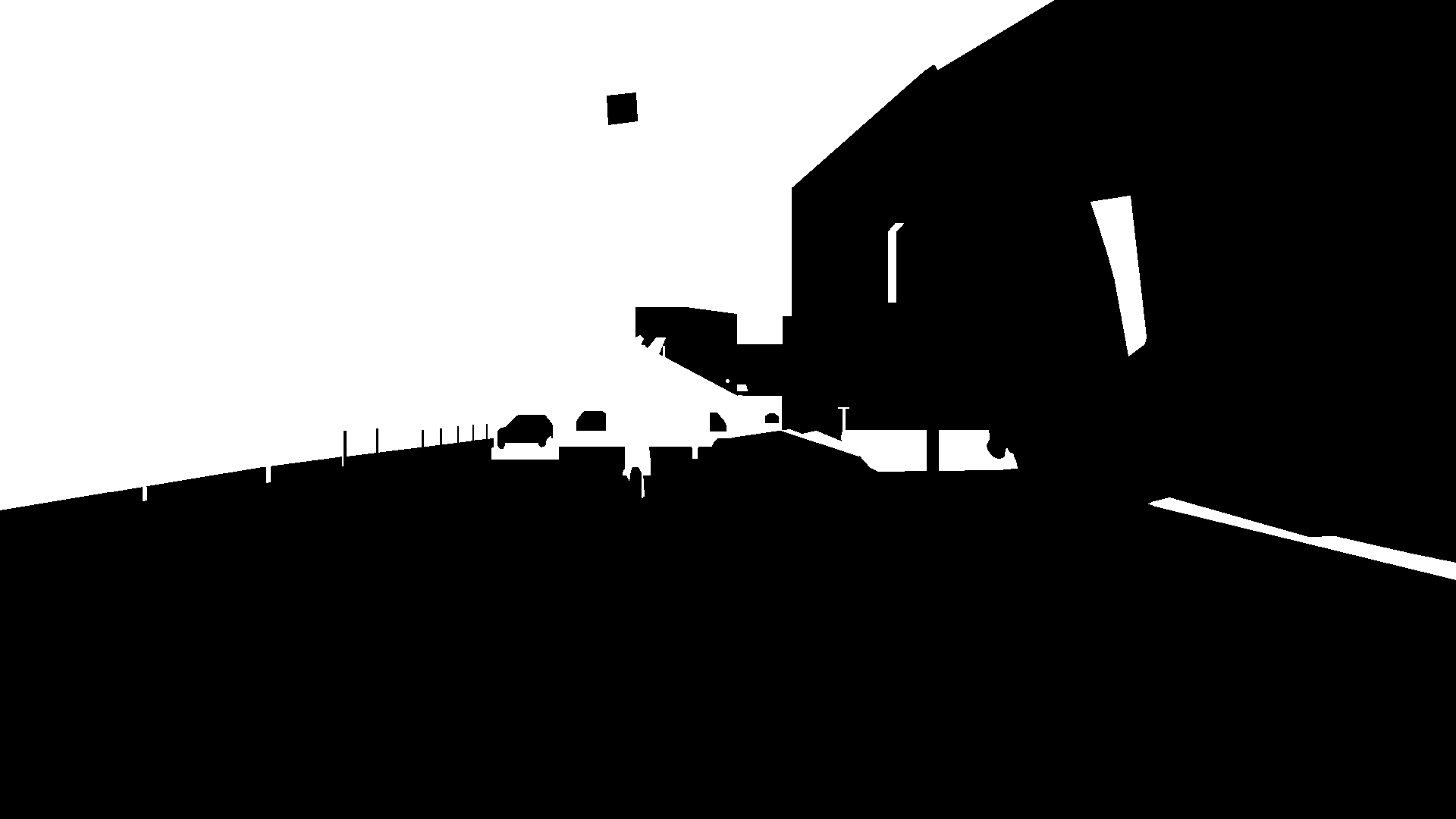}\\

\includegraphics[width=0.14\textwidth]{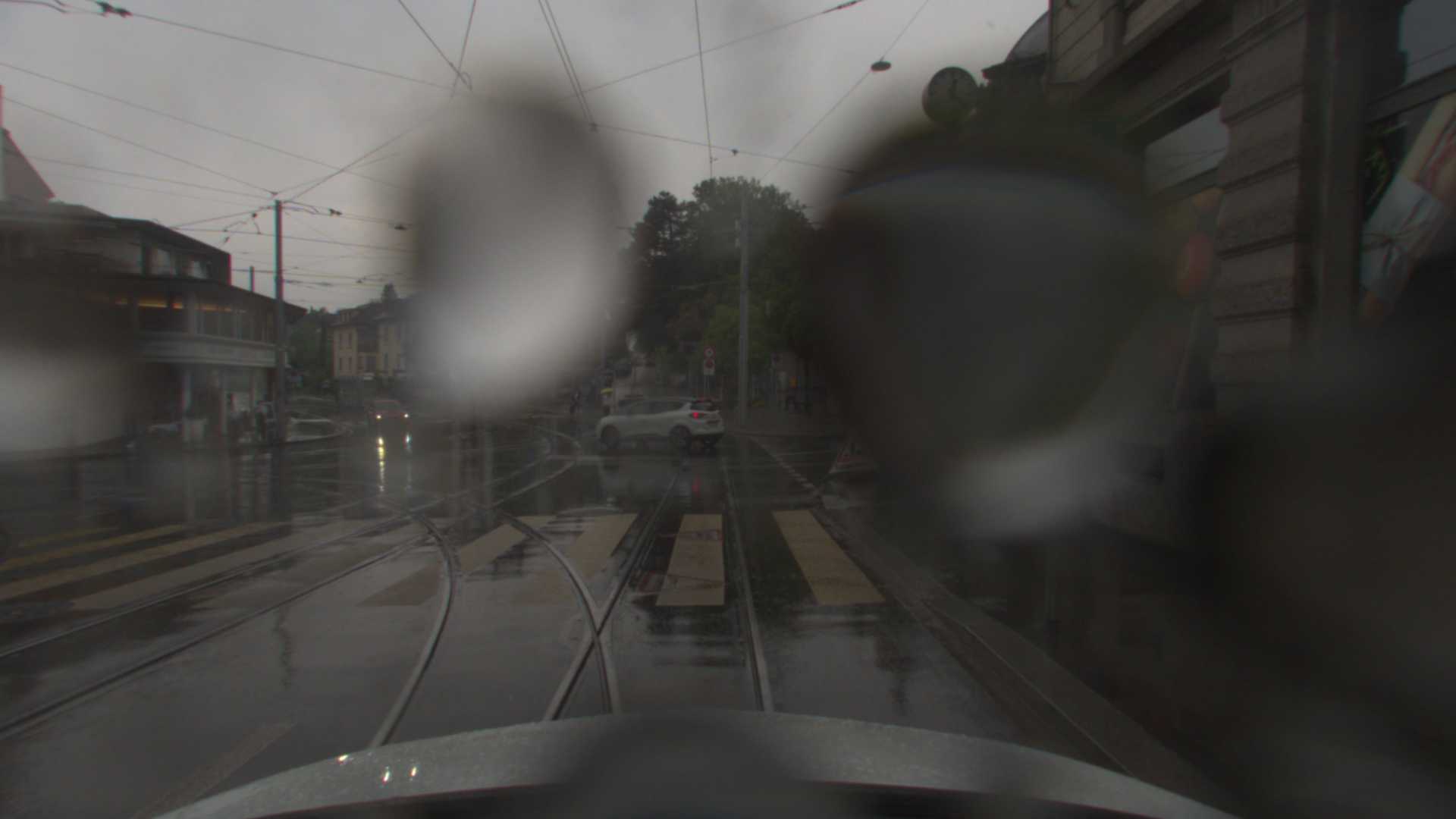} &
\includegraphics[width=0.14\textwidth]{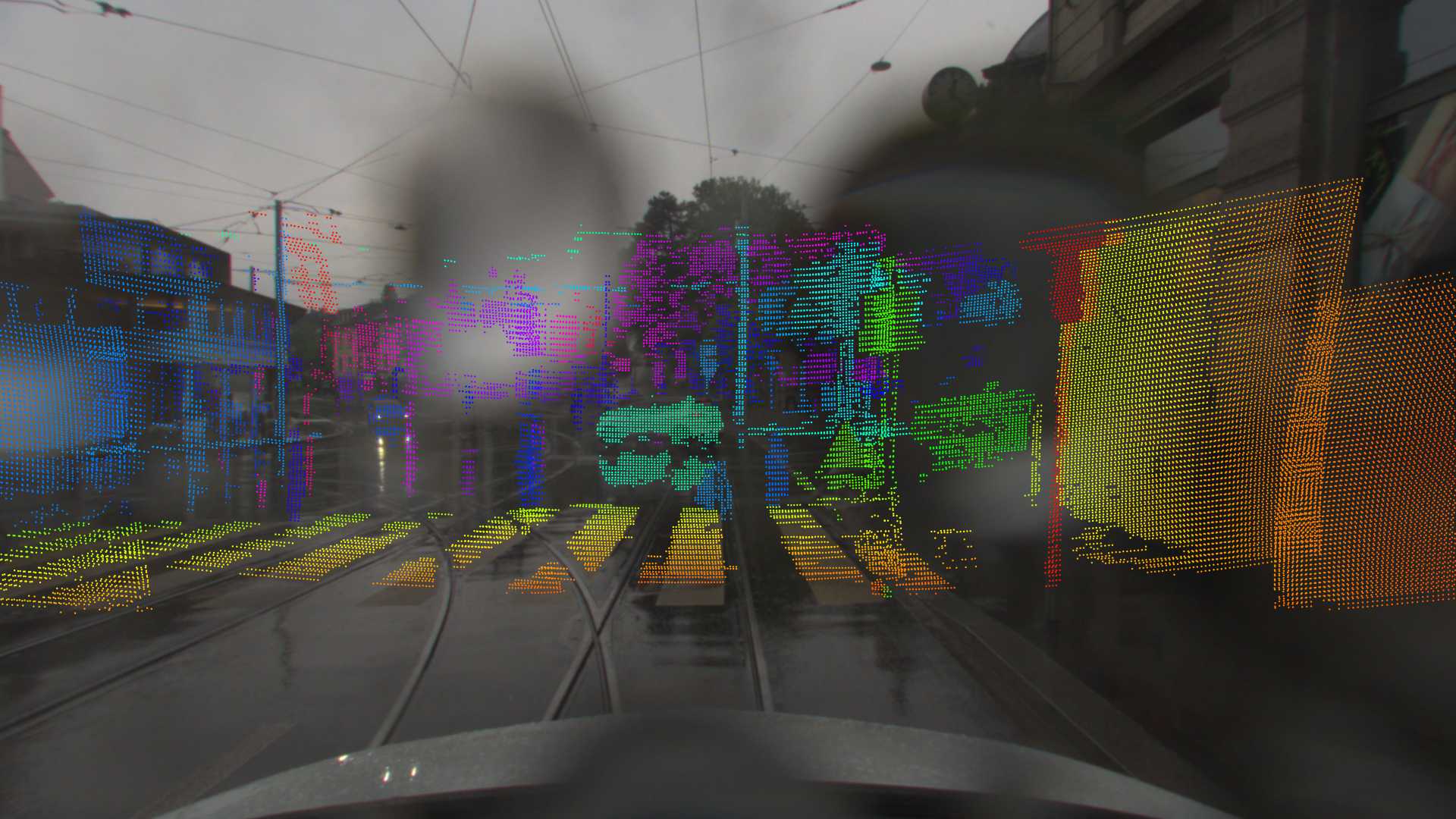} &
\includegraphics[width=0.14\textwidth]{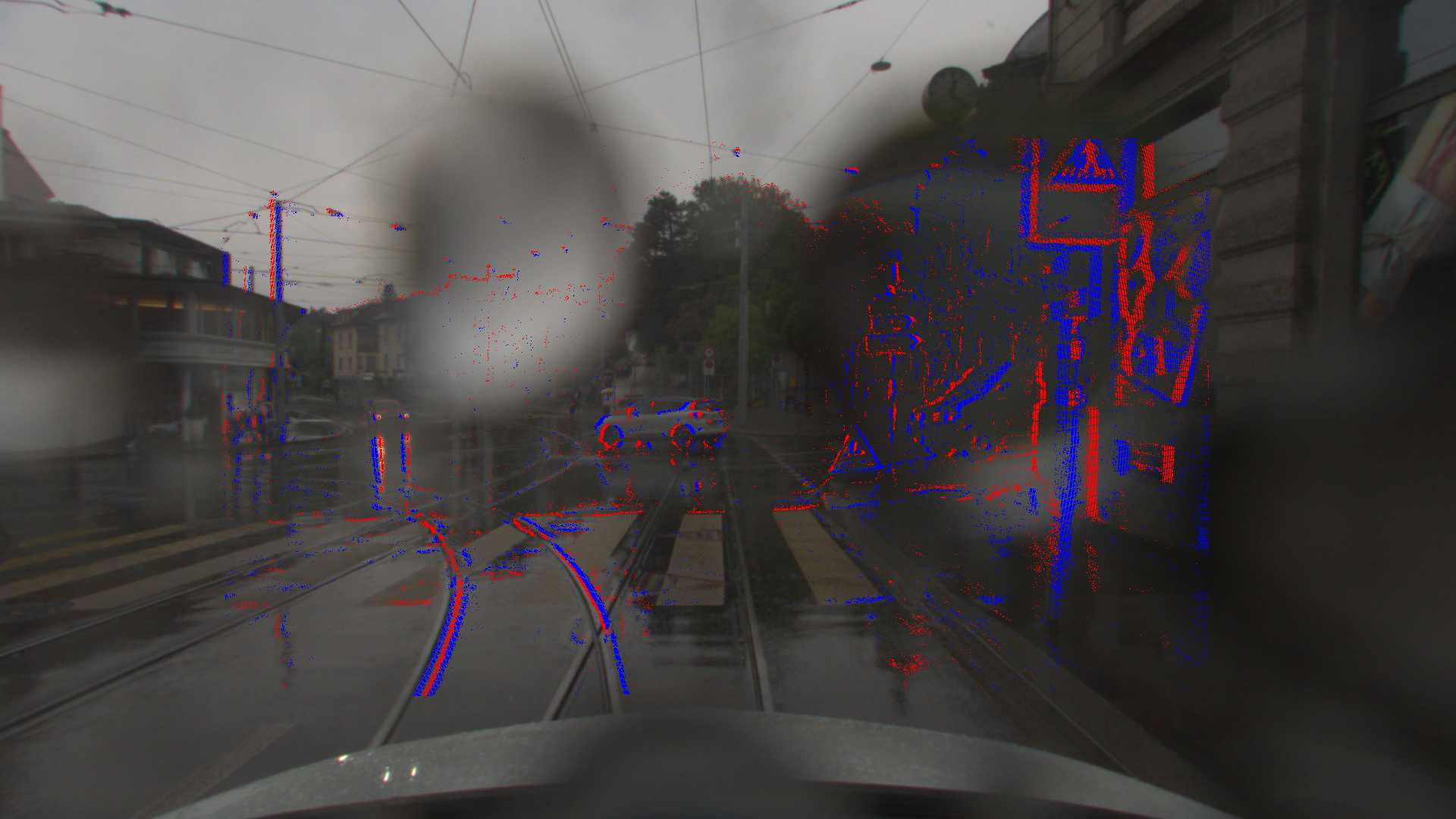} &
\includegraphics[angle=90, trim=0 6835 0 0, clip,width=0.14\textwidth]{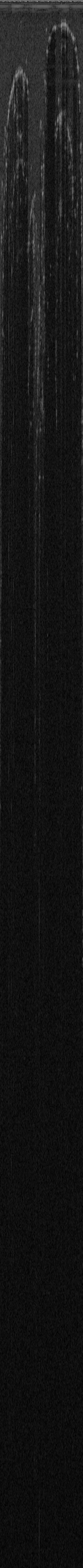}&
\includegraphics[width=0.14\textwidth]{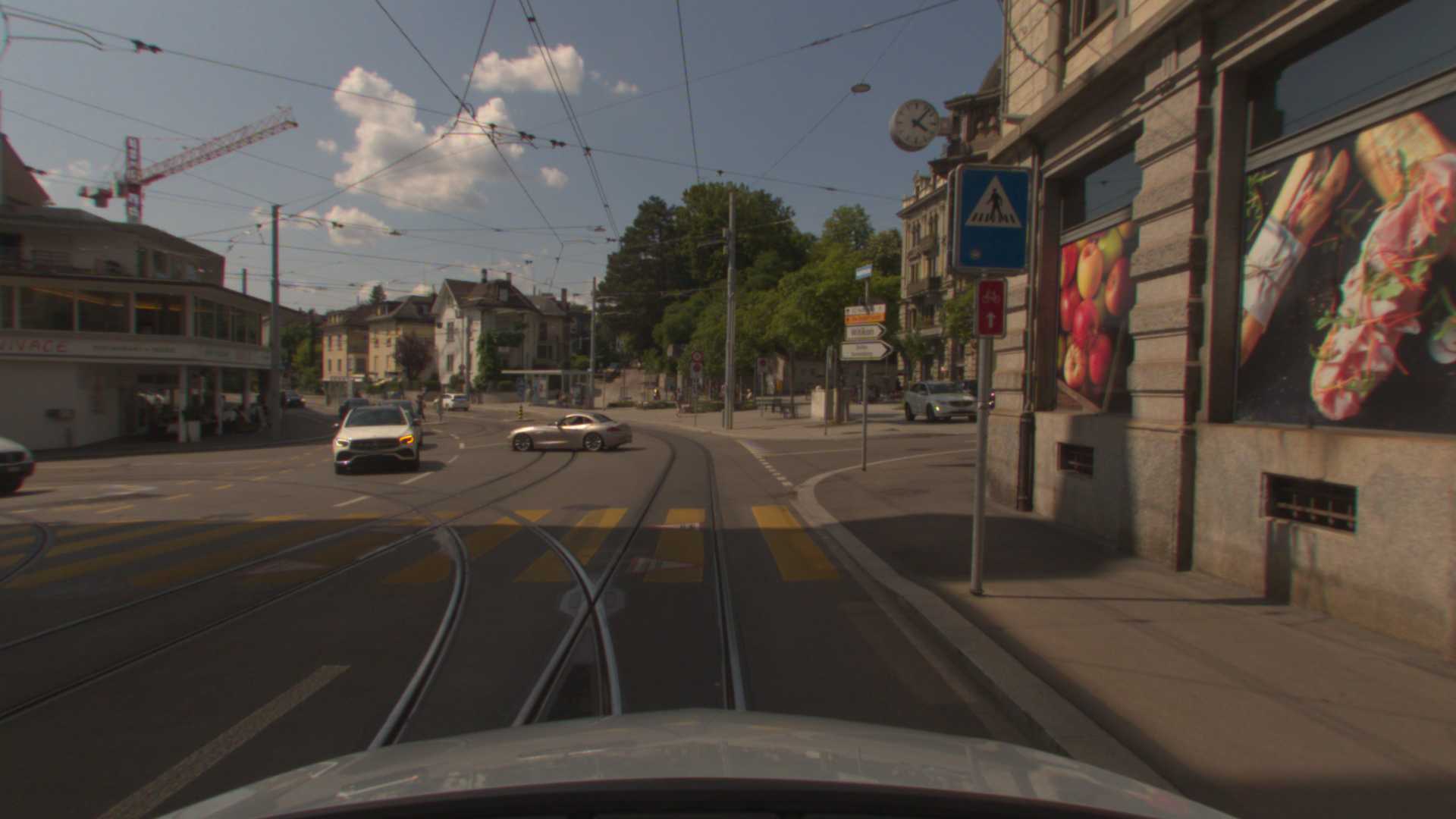} &
\includegraphics[width=0.14\textwidth]{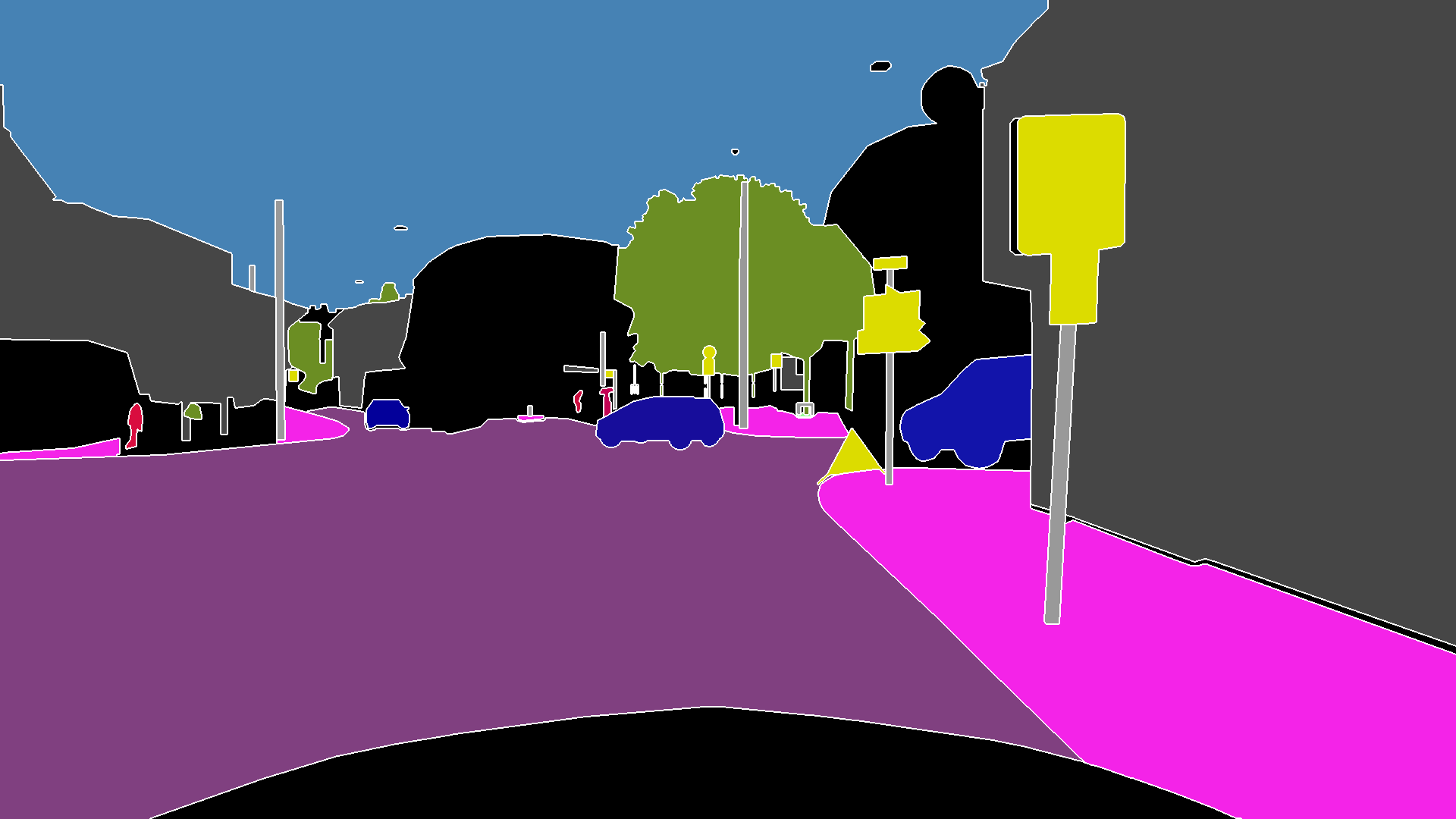} &
\includegraphics[width=0.14\textwidth]{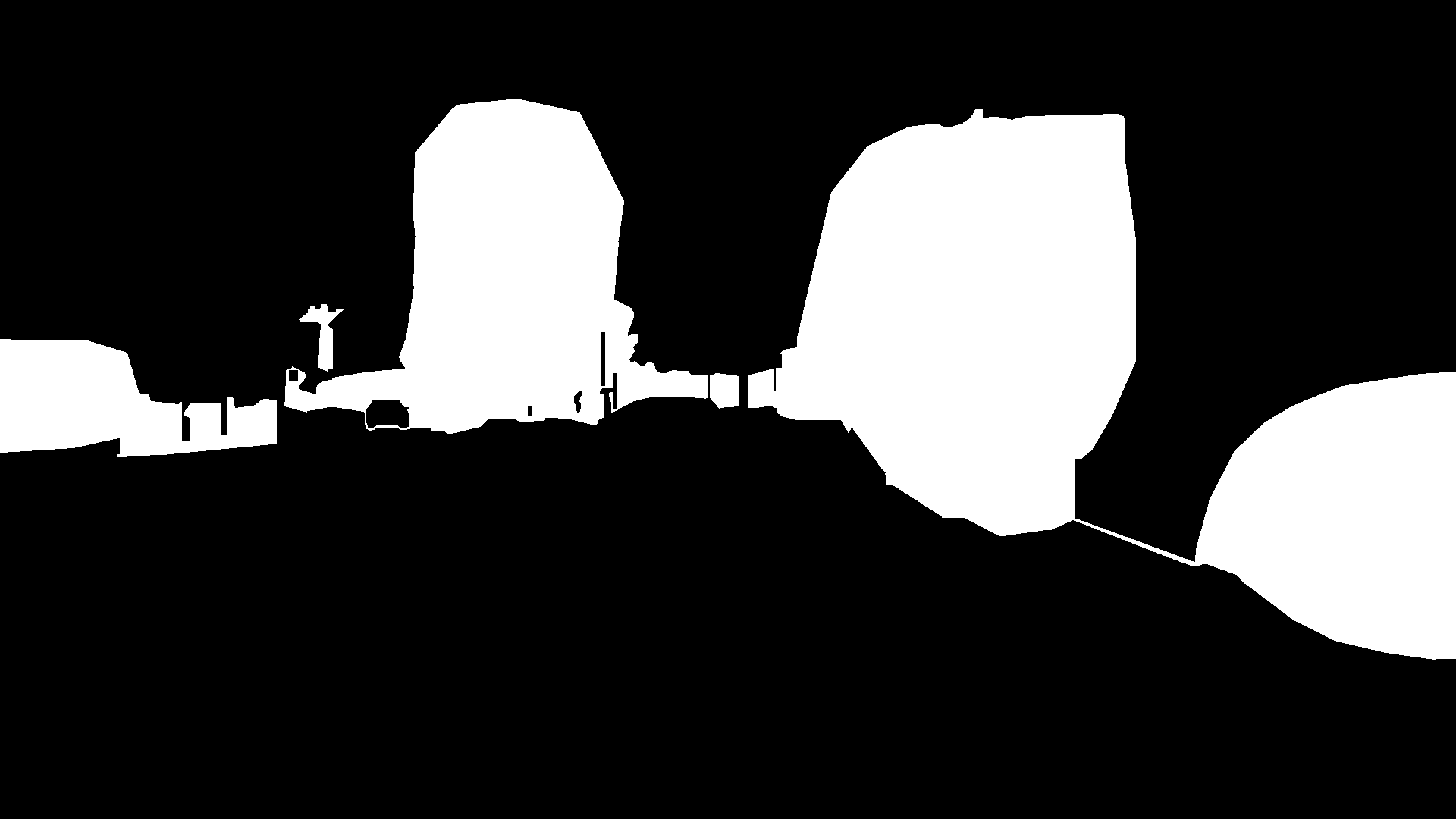}\\
\end{tabular}
% \vspace{-0.5em}
\caption{Visualization of two adverse-condition samples from \Ours{}. From left to right: RGB image; motion-compensated lidar points projected and overlaid with the image; events projected onto the image (assuming infinite distance); azimuth-range radar scan (with ranges above a threshold cropped out); corresponding normal-condition image; panoptic ground truth; difficulty map. Best viewed zoomed in.}
\label{fig:annotation:aux:data}
\end{figure}

\Ours{} consists of 2500 annotated multimodal samples, 500 of which are recorded in daytime and clear weather.
The remaining 2000 adverse-condition samples are split between 1000 daytime (333/334/333 samples in fog/rain/snow) and 1000 nighttime (250/250/250/250 samples in clear/fog/rain/snow) samples.
\Cref{fig:annotation:aux:data} shows two example adverse-condition scenes with 
a visualization of each recorded modality, the corresponding normal-condition images, and the annotation outputs.
We split the full dataset into 1500 training, 250 validation, and 750 test samples, stratified by condition.
The individual splits are strictly geographically separated across conditions.
The annotations of the test set are withheld, and the test set evaluation is only accessible through a submission system.
Public benchmarks will be provided for three tasks: 1) semantic segmentation, 2) panoptic segmentation, and 3) uncertainty-aware panoptic segmentation (see \cref{sec:upq}).
Each of the three benchmarks will have two different tracks: i) using RGB input only, and ii) using multimodal input.

%% file: sec/4_upq.tex
\section{Uncertainty-Aware Panoptic Segmentation}
\label{sec:upq}

The ternary difficulty levels assigned to each pixel during the annotation of \Ours{} (see \cref{subsec:annotation}) enable a novel task: uncertainty-aware panoptic segmentation.
In this task, a panoptic segmentation model is compensated for errors in difficult image regions if it predicts the difficulty level correctly.
To incorporate this idea into evaluation, we introduce the uncertainty-aware panoptic quality (UPQ) metric, an extension of panoptic quality~\cite{kirillov2019panoptic}.

\subsection{Uncertainty-Aware Panoptic Quality}

In addition to the panoptic prediction and ground truth, UPQ takes as input a binary class confidence prediction and a binary instance confidence prediction, which are evaluated against the ternary difficulty maps (see \cref{subsec:annotation}).
This confidence comparison transforms some pixels in the panoptic prediction to either \texttt{ANY} or \texttt{VOID}, as follows:
First, the binary class confidence prediction divides all pixels into class-unconfident (CU) and class-confident (CC) sets.
CU pixels are compared with the ground truth difficulty map:
If they are indeed \texttt{class\_difficult}, they are converted to \texttt{ANY} pixels in the panoptic prediction, otherwise, they are converted to \texttt{VOID}.
The CC pixels are instead divided into instance-unconfident (IU) and instance-confident (IC) sets according to the binary instance confidence prediction.
IU pixels with a correct class prediction are compared with the difficulty map:
If they are \texttt{difficult\_instance} or \texttt{difficult\_class}, they are converted to \texttt{ANY} pixels, otherwise, they are converted to \texttt{VOID}.
Thus, ANY masks pixels that are correctly predicted as uncertain (class or instance level), while VOID masks pixels that are erroneously predicted as uncertain. Note that ANY/VOID masks apply to the predictions, not the ground truth. 

Analogously to the standard PQ~\cite{kirillov2019panoptic}, UPQ is composed of two steps: segment matching and UPQ computation given the matched segments.
Compared to PQ, UPQ modifies the first step, by incorporating \texttt{ANY} pixels as wild cards and \texttt{VOID} pixels as wrong predictions. 
Hence, UPQ forgives errors for difficult pixels when predictions are correctly uncertain but penalizes predicting "easy" pixels as uncertain.
Note that, if a model predicts all pixels as confident, no \texttt{ANY} or \texttt{VOID} pixels are created and UPQ reduces to the standard PQ.

\begin{figure}[tb]
    \centering
    \begin{tikzpicture}[node distance=2.7cm]

    \node[inner sep=0pt] (pred) {\includegraphics[width=2.7cm]{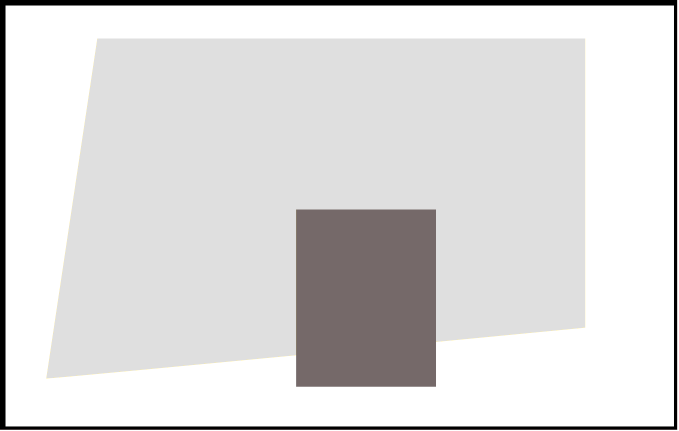}};
    \node[above] at (pred.north) {\textbf{Prediction}};
    
    \node[right of=pred, xshift=0.3cm, inner sep=0pt] (gt) {\includegraphics[width=2.7cm]{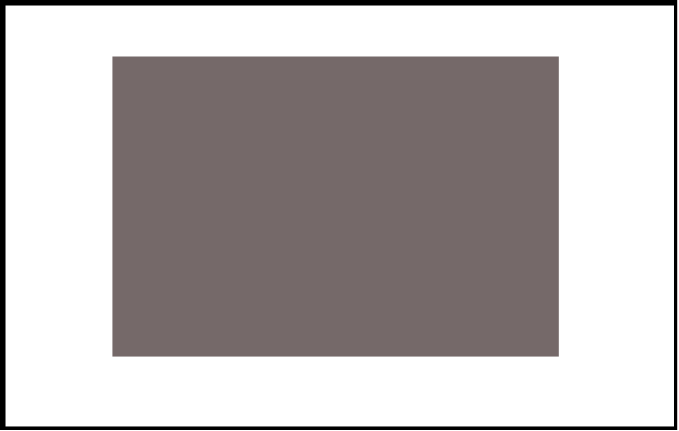}};
    \node[above] at (gt.north) {\textbf{Ground Truth}};
    
    \node[right of=gt, xshift=0.3cm, inner sep=0pt] (res) {\includegraphics[width=2.7cm]{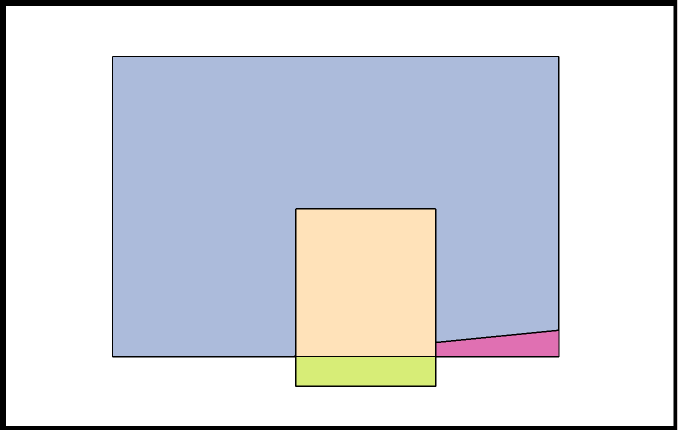}};
    \node[above] at (res.north) {\textbf{Computation}};

    \node[right of=res, xshift=0.4cm, yshift=0.08cm, inner sep=0pt, text width=3cm] (legend) {Segment matching:\\[5pt]$\text{IoU} = \frac{\hspace{0.3cm}}{ \hspace{0.3cm} + \hspace{0.3cm} + \hspace{0.3cm}}$\\[14pt] UPQ computation:\\[5pt]$\text{IoU} = \frac{\hspace{0.3cm} + \hspace{0.3cm}} {\hspace{0.3cm} + \hspace{0.3cm} + \hspace{0.3cm} + \hspace{0.3cm}}$};
    
    \node[inner sep=0pt] at (9.10, 0.65) {\includegraphics[width=0.25cm]{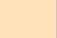}};;
    \node[inner sep=0pt] at (8.61, 0.40) {\includegraphics[width=0.25cm]{figs/orange.pdf}};;
    \node[inner sep=0pt] at (9.10, 0.40) {\includegraphics[width=0.25cm]{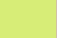}};;
    \node[inner sep=0pt] at (9.59, 0.40) {\includegraphics[width=0.25cm]{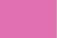}};;
    
    \node[inner sep=0pt] at (9.10, -0.62) {\includegraphics[width=0.25cm]{figs/orange.pdf}};;
    \node[inner sep=0pt] at (9.59, -0.62) {\includegraphics[width=0.25cm]{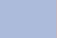}};;
    \node[inner sep=0pt] at (8.61, -0.87) {\includegraphics[width=0.25cm]{figs/orange.pdf}};;
    \node[inner sep=0pt] at (9.10, -0.87) {\includegraphics[width=0.25cm]{figs/blue.pdf}};;
    \node[inner sep=0pt] at (9.59, -0.87) {\includegraphics[width=0.25cm]{figs/green.pdf}};;
    \node[inner sep=0pt] at (10.08, -0.87) {\includegraphics[width=0.25cm]{figs/pink.pdf}};;

    \end{tikzpicture}
    \caption{UPQ computation between a prediction containing an instance (black), \texttt{ANY} region (gray), and background (white), and a ground-truth instance (black). For segment matching, the \texttt{ANY} regions are ignored, which leads to an IoU\textgreater0.5 in this example. Before UPQ computation after matching, the matched \texttt{ANY} regions are replaced with the ground truth.}
    \label{fig:upq_intuition}
\end{figure}

\PAR{Segment Matching.}
In standard PQ, matches between predicted and ground-truth segments are formed when their intersection over union (IoU) is strictly greater than 0.5, which guarantees unique matching.
To maintain this property while accounting for \texttt{ANY} pixels, UPQ matches segments in two steps:
\begin{enumerate}[noitemsep,topsep=0pt]
    \item Ignore all \texttt{ANY} pixels in both prediction and ground truth, then match remaining segments with $\text{IoU}>0.5$. After matching, copy the ground truth class/instance labels to the \texttt{ANY} pixels within matched segments.
    \item Check remaining unmatched ground-truth segment for $>50\%$ overlap with ANY pixels. For matches, replace the \texttt{ANY} label with the ground truth.
\end{enumerate}
\Cref{fig:upq_intuition} conveys the intuition for this process:
by ignoring the \texttt{ANY} pixels in step 1, segments surrounded by accurate confidence predictions are more easily matched.
Even though the predicted segment has an IoU of less than 0.5 with the ground-truth one, it is still matched because the confidence level of the surrounding pixels is correctly predicted as \texttt{difficult\_class} or \texttt{difficult\_instance} (which results in \texttt{ANY} pixels).
Note that ANY pixels never form segments and thus cannot create FP or FN segments.
If the confidence level of the entire instance were correctly classified as difficult, it would still be matched in step 2.

\PAR{UPQ Computation.}
Given segment matches, UPQ is computed as
\begin{equation}
    \text{UPQ} = \frac{\sum_{(p, g)\in TP}\text{IoU}(p, g)}{|TP| + \frac{1}{2}|FP| + \frac{1}{2}|FN|},
    \label{eq:upq}
\end{equation}
where $p$ and $g$ are matched segments, and $|TP|$, $|FP|$, and $|FN|$ are the numbers of true positive, false positive, and false negative segments respectively.
See~\cite{kirillov2019panoptic} for details and the motivation behind this formulation.

\subsection{Threshold-Agnostic Evaluation}

UPQ evaluation requires binary class confidence and instance confidence predictions.
For a model that produces confidence \emph{scores} for both cases, UPQ must thus be evaluated at a specific operating point, by defining confidence thresholds for class and instance confidences respectively.
To obtain a threshold-agnostic summary metric, we define a linear grid of 16\texttimes 16 thresholds for class and instance confidences, evaluate UPQ for each of the 256 threshold pairs, and report the average value\textemdash AUPQ.

%% file: sec/5_analysis.tex
\section{Analysis and Experiments}

In this section, we present a thorough analysis of \Ours{}.
We analyze its annotations in \cref{subsec:statistics}, corroborate the benefit of the non-camera modalities of our dataset for dense semantic perception in \cref{subsec:impact_of_sensors}, demonstrate the increased difficulty MUSES presents for state-of-the-art semantic segmentation approaches compared to competing central benchmarks in \cref{subsec:additional_segmentation}, and evaluate baselines and oracles for the novel uncertainty-aware panoptic segmentation task in \cref{subsec:upq_experiments}. 
Performance is always reported on the respective test sets.

\begin{figure}[tb]
  \centering
  \begin{subfigure}{.49\textwidth}
  \begin{tikzpicture}[baseline]
    \begin{axis}[
      ybar stacked,      
      width=\linewidth,
      height=4cm,
      bar width=0.3cm,
      axis lines=left,
      enlarge x limits=0.1,
      axis line style={-}, 
      legend style={at={(0.5,1.0)}, anchor=south,   draw=none}, % Updated legend style
      ylabel={Pixels (\%)},
      symbolic x coords={Clear, Rain, Snow, Fog, Day, Night, Total},
      xticklabel style={rotate=45,anchor=north east,text depth=1.0em},
      xtick=data,
      ymajorgrids,
      reverse legend,
      legend cell align={left},
      legend columns=-1
      nodes near coords=\empty, 
      ymin=0, % Added ymin
      ymax=100, % Added ymax
      ]
      \addplot+[ybar,fill=color1!20,draw=black] plot coordinates {(Clear,76.09) (Rain,68.60) (Snow,72.14) (Fog,50.79) (Day,80.58 ) (Night,47.93) (Total,67.52)};
      \addplot+[ybar,fill=color2!40,draw=black] plot coordinates {(Clear,5.53) (Rain,10.34) (Snow,10.34) (Fog,19.38) (Day,6.37) (Night,17.96) (Total,11.00)};
      \addplot+[ybar,fill=white,draw=black, postaction={pattern=north east lines, pattern color=black}] plot coordinates {(Clear,18.38) (Rain,21.06) (Snow,17.52) (Fog,29.83) (Day,12.74) (Night,34.11) (Total,21)};
      \legend{H1, H2, Unlabeled}
    \end{axis}
  \end{tikzpicture}
  \end{subfigure}%
  \begin{subfigure}{0.49\textwidth}
      \begin{tikzpicture}[baseline]
    \begin{axis}[
    ybar,
    width=\linewidth,
    height=4cm,
    bar width=3pt,    
    axis lines=left, % Remove top axis line
    x axis line style={-}, % Remove arrow at the end of the bottom axis
    symbolic x coords={n/a, 0-20m,20-40m,\textgreater40m},
    enlarge x limits=0.25,
    xticklabel style={rotate=45,anchor=north east},
    xtick=data,
    xtick align=outside,
    ylabel=Instances (\%),
    ymin=0,
    ymax=45,
    ymajorgrids,
    ytick={0, 10, 20, 30, 40},
    legend style={at={(0.5,1.0)},anchor=south,legend columns=-1, draw=none}, 
    legend cell align={left},
    cycle list={        {fill=color1!20},
    {fill=color2!40},
        {fill=color3!60},
        {fill=color4!80},
    },
]
\addplot coordinates {(n/a, 13.33) (0-20m, 16.01) (20-40m, 27.82) (\textgreater40m, 42.84) };
\addplot coordinates {(n/a, 9.80) (0-20m, 23.3) (20-40m, 31.68) (\textgreater40m, 35.21) };
\addplot coordinates {(n/a, 11.36) (0-20m, 22.12) (20-40m, 30.15) (\textgreater40m, 36.37) };
\addplot coordinates {(n/a, 22.11) (0-20m, 23.91) (20-40m, 24.92) (\textgreater40m, 29.06) };
\legend{Clear, Rain, Snow, Fog}
\end{axis}
\end{tikzpicture}
  \end{subfigure}
  \caption{\textbf{(left)} Percentages of pixels labeled in the two annotation stages for different conditions. \textbf{(right)} Distance distribution of instances across different weather conditions. H1: stage 1, H2: stage 2, ``n/a'': instances without any lidar returns.}
\label{fig:lidar:stats:plot}
  
\end{figure}
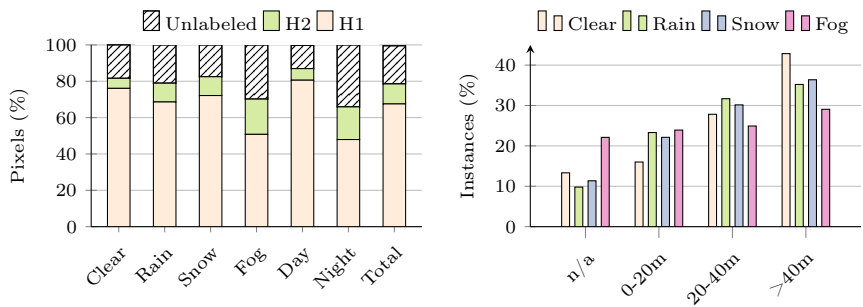

\subsection{Analysis of Annotations}
\label{subsec:statistics}

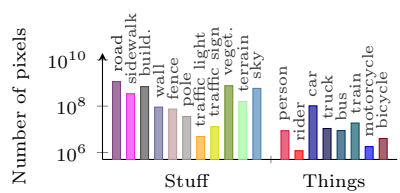
\begin{wrapfigure}{R}{5.5cm}
    \centering
    \begin{tikzpicture}
    \tikzstyle{every node}=[font=\fontsize{7}{8}\selectfont]
    \begin{axis}[
      ybar,
      ymode=log,
      width=\linewidth,
      height=3cm,
      bar width=3pt,
      xmin=0,
      xmax=21,
      ymin=5e5,
      ymax=2e10,
      % ymajorgrids=true,
      ylabel={Number of pixels},
      ytick={1e6,1e8,1e10},
      yticklabels={$10^6$,$10^8$,$10^{10}$,},
      xtick={6,16.5},
      minor xtick={12},
      xticklabels = {
        Stuff,
        Things,
      },
      major x tick style = {opacity=0},
      minor x tick num = 1,
      xtick pos=left,
      every node near coord/.append style={
      anchor=west,
      rotate=90,
      font=\fontsize{6}{7}\selectfont,
      }, 
      axis lines=left, % Remove top axis line
      x axis line style={-}, % Remove arrow at the end of the bottom axis
    ]
    
    % flat
    \addplot[bar shift=0pt,draw=road,          fill opacity=0.9,fill=road!80!white           , nodes near coords=road                 ] plot coordinates{ ( 1,     1149296885  ) };
    \addplot[bar shift=0pt,draw=sidewalk,      fill opacity=0.8,fill=sidewalk!80!white       , nodes near coords=sidewalk             ] plot coordinates{ ( 2,     344549542   ) };

    % construction
    \addplot[bar shift=0pt,draw=building,      fill opacity=0.8,fill=building!80!white       , nodes near coords=build.               ] plot coordinates{ ( 3,     695574861  ) };
    \addplot[bar shift=0pt,draw=wall,          fill opacity=0.8,fill=wall!80!white           , nodes near coords=wall                 ] plot coordinates{ ( 4,     90904470  ) };
    \addplot[bar shift=0pt,draw=fence,         fill opacity=0.8,fill=fence!80!white          , nodes near coords=fence                ] plot coordinates{ ( 5,     74789021   ) };
    
    % object
    \addplot[bar shift=0pt,draw=pole,          fill opacity=0.8,fill=pole!80!white           , nodes near coords=pole                 ] plot coordinates{ ( 6,    35554842   ) };
    \addplot[bar shift=0pt,draw=traffic light, fill opacity=0.8,fill=traffic light!80!white  , nodes near coords=traffic light        ] plot coordinates{ ( 7,    4898555    ) };
    \addplot[bar shift=0pt,draw=traffic sign,  fill opacity=0.8,fill=traffic sign!80!white   , nodes near coords=traffic sign         ] plot coordinates{ ( 8,    13066720   ) };

    % nature
    \addplot[bar shift=0pt,draw=vegetation,    fill opacity=0.8,fill=vegetation!80!white     , nodes near coords=veget.               ] plot coordinates{ ( 9,    762655939   ) };
    \addplot[bar shift=0pt,draw=terrain,       fill opacity=0.8,fill=terrain!80!white        , nodes near coords=terrain              ] plot coordinates{ ( 10,    159060217    ) };
    
    % sky
    \addplot[bar shift=0pt,draw=sky,           fill opacity=0.8,fill=sky!80!white            , nodes near coords=sky                  ] plot coordinates{ ( 11,    583031195 ) };

    % human
    \addplot[bar shift=0pt,draw=person,        fill opacity=0.8,fill=person!80!white         , nodes near coords=person        ] plot coordinates{ ( 13,    8709839     ) };
    \addplot[bar shift=0pt,draw=rider,         fill opacity=0.8,fill=rider!80!white          , nodes near coords=rider         ] plot coordinates{ ( 14,    1192011        ) };
    
    % vehicle
    \addplot[bar shift=0pt,draw=car,           fill opacity=0.8,fill=car!80!white            , nodes near coords=car           ] plot coordinates{ ( 15,    103145145   ) };
    \addplot[bar shift=0pt,draw=truck,         fill opacity=0.8,fill=truck!80!white          , nodes near coords=truck         ] plot coordinates{ ( 16,    10828634     ) };
    \addplot[bar shift=0pt,draw=bus,           fill opacity=0.8,fill=bus!80!white            , nodes near coords=bus           ] plot coordinates{ ( 17,    8848115     ) };
    \addplot[bar shift=0pt,draw=train,         fill opacity=0.8,fill=train!80!white          , nodes near coords=train         ] plot coordinates{ ( 18,    18672428        ) };
    \addplot[bar shift=0pt,draw=motorcycle,    fill opacity=0.8,fill=motorcycle!80!white     , nodes near coords=motorcycle    ] plot coordinates{ ( 19,    1800130        ) };
    \addplot[bar shift=0pt,draw=bicycle,       fill opacity=0.8,fill=bicycle!80!white        , nodes near coords=bicycle       ] plot coordinates{ ( 20,    4019864      ) };

    \end{axis}
    \end{tikzpicture}
    \caption{Number of annotated pixels per class in \Ours{}.}
    \label{fig:dataset:stats}
\end{wrapfigure}

\cref{fig:dataset:stats} summarizes the number of annotated pixels per class in \Ours{}.
Overall, 78.5\% of all pixels receive a valid label after stage 2 of the annotation (in H2), an increase of 10.3\% compared to H1.
Notably, H2 contains roughly 1.5\texttimes\ more instances than H1.
The increased level of detail and completeness in H2 supports the design choice of complementing the annotation with multi-sensor information.
When comparing the annotation density in H2 for images of different visual conditions in \cref{fig:lidar:stats:plot} (left), we observe that fog and nighttime constitute the most challenging conditions to annotate, showing the greatest increases in annotation density when incorporating auxiliary data in stage 2 of the annotation.
To measure the added \emph{difficulty} of stage 2 annotations, we train Mask2Former~\cite{cheng2022masked} with H2 and evaluate it on stage 1 (H1) and stage 2 (H2) labels. 
We observe a 
11.1\% drop in PQ, from 58.0\% (H1) to 46.9\% (H2), 
which indicates the substantial increase in the difficulty of the labels that are created thanks to our multimodal annotation. This highlights the challenging nature of ground truth in \Ours{} compared to the ground truth in 2D semantic perception datasets that lack auxiliary data\textemdash{}such as our additional modalities\textemdash{}for adverse-condition annotations. We provide a quantitative verification of this argument via a cross-dataset comparison in \cref{subsec:additional_segmentation}.
As a result of our two-stage annotation protocol, 24.5\% of pixels are labeled as \texttt{difficult\_class}, and 1.5\% of \texttt{things} pixels are labeled as \texttt{difficult\_instance}.

We investigate the range distribution of annotated \texttt{things} instances in \Ours{} by estimating their distances via the motion-compensated and outlier-filtered lidar point cloud.
\Cref{fig:lidar:stats:plot} (right) shows that annotated instances in adverse conditions are closer on average.
However, those results need to be interpreted cautiously, as the lidar can be heavily affected by adverse conditions, \eg, in fog 43.4\% of instances have less than 10 lidar returns and 22.3\% have none.
This underscores the advantage of 2D image labeling over point cloud labeling for adverse scenes.
More statistics for \Ours{} are provided in~\cref{sec:suppl:further:stats}.

\subsection{Benefit of Additional Modalities for Dense Semantic Perception}
\label{subsec:impact_of_sensors}

\begin{table}[tb]
  \caption{PQ of Mask2Former~\cite{cheng2022masked} models trained on different sets of input modalities. A Swin-T~\cite{liu2021swin} backbone is used in all cases.}  
  \label{table:weather_ablation}
  \centering
  \setlength\tabcolsep{3pt}
  \tablefontsize
  \begin{tabular*}{\linewidth}{@{\extracolsep{\fill}}ccccccccccc@{}}
  \toprule
  \multirow{2}{*}{\begin{tabular}{@{}c@{}}Frame\\camera\end{tabular}} & \multirow{2}{*}{\begin{tabular}{@{}c@{}}Event\\camera\end{tabular}} & \multirow{2}{*}{Radar} & \multirow{2}{*}{Lidar} & \multirow{2}{*}{Clear} & \multirow{2}{*}{Fog} & \multirow{2}{*}{Rain} & \multirow{2}{*}{Snow} & \multirow{2}{*}{Day} & \multirow{2}{*}{Night} & \multirow{2}{*}{All}\\\\
   \midrule
   % RGB
  \checkmark & & &   
  &48.8
  &46.5
  &45.4
  &42.2
  &49.4&39.4&46.9
  \\
  % Event
  \checkmark & \checkmark & & 
  &52.1
  &49.4
  &48.2
  &42.6
  &51.7&42.2&49.5
  \\
  % Radar
  \checkmark & & \checkmark & 
  &52.9
  &49.5
  &49.9
  &46.1
  &52.9&44.8&51.3
  \\
  % Lidar
  \checkmark & & & \checkmark
  &54.2
  &49.9
  &52.2
  &47.6
  &53.7&48.0&52.7
  \\
  %  All
  \checkmark & \checkmark & \checkmark &\checkmark 
  &\textbf{55.3}
  &\textbf{50.3}
  &\textbf{53.8}
  &\textbf{47.9}
  &\textbf{54.1}&\textbf{49.7}&\textbf{53.6}
  \\
  \bottomrule
  \end{tabular*}
\end{table}

In \cref{table:weather_ablation}, 
we show the benefit of the additional non-camera modalities which
\Ours{} offers compared to previous dense semantic driving perception benchmarks~\cite{Cityscapes,yu2020bdd100k,ACDC}, by comparing multimodal models to a camera-only model.
In particular, we train different Mask2Former-based networks~\cite{cheng2022masked} on \Ours{} for panoptic segmentation, each using different modalities as inputs.
For this, we construct simple multimodal networks by employing separate Swin-T~\cite{liu2021swin} backbones for each modality and fusing their outputs with a parallel cross-attention block~\cite{broedermann2023hrfuser}. 
For more implementation details, please refer to~\cref{sec:train:det}.

The key observation on the results of this comparison is
that the quadrimodal network outperforms the baseline camera-only network across all visual conditions.
This finding proves the value of our novel multimodal dense semantic perception dataset for developing better models for such pixel-level tasks.
In addition, we observe that all three examined bimodal networks, which fuse the camera input with one of the additional modalities respectively, also deliver consistent improvements over the camera-only network across conditions and trail the performance of the quadrimodal network. 
Each non-camera modality provides valuable complementary information for semantic perception on top of the camera. However, there is significant potential to improve upon our simple fusion method. Future multimodal dense semantic perception works based on MUSES could develop more specialized fusion architectures and enhance model performance across all domains.

\subsection{State of the Art in Semantic Image Segmentation on MUSES}
\label{subsec:additional_segmentation}

\begin{wraptable}{R}{6cm}
\caption{State of the art in semantic segmentation on \Ours{}.}
\label{table:sota:sem:seg}
\centering
\tablefontsize
\setlength\tabcolsep{3pt}
\begin{tabular}{lc}
\toprule
Architecture & mIoU\\
\midrule
DeepLabv3+ (ResNet101-D8)~\cite{chen2018encoder} & 70.5\\
OCRNet (HRNetV2p-W48)~\cite{yuan2020object} &  71.9\\ 
SETR (ViT-L)~\cite{zheng2021rethinking} &  71.1\\ 
SegFormer (MiT-B2)~\cite{xie2021segformer} & 72.5\\
SegFormer (MiT-B5)~\cite{xie2021segformer} & 74.7\\
Mask2Former (Swin-T)~\cite{cheng2022masked} &  70.7\\
Mask2Former (Swin-L)~\cite{cheng2022masked} & \textbf{77.1}\\
\bottomrule
\end{tabular}
\end{wraptable}

We evaluate the performance of state-of-the-art semantic segmentation architectures on \Ours{} in~\cref{table:sota:sem:seg}.
All backbones are pre-trained on ImageNet~\cite{deng2009imagenet}.
High-capacity models tend to perform better: Mask2Former~\cite{cheng2022masked} achieves the highest mIoU. %, 

\begin{table}[b]
  \caption{Generalization of semantic segmentation models trained on different datasets. Mask2Former~\cite{cheng2022masked} with a Swin-L~\cite{liu2021swin} backbone and camera-only input is used. We train and test on different combinations of datasets. $\mathrm{\Delta}$: mIoU difference to in-domain model.}
  \label{table:distribution_shift_v2}
  \centering
  \setlength\tabcolsep{5pt}
  \tablefontsize
  \resizebox{\linewidth}{!}{%
  \begin{tabular}{lccccccccccr} %{\linewidth}{@{\extracolsep{\fill}} lcccccr@{}}
  \toprule
  \multirow{2}{*}{Training Dataset} && \multicolumn{2}{c}{Cityscapes-val~\cite{Cityscapes}} && \multicolumn{2}{c}{ACDC-test~\cite{ACDC}} && \multicolumn{2}{c}{MUSES-test} && Mean \\
  \cmidrule{3-4} \cmidrule{6-7} \cmidrule{9-10} \cmidrule{12-12}
  && mIoU & $\mathrm{\Delta}$ && mIoU & $\mathrm{\Delta}$ && mIoU & $\mathrm{\Delta}$ && mIoU \\
  \midrule
  Cityscapes~\cite{Cityscapes} && 83.7 & --- && 65.7  & -10.6 && 58.9 & -18.2 && 69.4 \\
  ACDC~\cite{ACDC} && 70.9 & -12.8 && 76.3 & --- && 66.9 & \textbf{-10.2} && 71.4  \\
  \Ours{} && 73.1 & \textbf{-10.6} && 72.0 & \textbf{-4.3} && 77.1 & --- && \textbf{74.1} \\
  \bottomrule
\end{tabular}}
\end{table}

Because of its diverse range of conditions and high-quality ground truth, \Ours{} constitutes an ideal testbed for model adaptation and generalization experiments.
To compare different datasets along this axis, we train Mask2Former on three datasets\textemdash Cityscapes~\cite{Cityscapes}, ACDC~\cite{ACDC}, and \Ours{}\textemdash and cross-evaluate the resulting models.
Note that, of the three datasets, \Ours{} has the smallest training set.
The results in \cref{table:distribution_shift_v2} show that, despite this handicap, the model trained on \Ours{} achieves the highest average performance over all datasets, demonstrating its robustness.
We hypothesize that this can be attributed to the superior diversity and annotation quality of MUSES samples.
Nevertheless, the performance of all models drops significantly under domain shift, motivating future research efforts in this field.

\subsection{Uncertainty-Aware Panoptic Segmentation}
\label{subsec:upq_experiments}

\begin{table}
  \caption{Uncertainty-aware panoptic segmentation baselines and oracles. Mask2Former~\cite{cheng2022masked} with Swin-T~\cite{liu2021swin} is used.}
  \label{table:upq_evaluation}
  \centering
  \setlength\tabcolsep{5pt}
  \tablefontsize
  \begin{tabular}{lccccr} 
  \toprule
  \multirow{2}{*}{Confidence} &  \multirow{2}{*}{AUSQ\,$\uparrow$} &  \multirow{2}{*}{AURQ\,$\uparrow$} & \multicolumn{3}{c}{AUPQ\,$\uparrow$}  \\
  \cmidrule{4-6}   &&& Stuff & Things & All \\ \midrule
  Constant 100\%  & 78.3& 58.0& 58.2  & 31.3 & 46.9\\
  Marginalization & 80.2 & 53.6 & 54.3 & 30.7 & 44.3  \\
  Oracle & 89.7    & 83.7   & 78.0& 71.2   & 75.2\\
  \bottomrule
\end{tabular}
\end{table}

For the novel uncertainty-aware panoptic segmentation task introduced in \cref{sec:upq}, we present in \cref{table:upq_evaluation} two mask-classification baselines and an oracle approach for the unimodal setting.
The first baseline simply predicts 100\% class and instance confidence for all pixels.
In this case, the AUPQ is equivalent to the standard PQ.
The second baseline aims to extract class and instance confidence from the trained mask-classification model: we obtain the class confidence by marginalizing over probability-mask pairs, whereas the instance confidence is obtained by dividing the mask class score through the class confidence.
A more detailed description and motivation for this procedure is given in~\cref{subsec:sup:marg}.
The oracle approach is a confidence oracle: it uses ground-truth class and instance confidence maps for confidence prediction.
Its performance sets an upper bound for the examined models.
As shown in \cref{table:upq_evaluation}, the first baseline outperforms the marginalization approach, meaning that the confidence scores of those models are not well calibrated to the difficulty labels.
Furthermore, there is a substantial performance gap between the two baselines and the oracle, indicating that future custom-trained models have great room for improvement in this novel task.

%% file: sec/6_conclusion.tex
\section{Conclusion}
\label{sec:conclusion}

We have presented \Ours{}, the first multi-sensor semantic perception dataset for autonomous driving in adverse conditions.
\Ours{} provides 2500 samples, each consisting of synchronized and calibrated RGB images, MEMS lidar scans, HD event sequences, FMCW radar scans, IMU/GNSS readings, and an image-level corresponding normal-condition image.
Our new two-stage panoptic annotation protocol yields high-quality 2D panoptic annotations for each image and class- and instance-level difficulty maps, which enable the novel task of uncertainty-aware panoptic segmentation.
Our experiments highlight the benefits of the additional non-camera modalities for dense semantic perception and prove that \Ours{} is effective for training and provides a challenging evaluation for models under diverse visual conditions.
acrossThis dataset motivates many possible lines of research: sensor fusion for dense semantic perception tasks, the exploration of strengths and weaknesses of individual sensors, \eg, event cameras, for semantic perception in diverse visual conditions, and model generalization and/or adaptation across conditions or modalities.

%% file: sec/X_suppl.tex
\clearpage

\appendix

\section{Annotation Time}

The total annotation time over the 2500 images of the \Ours{} dataset was 11\,827 hours, which translates to more than 15 months of 24-hour/day labeling for one person. During annotation, we observed a large diversity in annotation difficulty, leading to varying annotation times from 3 hours to 8 hours and 40 minutes per image. We provide a detailed breakdown of the annotation times in \cref{tab:sup:time-breakdown-hhmm}, where we list the average annotation times for each of our two stages. We thereby separate the initial drawing of the annotation and the subsequent quality control step, where a different annotator refines the annotations by trying to find mistakes in the initially drawn annotations. Quality control contributes significantly to the overall annotation time, emphasizing the significant effort needed to ensure high-quality annotations.

\begin{table}[h]
    \centering
    \caption{\textbf{Annotation time breakdown by annotation difficulty} in HH:MM format. ``Draw'': Annotation drawing. ``QC'': Quality control.}
    \label{tab:sup:time-breakdown-hhmm}    
    \footnotesize
    \begin{tabular*}{\columnwidth}{@{\extracolsep{\fill}}lcccc@{}}
        \toprule
        Difficulty & Easy & Medium & Hard & Very hard \\        
        \midrule
        \# Scenes & 728	&1345	&265	&9 \\
        \midrule
        Draw stage 1 & 00:45 & 01:45 & 02:30 & 03:00 \\
        QC stage 1 & 00:30 & 01:00 & 01:20 & 01:40 \\
        Draw stage 2 & 00:45 & 01:15 & 01:30 & 01:50 \\
        QC stage 2 & 01:00 & 01:40 & 02:00 & 02:10 \\
        \midrule
        Total average time & 03:00 & 05:40 & 07:20 & 08:40 \\
        \bottomrule
    \end{tabular*}
\end{table}

\section{Recording Platform}
\begin{figure}
    \centering
    \includegraphics[width=0.8\linewidth, trim=0 50 0 30, clip]{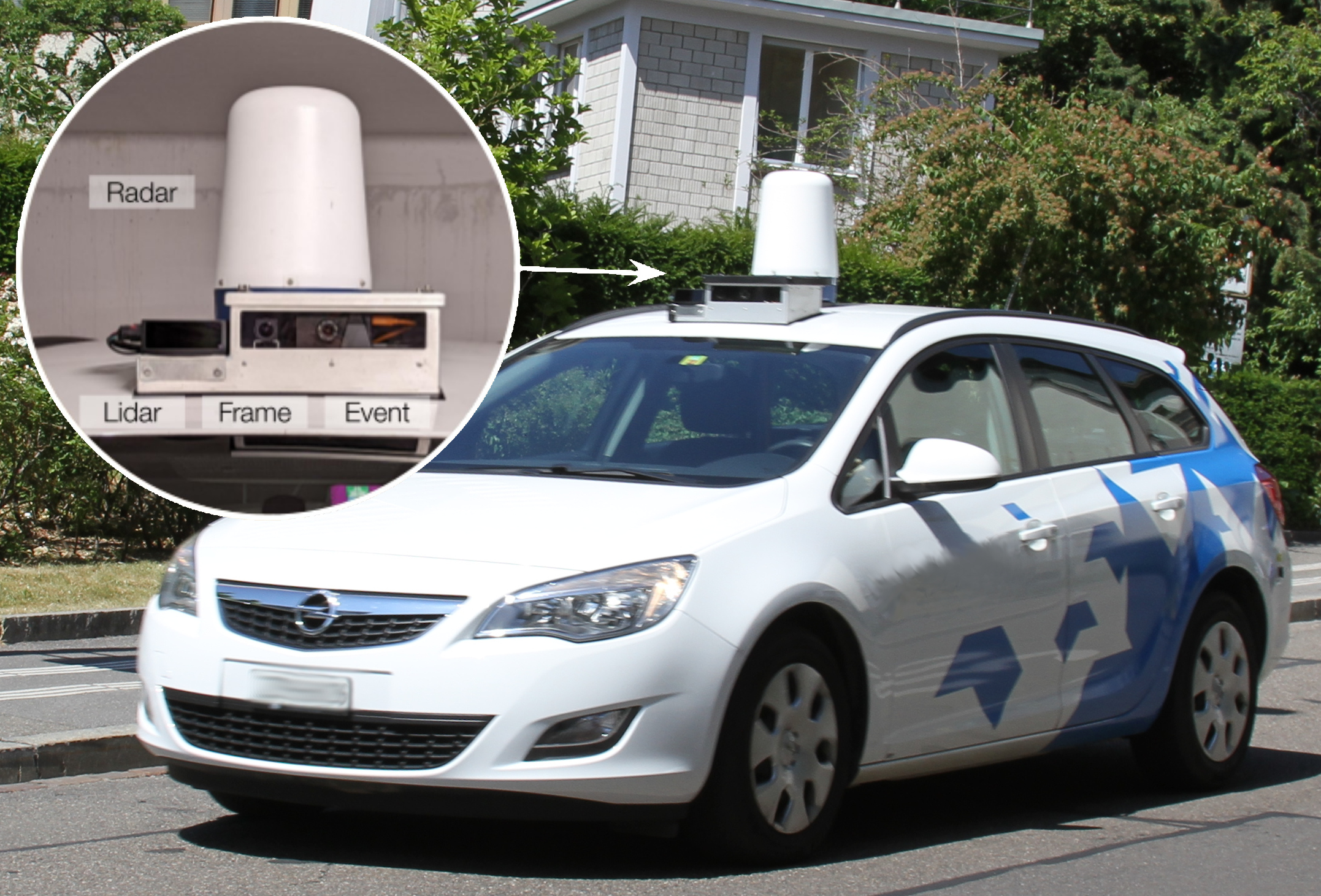}
    \caption{\textbf{Recording car with waterproof sensor rig.}}
    \label{fig:car_pic}
\end{figure}

Fig.~\ref{fig:car_pic} shows a picture of the recording car with the mounted waterproof sensor rig.

\section{Calibration}
\label{subsec:calibration}

To fuse multi-sensor information accurately, we need to calibrate the sensors both geometrically and temporally.

\PAR{Geometric calibration.} 
The intrinsic parameters for the lidar and radar are vendor-calibrated.
For the event camera, we reconstruct frames following~\cite{muglikar2021calibrate} and use the same intrinsic calibration procedure as for the frame camera, using a metrology-grade checkerboard and OpenCV~\cite{opencv_library}.
For the extrinsic calibration, we create a consistent transform graph between frame camera, event camera, and lidar:
First, we perform pairwise calibrations: Stereo-calibration between the frame and event camera according to~\cite{muglikar2021calibrate}, mutual information maximization between lidar and frame camera according to~\cite{pandey2015automatic}, and mutual information maximization between lidar and event camera according to~\cite{kevinta2023:l2e}.
Next, we formulate a triangular pose graph and optimize it for loop closure (Powell's dog leg method) to get a consistent set of extrinsics.Author
For radar-lidar calibration we follow~\cite{burnett2022boreas}: the rotation is estimated via correlative scan matching using a Fourier Mellin transform~\cite{checchin2010radar} and the translation is simply measured.
The IMU/GNSS is calibrated with u-center~\cite{u:center} and subsequent point cloud consistency optimization.

\PAR{Synchronization.}
As all sensors record at different frequencies (see 
% Tab. 2),
\cref{tab:sensors:perception}), 
we synchronize their internal clocks and record asynchronously. 
In post-processing, we match the camera frame with the lidar and radar using their mid-exposure timestamps and choose samples that minimize this delta. For the event camera, we consider the 3 seconds up until the frame camera's exposure end time, and the GNSS is interpolated to match the frame camera's time. For all sensors, timestamps in $\mu$s are provided.

Our data-recording computer is synced to a web-based GPS-time server via the network time protocol (NTP) and functions as a master clock. The radar is synced via the NTP. The lidar and frame camera are synced via the precision time protocol (PTP) with software timestamping. The event camera receives synthetic events from the frame camera at exposure start, used for temporal alignment in post-processing. The GNSS clock naturally syncs with GPS satellite atomic clocks.
We will release an SDK for lidar and radar ego-motion compensation and projection of all modalities to the frame camera.

\section{Anonymization}
\label{sec:anonymization}

To protect the privacy of all individuals in our dataset, we use a semi-automatic anonymization pipeline to blur faces and license plates.
Specifically, we manually draw bounding boxes over all recognizable faces and license plates in the images and use an off-the-shelf object tracker and segmenter~\cite{wang2019fast} to refine the blur mask.
Finally, we applied Gaussian blurring to all masks and checked the images individually.

\section{Uncertainty Type}

Assuming human annotation as the gold standard—given our intensive quality control—any human uncertainty is attributed to aleatoric uncertainty (data-inherent and irreducible). “Difficult” pixels are those explained only by additional data available at annotation stage 2 (e.g., videos, lidar). Hence, a model that only has access to stage-1 (camera) data should not be confident in predicting the labels of “difficult” pixels, as their semantics cannot be explained by this data alone. UPQ allows the model to acknowledge uncertainty alongside making a correct prediction, thus encouraging an uncertainty-aware prediction. This rationale only holds for UPQ evaluation of camera-based models, and not multimodal ones.

\section{Training Details}
\label{sec:train:det}

\subsection{Experiment Implementation Details}
\label{sup:sub:sec:exp:imp:det}

For our panoptic segmentation experiments, we train on 2 NVIDIA A100 GPUs with a batch size of 8 with all input
channels normalized over the entire dataset. We train a Mask2Former~\cite{cheng2022masked} with an ImageNet-1K pre-trained Swin-T~\cite{liu2021swin} backbone and following the mmdetection~\cite{mmdetection}
configuration for the frame camera networks. 

\begin{figure*}
    \centering
    \begin{tabular}{@{}c@{\hspace{0.05cm}}c@{\hspace{0.05cm}}c@{\hspace{0.05cm}}c@{}}    
        \subfloat{\footnotesize Frame Camera} &
        \subfloat{\footnotesize Lidar} &
        \subfloat{\footnotesize Event Camera} &
        \subfloat{\footnotesize Radar} \\
        \vspace{-0.07cm}
    \includegraphics[ width=0.248\textwidth]{figs/Annotation/REC0224_frame_325632/REC0224_frame_325632.jpg} &
    \includegraphics[ width=0.248\textwidth]{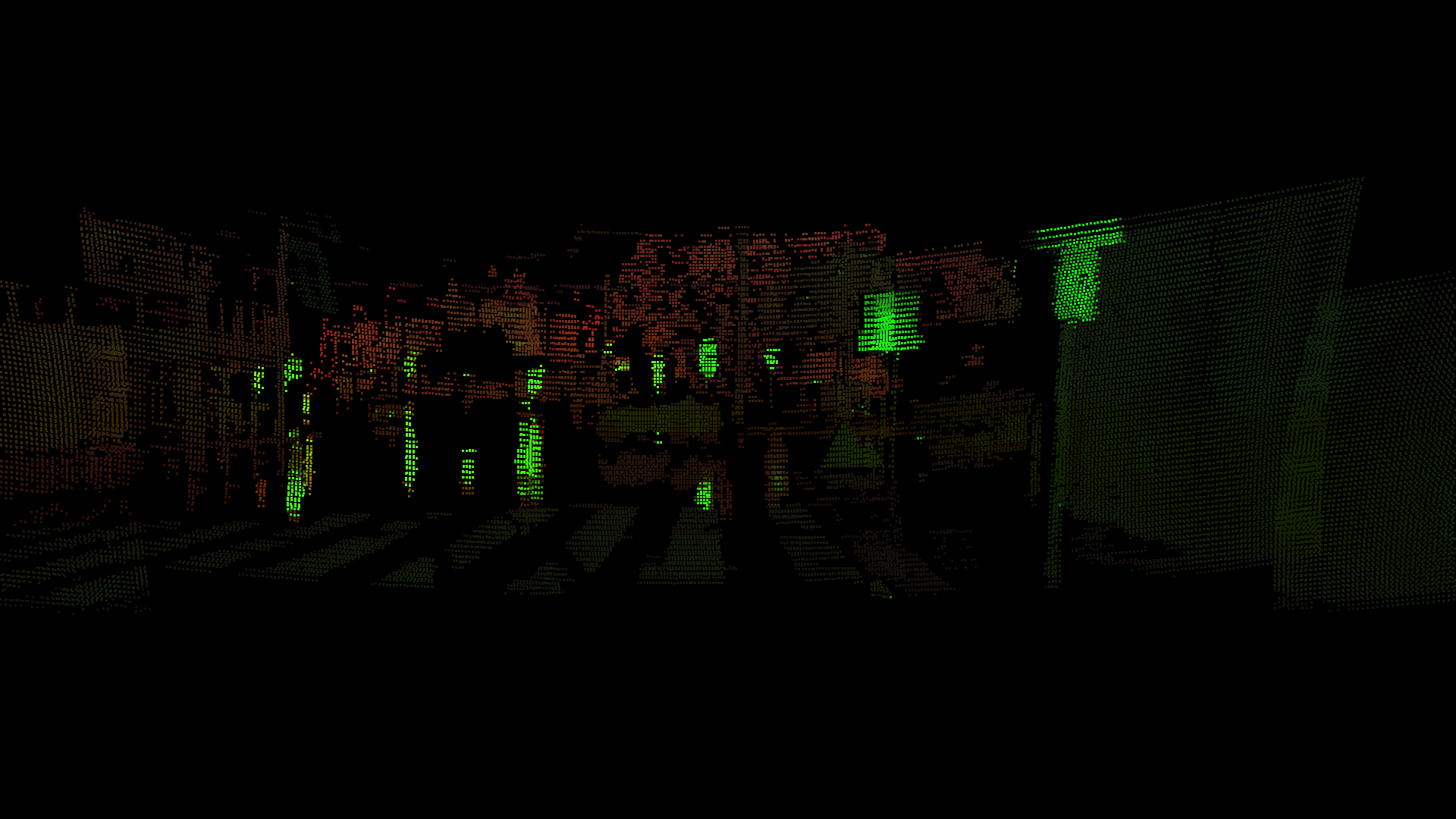} &
    \includegraphics[width=0.248\textwidth]{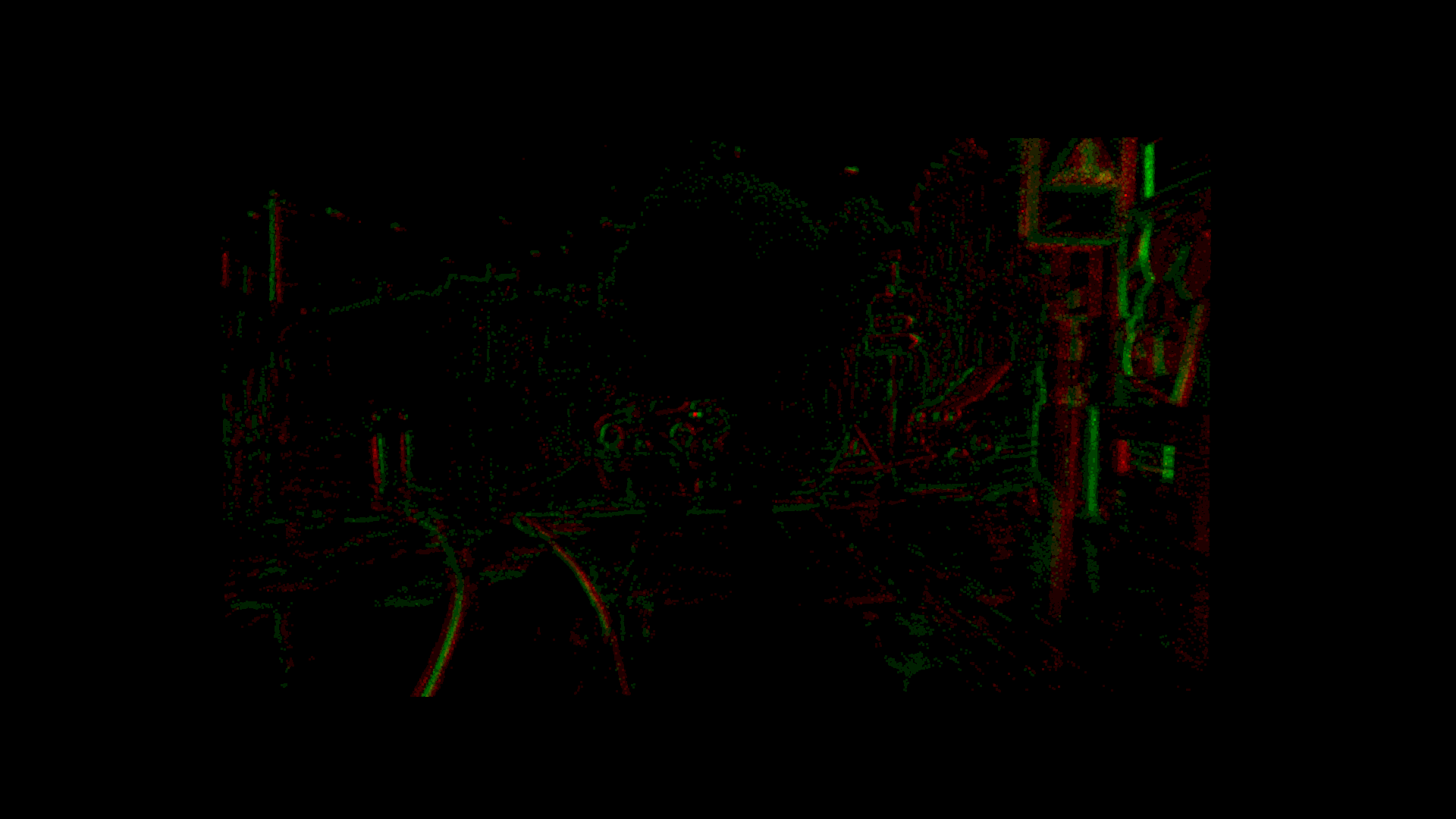} &
    \includegraphics[width=0.248\textwidth]{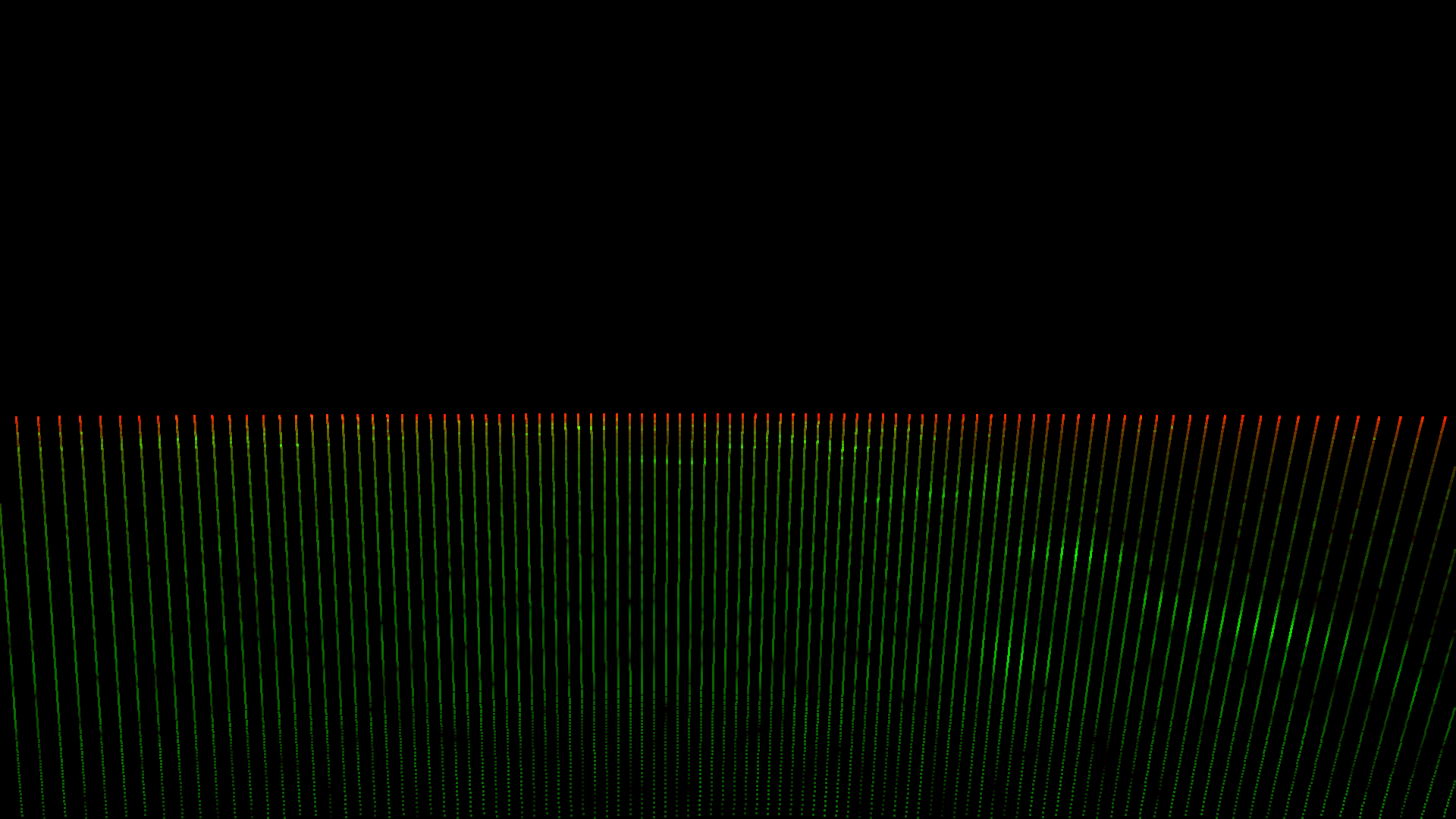} \\
    \vspace{-0.07cm}    
    \includegraphics[ width=0.248\textwidth]{figs/Annotation/REC0019_frame_2378703/REC0019_frame_2378703.jpg} &
    \includegraphics[width=0.248\textwidth]{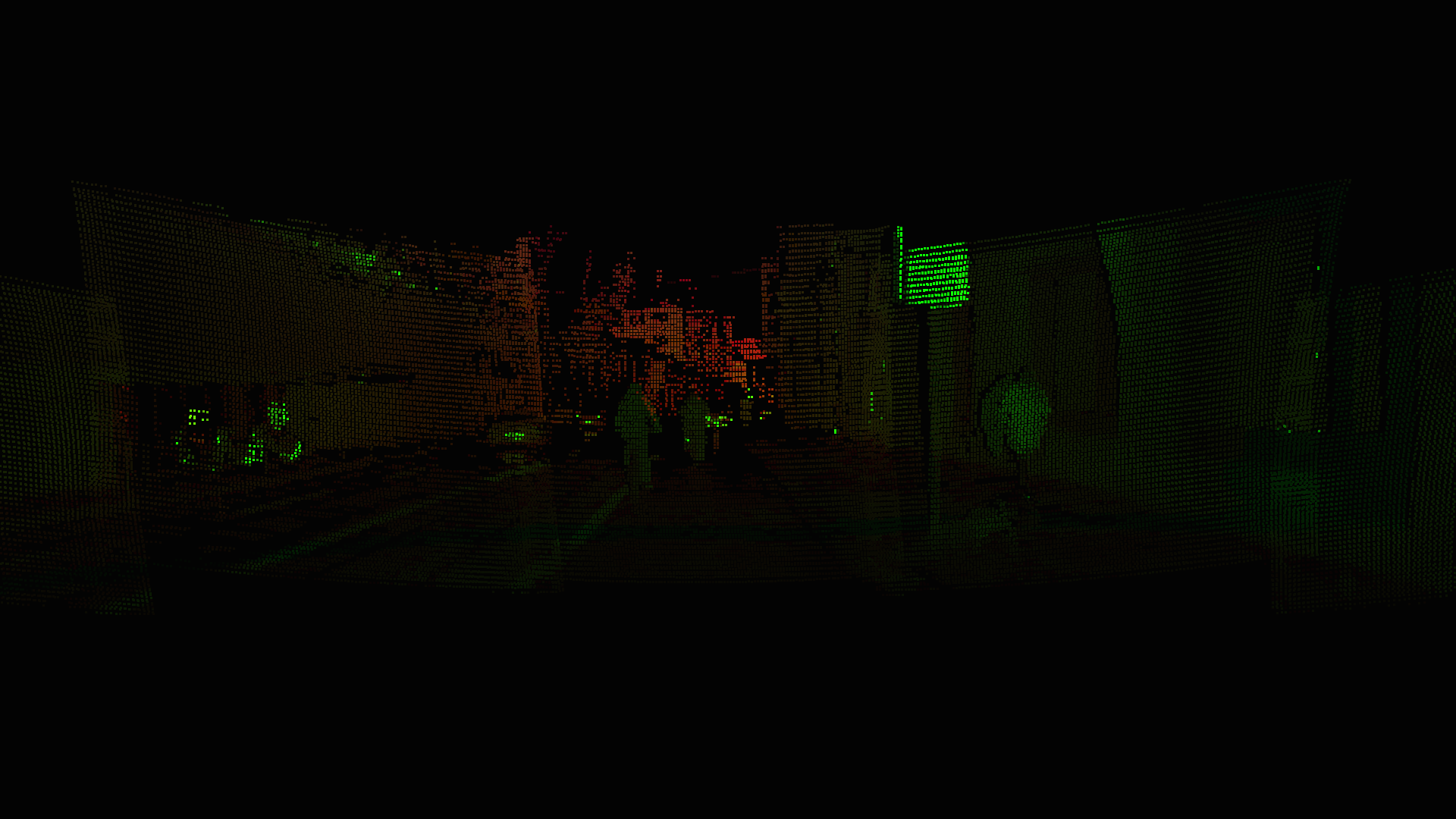} &
    \includegraphics[ width=0.248\textwidth]{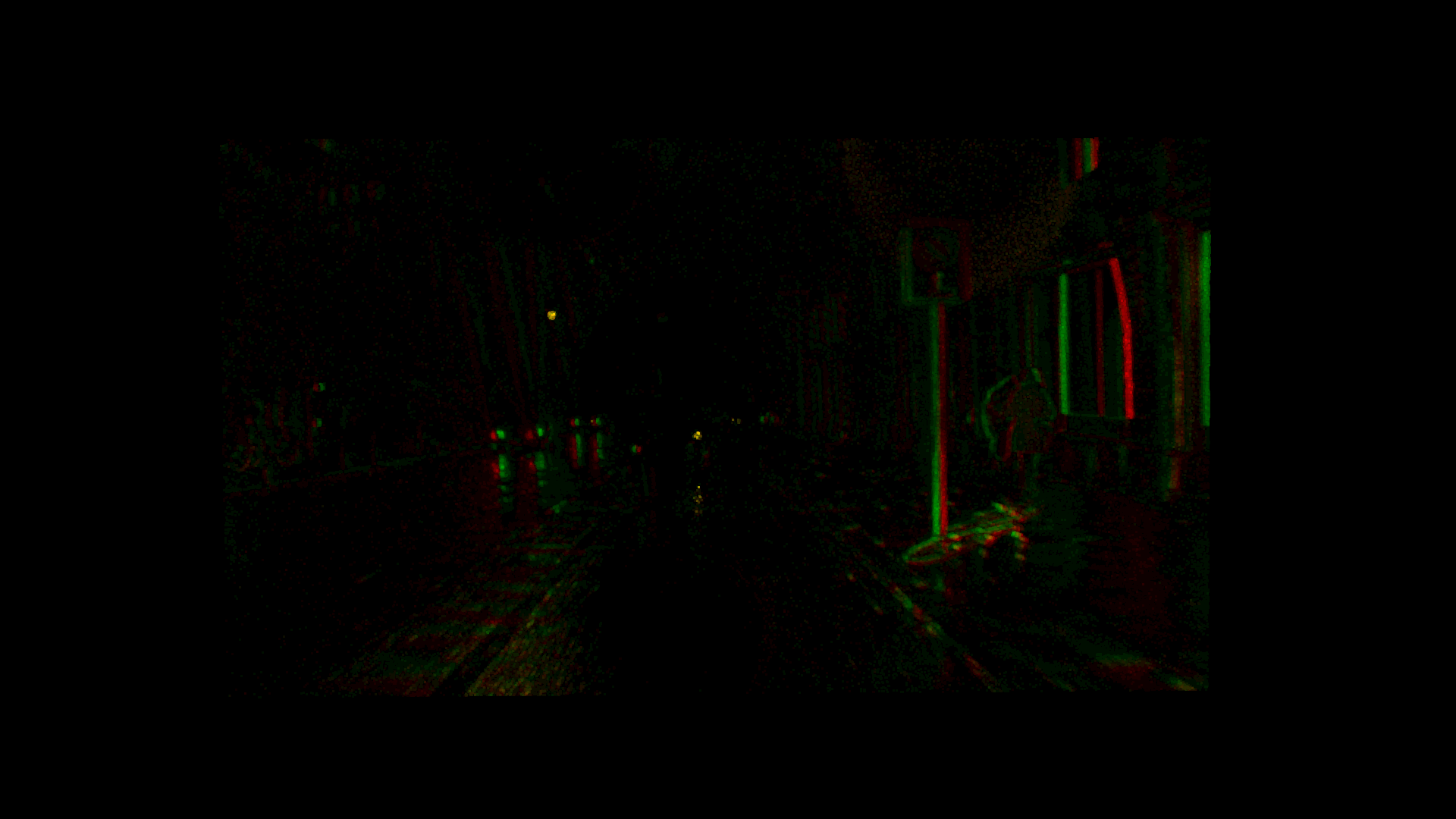} &
    \includegraphics[width=0.248\textwidth]{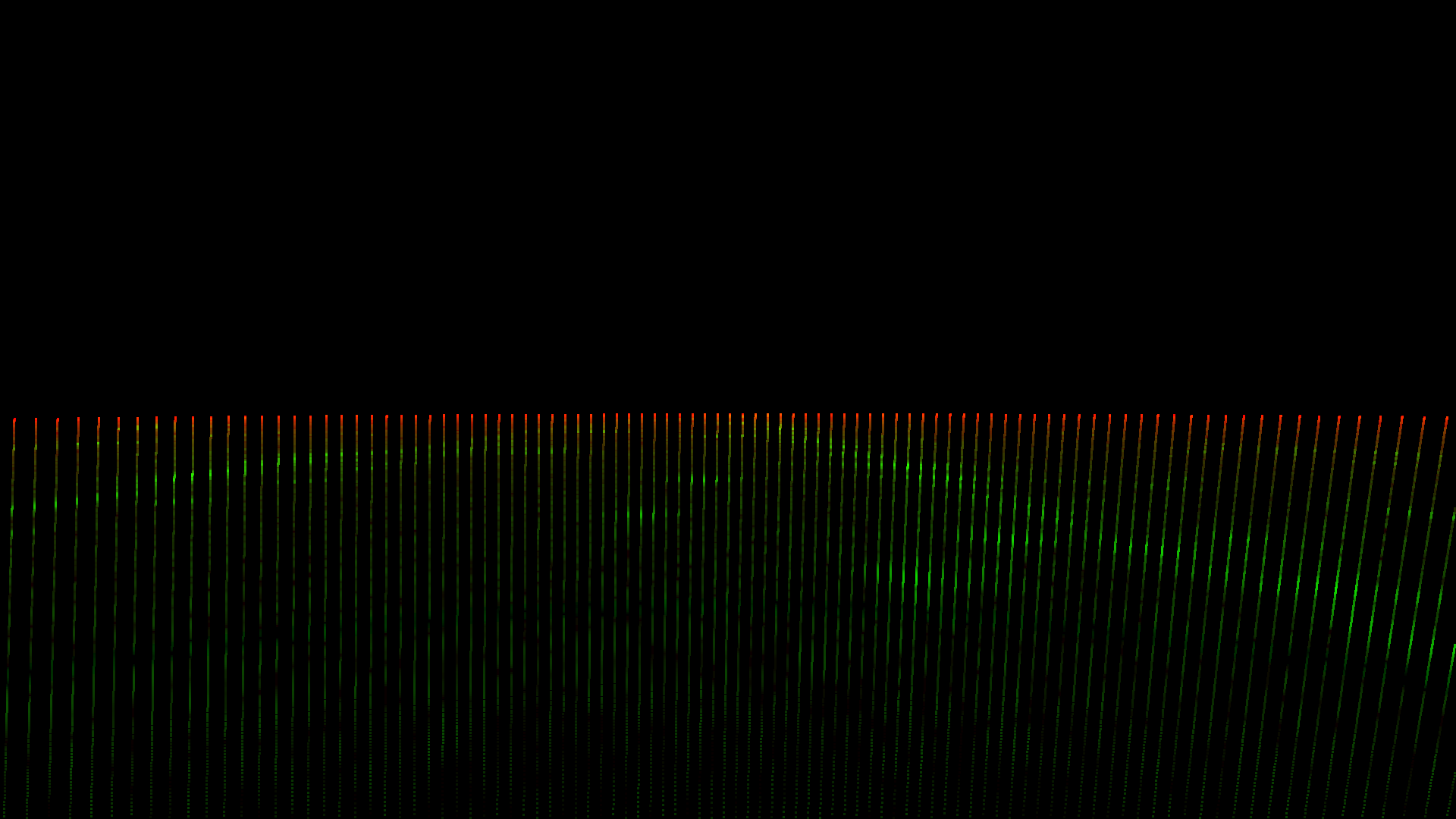} \\
    \end{tabular}        
  \caption{\textbf{Example inputs to the quadrimodal Mask2Former}. From left to right: Frame camera, projected and ego-motion corrected lidar points, projected event camera, projected and ego-motion corrected radar points. The projections are highlighted for better visualization. Best viewed on a screen at full zoom.}
  \label{fig:supl:example:inputs}
\end{figure*}

For the multimodal networks, we use a learning rate of 0.0002 and follow~\cite{broedermann2023hrfuser} in projecting all secondary modalities onto the 1920$\times$1080 image plane of the frame camera. 
The lidar points are ego-motion-compensated and projected onto the image plane with 3 channels: range, intensity, and height.
As the radar provides the full azimuth-range spectrum, we project every ego-motion-compensated intensity reading up to 150m range as its own individual point, assuming the ground level as height, into the camera plane with 2 channels: range and intensity.
For the event camera, we accumulate positive and negative events in individual channels over 30ms, resulting in a 2-channel image. 
To avoid overly sparse input images, we dilate the projected points. Exemplary inputs for the quadrimodal Mask2Former are visualized in \cref{fig:supl:example:inputs}.
To have a fair comparison with the frame-camera-only network, we likewise use ImageNet-1K pre-training for the Swin-T backbones. As the pre-trained Swin backbones expect a 3-channel input, we add empty channels where necessary. 
To preserve the pixel values, we apply nearest neighbor interpolation in the random resizing operation during training.
All inputs of one modality are randomly set to zero during training with a chance of 20\%, to discourage an overreliance on individual modalities~\cite{broedermann2023hrfuser}.

For the semantic segmentation experiments, we use the mmsegmentation~\cite{mmseg2020}
framework and train Mask2Former~\cite{cheng2022masked} on 8 A100 GPUs with a batch size of 16.
We use ImageNet-22K pre-trained weights for the Swin-L~\cite{liu2021swin} backbone and train the network for 70000 iterations, following the hyperparameters used in mmsegmentation.

\subsection{Multimodal Architecture Details}
\label{sup:subsec:mm:impl:detail}

For our multimodal experiments, we use a Mask2Former model with different input combinations, utilizing separate pre-trained Swin-T backbones for each modality. For the bimodal networks, we therefore have 2 parallel running backbones and for the quadrimodal network, we have 4 parallel backbones. Each backbone gets a 3-channel input image from a single modality. 
We fuse each of the 4 outputs (feature pyramid) of the backbones individually with a parallel cross-attention block~\cite{broedermann2023hrfuser} before passing the fused features to the pixel decoder of Mask2Former. This fusion block allows for parallel fusion of an arbitrary amount of different input modalities. Whereby one backbone has to be picked as the \emph{primary modality} where all other features are fused in parallel by performing standard cross attention between the \emph{primary modality} and each \emph{secondary modality} individually, including a skip connection. The frame camera features thereby serve as the \emph{primary modality} and all other modalities are treated as \emph{secondary modalities}.

\subsection{Mask-Classification Baseline for Uncertainty-Aware Panoptic Segmentation}
\label{subsec:sup:marg}

We construct a simple baseline for predicting pixel-level class and instance uncertainty scores with trained mask classification networks, such as Mask2Former~\cite{cheng2022masked}.
During inference, mask-classification approaches predict pairs $\{(p_i, m_i)\}_{i=1}^{N}$ for each of the $N$ masks (see~\cite{cheng2021per}).
$p_i \in \Delta^{K + 1}$ denotes the class probability distribution over $K + 1$ classes for mask $i$ (the $K$ object classes plus one no-object class $\varnothing$), $m_i \in [0,1]^{H\times W}$ the soft mask prediction over the image with dimensions $H \times W$.
Each pixel is assigned to a probability-mask pair $i^*$ according to 
\begin{equation}
    i^* = \mathrm{arg\,max}_{i: c_i \neq\varnothing} p_i(c_i) \cdot m_i[h, w]
    \label{eq:pan_inf}
\end{equation}
where $c_i$ is the most likely class for each probability-mask pair.
To obtain a class confidence score, we first normalize the mask predictions $m_i$ to sum up to one, overall $N$ masks.
We then marginalize overall probability-mask pairs to find a class score 
\begin{equation}
s_{class}[h, w] = \sum_{i=1}^N p_i(c_{i^*}) \cdot \bar{m}_i[h, w]
\end{equation}
where $\bar{m}_i$ denotes the normalized mask predictions and $c_{i^*}$ is the predicted class at that pixel.
For the instance confidence, we tease out the class influence as follows:
\begin{equation}
\begin{split}
    s_{inst}[h, w] &= \frac{p_{i^*}(c_{i^*}) \cdot \bar{m}_{i^*}[h,w]}{\sum_{i=1}^N p_i(c_{i^*}) \cdot \bar{m}_i[h, w]} \\
    &= \frac{p_{i^*}(c_{i^*}) \cdot \bar{m}_{i^*}[h,w]}{s_{class}[h,w]}
\end{split}
\end{equation}
The denominator corresponds exactly to the class confidence score, whereas the nominator corresponds to the assigned probability-mask pair score during panoptic inference in \cref{eq:pan_inf}.
This reveals an interesting interpretation: the probability-mask pair score used for panoptic inference can be decomposed into a product of class and instance confidence scores.

\section{Stage 1 vs.\ Stage 2: Detailed Results}
\label{suppl:sec:stage1:vs:2}

As mentioned in 
% Sec. 5.1
\cref{subsec:statistics} 
of the main paper, we try to quantify the added difficulty\textemdash{}and implied subsequent quality\textemdash{}of our second labeling stage onto annotations. We train frame-camera-only Mask2Former with a Swin-T backbone on our final panoptic ground truth (equivalent to H2 labels) and evaluate it on stage 1 (H1) and stage 2 (H2) labels. The results by condition presented in 
\cref{table:exp:H1:H2} 
show a 11.1\% drop from H2 to H1, indicating substantially more difficult ground truth after the second stage of the annotation. 

To further investigate the quality of the additional annotations, we train a semantic segmentation Mask2Former (Swin-L) exclusively on H1 annotations. The performance of this model is shown in \cref{table:exp:seg:H1:H2}. The model trained on H1 annotations shows a significant performance drop on ACDC and MUSES datasets. Specifically, there is a drop of 3.3 mIoU on MUSES and 2.0 mIoU on ACDC compared to the model trained on H2 annotations. On the other hand, the scores on Cityscapes are similar for both H1 and H2-trained models. This result is expected because the additional H2 labels primarily address indistinct areas in adverse scenes, which are not present in Cityscapes. Cityscapes labels do not include adverse conditions or auxiliary data, making them more aligned with the H1 labels. As a result, the performance of the H1 and H2 models on Cityscapes is similar. In contrast, the ACDC dataset, which includes some auxiliary data, benefits significantly from the higher-quality H2 labels.

These findings collectively underscore the high quality of the additional H2 labels, as they enhance model performance in challenging conditions where auxiliary data is crucial.
Together with the good generalization results in 
% Sec. 5.3,
\cref{subsec:additional_segmentation}, 
these results indicate that the additional labeled portion of the images (see 
% Fig. 7a) 
\cref{fig:lidar:stats:plot} left) 
is accurately labeled. As these are also more difficult-to-predict areas, these labels guide models effectively during training, leading to improved generalization (see 
% Sec. 5.3).
\cref{subsec:additional_segmentation}).

\begin{table}
\caption{\textbf{Annotation stage wise PQ of Mask2Former~\cite{cheng2022masked}} (Swin-T~\cite{liu2021swin} backbone, frame camera input only) evaluated on stage 1 labels (H1) and stage 2 labels (H2).}
\label{table:exp:H1:H2}
\centering
\setlength\tabcolsep{4pt}
\footnotesize
\begin{tabular*}{\linewidth}{@{\extracolsep{\fill}}lccccccc@{}}
\toprule
 & Clear & Fog & Rain & Snow & Day & Night & All \\
\midrule
H1&57.9&57.2&57.0&52.4&58.3&51.7&58.0\\
H2&48.8&46.5&45.4&42.2&49.4&39.4&46.9\\
\bottomrule
\end{tabular*}
\end{table}

\begin{table}
\caption{\textbf{
Mask2Former~\cite{cheng2022masked} (Swin-L~\cite{liu2021swin}) performance trained on H2 versus H1 annotations.}}
\label{table:exp:seg:H1:H2}
\centering
\setlength\tabcolsep{4pt}
\footnotesize
\begin{tabular*}{\linewidth}{@{\extracolsep{\fill}} lcccr@{}}
\toprule
mIoU $\uparrow$ & Cityscapes & ACDC & MUSES \\
\midrule
Train on MUSES--H2 & 73.1 & 72.0 & 77.1\\
Train on MUSES--H1 & 73.3 & 70.0 & 73.8\\
\bottomrule
\end{tabular*}
\end{table}

\section{Detailed Class-Level Results}

For the models presented in 
% Sec. 5.2,
\cref{subsec:impact_of_sensors}, 
we present detailed class-level PQ results in \cref{table:supp:class:condition:all,table:supp:class:condition:clear,table:supp:class:condition:fog,table:supp:class:condition:rain,table:supp:class:condition:snow,table:supp:class:condition:day,table:supp:class:condition:night}. 
Adding an event camera to the frame-based camera performs well across the board, with the event camera showing large improvement on small dynamic classes like ``motorcycle'', ``bicycle'' and ``person''. 
Radar seems to be specifically good for larger metallic objects like ``bus'', ``train'' and ``car''. Their metal parts make them easier to detect with radar, due to their large radar cross-section.
The lidar is generally very helpful in identifying large continuous objects like ``building'' and ``truck''. These larger objects usually have good lidar returns, even in more challenging weather conditions.
Further, we can also observe an especially large gap of 7.3\% PQ from lidar to the other modalities for ``sign''. This is because traffic signs, which have reflective coatings, give off high-intensity readings for lidar.

For the models presented in 
% Sec. 5.4,
\cref{subsec:upq_experiments}, 
we present detailed class-level UPQ results in \cref{table:supl:pq:upq:oracle}.
The largest performance gaps between the baselines and the oracle exist for small \emph{things} classes, such as ``person'', ``rider'', and ``motorcycle''.

Class-level results for the semantic segmentation experiments are shown in \cref{table:distribution_shift_supp}.
The models are trained on Cityscapes~\cite{Cityscapes}, 
ACDC~\cite{ACDC}, or MUSES, and evaluated on the MUSES test set.
The class-level results suggest that the ``rider'' and ``motorcycle'' classes are the most difficult, potentially due to their rarity.  

\begin{table*}[!tb]
  \caption{\textbf{Test set class-wise PQ of Mask2Former~\cite{cheng2022masked} for different variations of input sensors}. A Swin-T~\cite{liu2021swin} backbone is used in all cases.}
  \label{table:supp:class:condition:all}
  \centering
  \footnotesize
  \setlength{\tabcolsep}{3pt}
  \resizebox{\textwidth}{!}{
  \begin{tabular}{*{24}{c}}
  \toprule
  \rotatebox[origin=c]{90}{Frame camera} & \rotatebox[origin=c]{90}{Event camera} & \rotatebox[origin=c]{90}{Lidar} & \rotatebox[origin=c]{90}{Radar}  && \rotatebox[origin=c]{90}{road} & \rotatebox[origin=c]{90}{sidewalk} & \rotatebox[origin=c]{90}{building} & \rotatebox[origin=c]{90}{wall} & \rotatebox[origin=c]{90}{fence} & \rotatebox[origin=c]{90}{pole} & \rotatebox[origin=c]{90}{light} & \rotatebox[origin=c]{90}{sign} & \rotatebox[origin=c]{90}{vegetation} & \rotatebox[origin=c]{90}{terrain} & \rotatebox[origin=c]{90}{sky} & \rotatebox[origin=c]{90}{person} & \rotatebox[origin=c]{90}{rider} & \rotatebox[origin=c]{90}{car} & \rotatebox[origin=c]{90}{truck} & \rotatebox[origin=c]{90}{bus} & \rotatebox[origin=c]{90}{train} & \rotatebox[origin=c]{90}{motorcycle} & \rotatebox[origin=c]{90}{bicycle} \\
  \midrule
  \yes & \no & \no & \no &&95.1 & 72.1 & 76.6 & 38.3 & 29.0 & 39.9 & 37.5 & 48.7 & 69.7 & 49.6 & \textbf{84.0} & 34.0 & 19.3 & 55.7 & 28.6 & 42.6 & 38.2 & 14.9 & 17.1\\
  \yes & \yes & \no & \no &&95.3 & 73.8 & 79.4 & 39.6 & 31.6 & 41.9 & 39.9 & 52.8 & 71.2 & 51.5 & 82.3 & 39.2 & 24.4 & 59.2 & 33.4 & 42.2 & 35.9 & 23.3 & 24.3\\
  \yes & \no & \yes & \no && 95.7 & 75.0 & 79.4 & 41.4 & 32.8 & 45.0 & 39.2 & 52.6 & 71.8 & 52.4 & 83.9 & 40.2 & 25.9 & 61.2 & 35.7 & \textbf{52.2} & 43.2 & 24.4 & 22.3\\
  \yes & \no & \no & \yes &&95.3 & 76.0 & \textbf{80.6} & 47.1 & 34.8 & \textbf{45.9} & \textbf{43.4 }& 60.1 &\textbf{ 75.5 }& 53.1 & 83.6 & 42.1 & 29.7 & 63.5 & \textbf{40.9} & 45.5 & \textbf{44.4} & 26.7 & \textbf{24.8}\\
  \yes & \yes & \yes & \yes &&\textbf{95.9} & \textbf{76.4} & 80.4 & \textbf{47.4} & \textbf{37.6} & 45.6 & 42.8 & \textbf{60.2} & 75.0 &\textbf{ 53.8} & 83.7 & \textbf{42.2} & \textbf{33.5 }&\textbf{ 63.7} & 40.0 & 45.1 & 42.6 & \textbf{28.5} & 23.9\\
  \bottomrule
  \end{tabular}}
\end{table*}

\begin{table*}[!tb]
  \caption{\textbf{Uncertainty-aware panoptic segmentation baselines and oracles by class in UPQ.} Mask2Former~\cite{cheng2022masked} with Swin-T~\cite{liu2021swin} is used.}
  \label{table:supl:pq:upq:oracle}
  \centering
  \footnotesize
  \setlength{\tabcolsep}{3pt}
  \resizebox{\textwidth}{!}{
  \begin{tabular}{*{21}{c}}
  \toprule  
\rotatebox[origin=c]{90}{Multimodal} & \rotatebox[origin=c]{90}{Confidence} & \rotatebox[origin=c]{90}{road} & \rotatebox[origin=c]{90}{sidewalk} & \rotatebox[origin=c]{90}{building} & \rotatebox[origin=c]{90}{wall} & \rotatebox[origin=c]{90}{fence} & \rotatebox[origin=c]{90}{pole} & \rotatebox[origin=c]{90}{light} & \rotatebox[origin=c]{90}{sign} & \rotatebox[origin=c]{90}{vegetation} & \rotatebox[origin=c]{90}{terrain} & \rotatebox[origin=c]{90}{sky} & \rotatebox[origin=c]{90}{person} & \rotatebox[origin=c]{90}{rider} & \rotatebox[origin=c]{90}{car} & \rotatebox[origin=c]{90}{truck} & \rotatebox[origin=c]{90}{bus} & \rotatebox[origin=c]{90}{train} & \rotatebox[origin=c]{90}{motorcycle} & \rotatebox[origin=c]{90}{bicycle} \\
\midrule
  \no & ~Constant 100\%  & 95.1 & 72.1 & 76.6 & 38.3 & 29.0 & 39.9 & 37.5 & 48.7 & 69.7 & 49.6 & 84.0 & 34.0 & 19.3 & 55.7 & 28.6 & 42.6 & 38.2 & 14.9 & 17.1\\
  \no & ~Marginalization  & 89.3 & 68.6 & 72.0 & 36.7 & 28.9 & 35.2 & 27.1 & 40.6 & 67.5 & 50.0 & 81.4 & 29.7 & 21.8 & 52.7 & 30.1 & 37.1 & 37.2 & 16.0 & 20.7 \\
  \no & ~Oracle &  96.6 & 84.0 & 85.9 & 67.7 & 59.5 & 58.5 & 73.0 & 69.2 & 88.4 & 78.2 & 97.1 & 71.8 & 71.5 & 76.7 & 66.1 & 68.7 & 80.0 & 68.9 & 66.3 \\
  \bottomrule
\end{tabular}}
\end{table*}

\begin{table*}[!tb]
  \caption{\textbf{Class-level results for semantic segmentation with Mask2Former~\cite{cheng2022masked}} (Swin-L~\cite{liu2021swin}, RGB input only). All models are evaluated on the test set of \Ours{}.}
  \label{table:distribution_shift_supp}
  \centering
  \footnotesize
  \setlength{\tabcolsep}{3pt}
  \resizebox{\textwidth}{!}{
  \begin{tabular}{l*{20}{c}}
  \toprule
  \multirow{2}{*}{Training Dataset} && \multicolumn{19}{c}{\Ours{} IoU\,$\uparrow$} \\
  \cmidrule{3-21} && \rotatebox[origin=c]{90}{road} & \rotatebox[origin=c]{90}{sidew.} & \rotatebox[origin=c]{90}{build.} & \rotatebox[origin=c]{90}{wall} & \rotatebox[origin=c]{90}{fence} & \rotatebox[origin=c]{90}{pole} & \rotatebox[origin=c]{90}{light} & \rotatebox[origin=c]{90}{sign} & \rotatebox[origin=c]{90}{veget.} & \rotatebox[origin=c]{90}{terrain} & \rotatebox[origin=c]{90}{sky} & \rotatebox[origin=c]{90}{person} & \rotatebox[origin=c]{90}{rider} & \rotatebox[origin=c]{90}{car} & \rotatebox[origin=c]{90}{truck} & \rotatebox[origin=c]{90}{bus} & \rotatebox[origin=c]{90}{train} & \rotatebox[origin=c]{90}{motorc.} & \rotatebox[origin=c]{90}{bicycle} \\ \midrule
  Cityscapes~\cite{Cityscapes} && 84.1 & 55.8 & 81.2 & 40.7 & 40.7 & 53.5 & 56.1 & 55.0 & 77.2 & 43.7 & 84.0 & 47.3 & 55.9 & 67.3 & 53.7 & 66.6 & 69.8 & 32.4 & 55.0  \\
  ACDC~\cite{ACDC} && 91.7 & 73.7 & 85.8 & 53.6 & 42.6 & 61.4 & 69.1 & 68.6 & 79.0 & 53.3 & 88.6 & 59.4 & 46.8 & 87.6 & 66.4 & 78.5 & 85.0 & 27.2 & 52.6 \\
  \Ours{} && 96.5 & 84.9 & 91.8 & 73.3 & 59.5 & 68.1 & 76.7 & 74.2 & 87.5 & 74.5 & 96.3 & 72.8 & 56.5 & 93.2 & 67.6 & 90.2 & 86.8 & 48.7 & 66.2 \\
  \bottomrule
\end{tabular}}
\end{table*}

\section{Further Dataset Statistics}
\label{sec:suppl:further:stats}

\subsection{Things-Classes Statistics}
\label{suppl:sub:sec:thing:stats}

By projecting the lidar points onto the ground truth, we calculate distance statistics on the \emph{things} classes. For each instance, we filter the points for outliers with a z-score of 1 and average the lidar distance. 
The resulting average distance per condition is presented in \cref{table:things:statistics:distance}. As expected by its degrading nature onto the lidar, the fog has the lowest average \emph{things} instance distance. We also observe a shift along the day-night axis, attributed to worse visual conditions at night making distant objects harder to identify and label. The average number of instances per image also varies largely between the different conditions. Fog does only have 2.06 instances per image, caused by two reasons. Firstly, it is harder to identify and annotate individual instances doubt-free in this condition, and secondly, heavy fog is mostly present in rural areas with less densely populated scenes.

\begin{table*}
\caption{\textbf{Average \emph{things} class statistics by condition.}}
\label{table:things:statistics:distance}
\centering
\setlength\tabcolsep{3.2pt}
\footnotesize
\begin{tabular*}{\linewidth}{ @{\extracolsep{\fill}}lccccccc@{}}
\toprule
                 & Clear   & Fog      & Rain     & Snow      & Day      & Night    & All    \\
\midrule
Distance [m]~~ & 46.64    & 38.23    & 39.25    & 40.40    & 44.22    & 38.44    & 42.05    \\
\# Instances in image      & \phantom{0}9.56   & \phantom{0}2.06      & 12.13    & \phantom{0}7.92   & \phantom{0}8.49     & \phantom{0}7.34     & \phantom{0}8.03     \\
\bottomrule
\end{tabular*}
\end{table*}

\subsection{Lidar Point Cloud Statistics}
\begin{figure}
\centering
\begin{tikzpicture}
\begin{axis}[
    ybar,
    width=0.79\textwidth,
    height=4.5cm,
    bar width=10pt,
    axis lines=left, 
    axis line style={-}, 
    symbolic x coords={0-20m,20-40m,\textgreater40m},
    enlarge x limits=0.25,
    xtick=data,
    ylabel=Lidar Points (\%),
    ymin=0,
    ymax=100,
    legend cell align={left},
    cycle list={        {fill=color1!20,draw=black},
    {fill=color2!40,draw=black},
        {fill=color3!60,draw=black},
        {fill=color4!80,draw=black},
    },
]
\addplot coordinates {(0-20m, 64.31) (20-40m, 23.02) (\textgreater40m, 12.66)};
\addplot coordinates {(0-20m, 80.69) (20-40m, 16.03) (\textgreater40m, 3.28)};
\addplot coordinates {(0-20m, 61.45) (20-40m, 24.30) (\textgreater40m, 14.25)};
\addplot coordinates {(0-20m, 70.97) (20-40m, 20.44) (\textgreater40m, 8.59)};
\legend{Clear, Fog, Rain, Snow}
\end{axis}
\end{tikzpicture}
\caption{\textbf{\Ours{} share of lidar points} in specific distance bins by weather condition. }
\label{fig:lidar:stats:plot:pc:dist}
\end{figure}
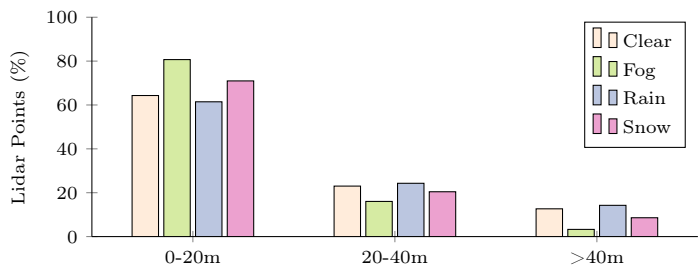

We present statistics on the lidar point clouds in \cref{table:statistics:lidar} and \cref{fig:lidar:stats:plot:pc:dist}. Notably, fog significantly impacts the point cloud, reducing the average points in a single lidar scan by one-third. In foggy conditions, only 3.28\% of points are farther away than 40m, in contrast to 12.66\% in clear weather. This is expected due to the squared attenuation effect of fog particles in the air. It underscores the limitations of lidar in highly adverse conditions, where fewer points are returned, and distant objects become imperceptible in the lidar point cloud.

\begin{table*}
\caption{\textbf{\Ours{} lidar point cloud statistics} by weather condition.}
\label{table:statistics:lidar}
\smallskip
\centering
\setlength\tabcolsep{4pt}
\footnotesize
\begin{tabular*}{\linewidth}{@{\extracolsep{\fill}}  lcccccc@{}}
\toprule
\multirow{2}{*}{Condition} & 
\multirow{2}{*}{0-20m} & 
\multirow{2}{*}{20-40m} & 
\multirow{2}{*}{$>$40m} & 
Average point  & Average \# of &  Average \# of \\
& & & & distance [m]  &points&  points in image \\
\midrule
Clear          & 64.31\% &	23.02\%		&12.66\% & 21.95 & 62\,862   & 42\,331 \\
Fog       & 80.69\%	& 16.03\%	&	\phantom{0}3.28\% & 14.60 &  41\,377 &   27\,346\\
Rain        & 61.45\%	& 24.30\%	&	14.25\%& 22.91 & 53\,563  & 35\,055\\
Snow       & 70.97\%	& 20.44\%	&	\phantom{0}8.59\% & 18.63 &  63\,713  & 42\,679\\
\bottomrule
\end{tabular*}
\end{table*}

\subsection{Difficulty Map Distribution}

We present the distribution by condition of the difficulty map in \cref{table:statistics:difficulty:condition}. 23.49\% of off all pixels have a \emph{difficult\_class} label and 6.57\% of all \emph{things} pixels have an implicit \emph{difficult\_instance} label.

We present the distribution by condition of the difficulty map in \cref{table:statistics:difficulty:class}. Terrain and fences have the largest share of \emph{difficult\_class} labels. From the \emph{things} classes, bicycle was specifically difficult to label with 21.6\% of the class having a \emph{difficult\_class} and an additional 30.6\% having an explicit \emph{difficult\_class} label.

\begin{table*}
\caption{\textbf{Difficulty map distribution by condition in \%.} \emph{Things} \emph{difficult\_class} regions are implicitly also \emph{difficult\_instance}. 
``\emph{difficult\_class}'': share of pixels with \emph{difficult\_class} entries in difficulty map.
``\emph{difficult\_class} excl.\ unlabeled'': share of pixels with \emph{difficult\_class} entries in difficulty map, excluding unlabeled pixels.
``\emph{difficult\_instance}'': share of all pixels with explicit \emph{difficult\_instance} entries in the difficulty map. 
``\emph{things} w/ exp.\ \emph{difficult\_instance}'': share of all \emph{things} pixels with explicit \emph{difficult\_instance} entries in the difficulty map. 
``\emph{things} w/ imp.\ \emph{difficult\_instance}'': share of all \emph{things} pixels with \emph{difficult\_instance} label, including \emph{difficult\_class} labels that are implicitly also \emph{difficult\_instance}.
}
\label{table:statistics:difficulty:condition}
\centering
\setlength\tabcolsep{4pt}
\footnotesize
\resizebox{\textwidth}{!}{
\begin{tabular}{l*{5}{c}}
\toprule
\multirow{2}{*}{Condition} & 
\multirow{2}{*}{\emph{difficult\_class}} & 
\emph{difficult\_class} & \multirow{2}{*}{\emph{difficult\_instance}} & \emph{things} w/ exp.\ & \emph{things} w/ imp.\ \\ 
& & excl.\ unlabeled &  & \emph{difficult\_instance} &  \emph{difficult\_instance}\\ 
\midrule
Clear        & 13.97 & \phantom{0}5.04 &0.03 & 1.08 & \phantom{0}4.71\\
Rain         & 22.71 &\phantom{0}9.65 &0.12 & 2.64 &  \phantom{0}9.58 \\
Snow        & 19.35 & \phantom{0}9.69&0.03 & 0.80 & \phantom{0}3.51\\
Fog         & 40.68 &18.59 &0.01 & 0.62 & 11.46 \\
Day        & 10.12 &\phantom{0}5.86 &0.04 & 1.23 & \phantom{0}4.57 \\
Night  & 43.56 &17.11 &0.06  & 1.95 & \phantom{0}9.63 \\
Total   & 23.49 &10.36 & 0.05 &  1.51& \phantom{0}6.57\\
\bottomrule
\end{tabular}}
\end{table*}

\begin{table*}
\caption{\textbf{Difficulty map distribution by class in \%.} Share of explicit difficulty labels compared to the total labeled pixels of a given class.}
\label{table:statistics:difficulty:class}
\centering
\setlength\tabcolsep{4pt}
\footnotesize
\resizebox{\textwidth}{!}{
\begin{tabular}{@{\extracolsep{\fill}}  lccccccccccccccccccc@{}}
\toprule
& \rotatebox[origin=c]{90}{road} & \rotatebox[origin=c]{90}{sidew.} & \rotatebox[origin=c]{90}{build.} & \rotatebox[origin=c]{90}{wall} & \rotatebox[origin=c]{90}{fence} & \rotatebox[origin=c]{90}{pole} & \rotatebox[origin=c]{90}{light} & \rotatebox[origin=c]{90}{sign} & \rotatebox[origin=c]{90}{veget.} & \rotatebox[origin=c]{90}{terrain} & \rotatebox[origin=c]{90}{sky} & \rotatebox[origin=c]{90}{person} & \rotatebox[origin=c]{90}{rider} & \rotatebox[origin=c]{90}{car} & \rotatebox[origin=c]{90}{truck} & \rotatebox[origin=c]{90}{bus} & \rotatebox[origin=c]{90}{train} & \rotatebox[origin=c]{90}{motorc.} & \rotatebox[origin=c]{90}{bicycle} \\
\midrule
\emph{difficult\_class} 
 & 3.4 & 13.0 & 9.1 & 17.4  & 23.8 & 12.9 & 15.8 & 10.3 & 20.7 & 37.2 & 21.3 &  16.6 & 15.9 & 3.4 &5.0  & 4.6 & 3.7 & 15.3 & 21.6\\
\emph{difficult\_instance}
& n/a & n/a & n/a & n/a & n/a & n/a & n/a & n/a & n/a & n/a & n/a & \phantom{0}2.0 & \phantom{0}0.1 & 0.6 & 0.0 & 0.0 & 1.5 & \phantom{0}4.9 & 30.6 \\
\bottomrule
\end{tabular}}
\end{table*}

\section{European Domain Bias}

The MUSES dataset consists exclusively of driving scenes from Switzerland, which may introduce a geographical bias towards Western European environments. This limitation is inherent in the dataset's design and location. While this regional focus ensures consistency and depth within a specific context, it may affect the generalizability of the results to other regions with different driving conditions, infrastructure, and weather patterns.

\section{Visualization of \Ours{} Samples }

We show further visualizations of \Ours{} samples in \cref{fig:suppl:annotation:aux:data:1,fig:suppl:annotation:aux:data:2}. The lidar and event camera are thereby projected onto the frame camera for easier inspection. In many of the scenes, the lidar and event camera help identify and clarify unclear areas. This is especially noticeable in rainy conditions, where the frame camera is often blurred by droplets.

%% file: sec/X_supplement_qual_figs.tex
\begin{figure}
\centering
\begin{tabular}{@{}c@{\hspace{0.03cm}}c@{\hspace{0.03cm}}c@{\hspace{0.03cm}}c@{\hspace{0.03cm}}c@{\hspace{0.03cm}}c@{\hspace{0.03cm}}c@{}}
\subfloat{\tiny RGB Image} &
\subfloat{\tiny Lidar} &
\subfloat{\tiny Events}&
\subfloat{\tiny Radar}&
\subfloat{\tiny Corr. Image}&
\subfloat{\tiny Panoptic GT}&
\subfloat{\tiny Difficulty Map}  \\

% Clear Day 1
\includegraphics[width=0.14\textwidth]{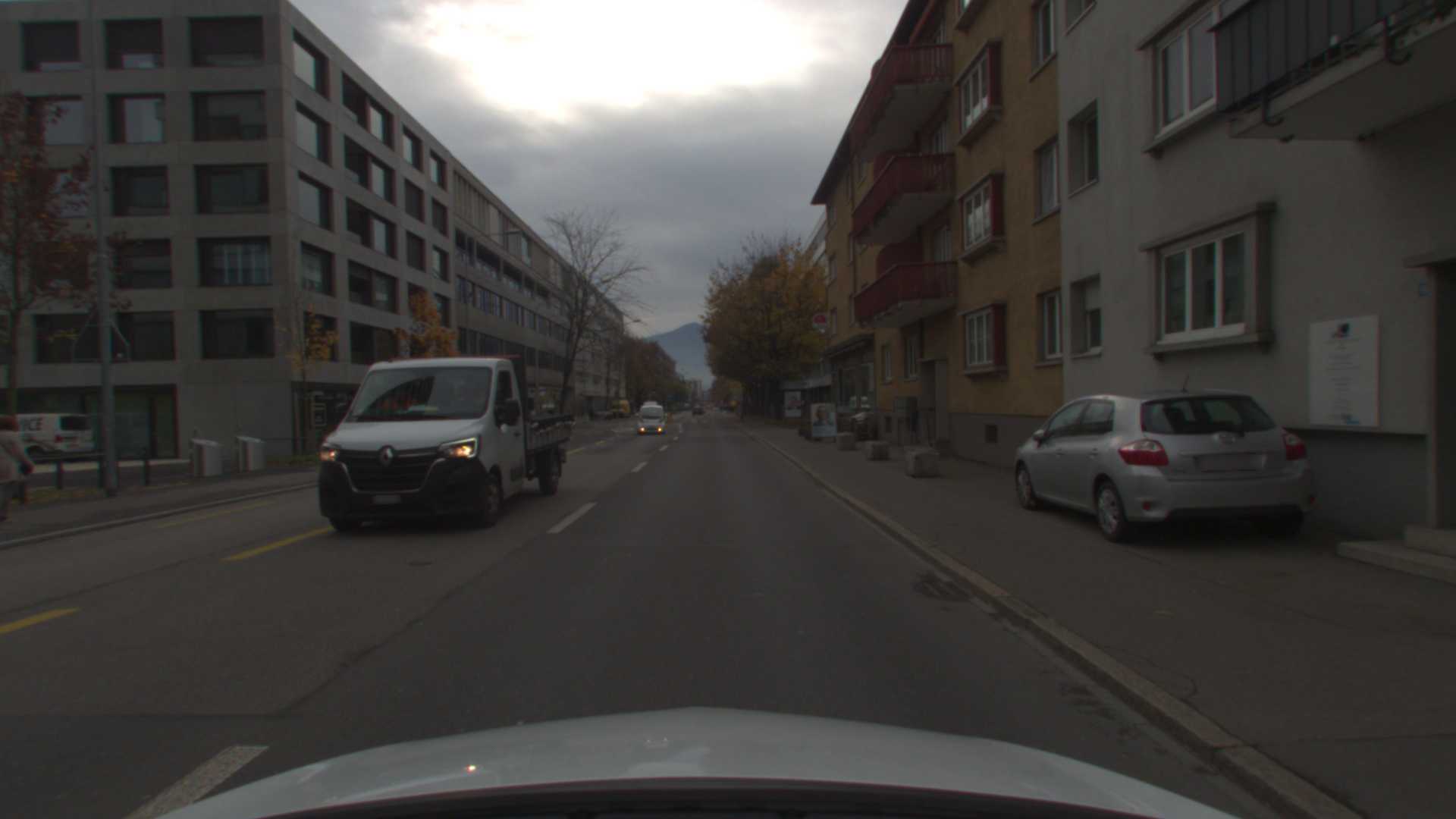} &
\includegraphics[width=0.14\textwidth]{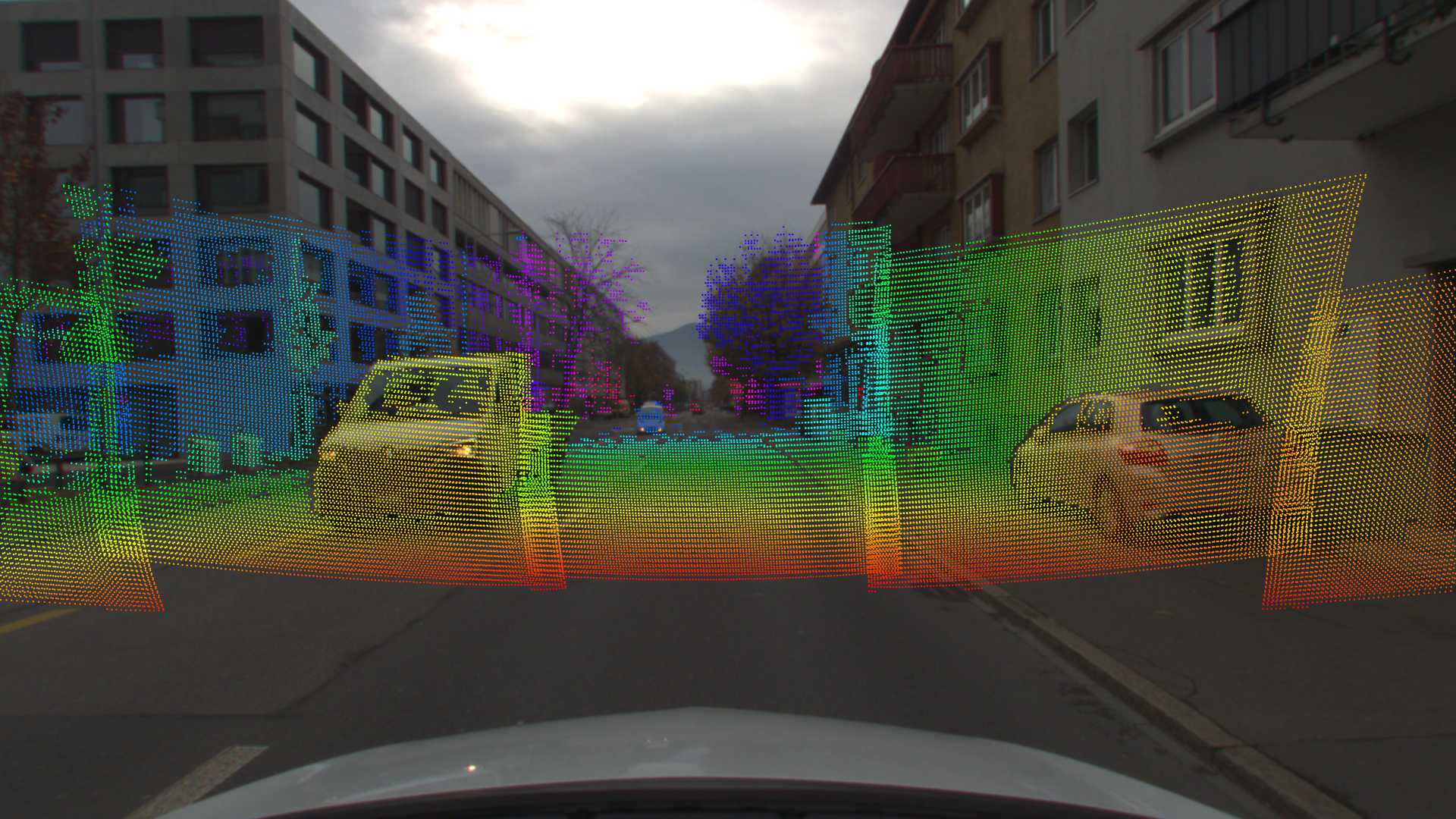} &
\includegraphics[width=0.14\textwidth]{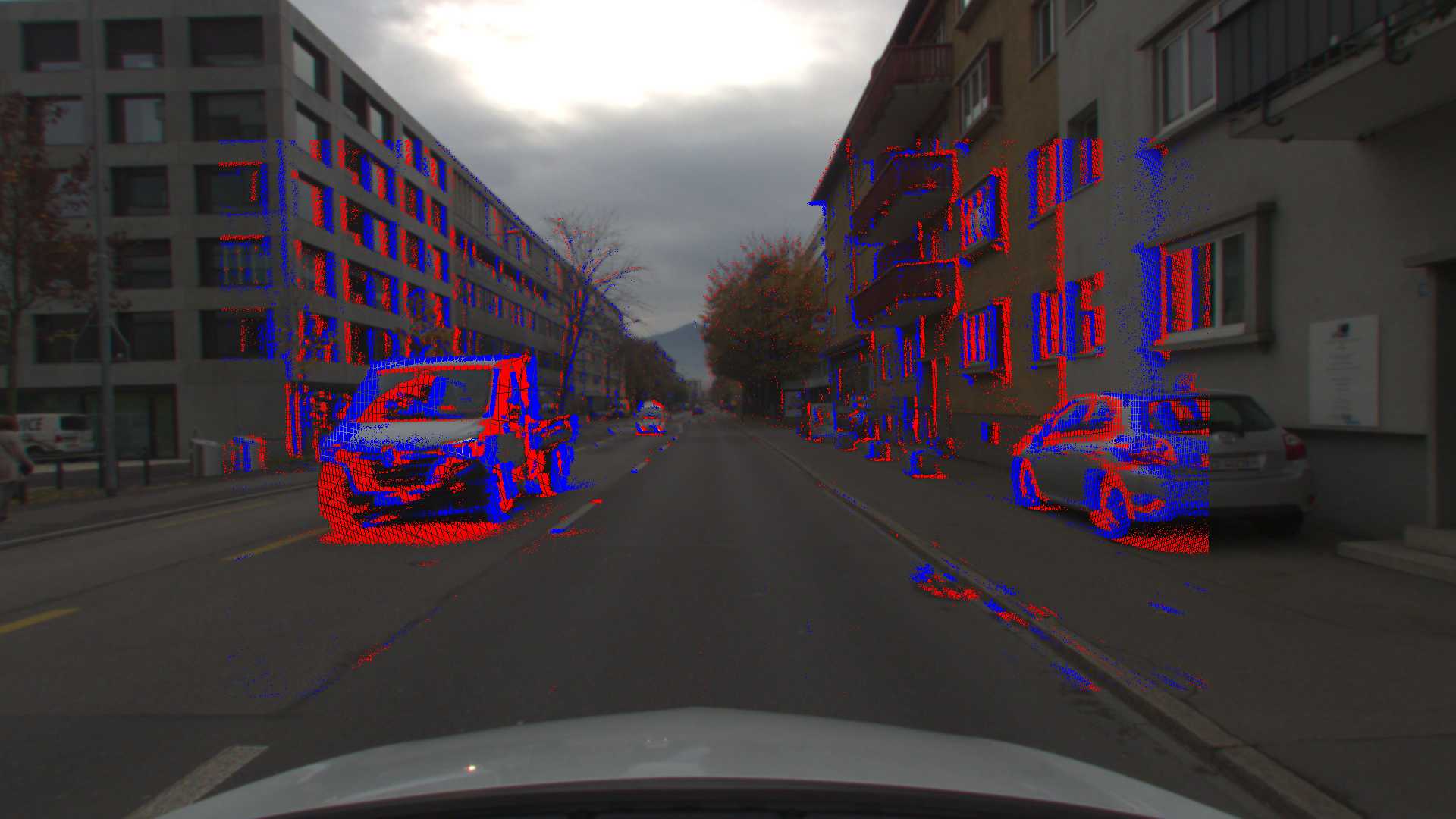} &
\includegraphics[angle=90, trim=0 6835 0 0, clip,width=0.14\textwidth]{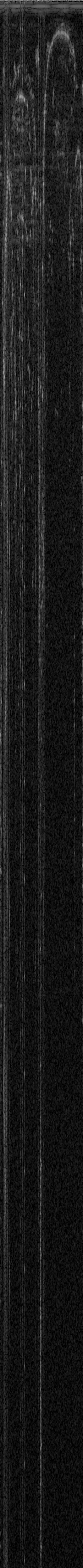}&
\includegraphics[width=0.14\textwidth]{figs/Annotation/supplement/REC0008_frame_080289/REC0008_frame_080289.jpg} &
\includegraphics[width=0.14\textwidth]{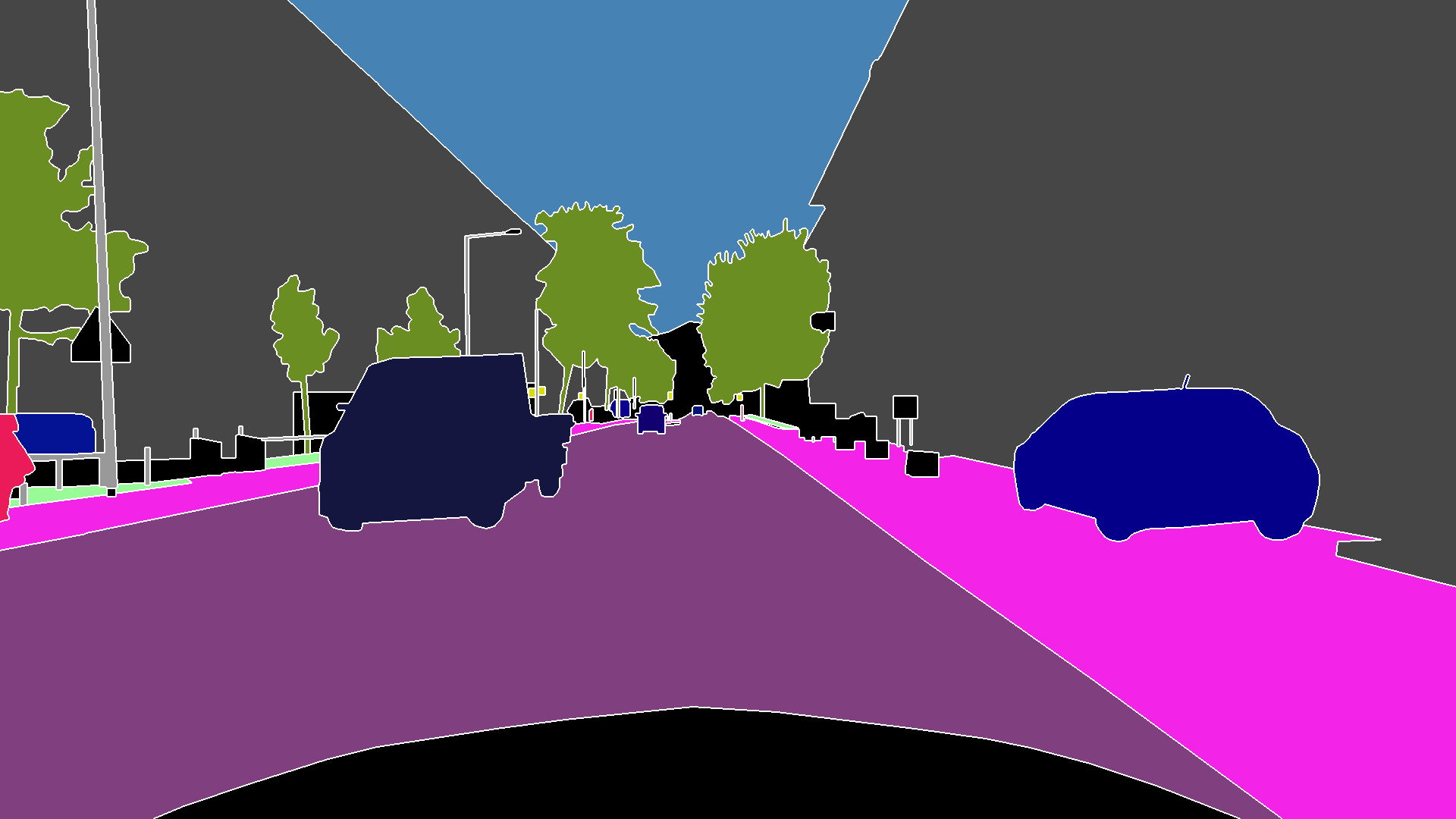} &
\includegraphics[width=0.14\textwidth]{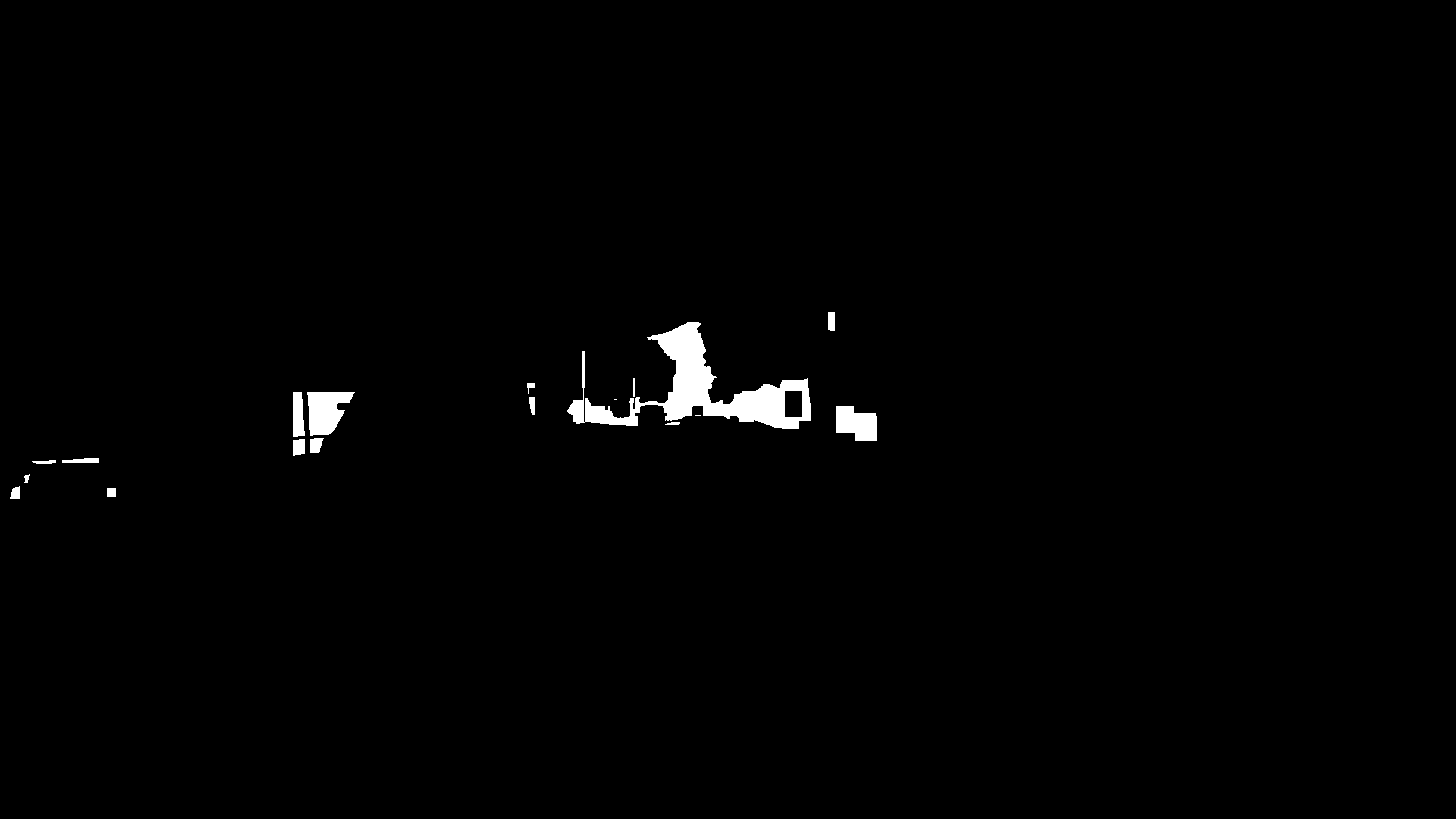}\\

% Clear Day 2
\includegraphics[width=0.14\textwidth]{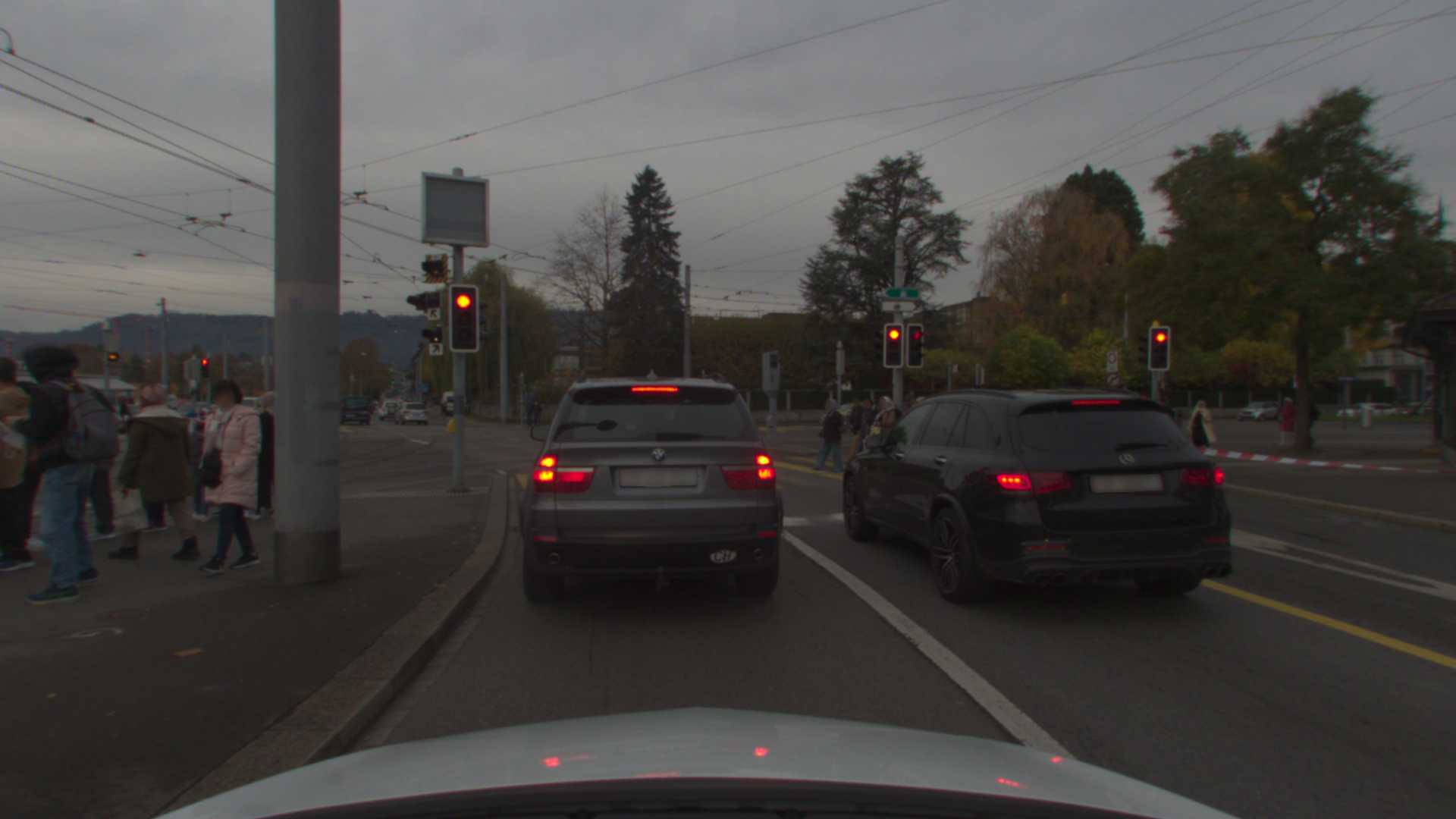} &
\includegraphics[width=0.14\textwidth]{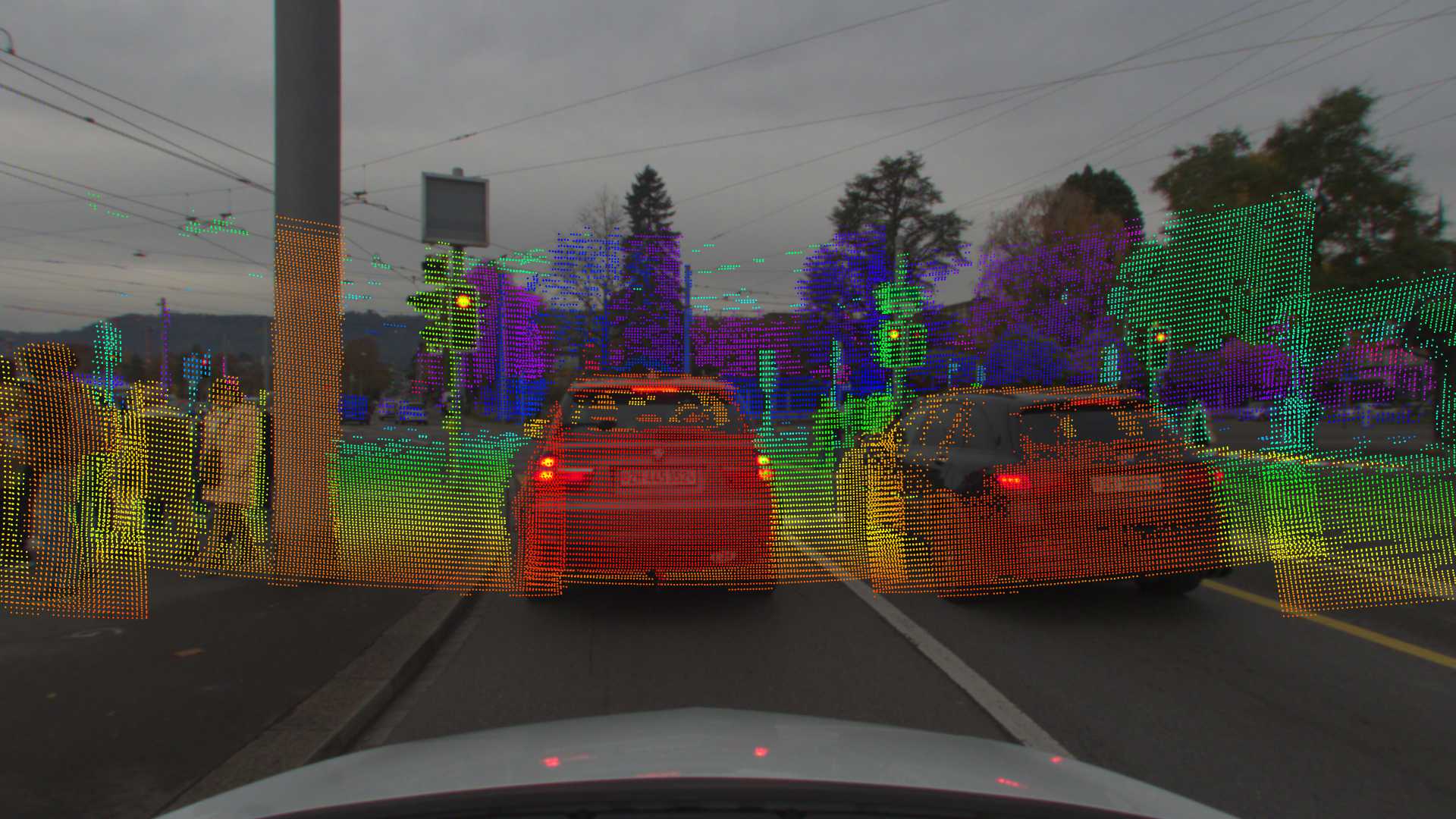} &
\includegraphics[width=0.14\textwidth]{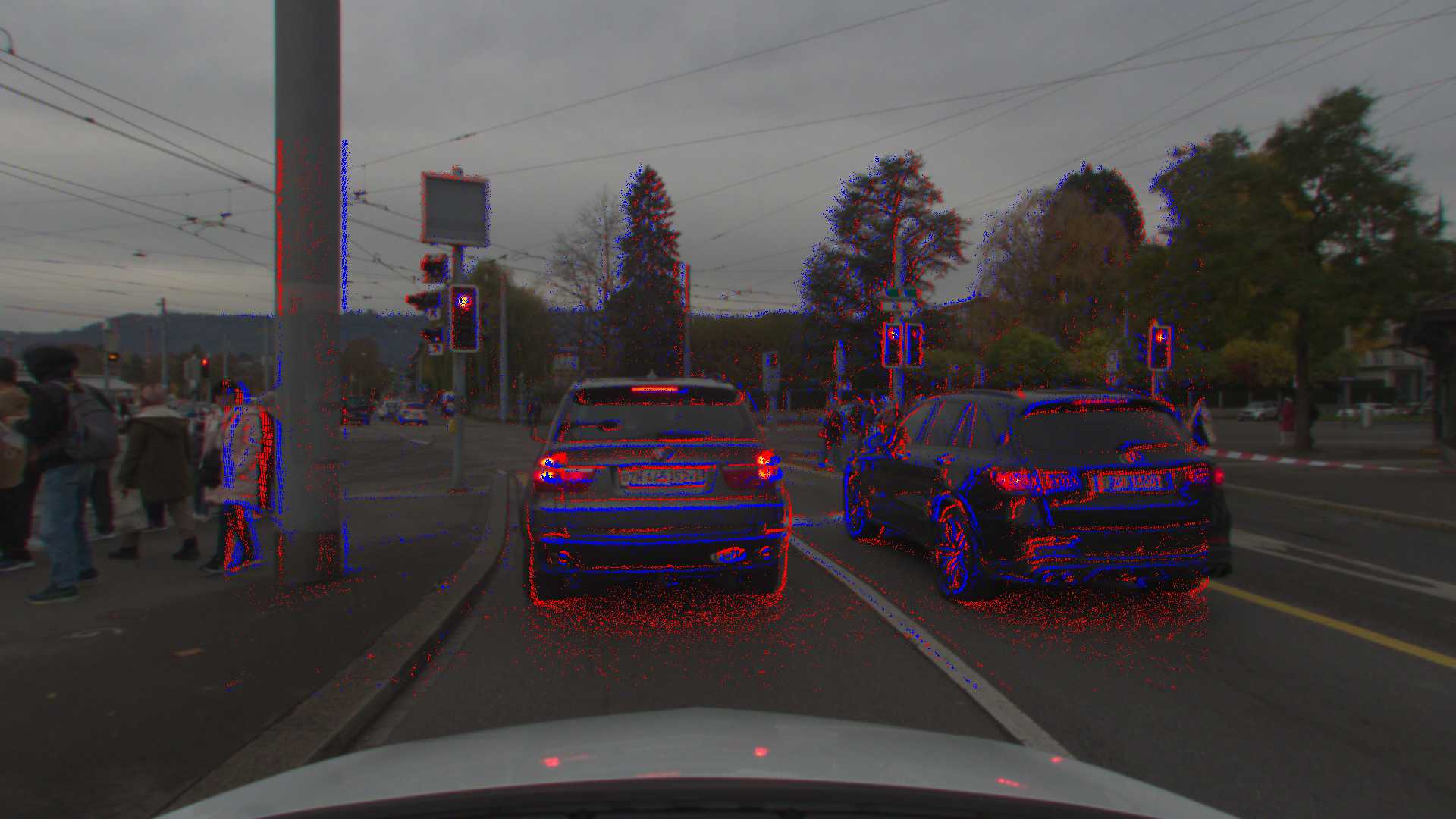} &
\includegraphics[angle=90, trim=0 6835 0 0, clip,width=0.14\textwidth]{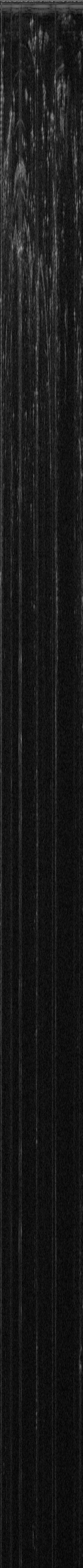}&
\includegraphics[width=0.14\textwidth]{figs/Annotation/supplement/REC0006_frame_043790/REC0006_frame_043790.jpg} &
\includegraphics[width=0.14\textwidth]{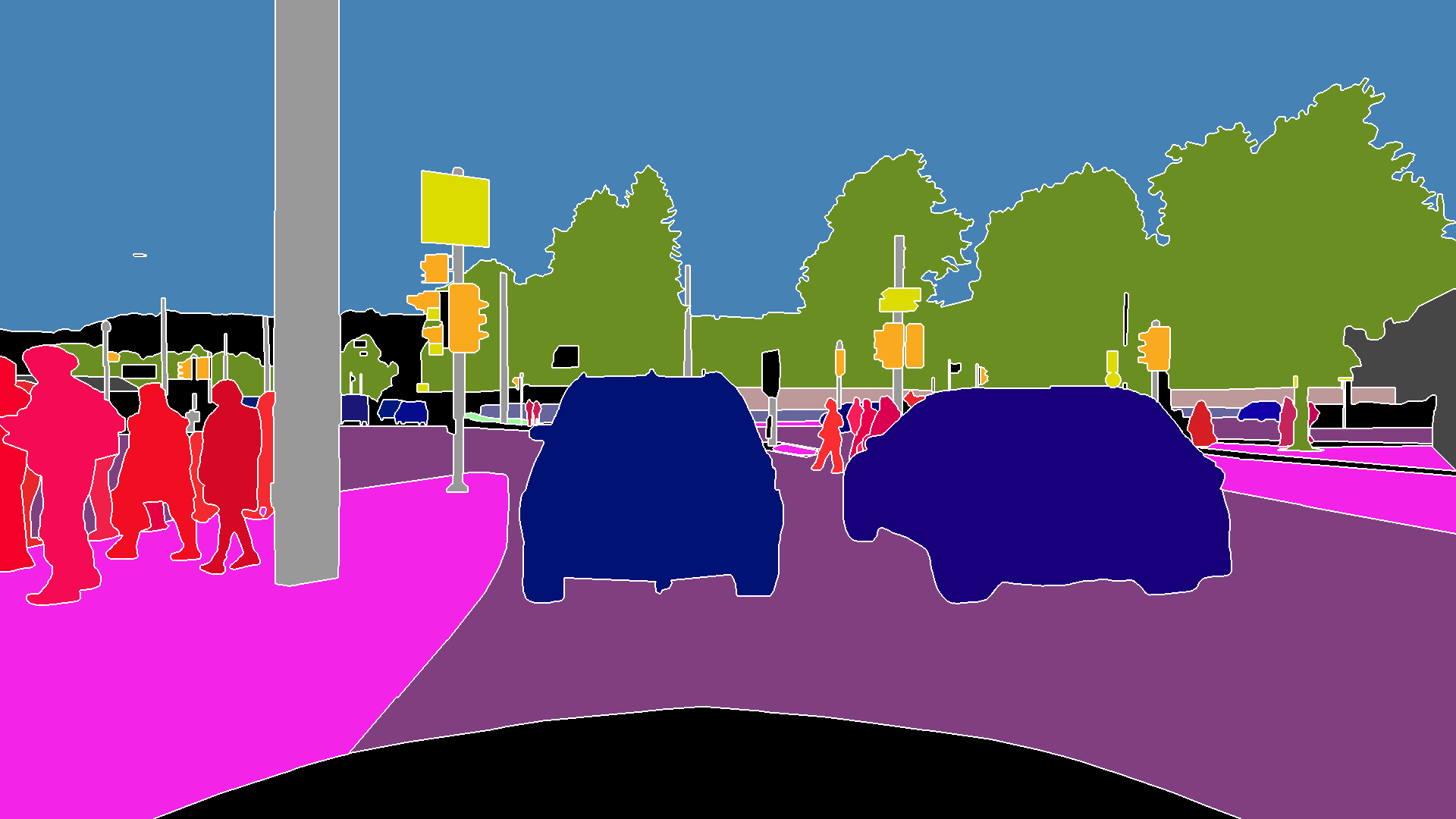} &
\includegraphics[width=0.14\textwidth]{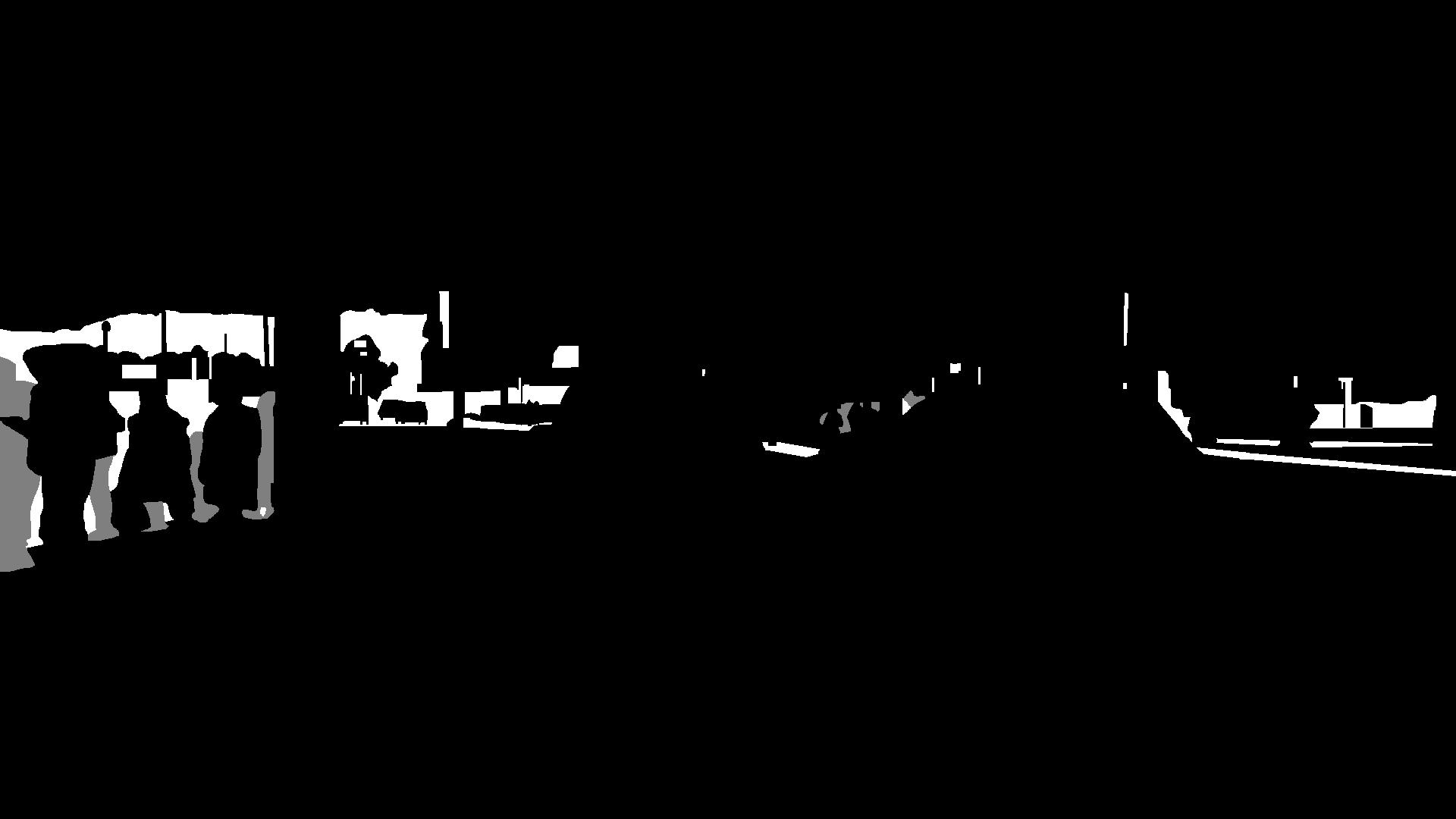}\\

% Clear Night 1
\includegraphics[width=0.14\textwidth]{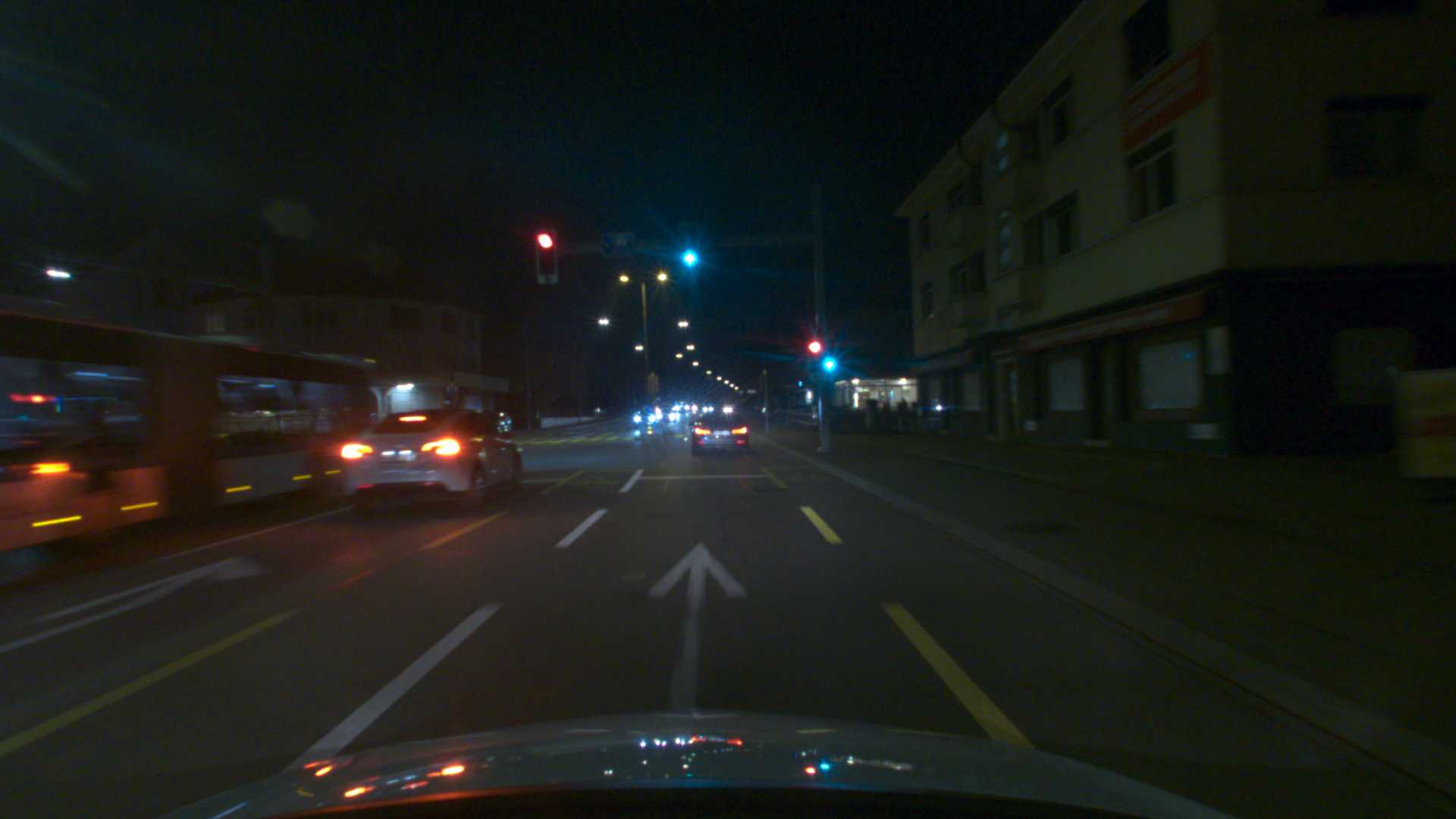} &
\includegraphics[width=0.14\textwidth]{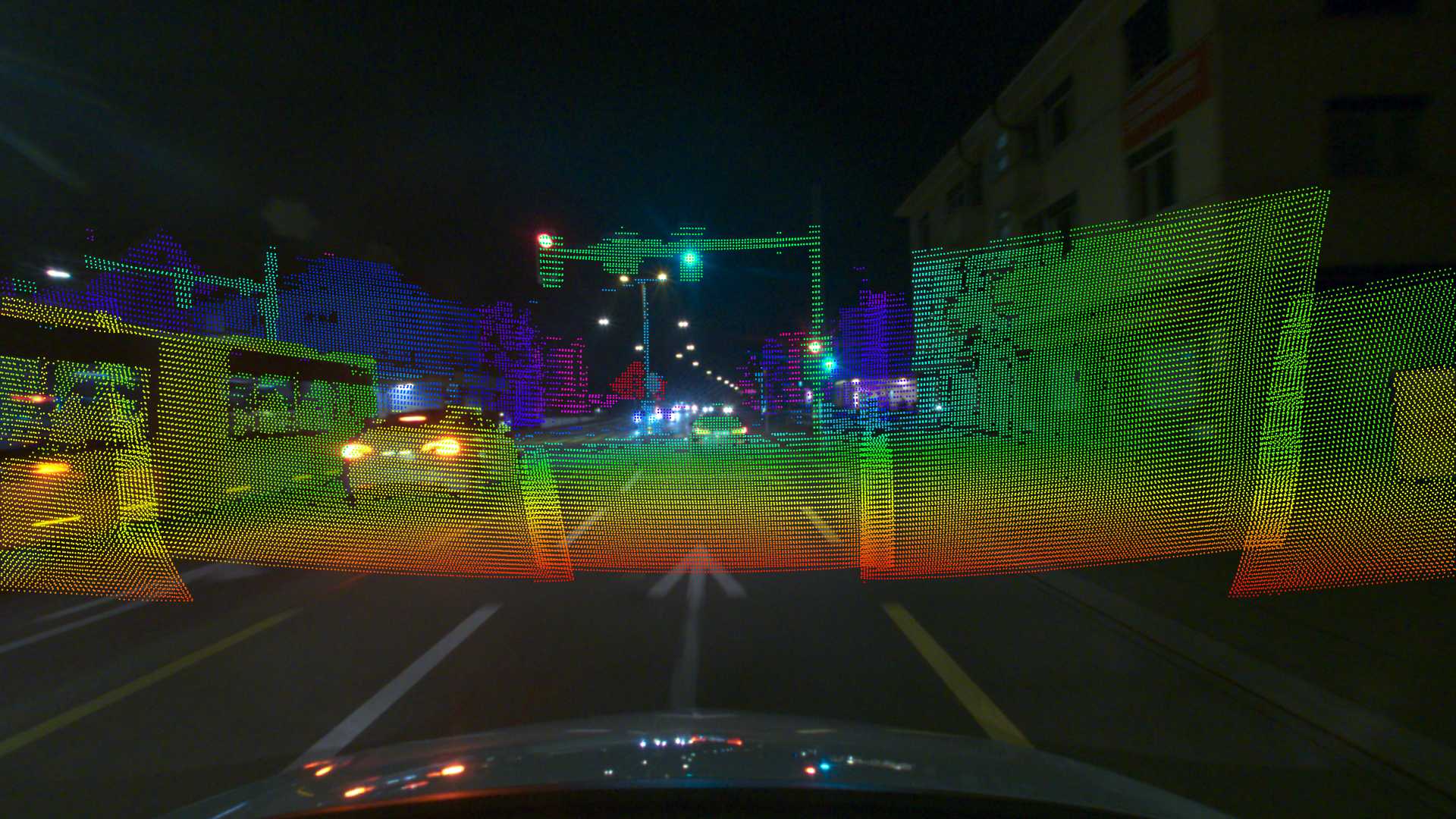} &
\includegraphics[width=0.14\textwidth]{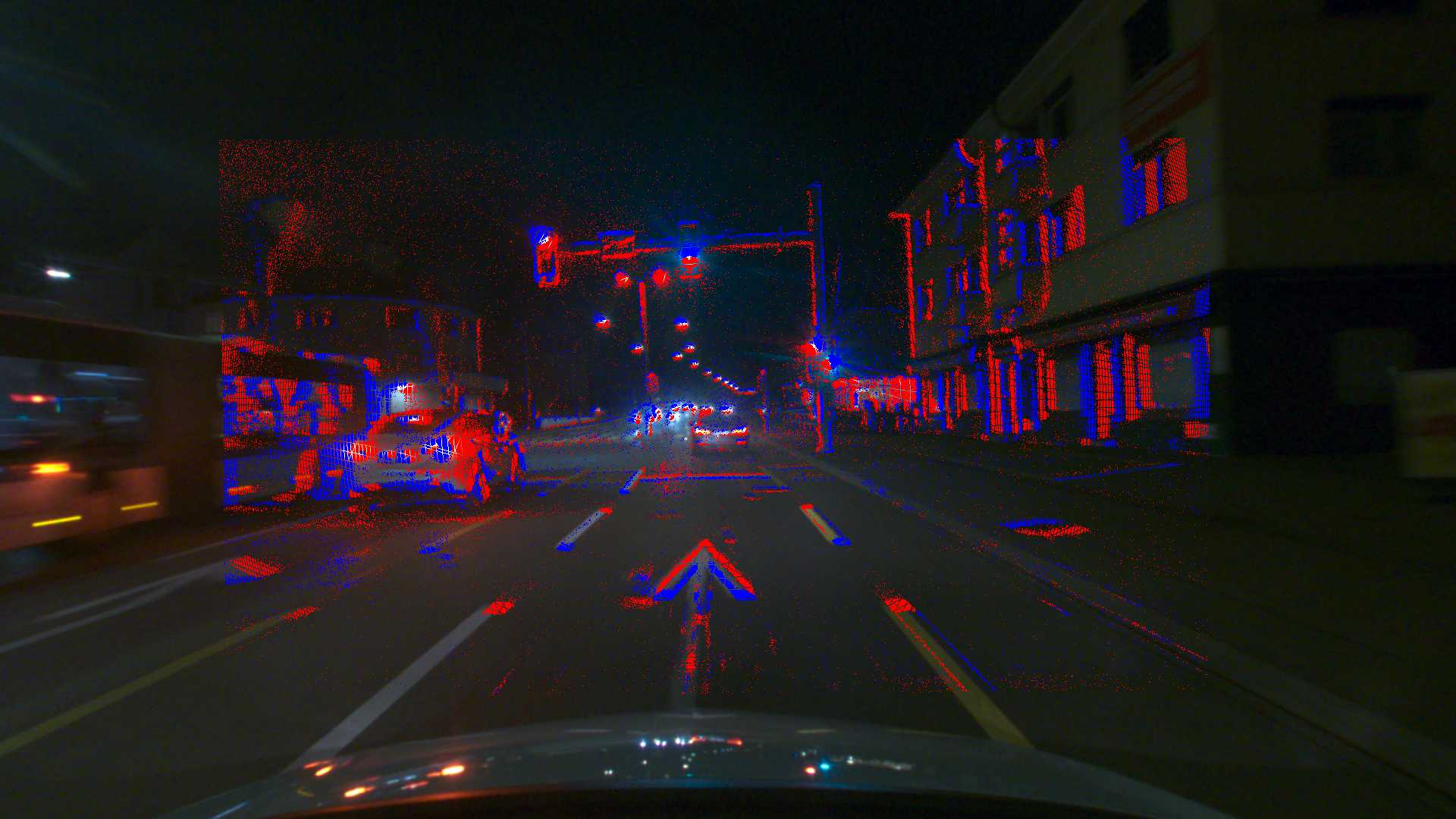} &
\includegraphics[angle=90, trim=0 6835 0 0, clip,width=0.14\textwidth]{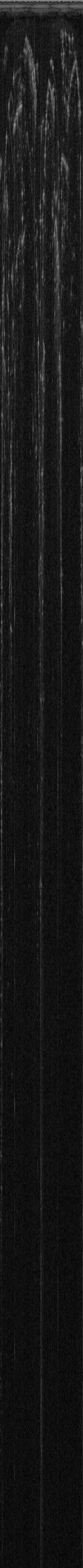}&
\includegraphics[width=0.14\textwidth]{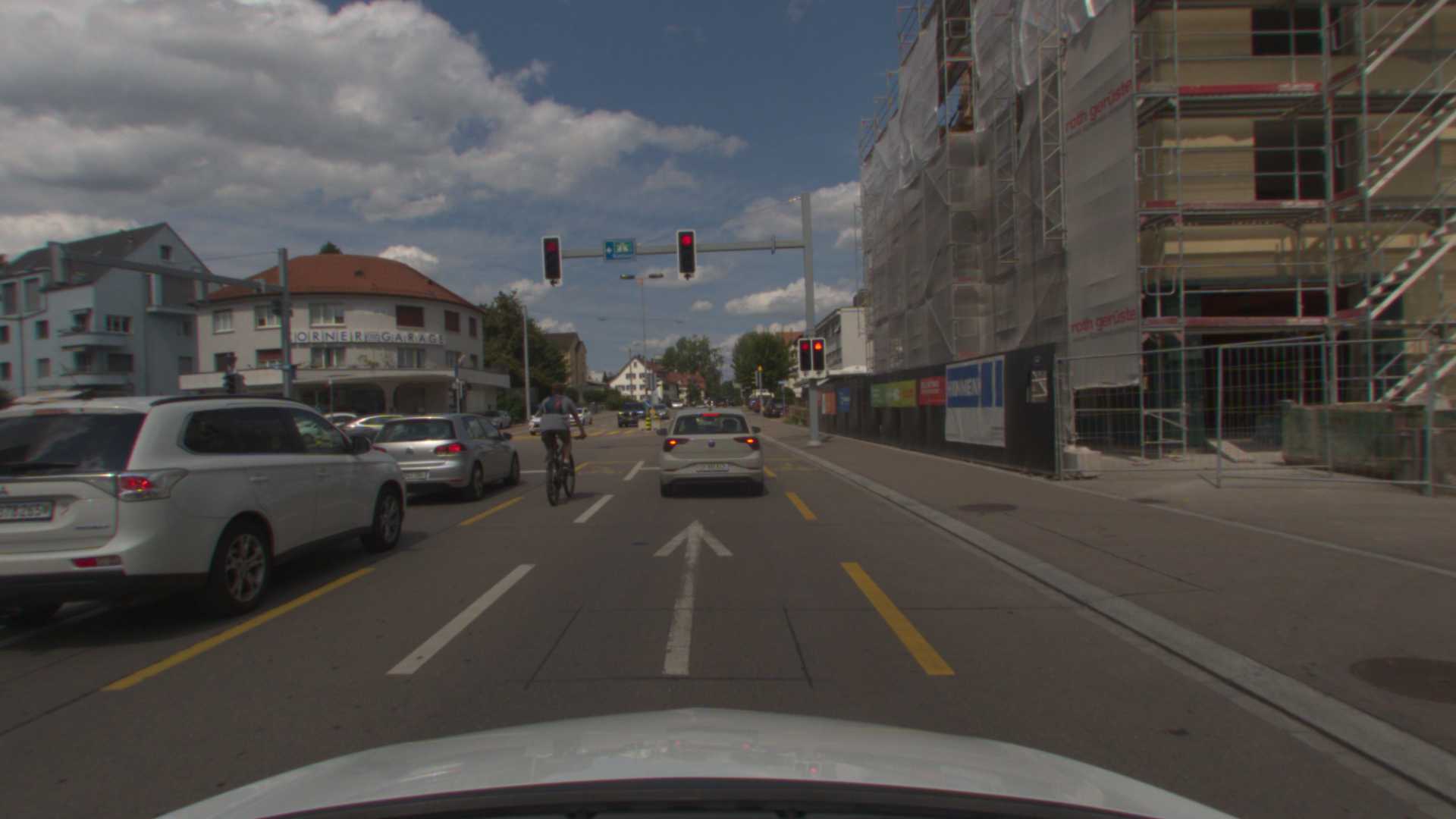} &
\includegraphics[width=0.14\textwidth]{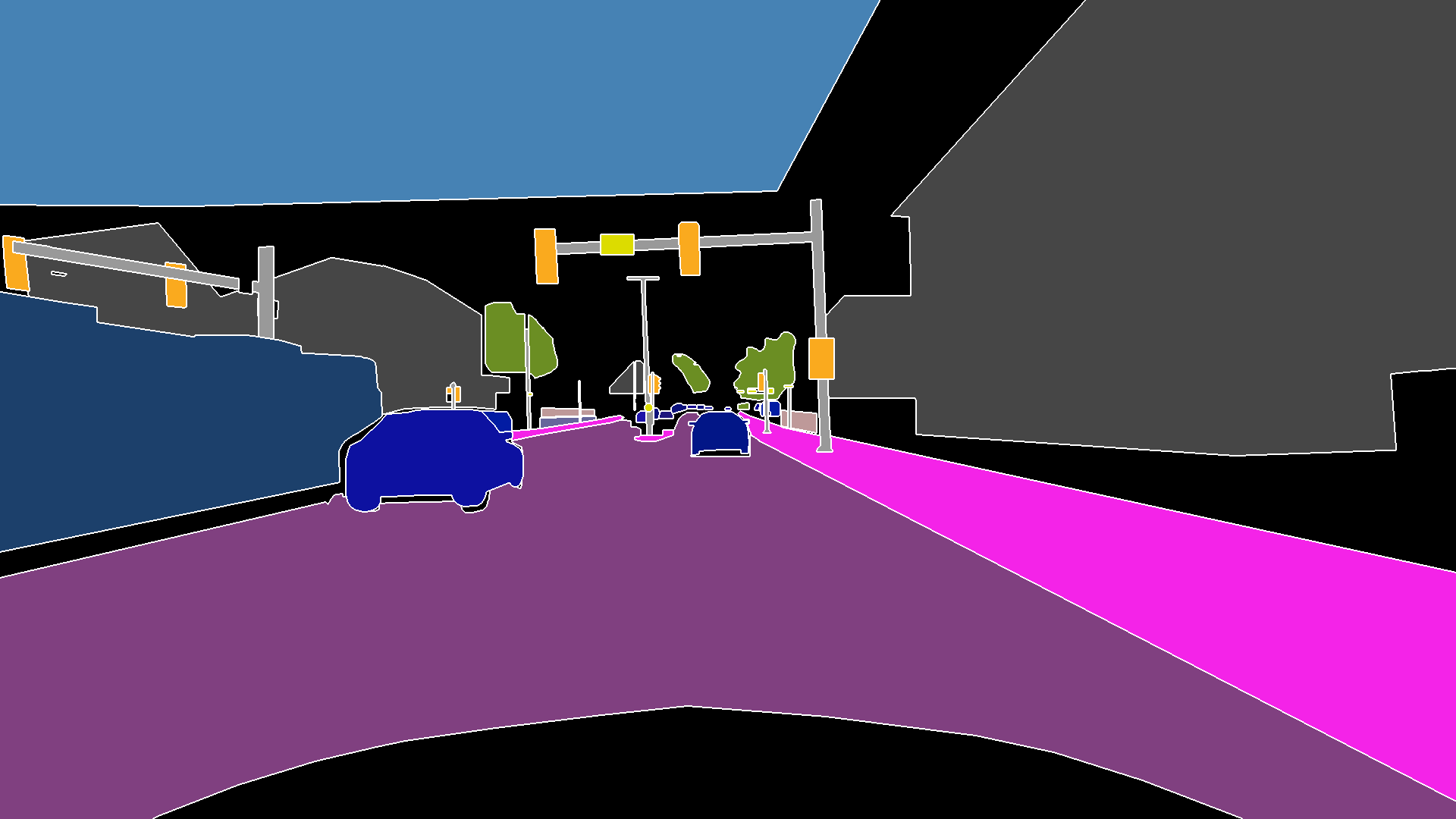} &
\includegraphics[width=0.14\textwidth]{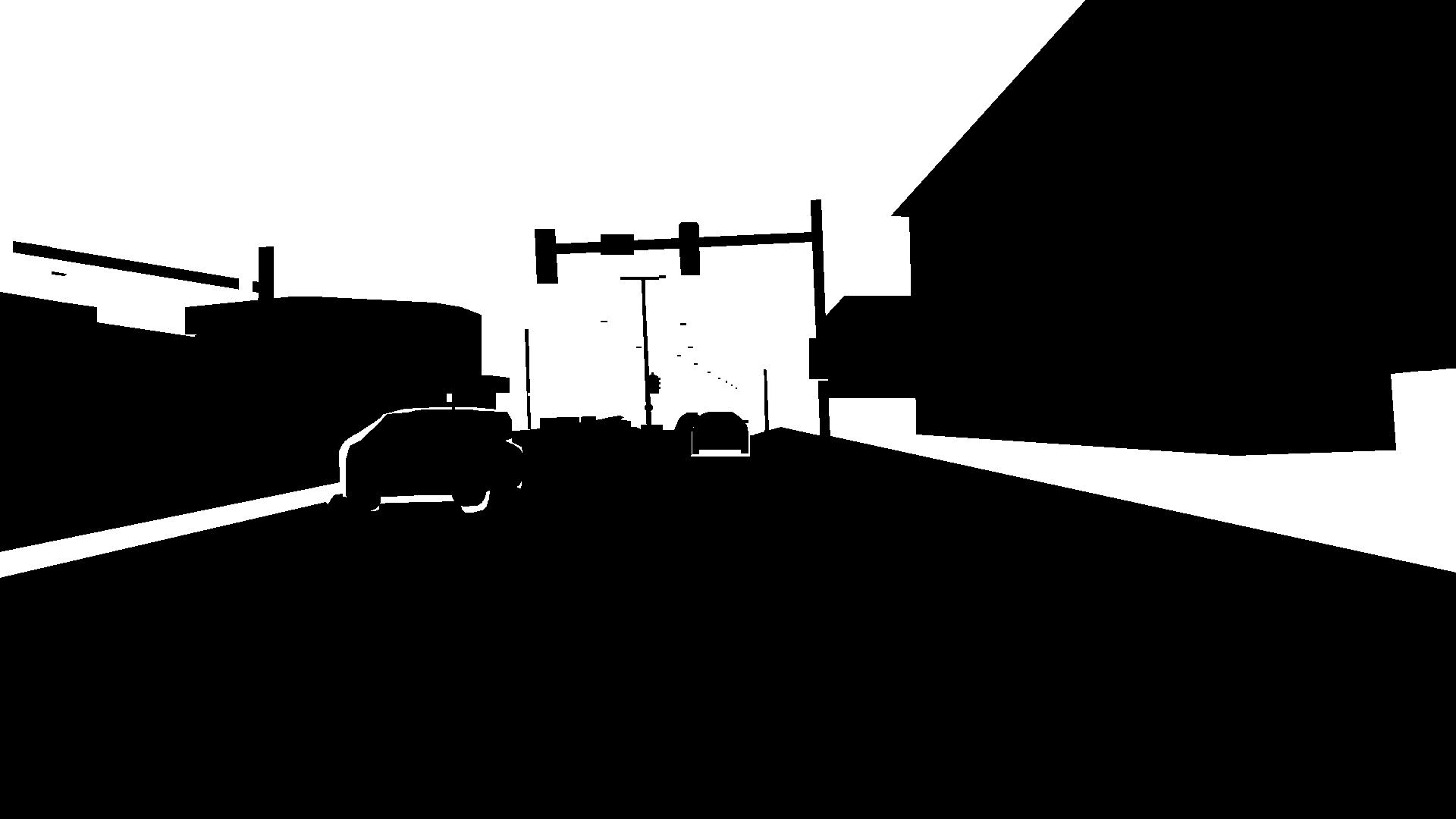}\\

% Clear Night 2
\includegraphics[width=0.14\textwidth]{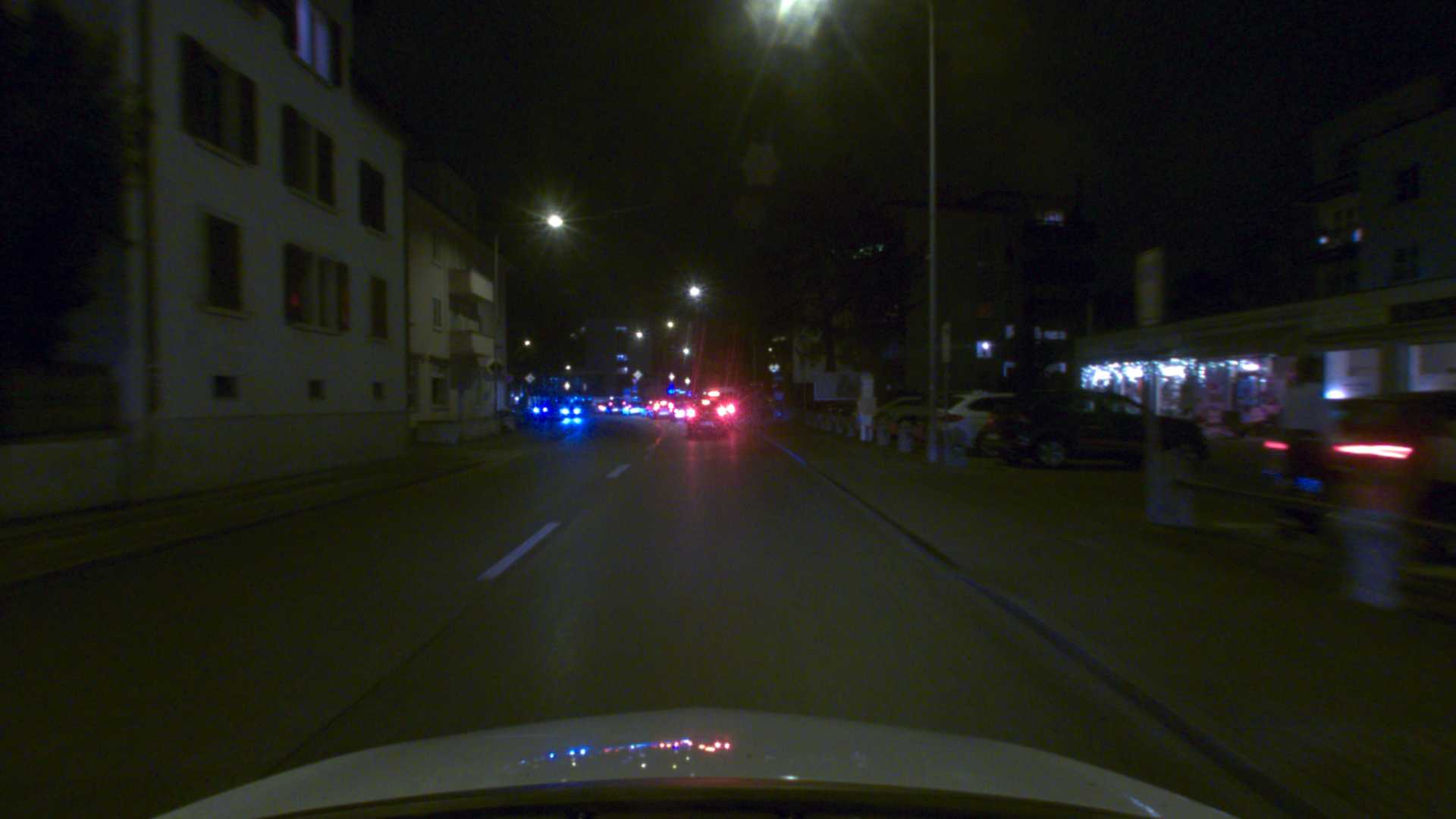} &
\includegraphics[width=0.14\textwidth]{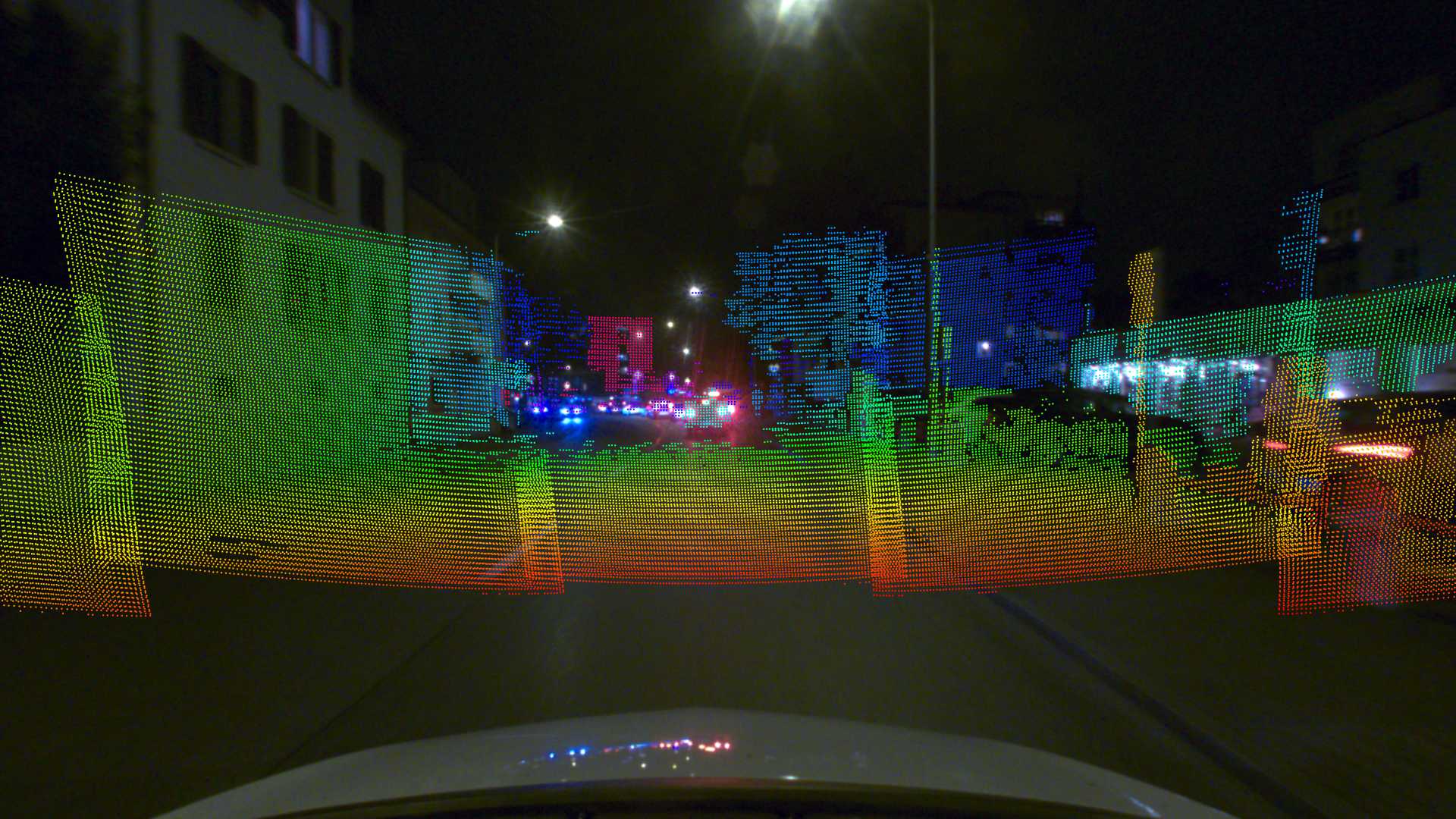} &
\includegraphics[width=0.14\textwidth]{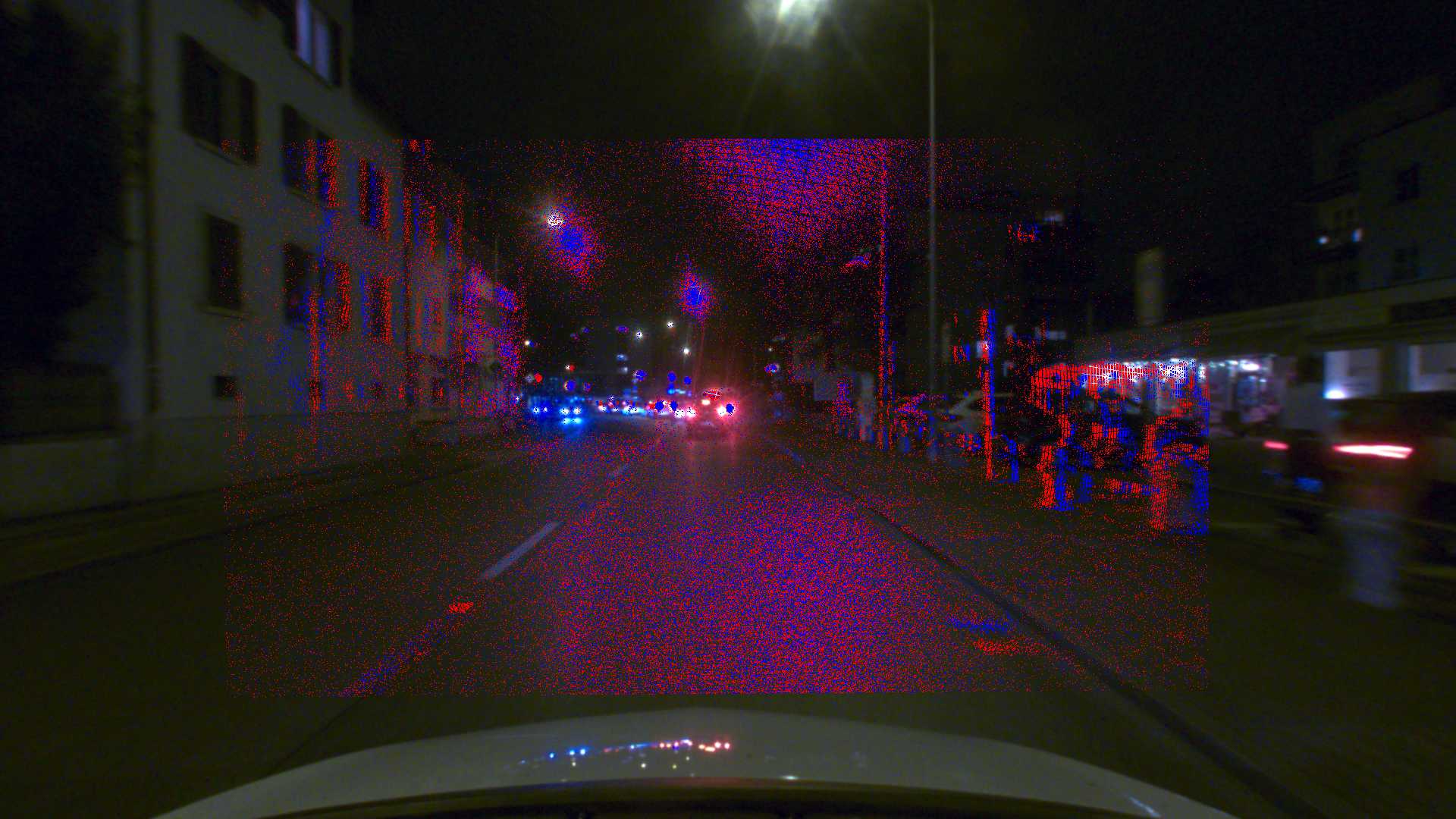} &
\includegraphics[angle=90, trim=0 6835 0 0, clip,width=0.14\textwidth]{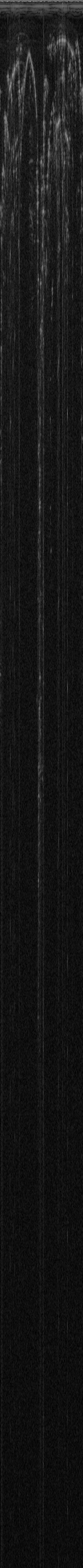}&
\includegraphics[width=0.14\textwidth]{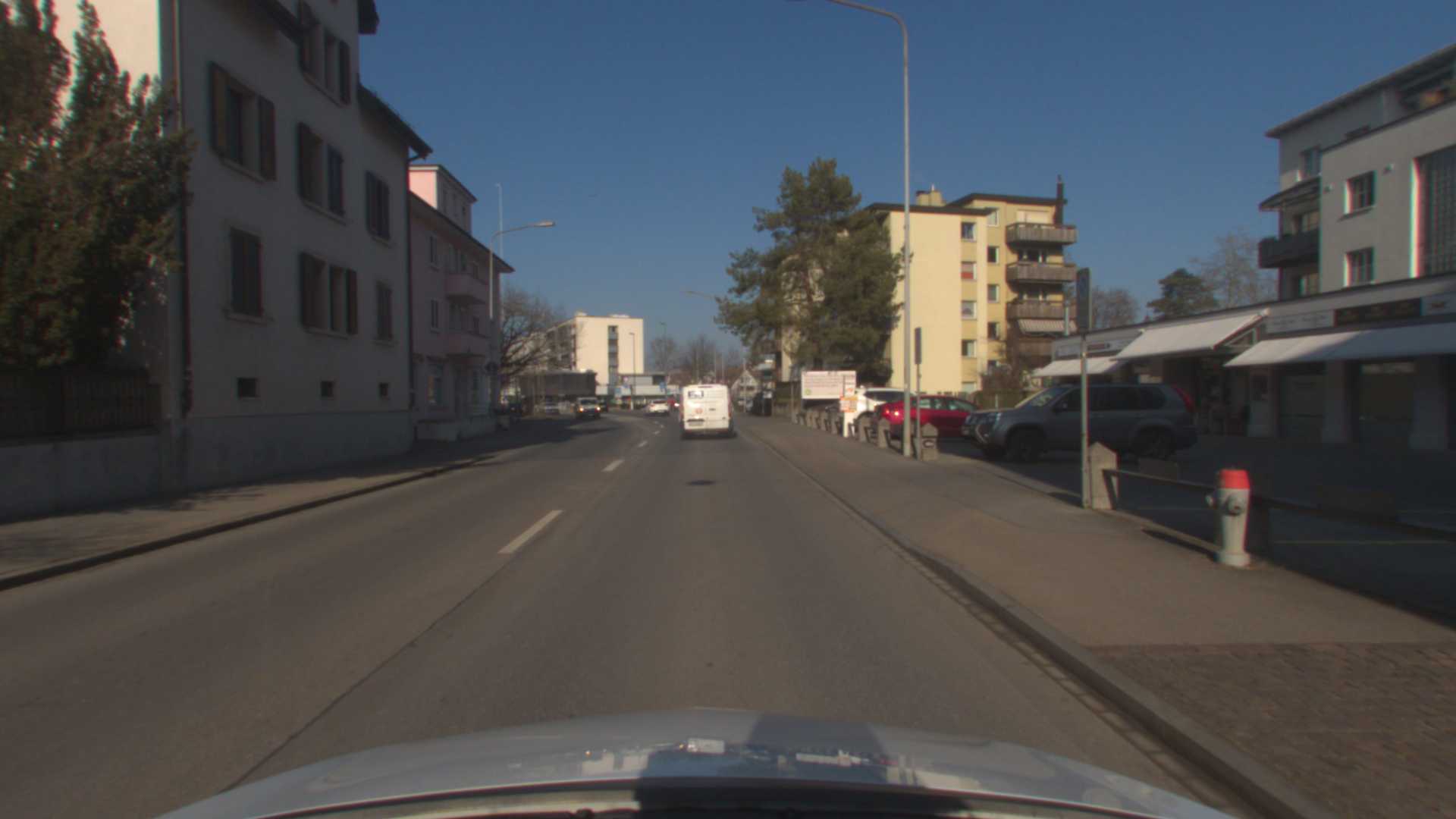} &
\includegraphics[width=0.14\textwidth]{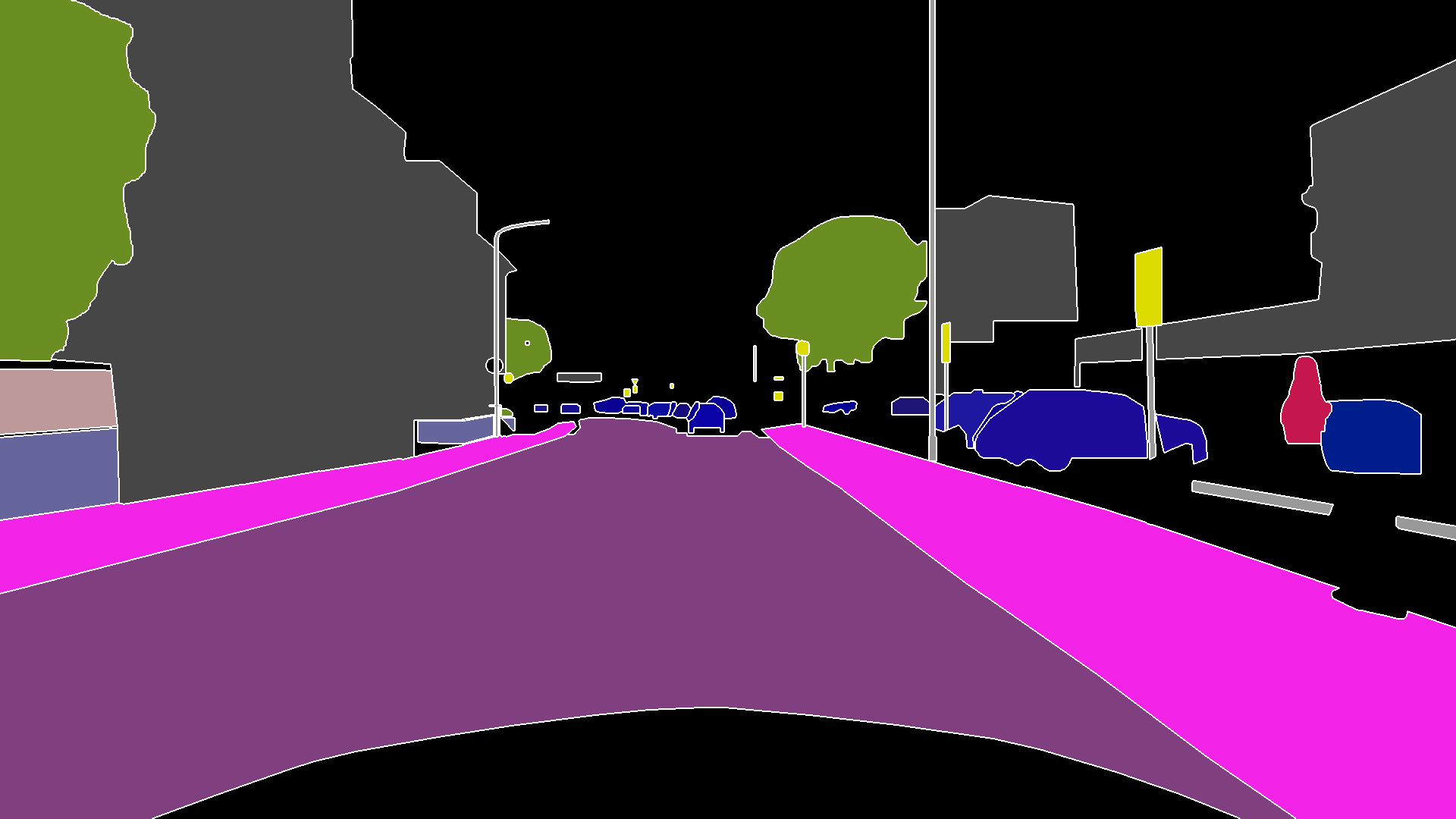} &
\includegraphics[width=0.14\textwidth]{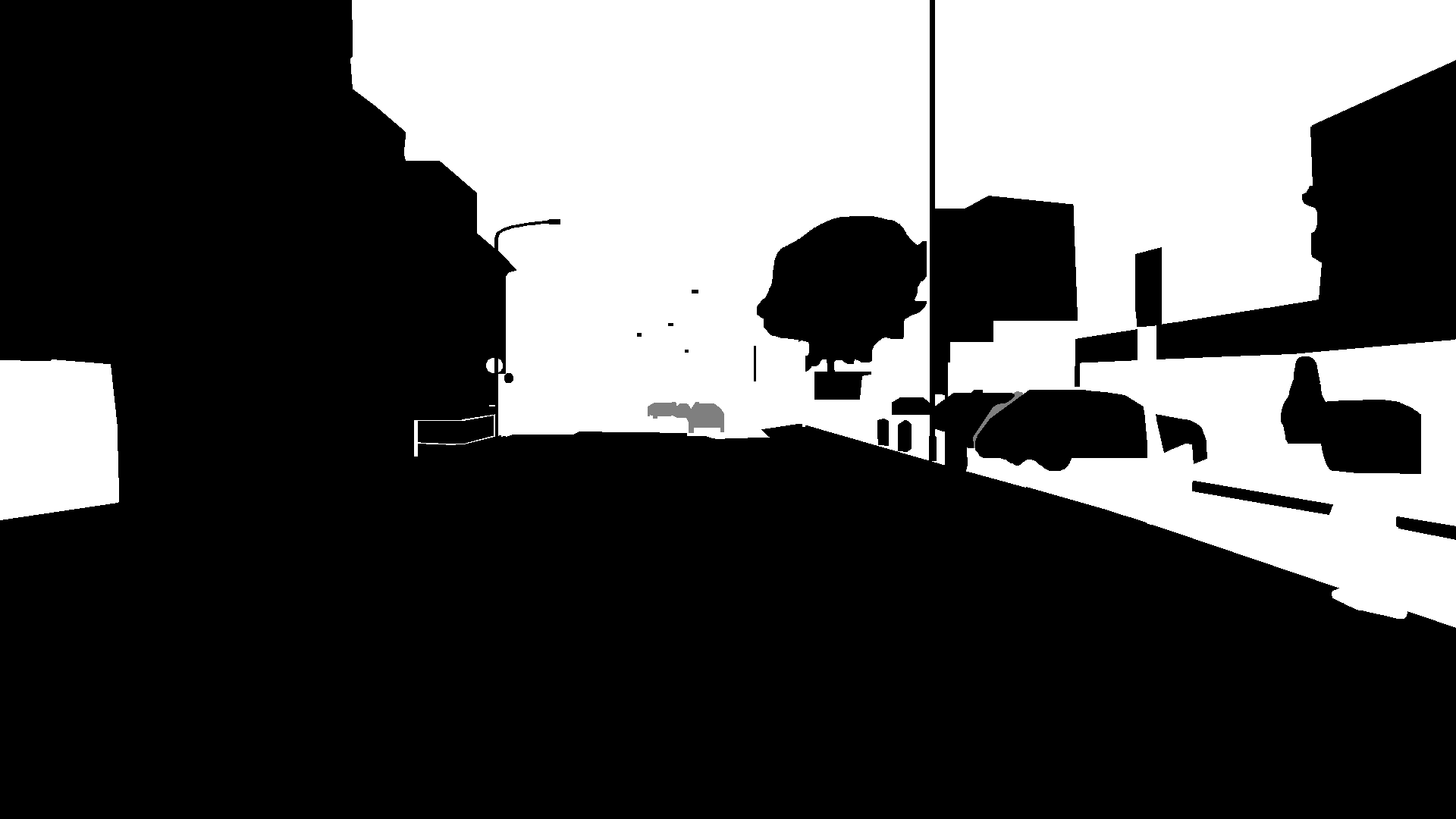}\\

% Fog Day 1
\includegraphics[width=0.14\textwidth]{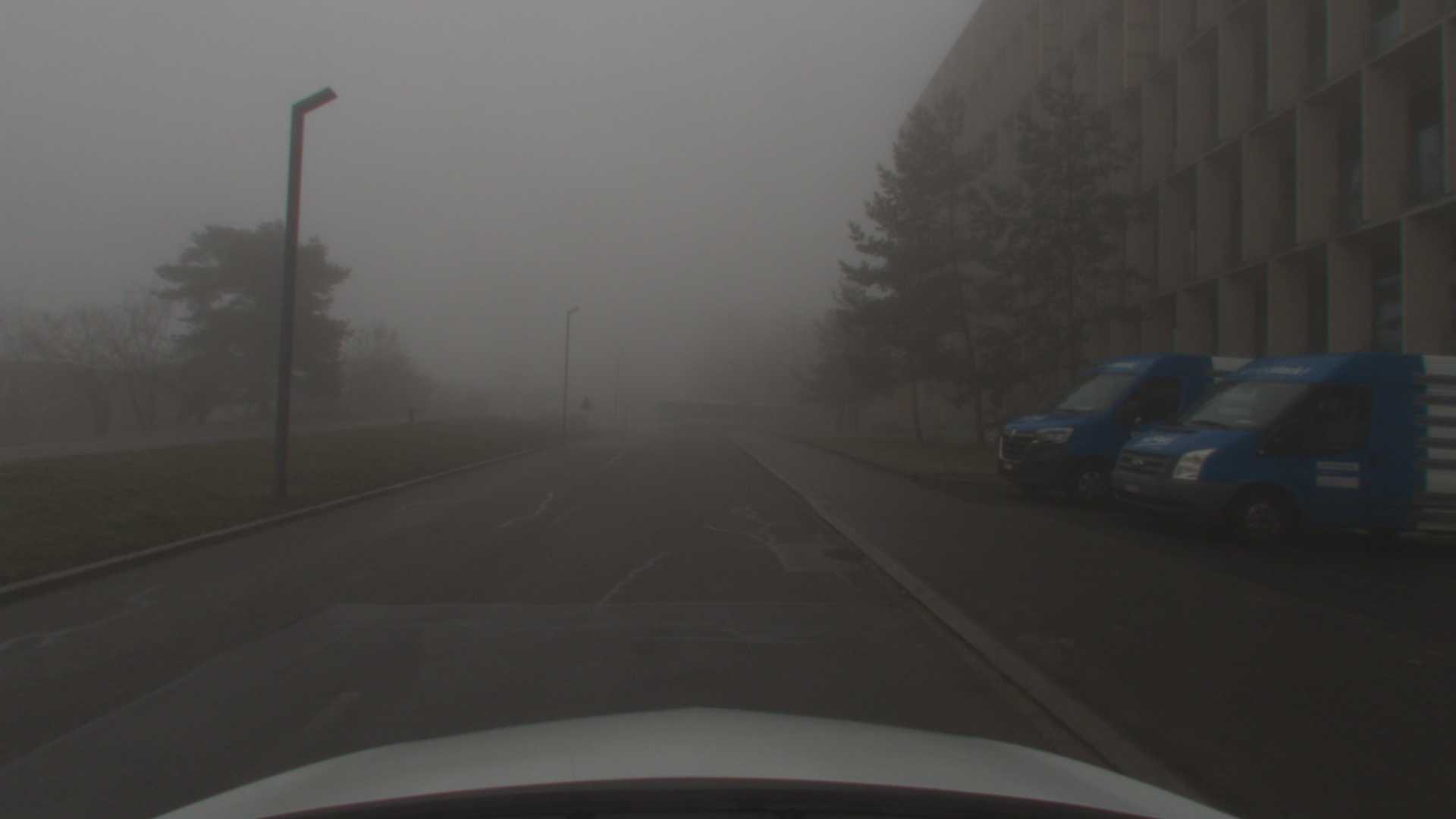} &
\includegraphics[width=0.14\textwidth]{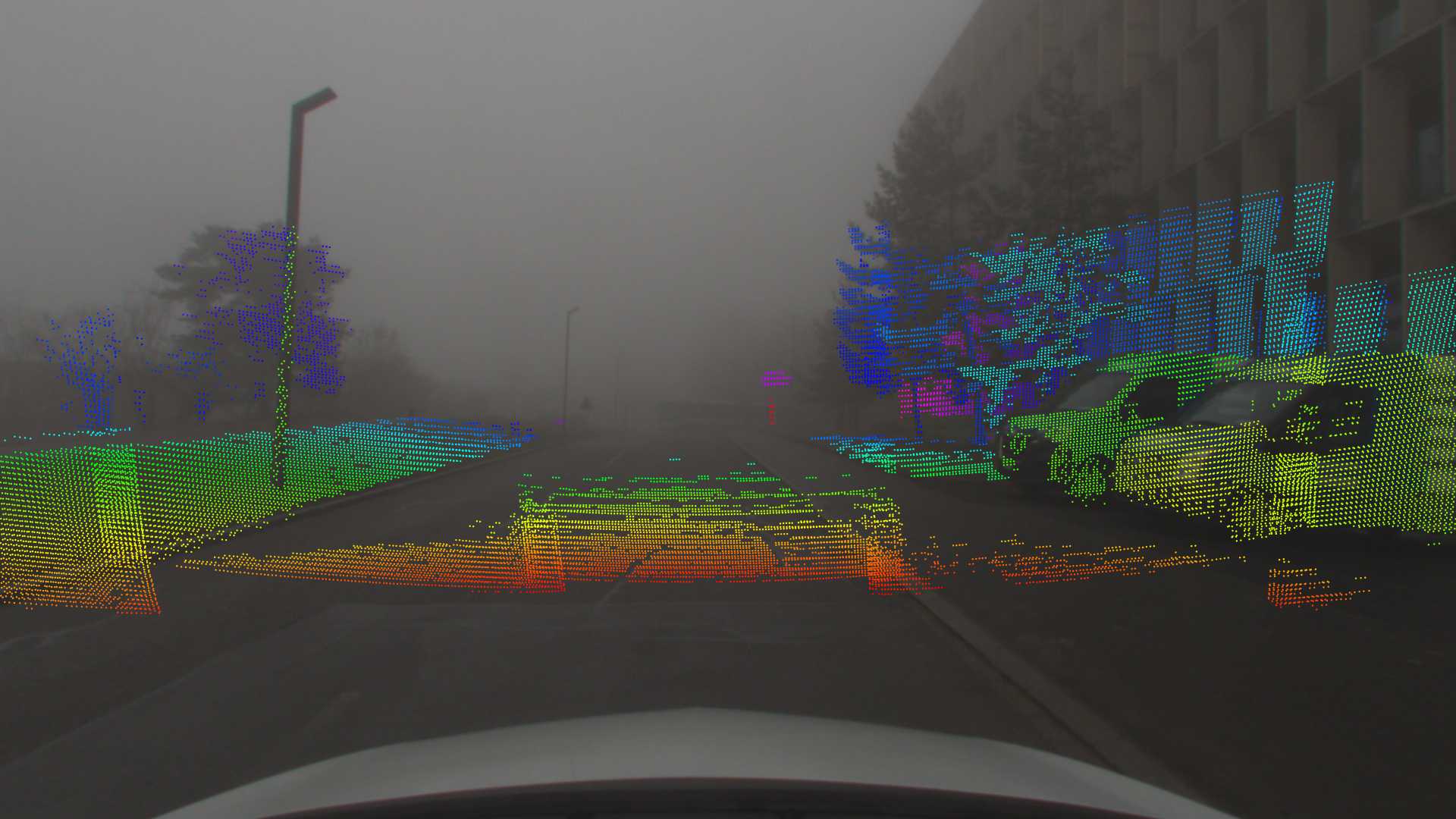} &
\includegraphics[width=0.14\textwidth]{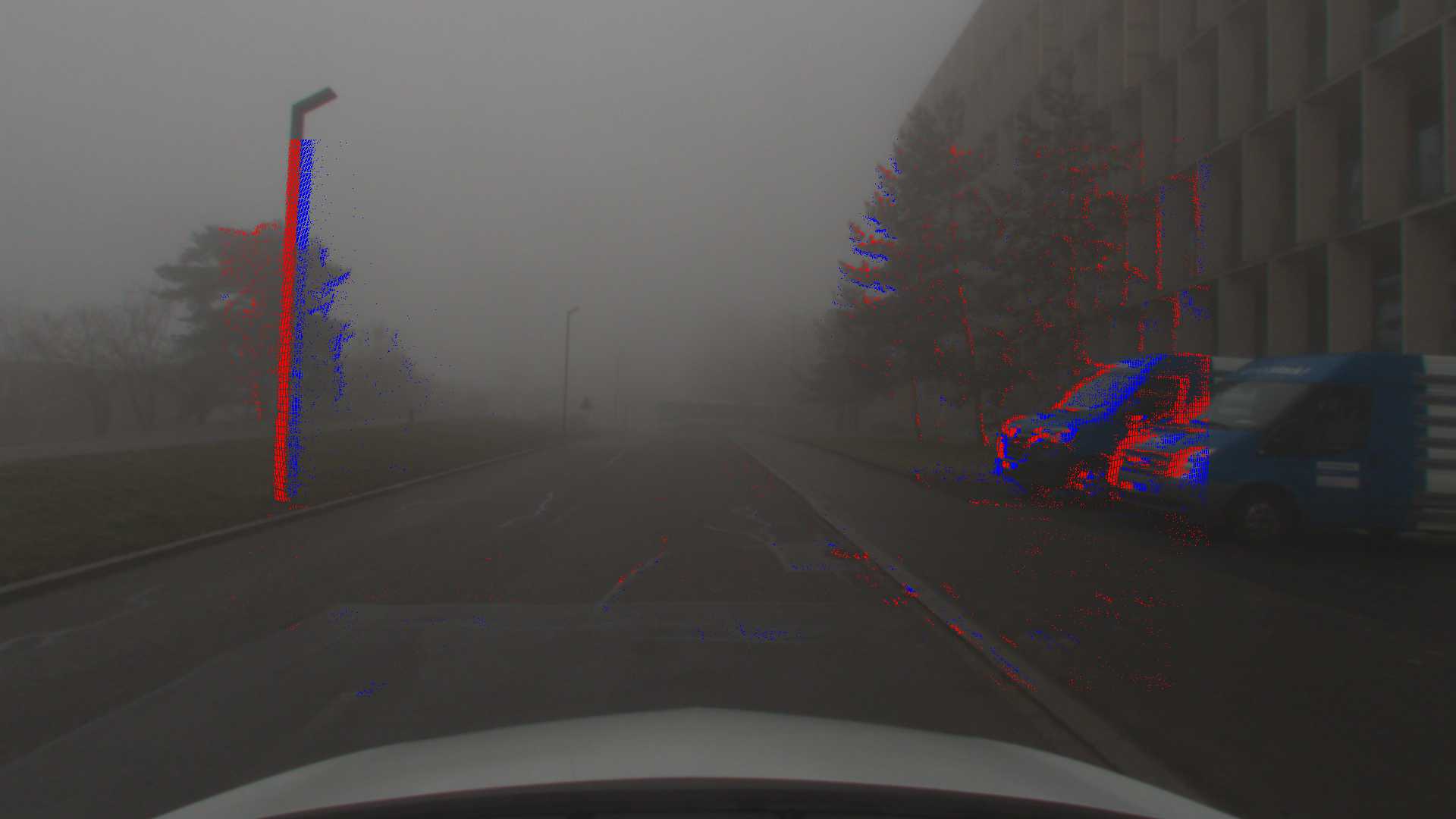} &
\includegraphics[angle=90, trim=0 6835 0 0, clip,width=0.14\textwidth]{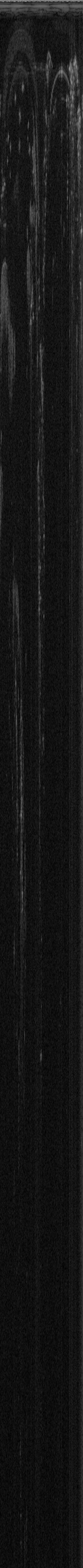}&
\includegraphics[width=0.14\textwidth]{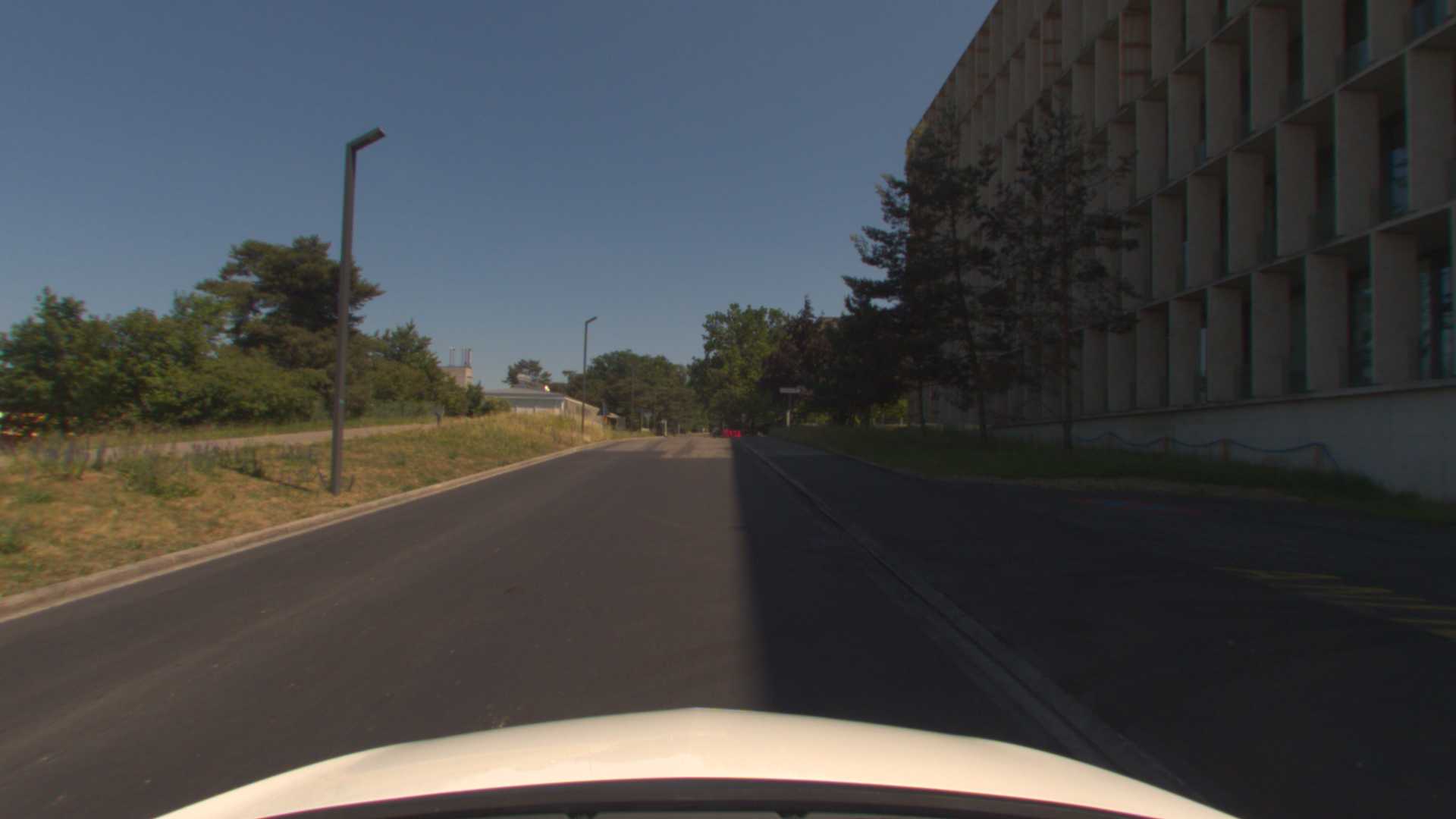} &
\includegraphics[width=0.14\textwidth]{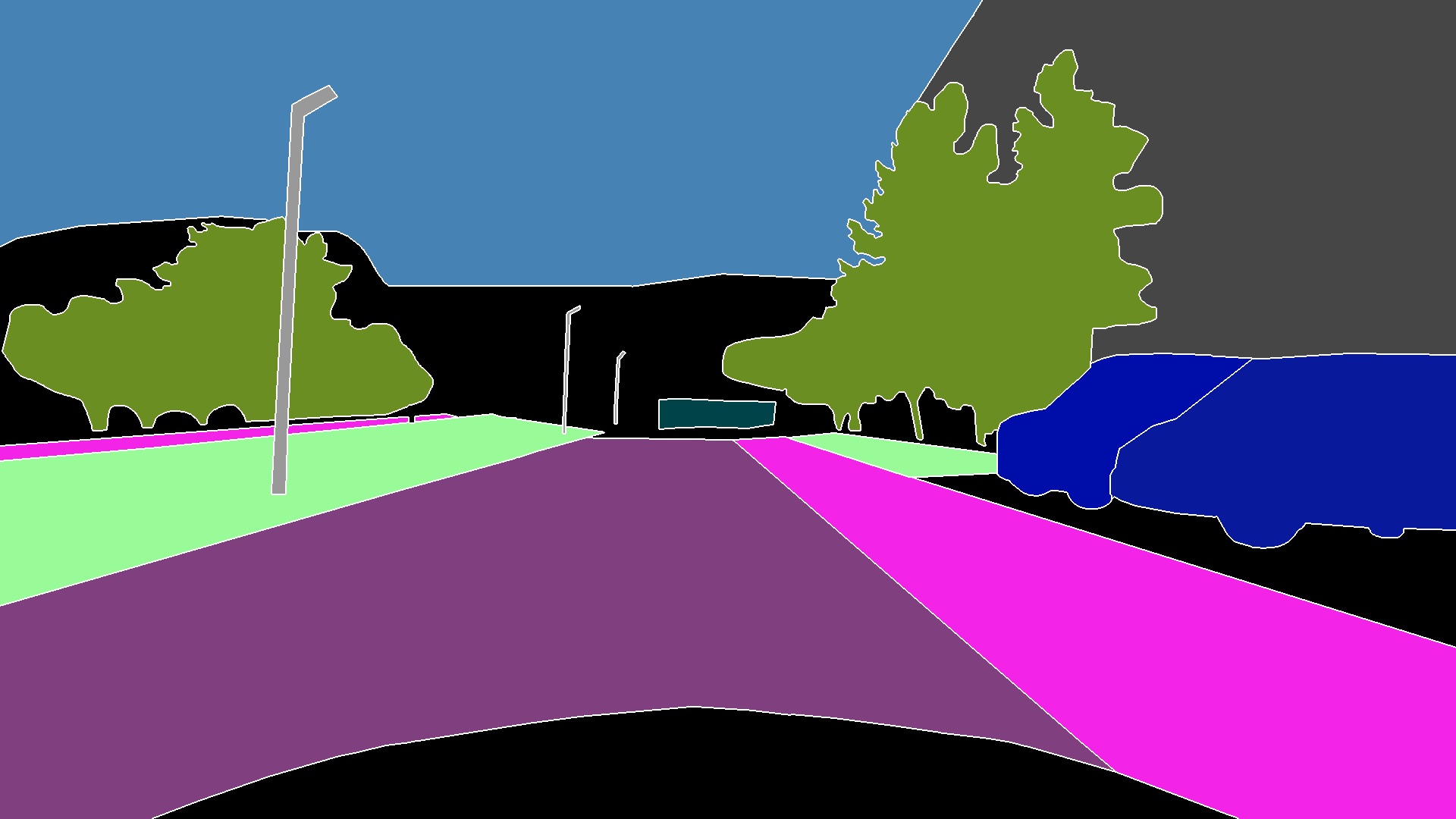} &
\includegraphics[width=0.14\textwidth]{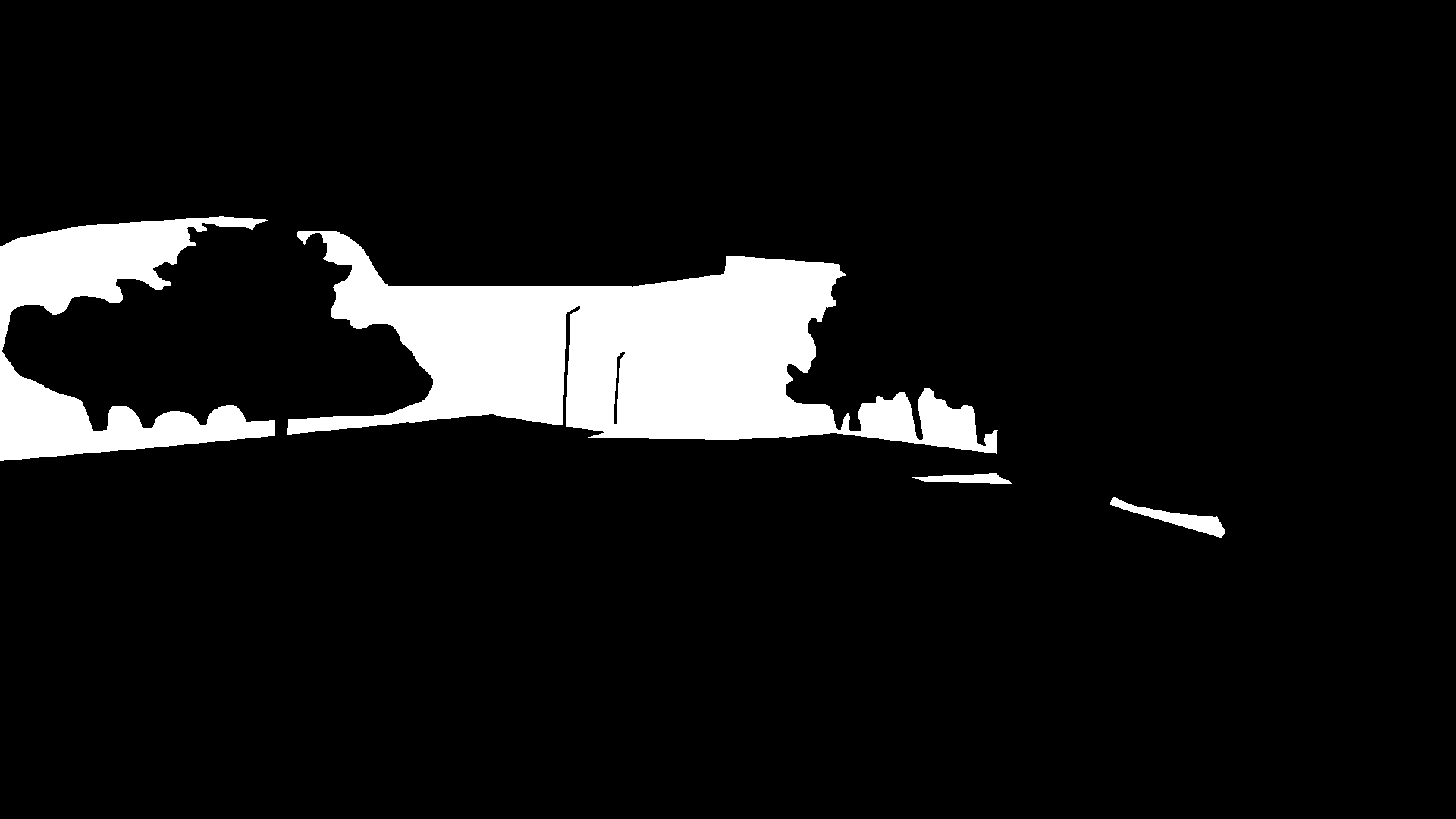}\\

% Fog Day 2
\includegraphics[width=0.14\textwidth]{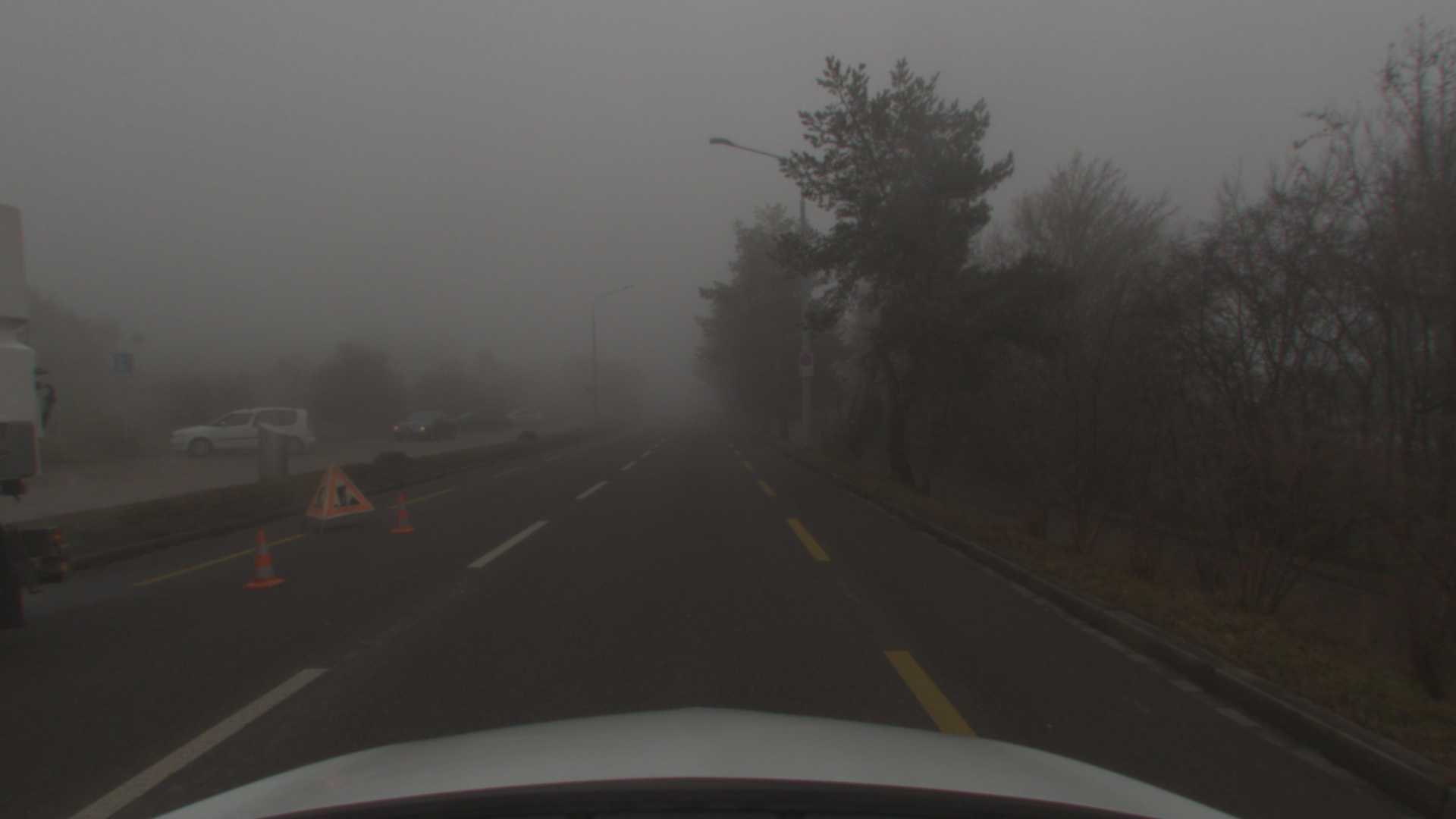} &
\includegraphics[width=0.14\textwidth]{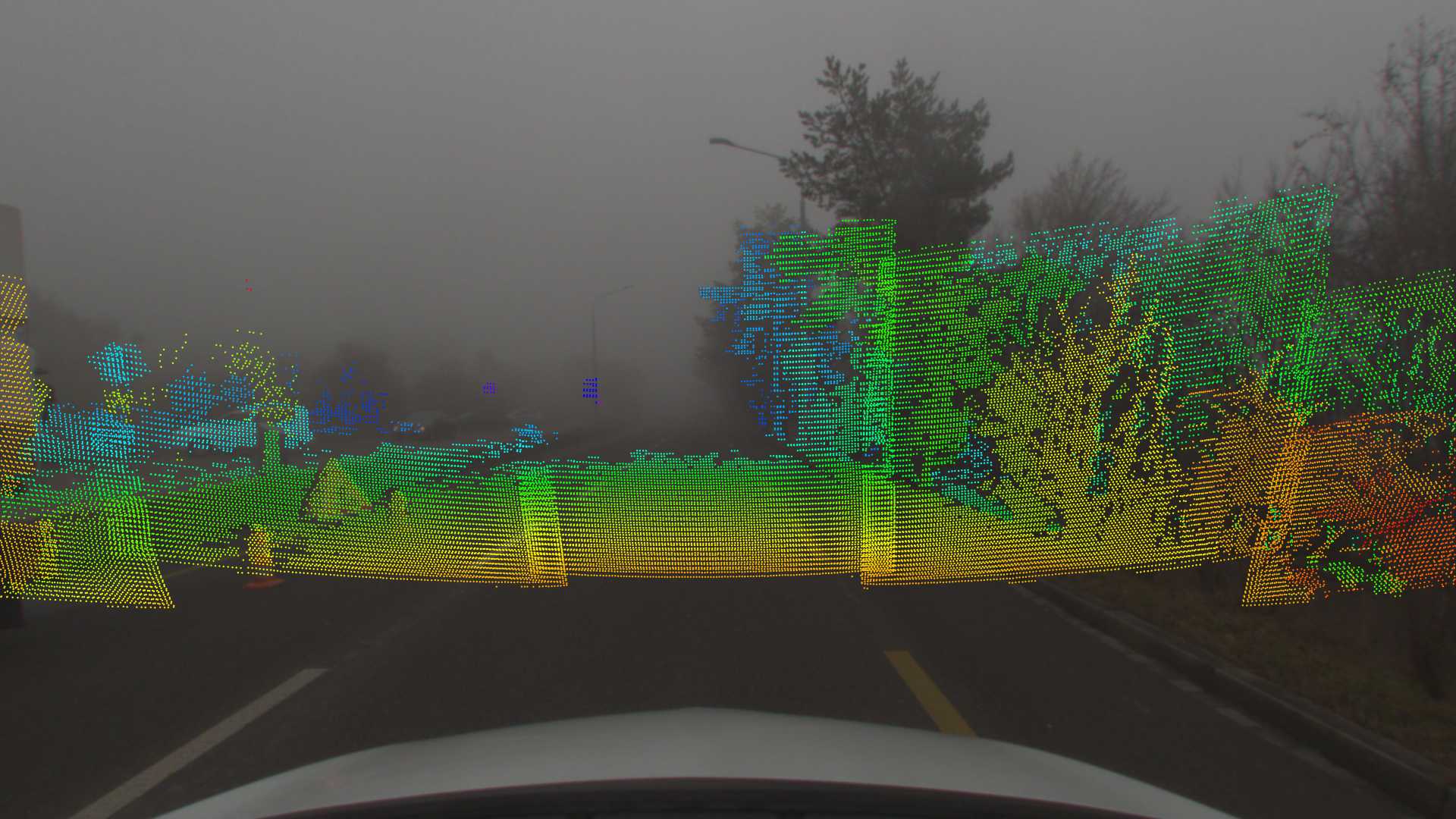} &
\includegraphics[width=0.14\textwidth]{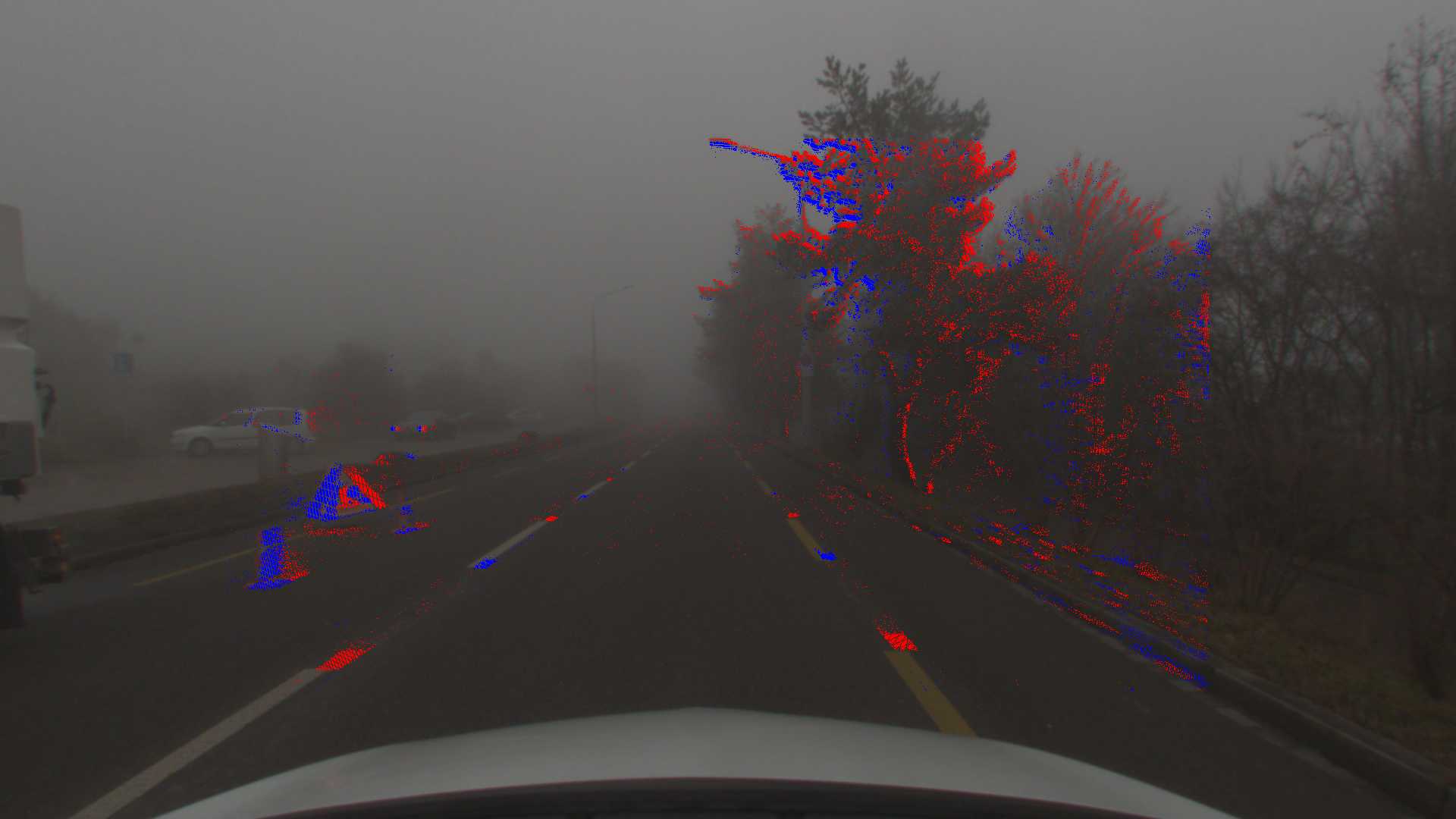} &
\includegraphics[angle=90, trim=0 6835 0 0, clip,width=0.14\textwidth]{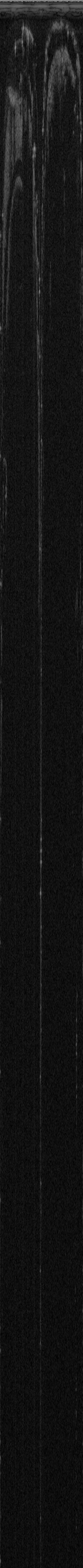}&
\includegraphics[width=0.14\textwidth]{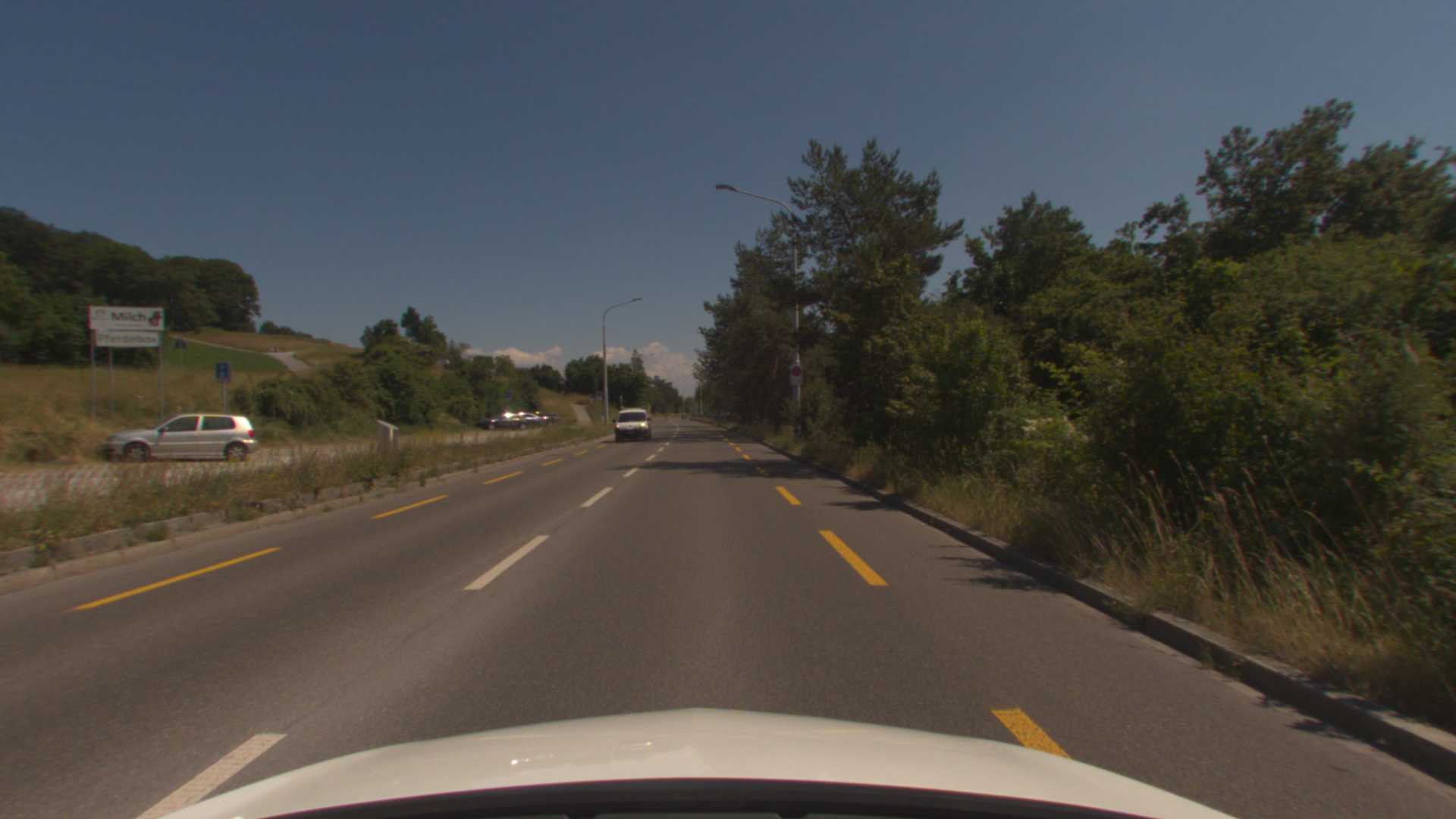} &
\includegraphics[width=0.14\textwidth]{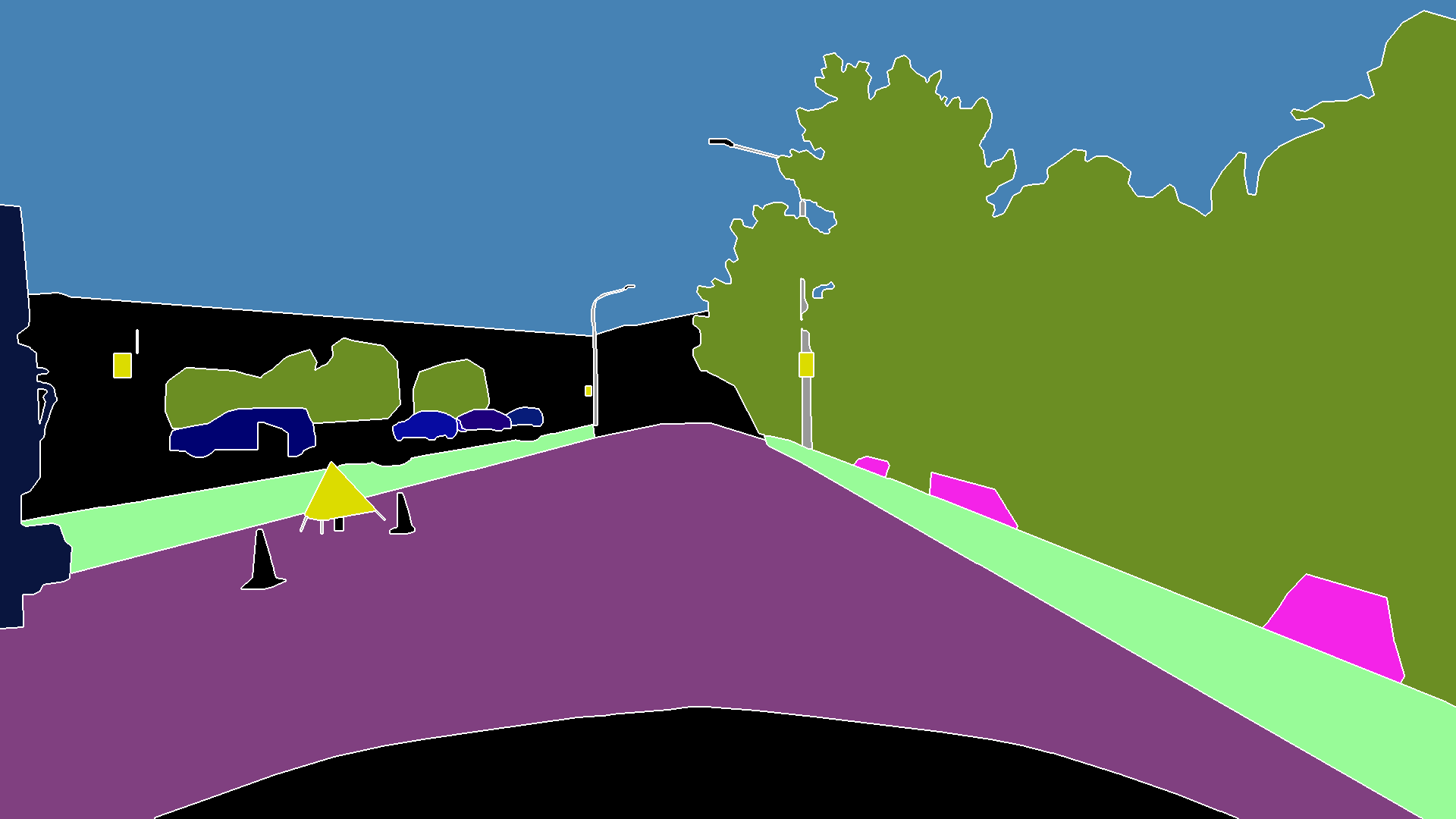} &
\includegraphics[width=0.14\textwidth]{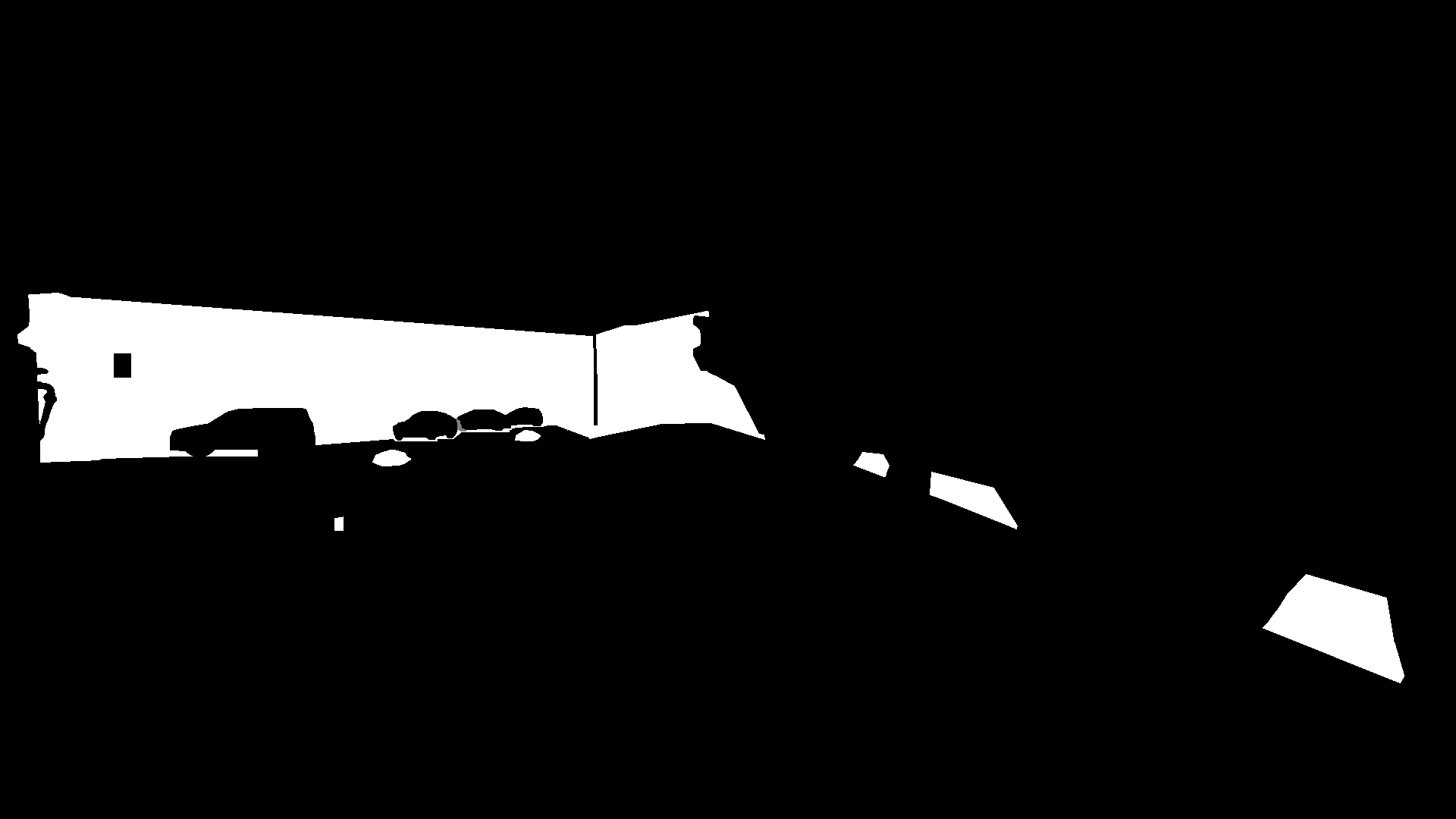}\\

% Fog Night 1
\includegraphics[width=0.14\textwidth]{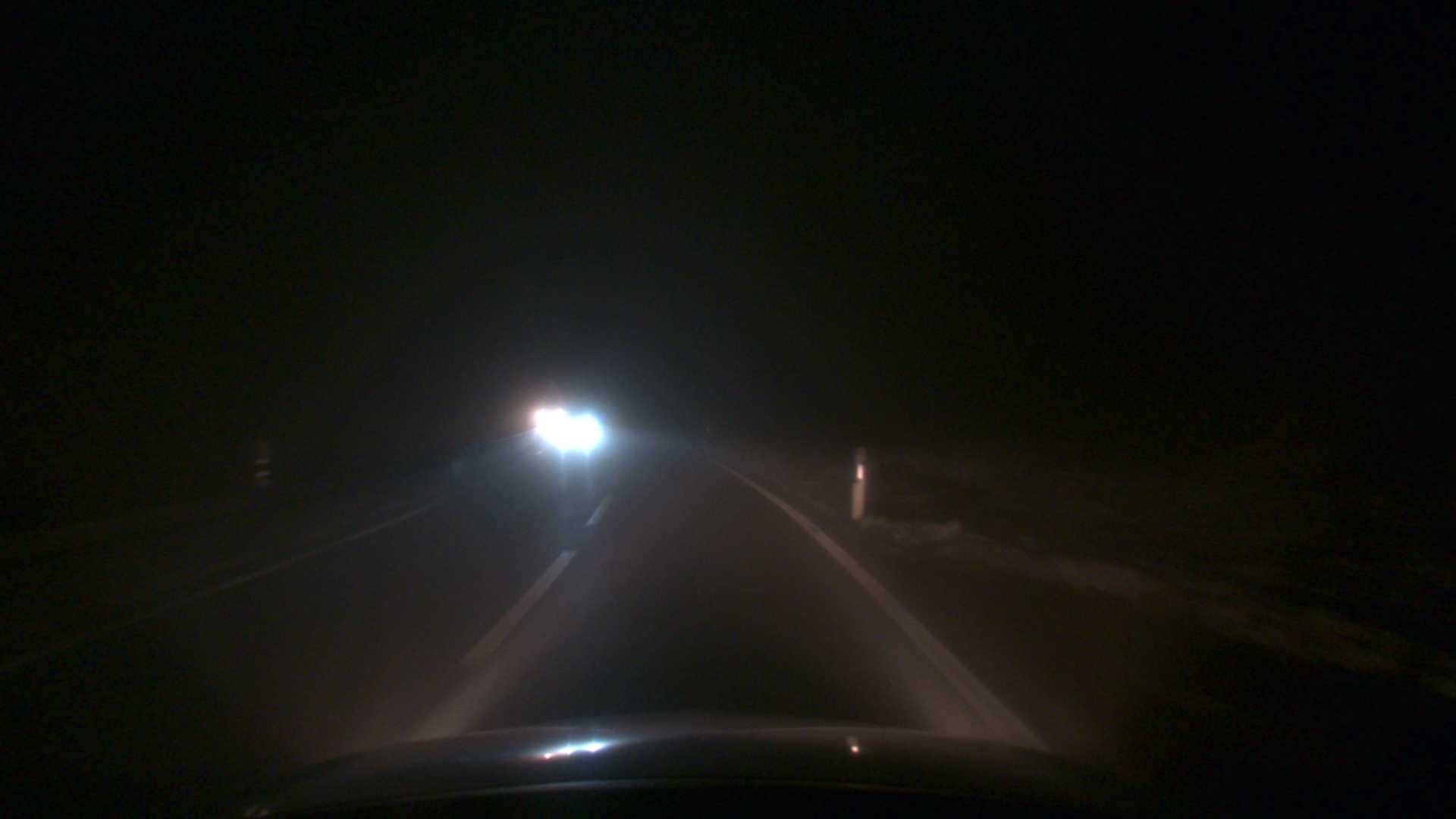} &
\includegraphics[width=0.14\textwidth]{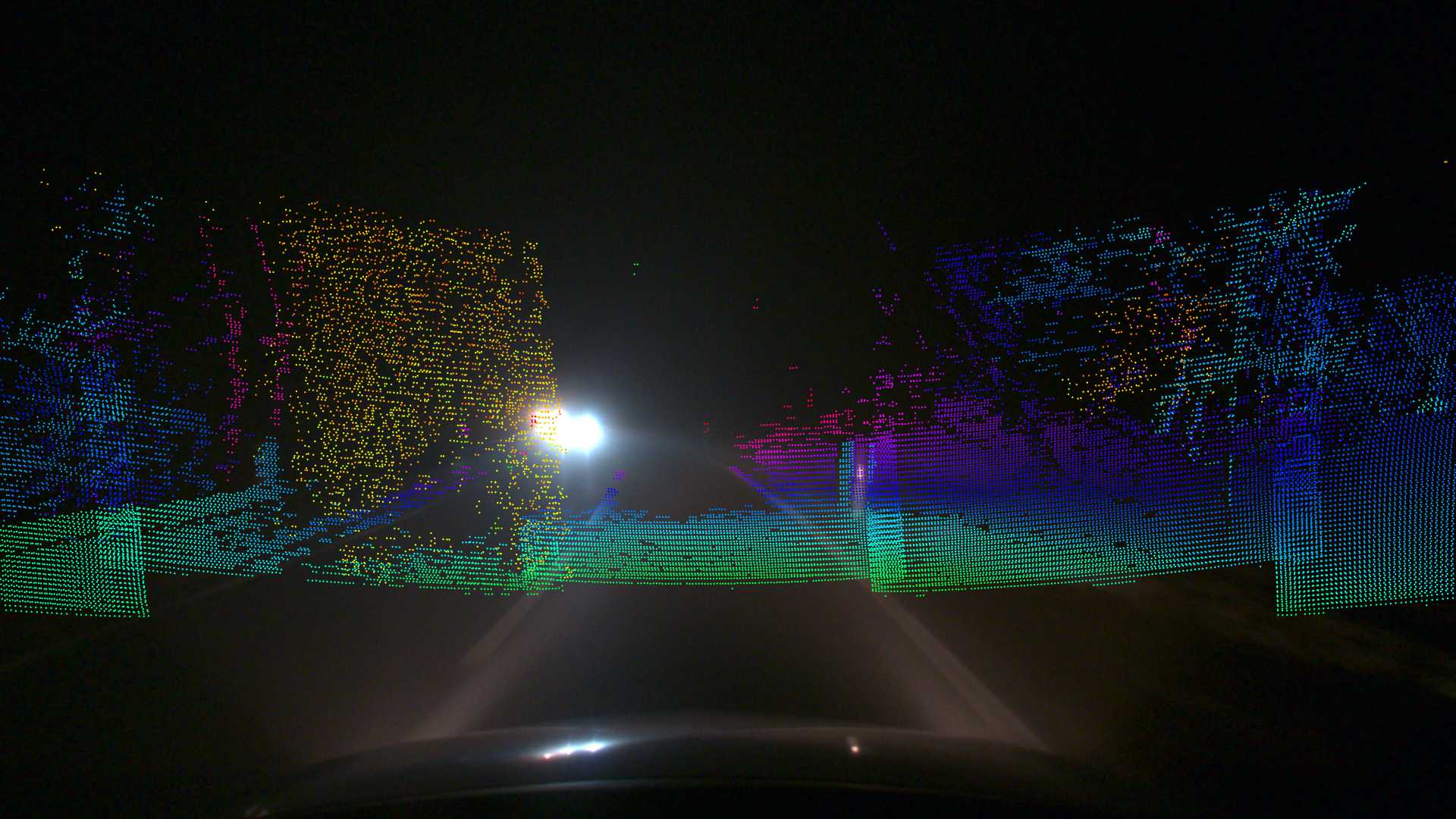} &
\includegraphics[width=0.14\textwidth]{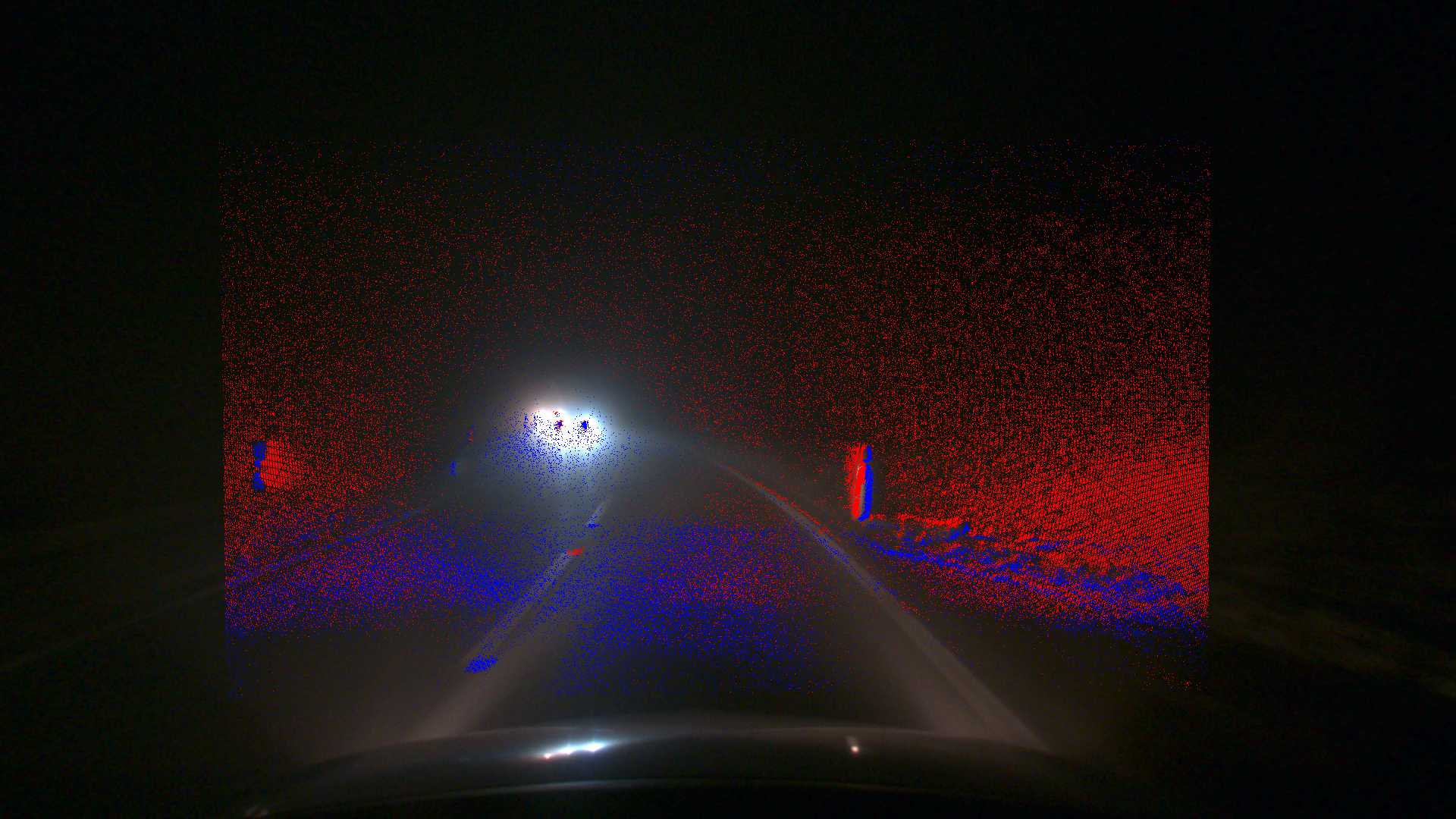} &
\includegraphics[angle=90, trim=0 6835 0 0, clip,width=0.14\textwidth]{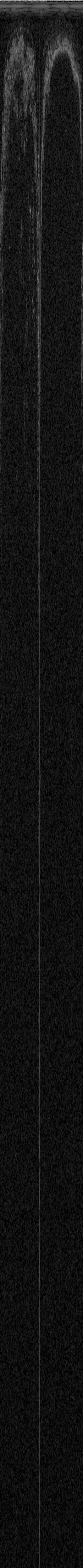}&
\includegraphics[width=0.14\textwidth]{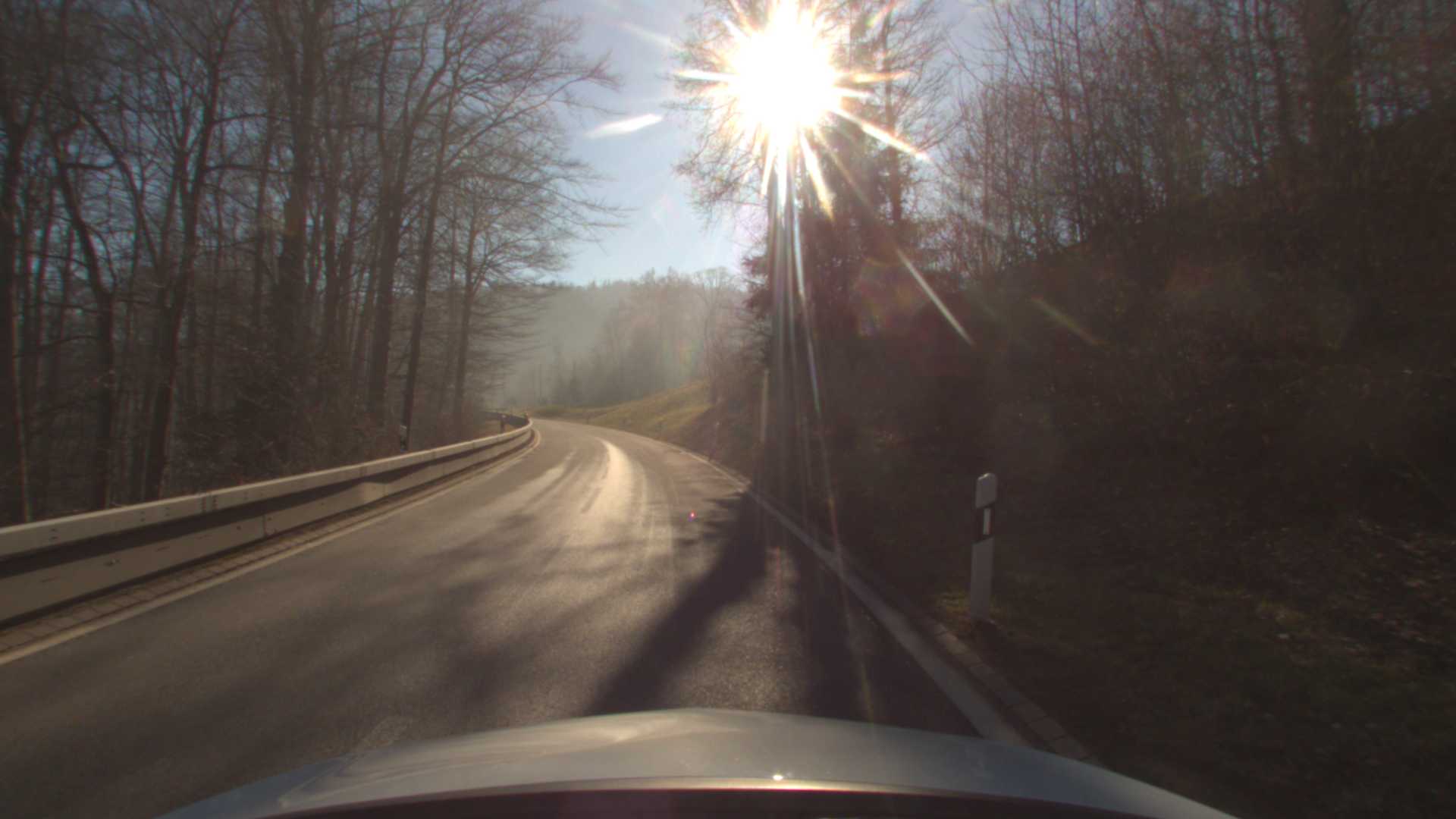} &
\includegraphics[width=0.14\textwidth]{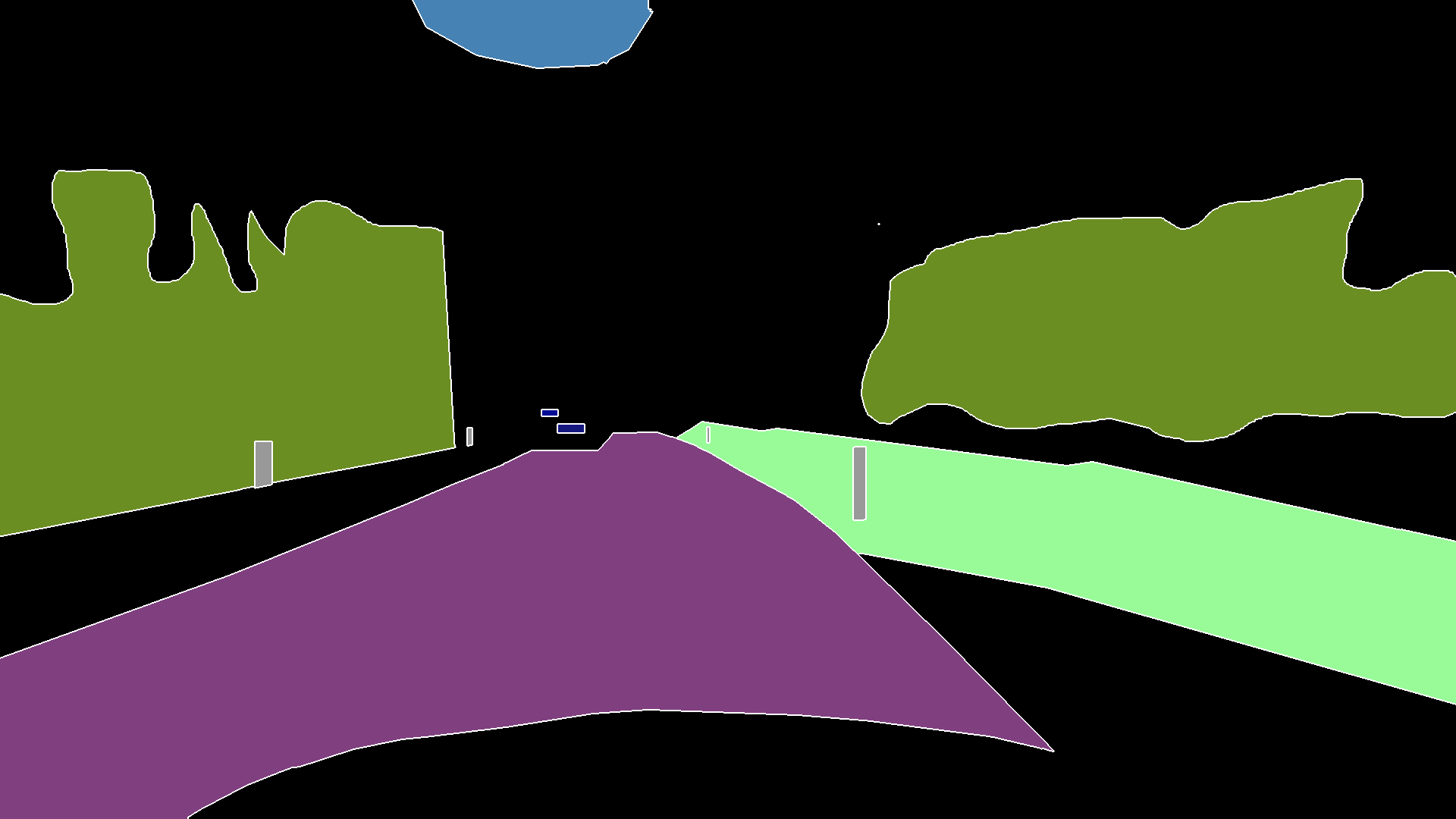} &
\includegraphics[width=0.14\textwidth]{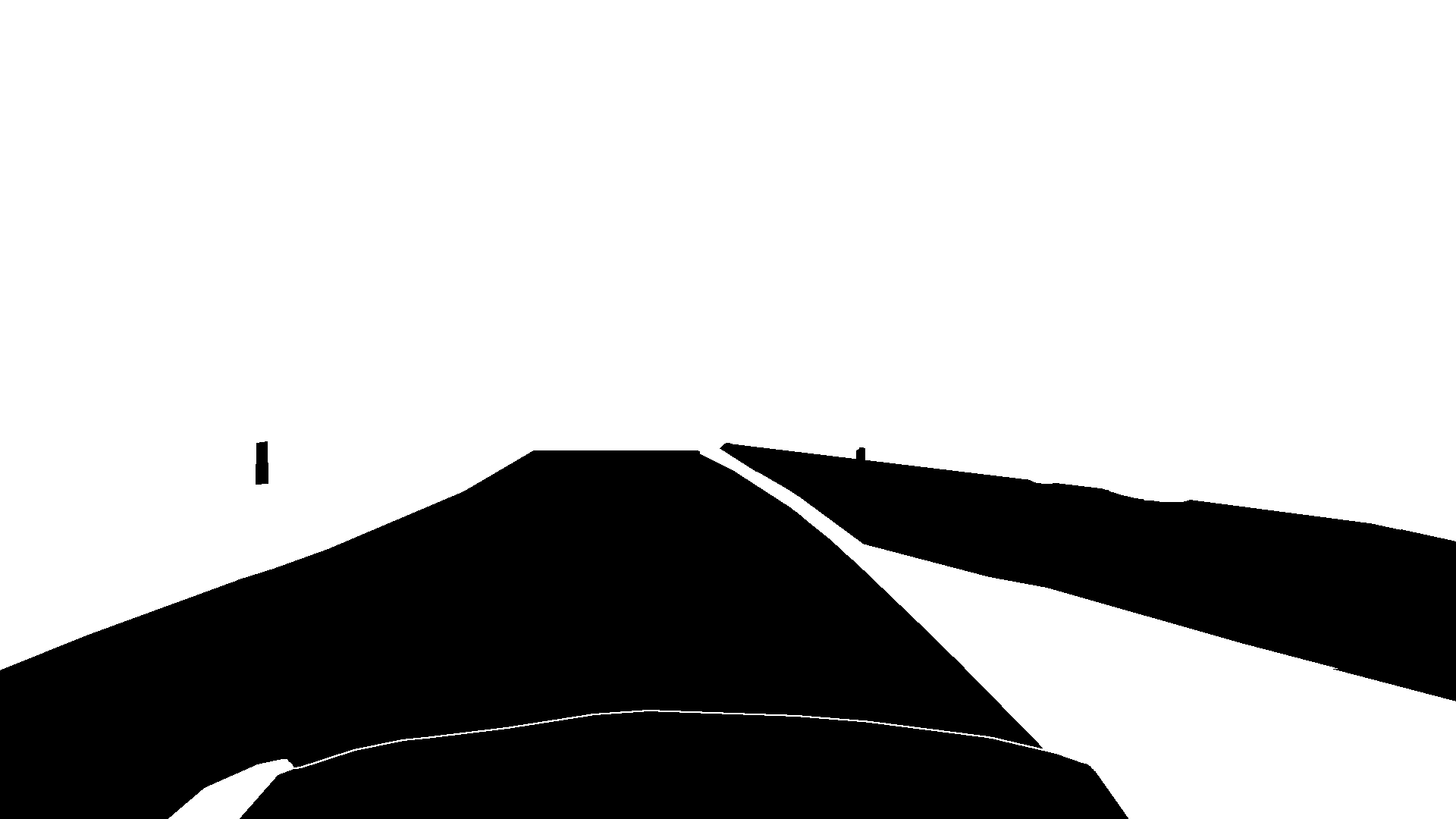}\\

% Fog Night 2
\includegraphics[width=0.14\textwidth]{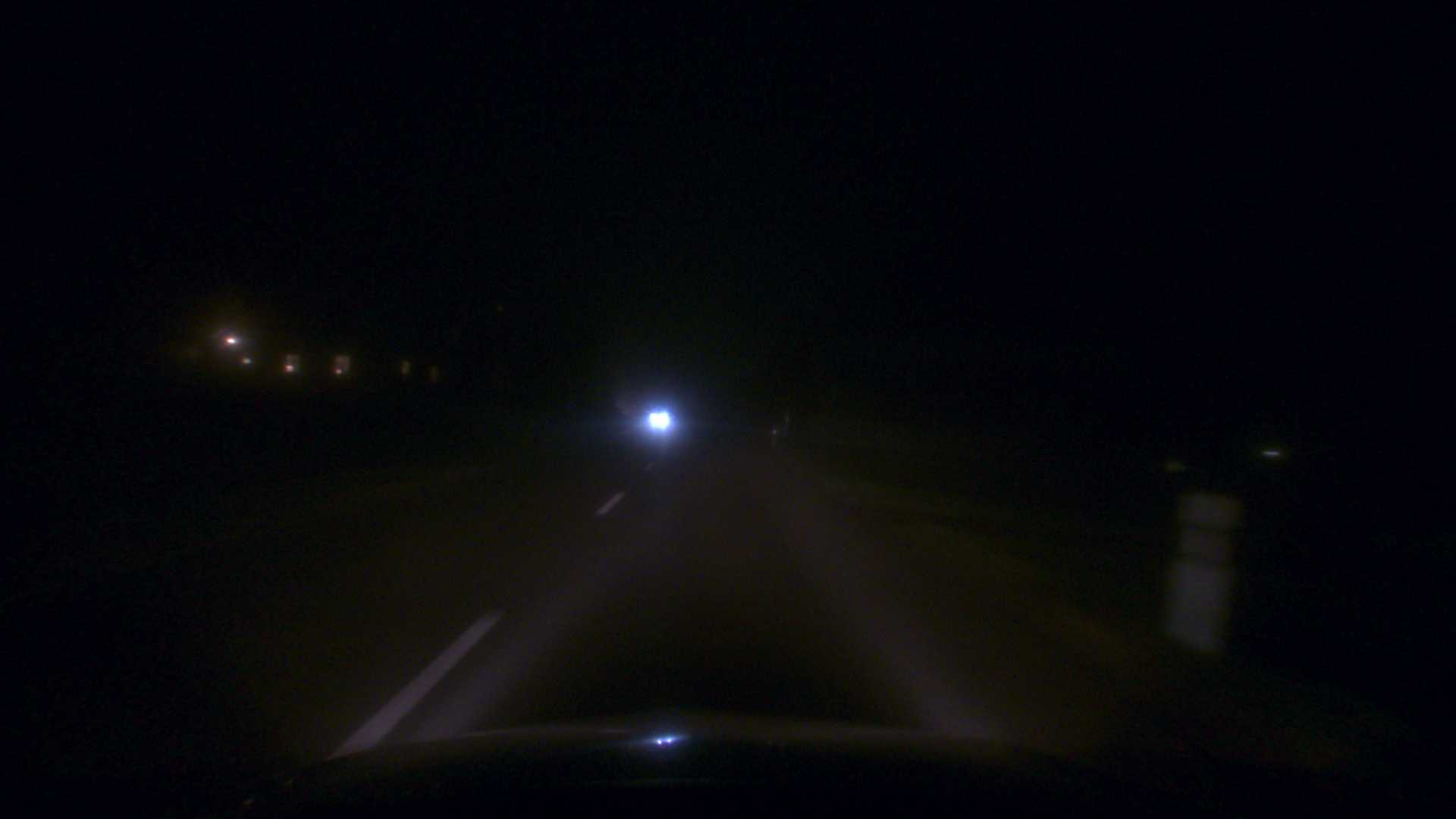} &
\includegraphics[width=0.14\textwidth]{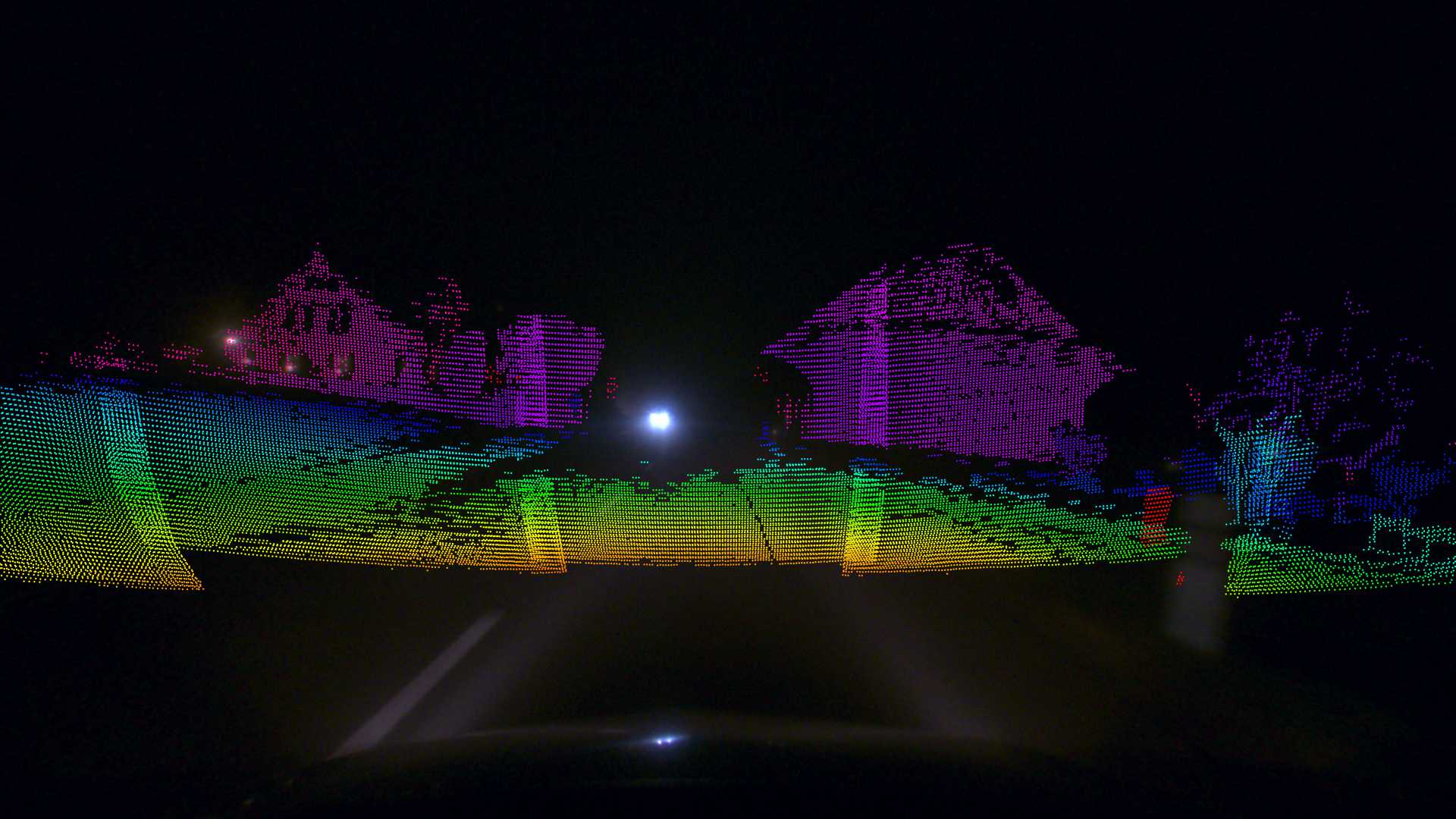} &
\includegraphics[width=0.14\textwidth]{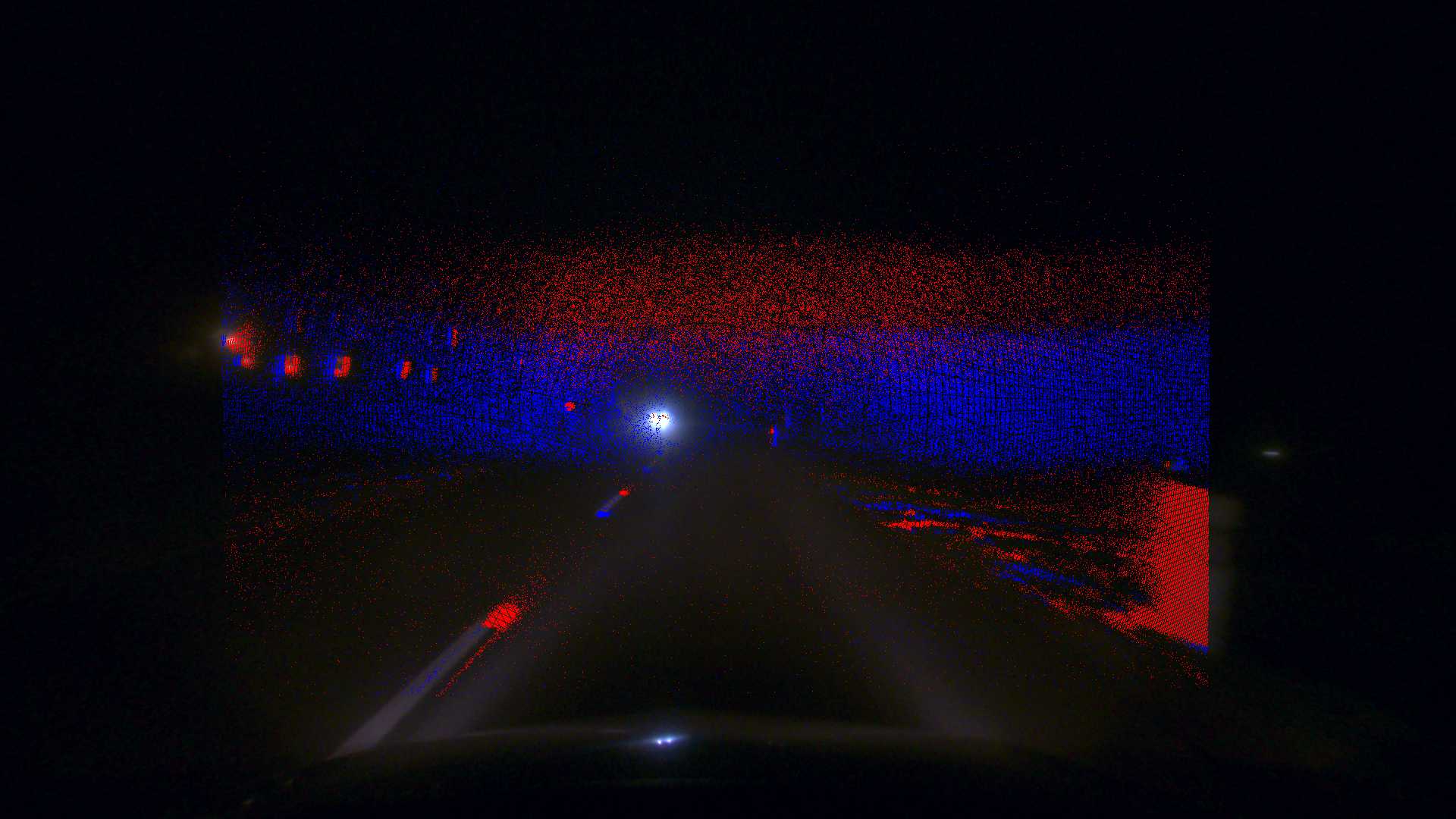} &
\includegraphics[angle=90, trim=0 6835 0 0, clip,width=0.14\textwidth]{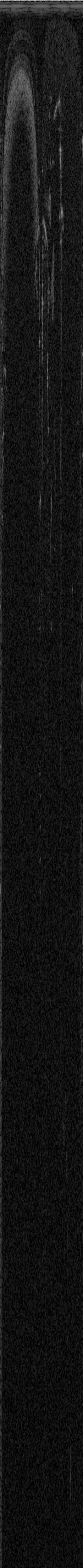}&
\includegraphics[width=0.14\textwidth]{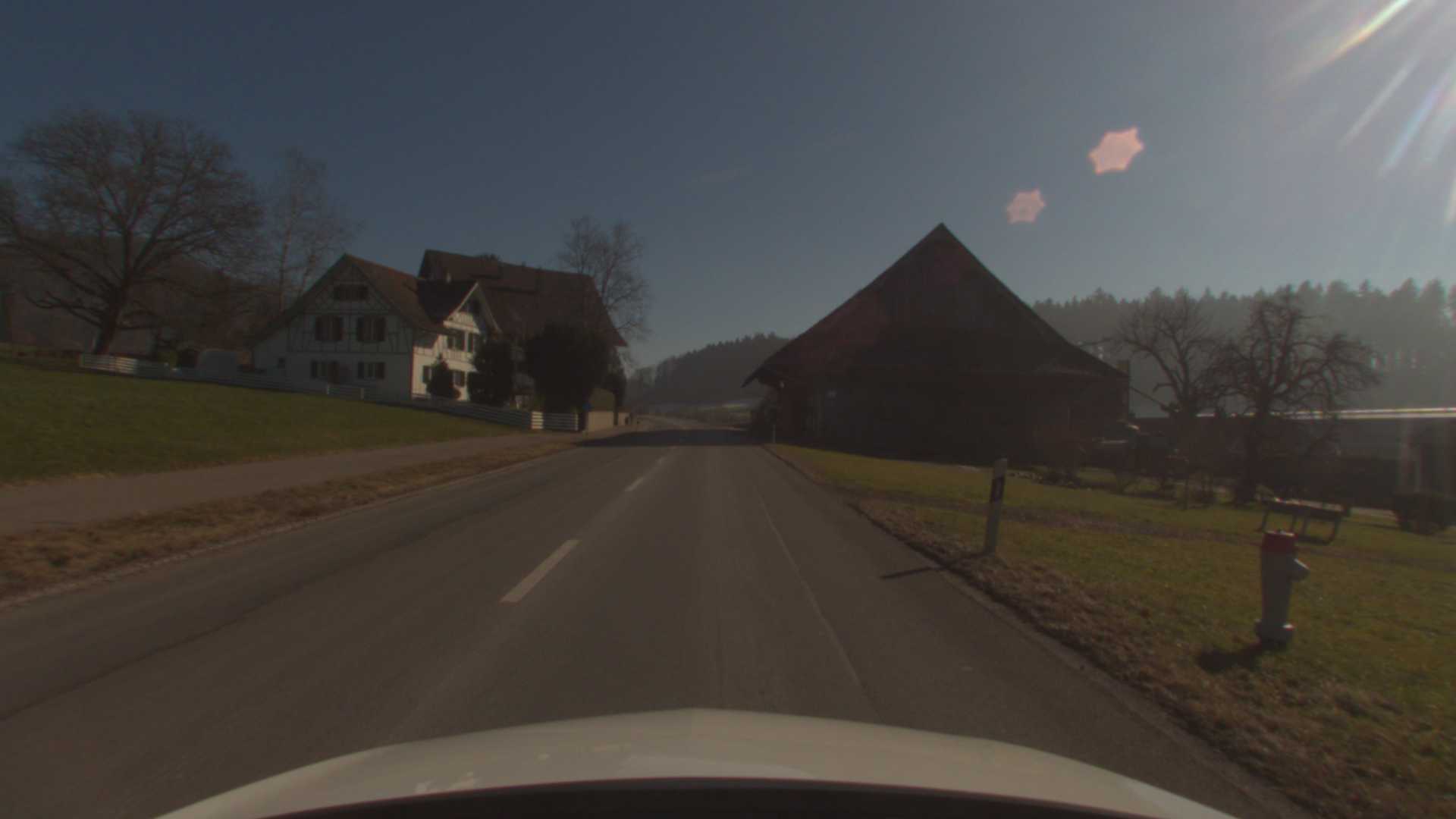} &
\includegraphics[width=0.14\textwidth]{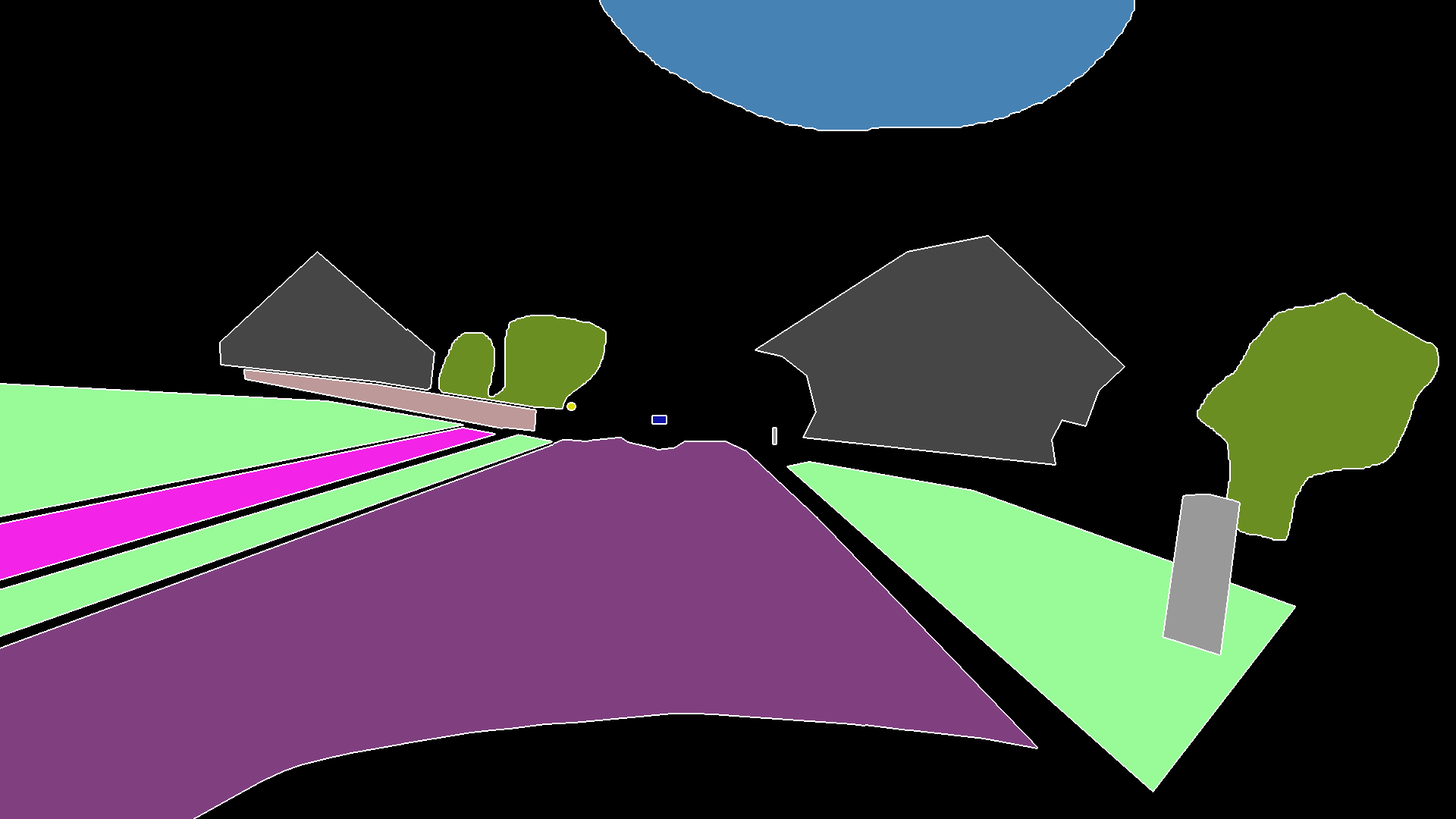} &
\includegraphics[width=0.14\textwidth]{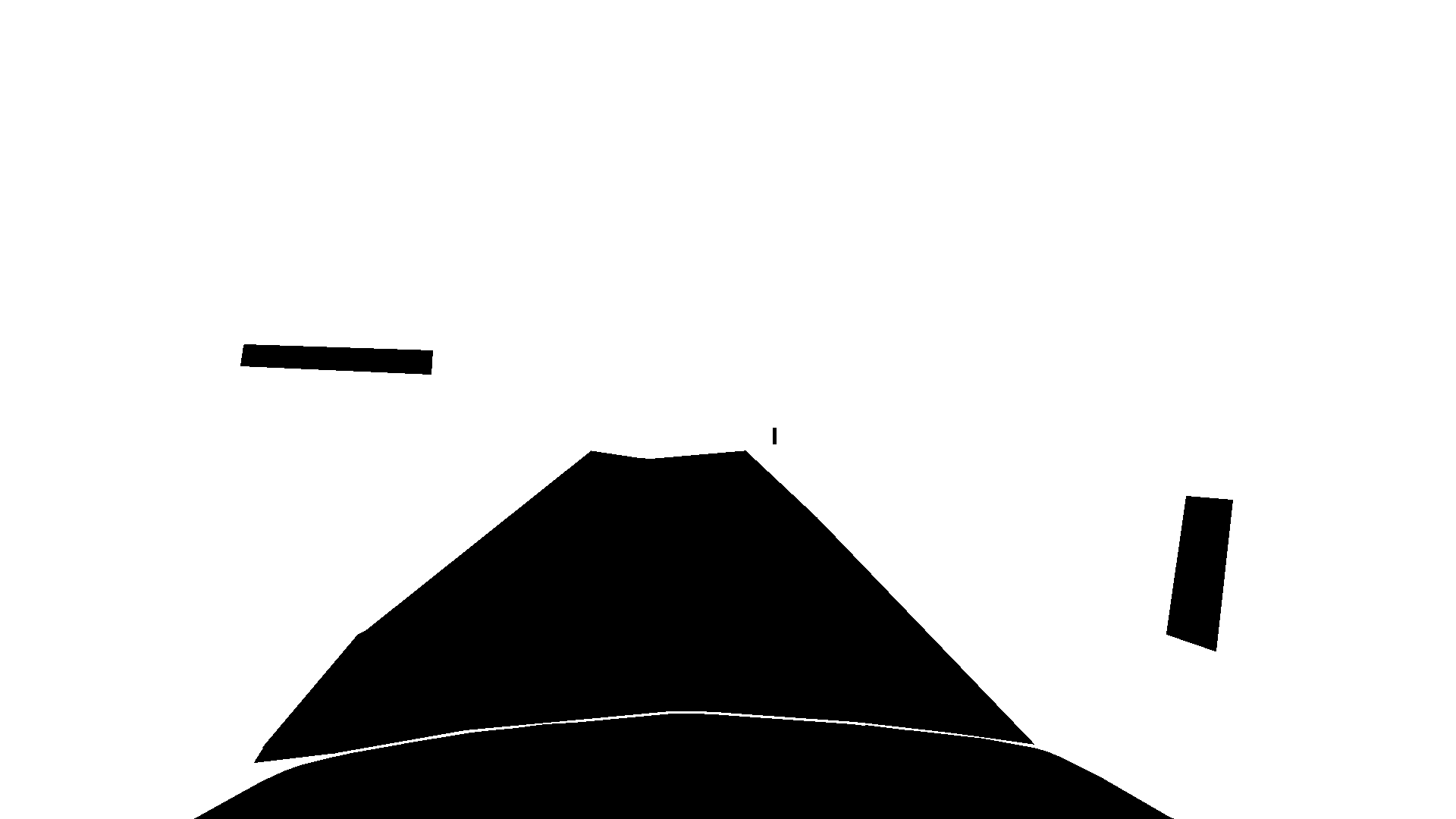}\\

\end{tabular}
\vspace{-0.5em}
\caption{\textbf{Visualization of \Ours{} samples.} From left to right: RGB image; motion-compensated lidar points projected and overlaid with the image; events projected onto the image (assuming infinite distance); azimuth-range radar scan (with ranges above a threshold cropped out); corresponding normal-condition image; panoptic ground truth; difficulty map. Best viewed zoomed in.
}
\label{fig:suppl:annotation:aux:data:1}
\end{figure}

\begin{figure*}
\centering
\begin{tabular}{@{}c@{\hspace{0.03cm}}c@{\hspace{0.03cm}}c@{\hspace{0.03cm}}c@{\hspace{0.03cm}}c@{\hspace{0.03cm}}c@{\hspace{0.03cm}}c@{}}
\subfloat{\tiny RGB Image} &
\subfloat{\tiny Lidar} &
\subfloat{\tiny Events}&
\subfloat{\tiny Radar}&
\subfloat{\tiny Corr. Image}&
\subfloat{\tiny Panoptic GT}&
\subfloat{\tiny Difficulty Map}  \\

% Rain Day 1
\includegraphics[width=0.14\textwidth]{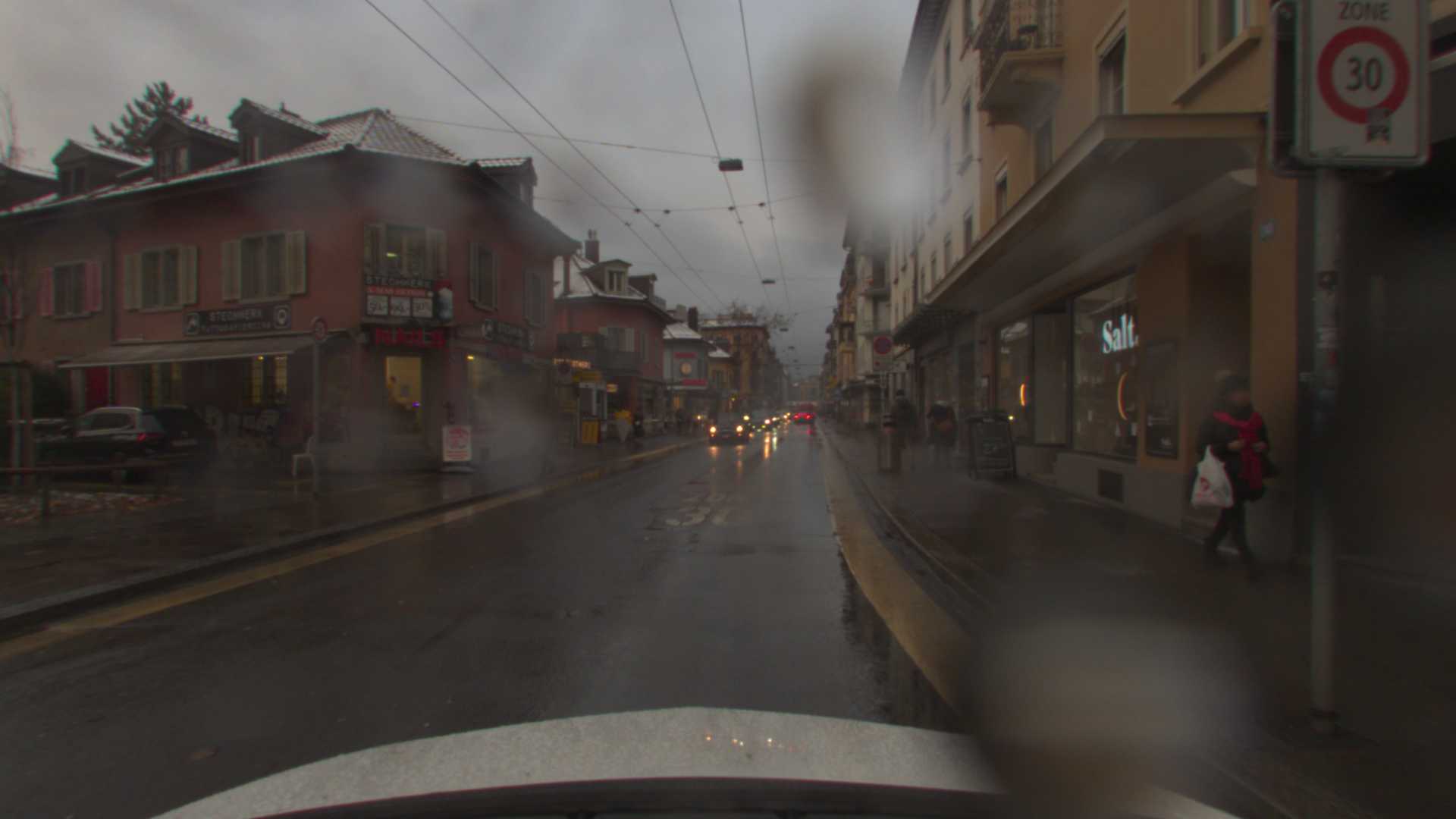} &
\includegraphics[width=0.14\textwidth]{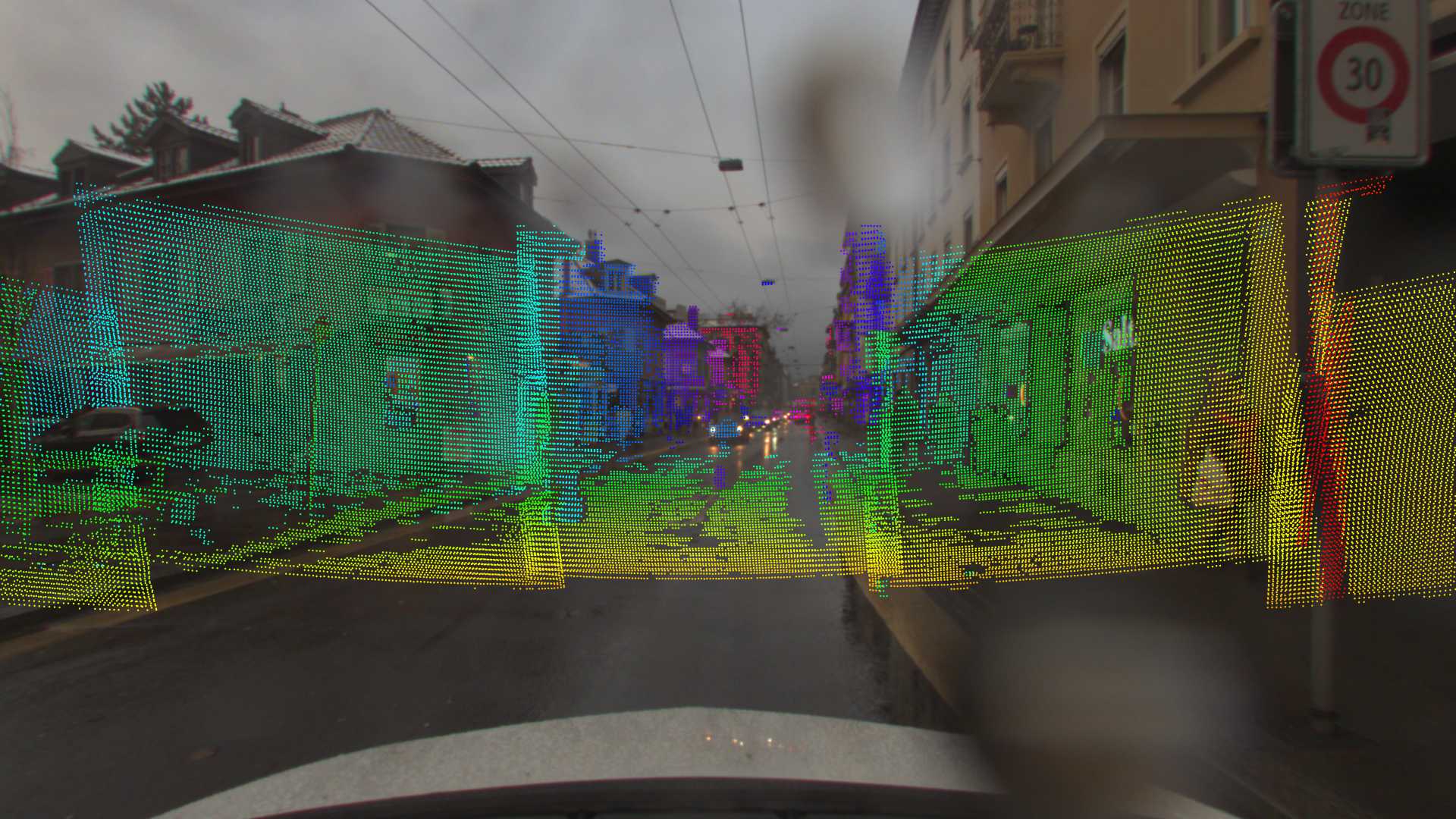} &
\includegraphics[width=0.14\textwidth]{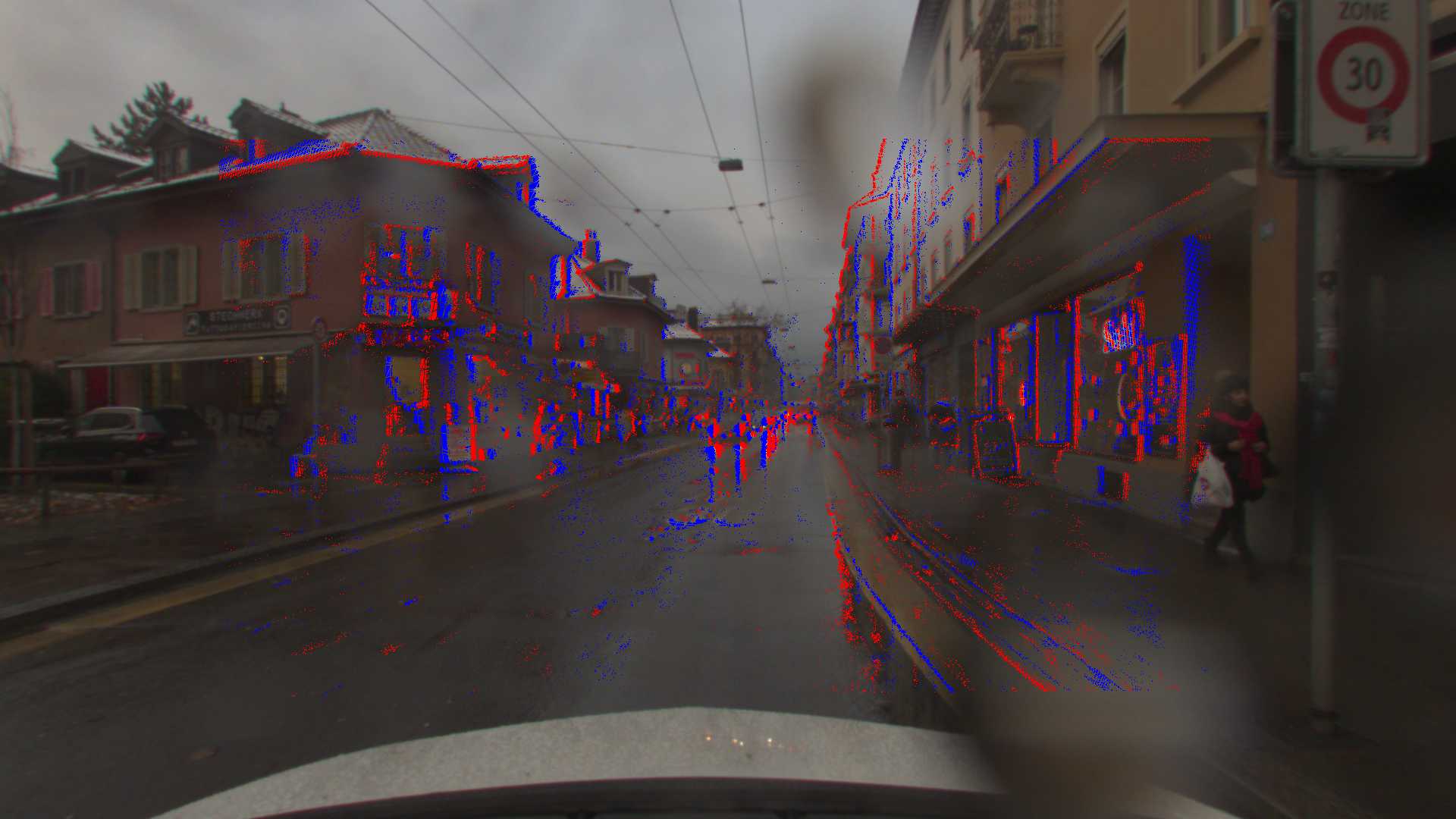} &
\includegraphics[angle=90, trim=0 6835 0 0, clip,width=0.14\textwidth]{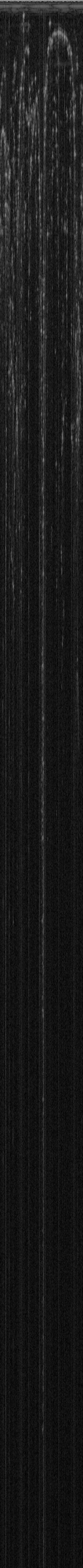}&
\includegraphics[width=0.14\textwidth]{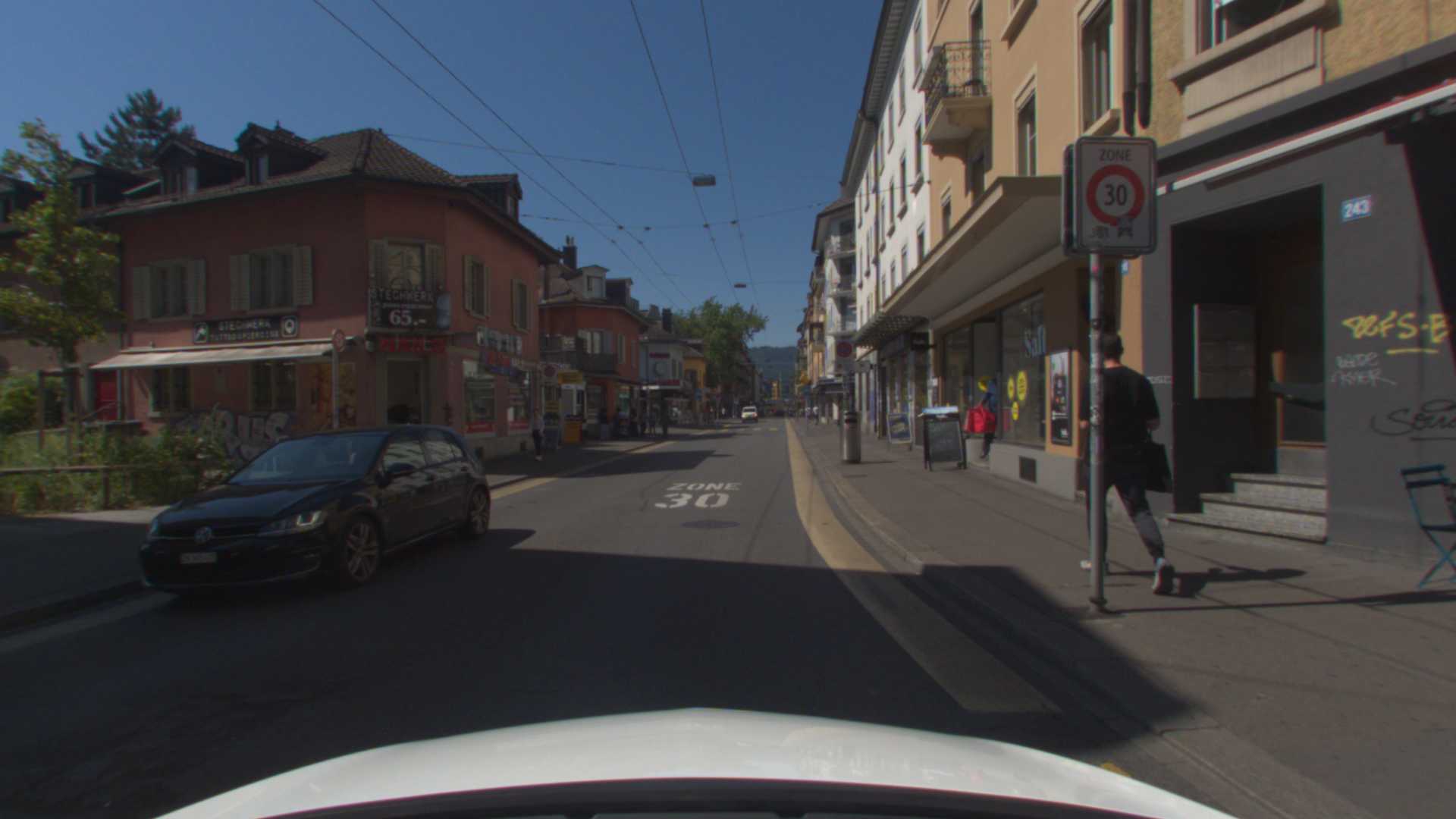} &
\includegraphics[width=0.14\textwidth]{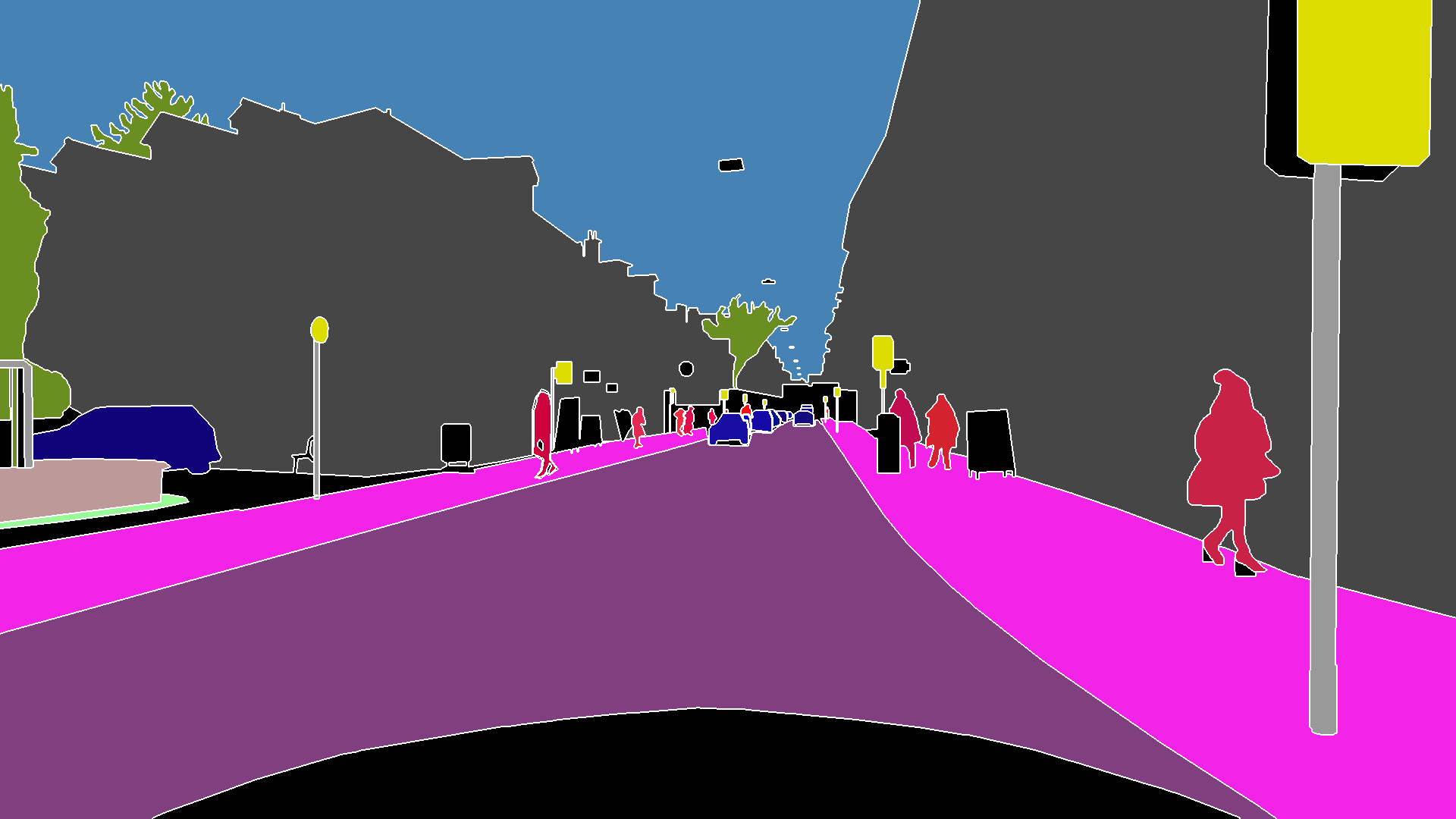} &
\includegraphics[width=0.14\textwidth]{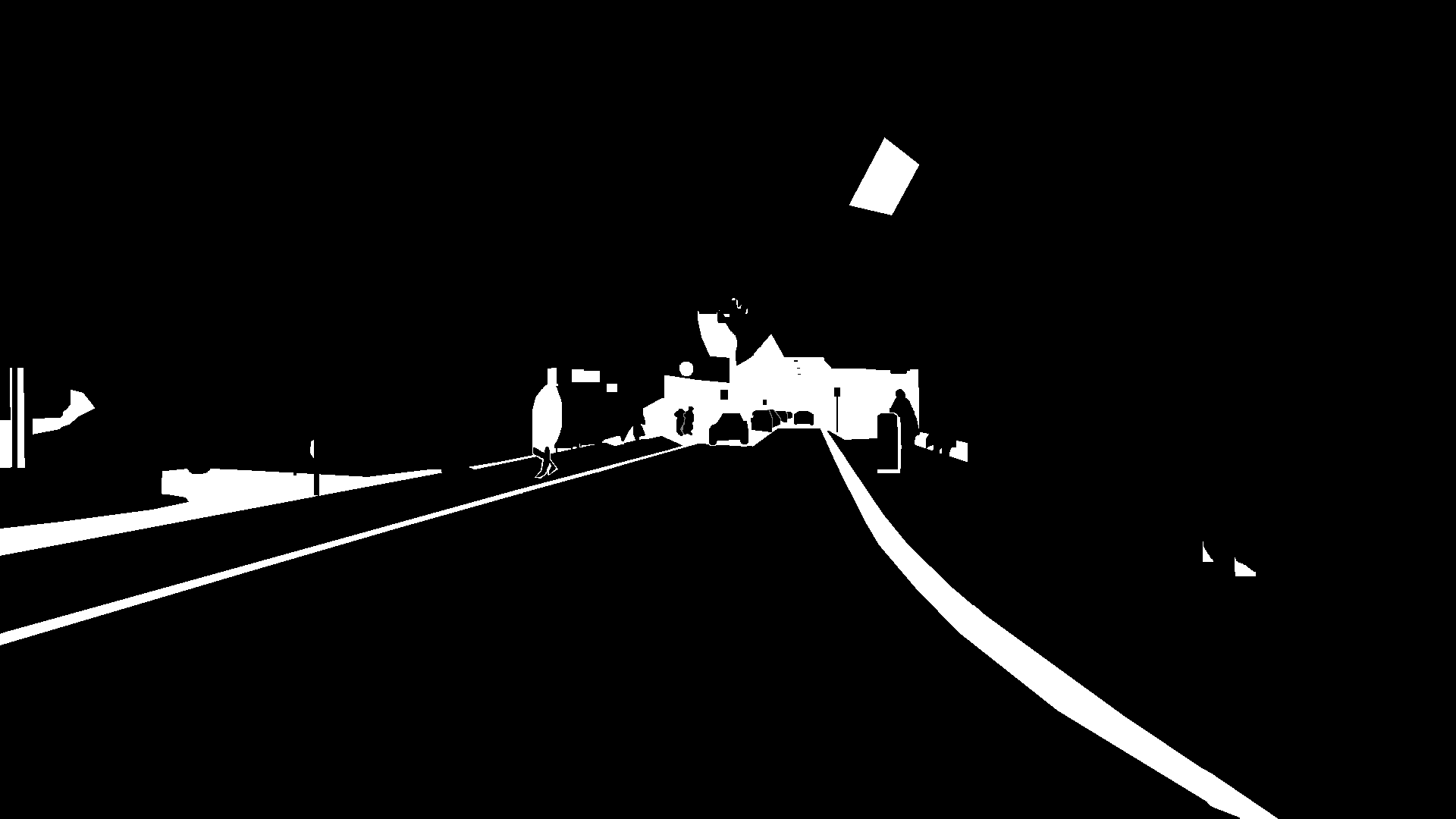}\\

% Rain Day 2
\includegraphics[width=0.14\textwidth]{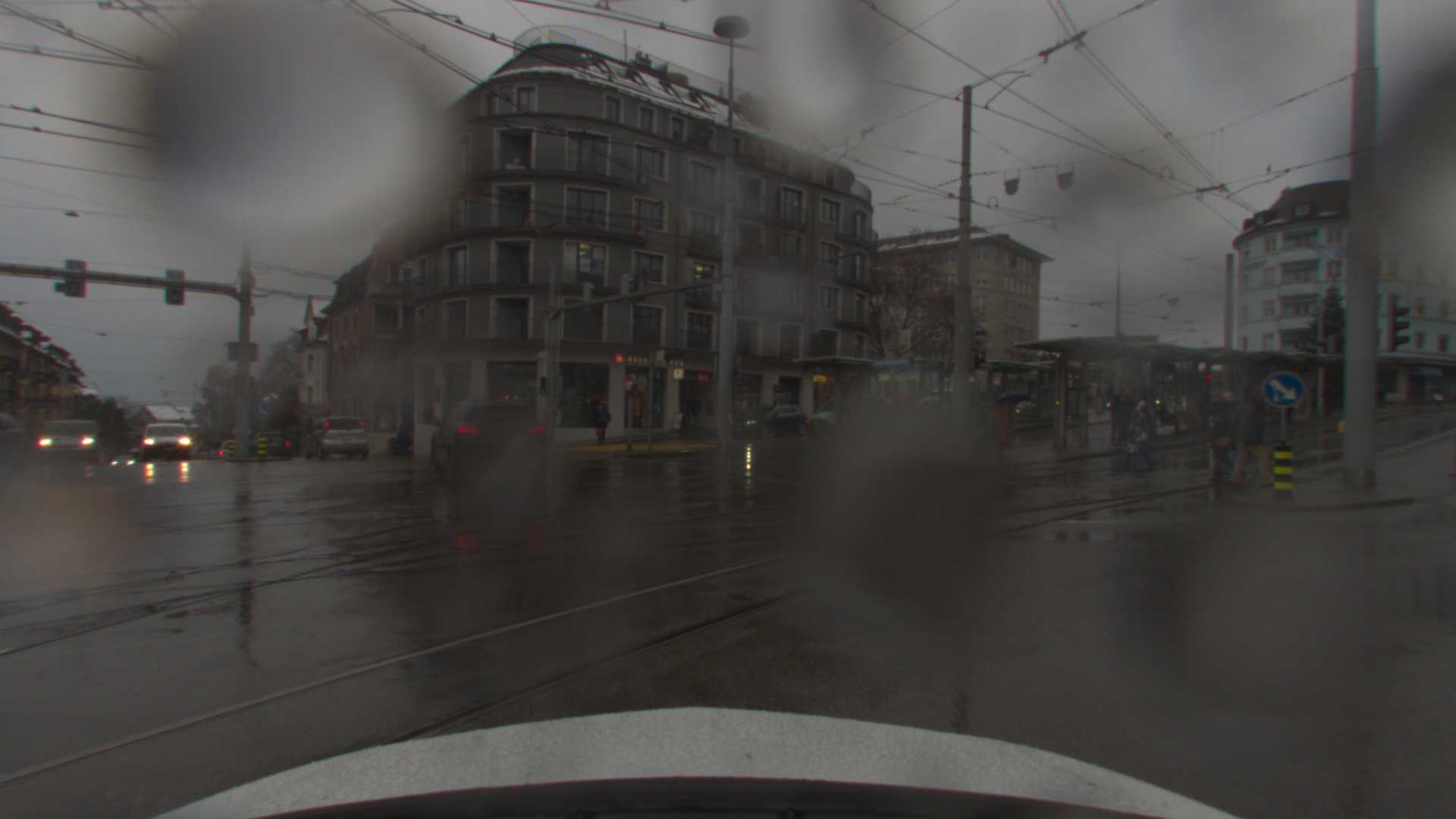} &
\includegraphics[width=0.14\textwidth]{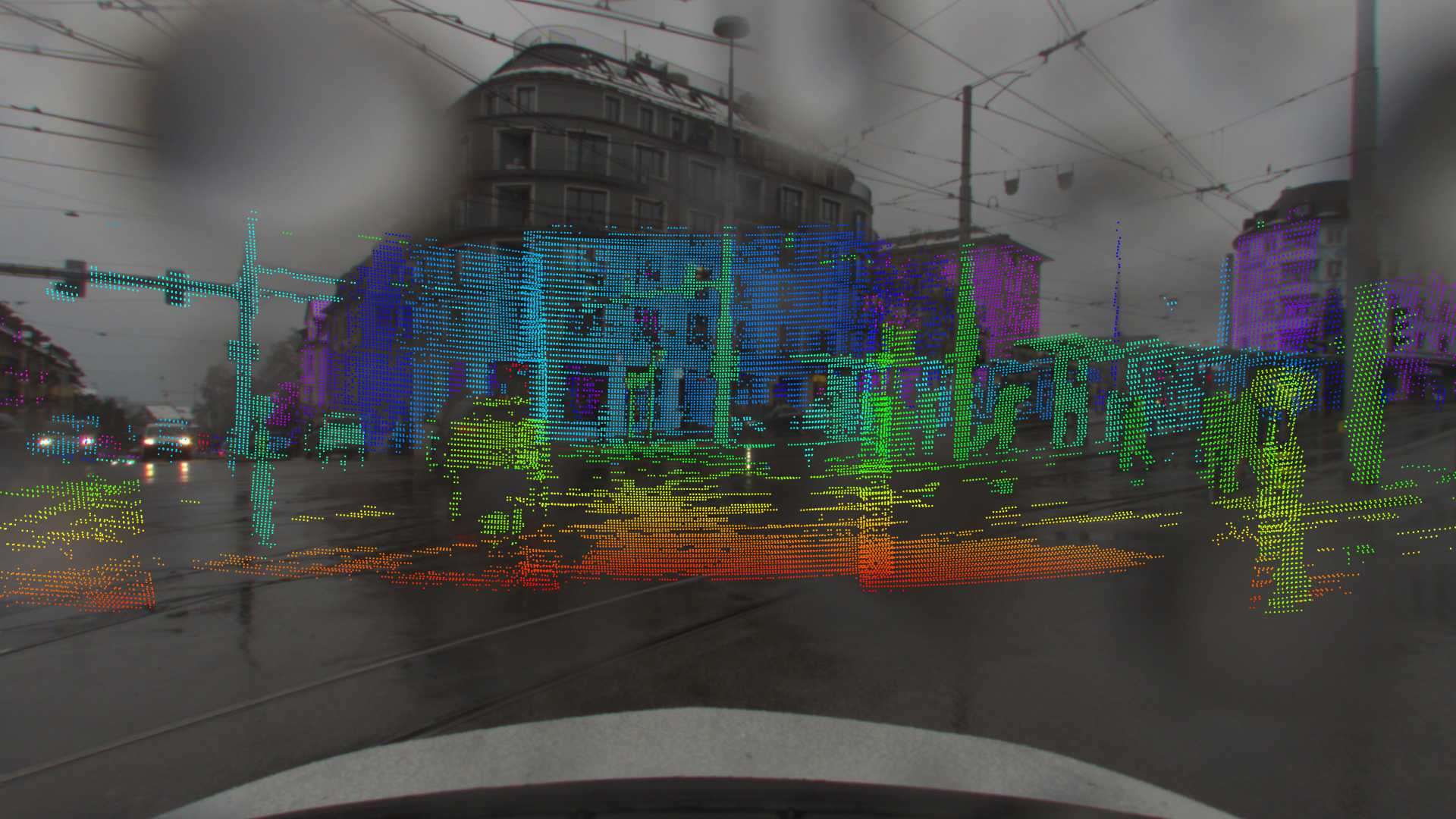} &
\includegraphics[width=0.14\textwidth]{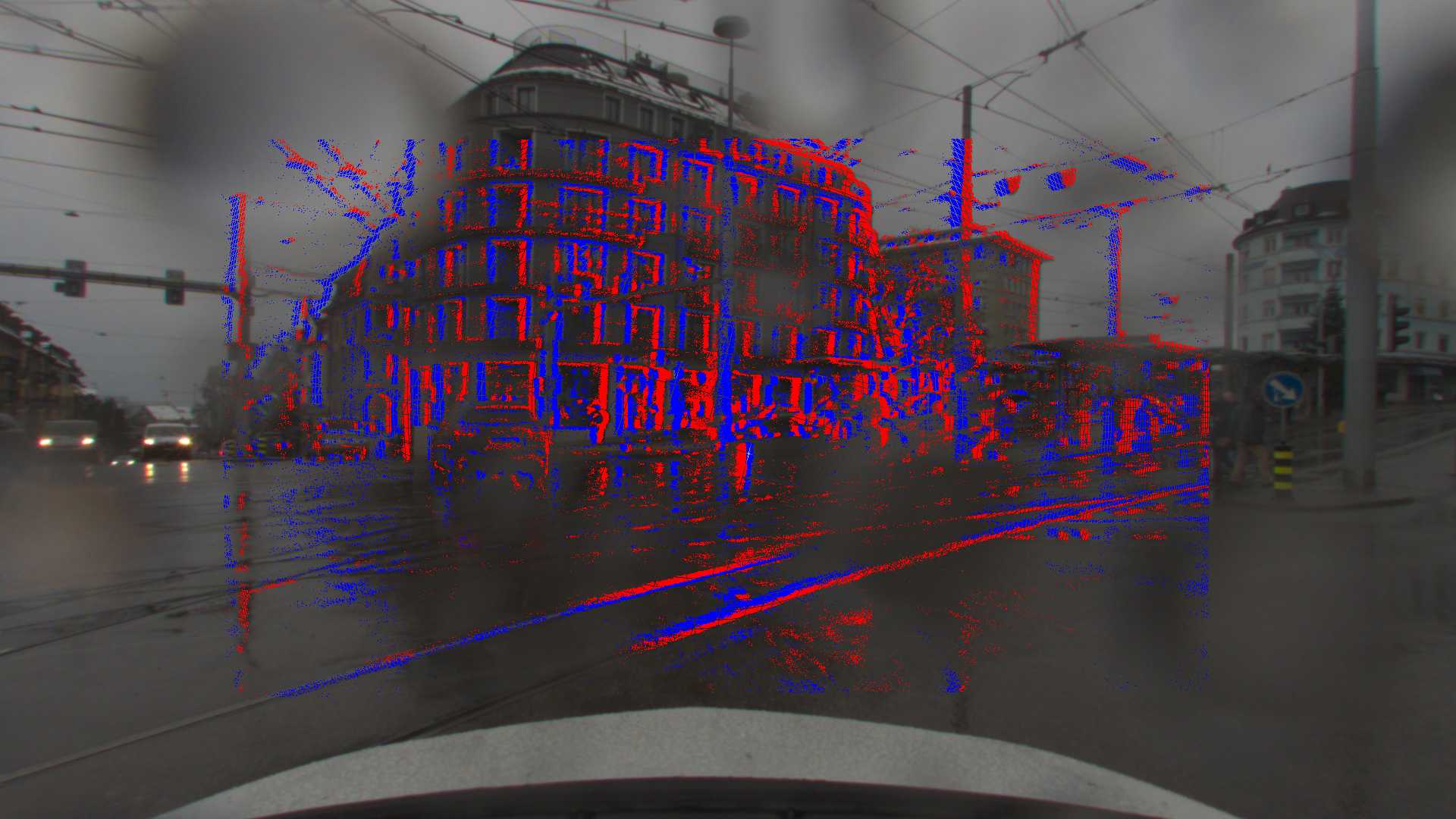} &
\includegraphics[angle=90, trim=0 6835 0 0, clip,width=0.14\textwidth]{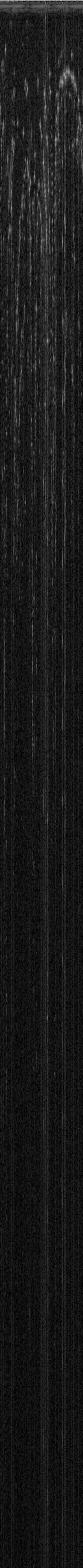}&
\includegraphics[width=0.14\textwidth]{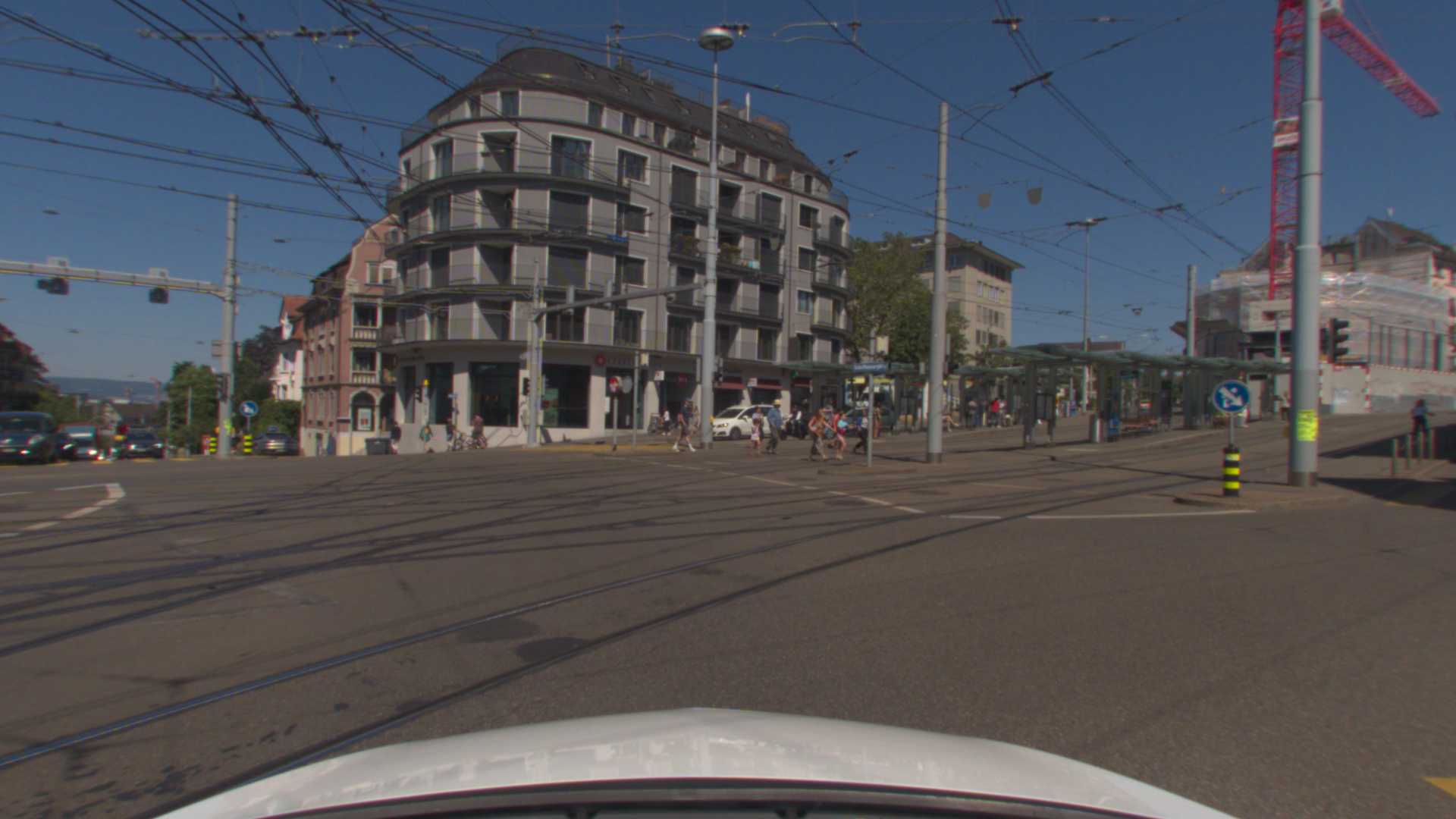} &
\includegraphics[width=0.14\textwidth]{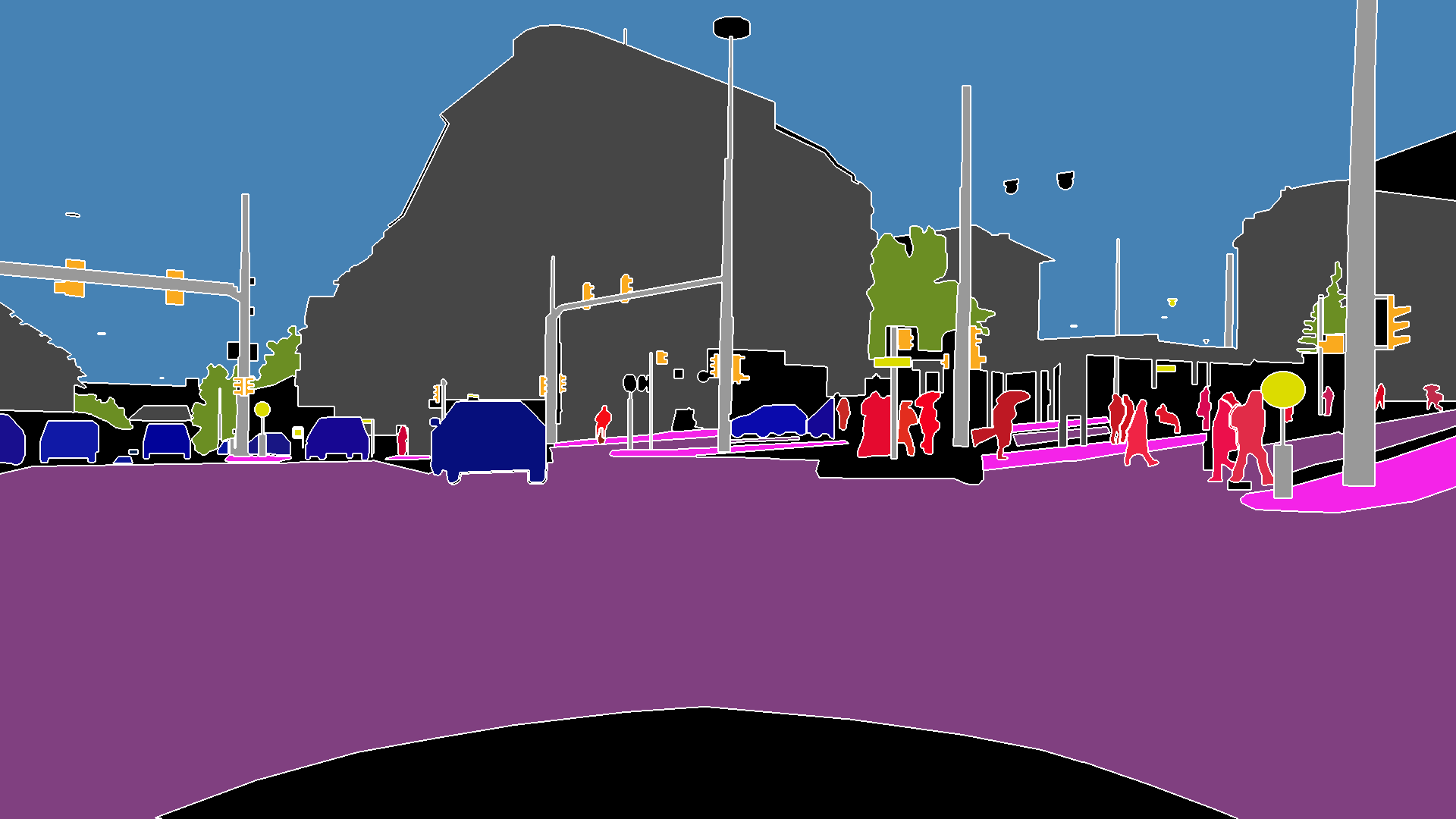} &
\includegraphics[width=0.14\textwidth]{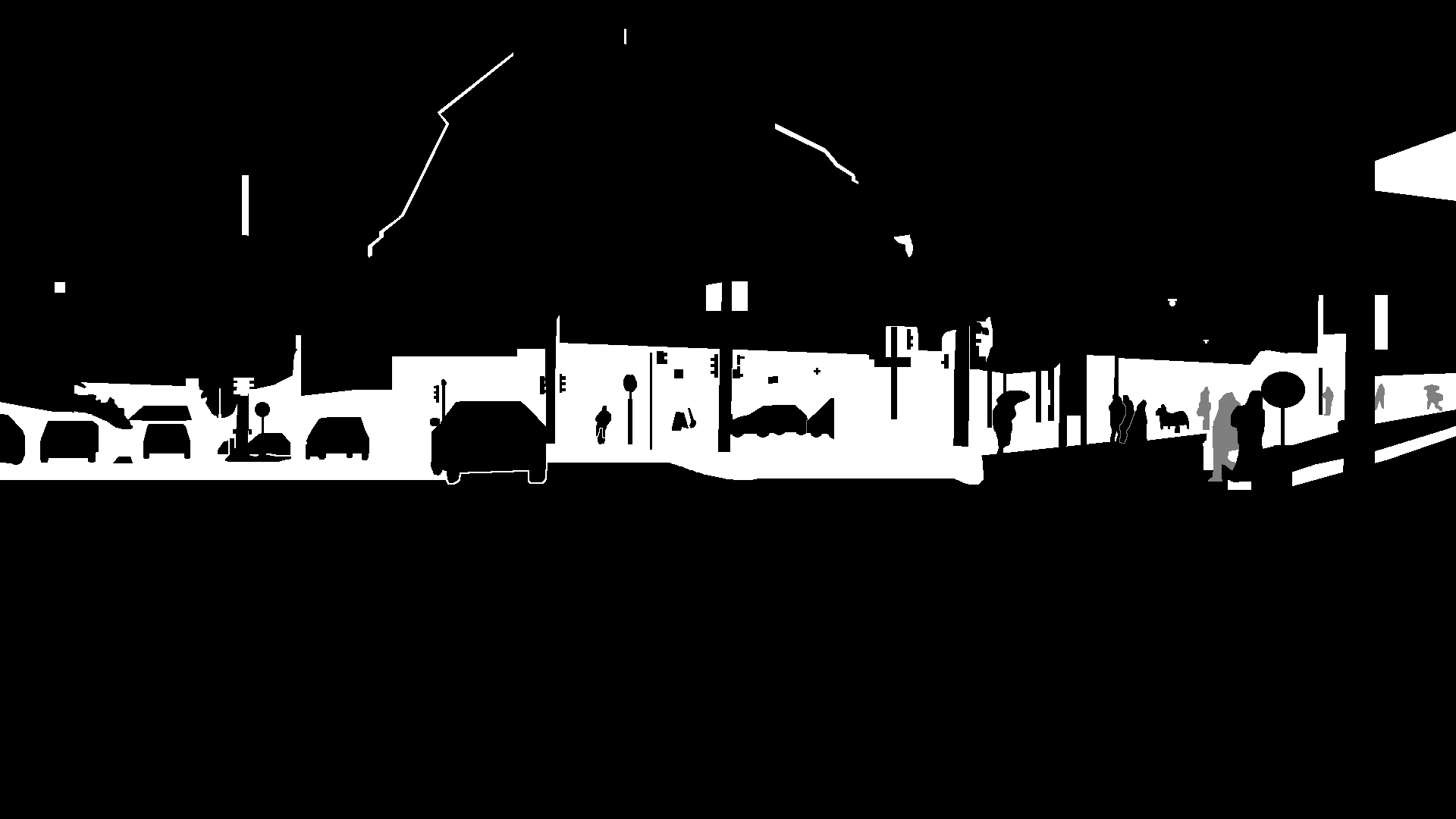}\\

% Rain Night 1
\includegraphics[width=0.14\textwidth]{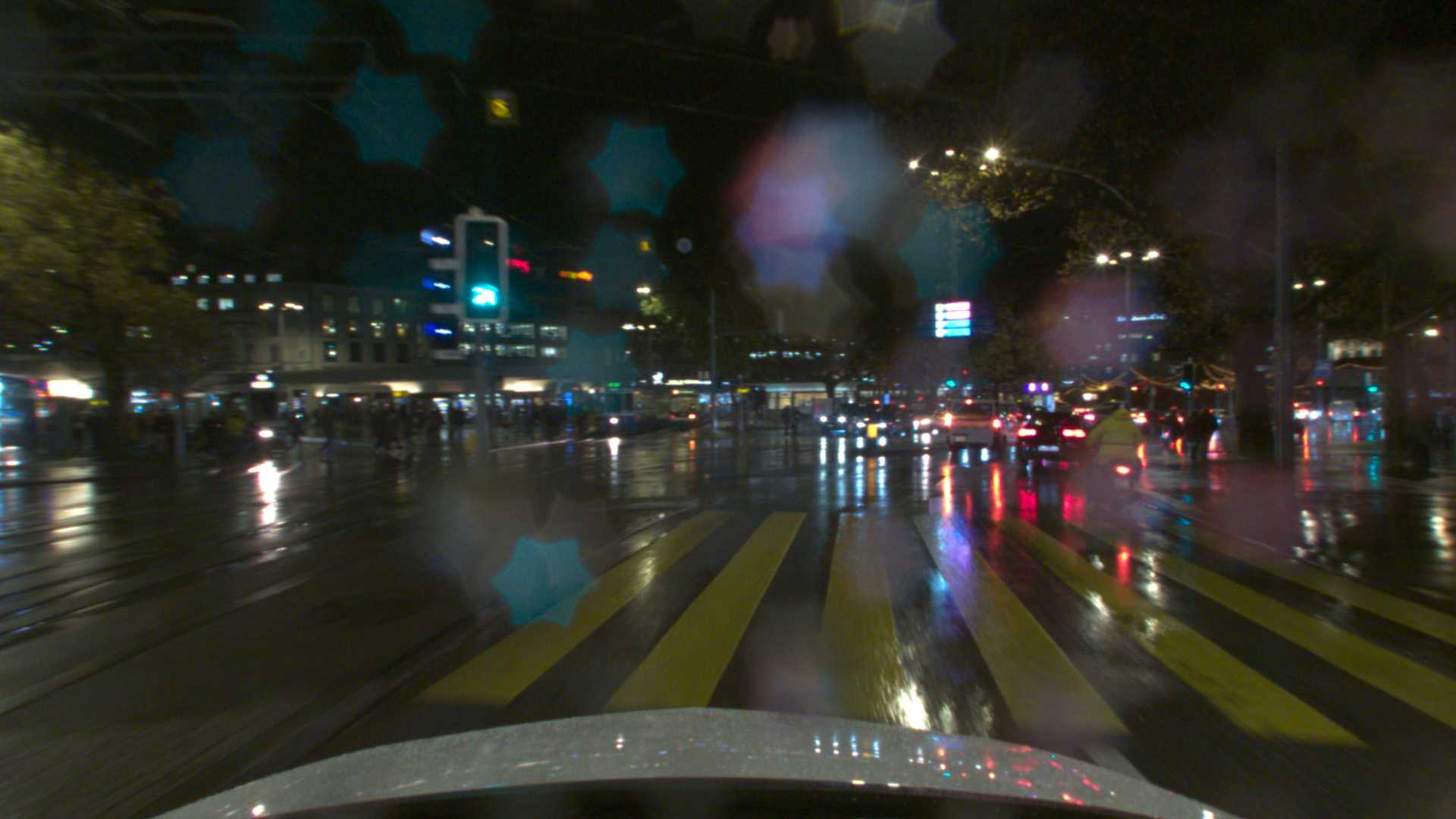} &
\includegraphics[width=0.14\textwidth]{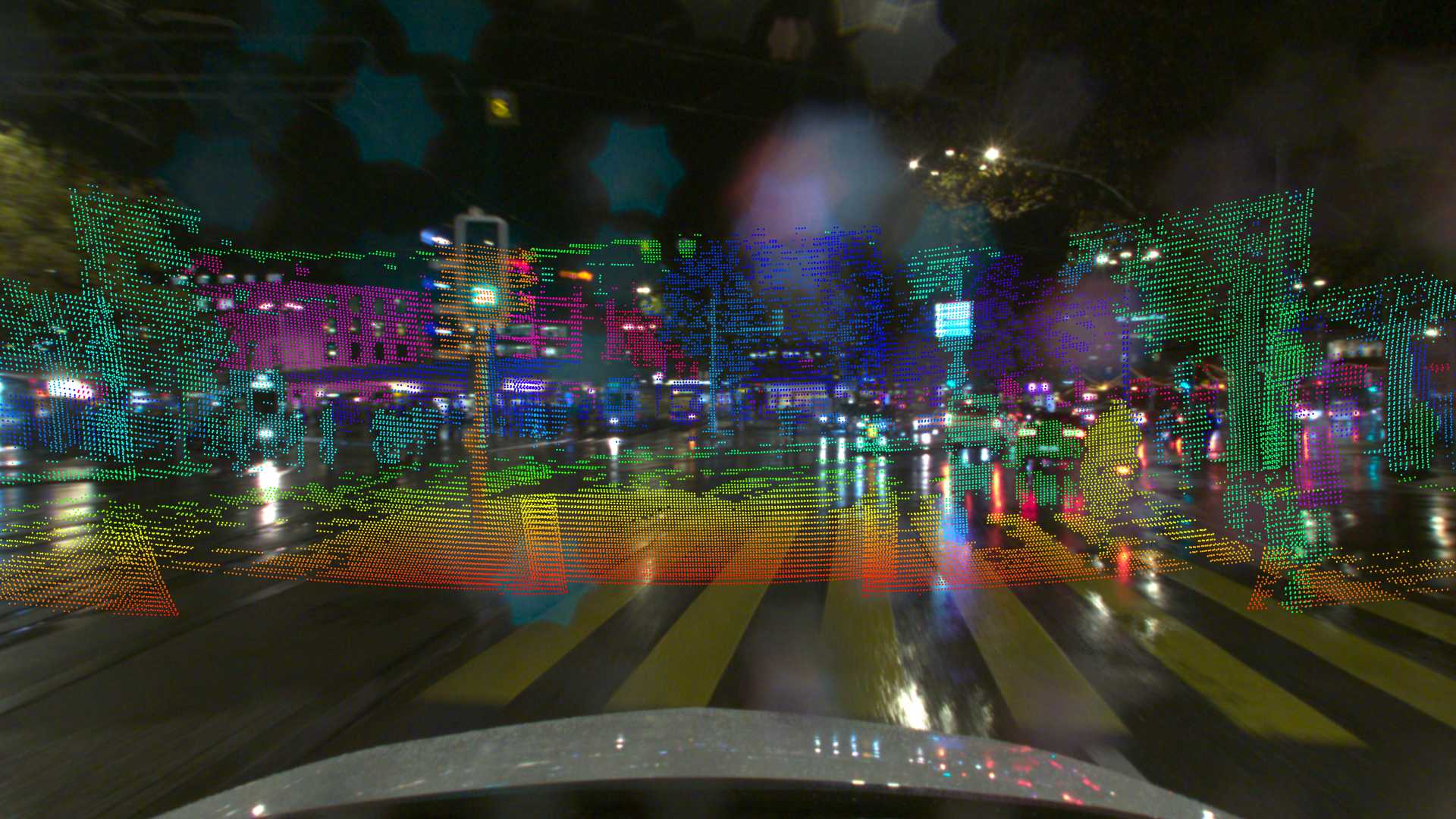} &
\includegraphics[width=0.14\textwidth]{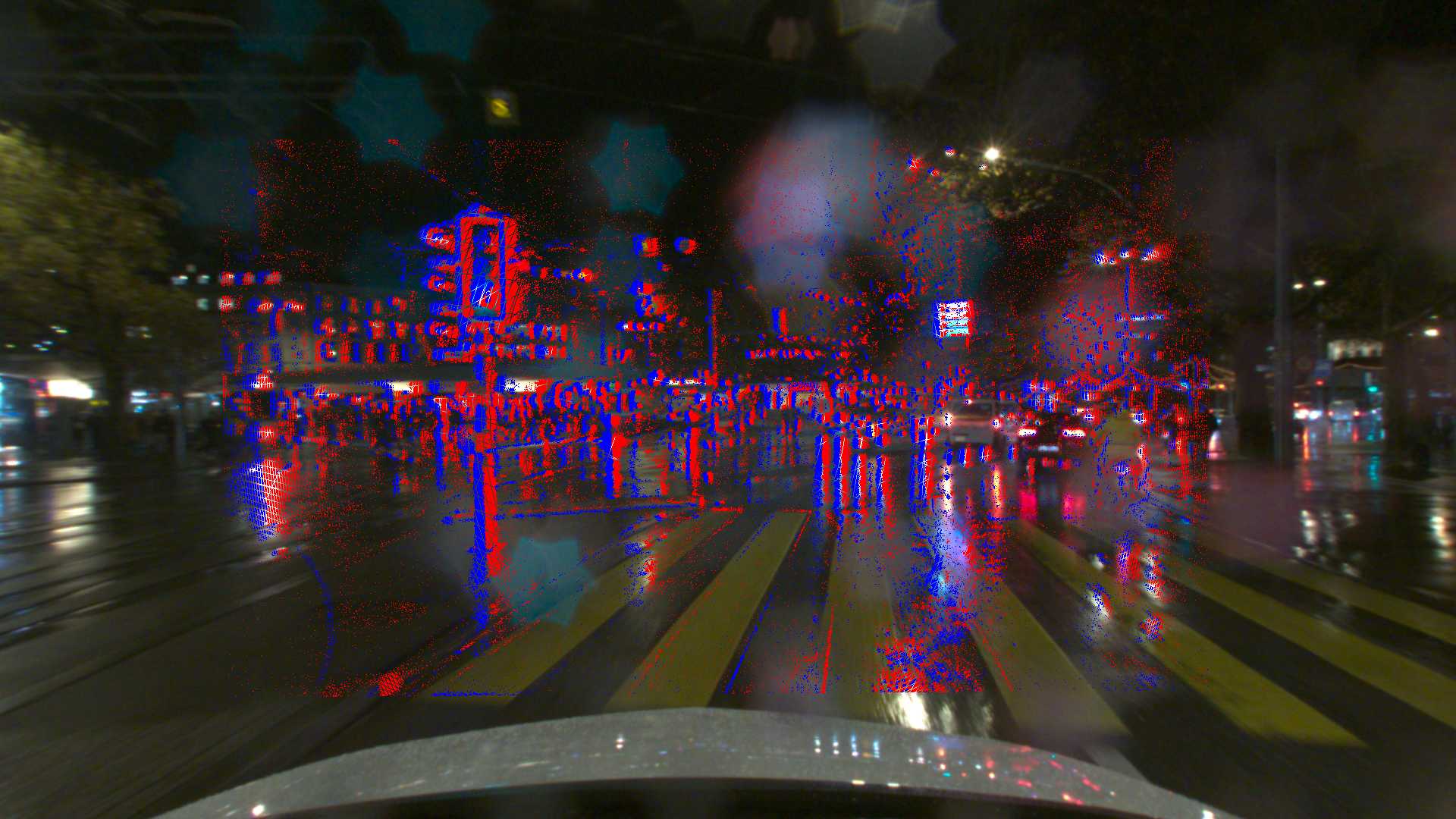} &
\includegraphics[angle=90, trim=0 6835 0 0, clip,width=0.14\textwidth]{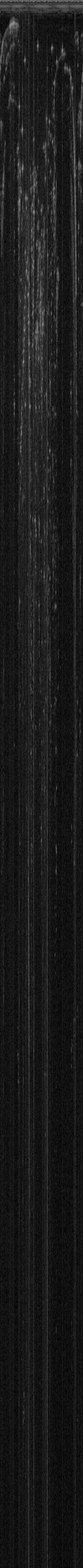}&
\includegraphics[width=0.14\textwidth]{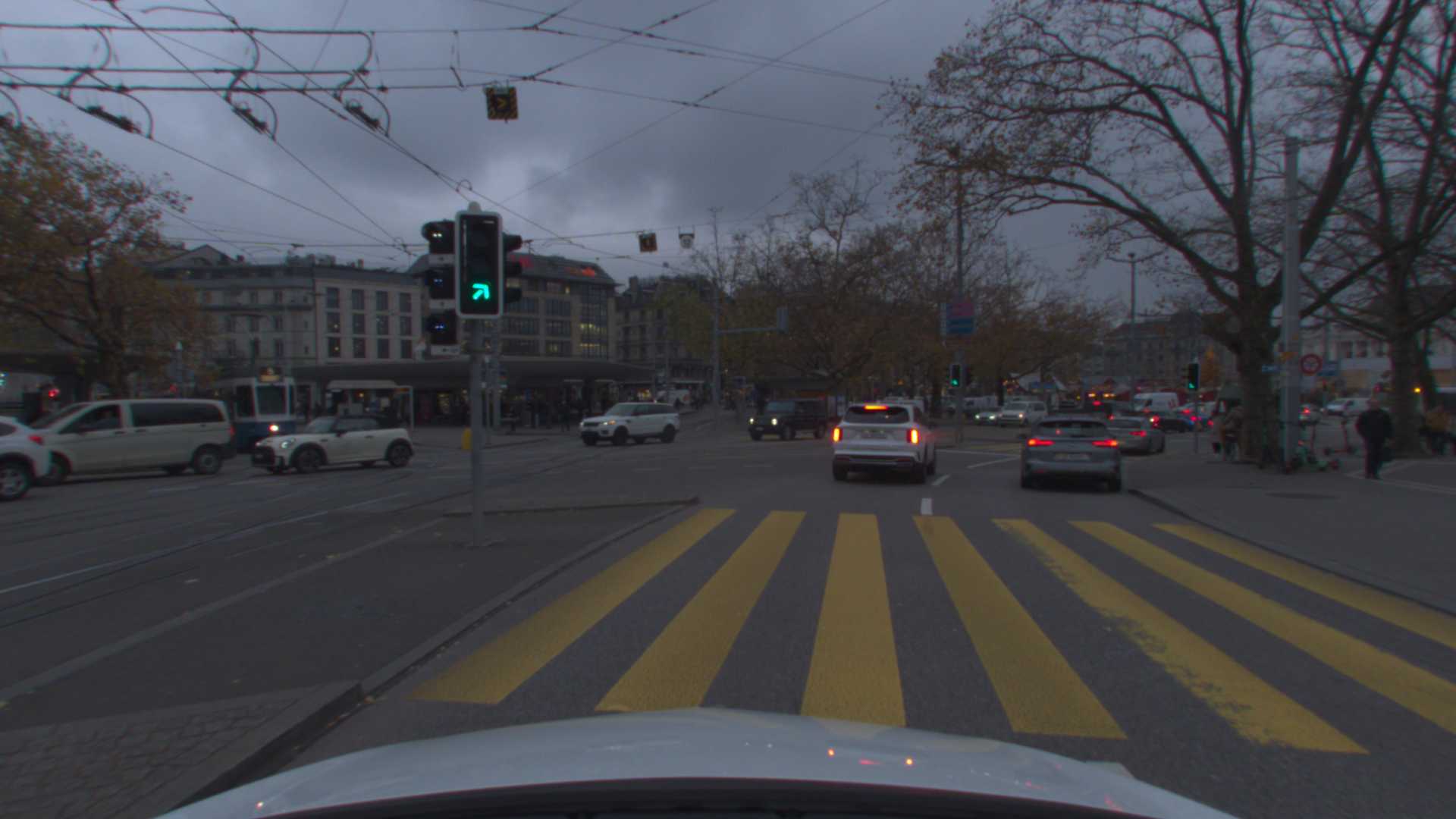} &
\includegraphics[width=0.14\textwidth]{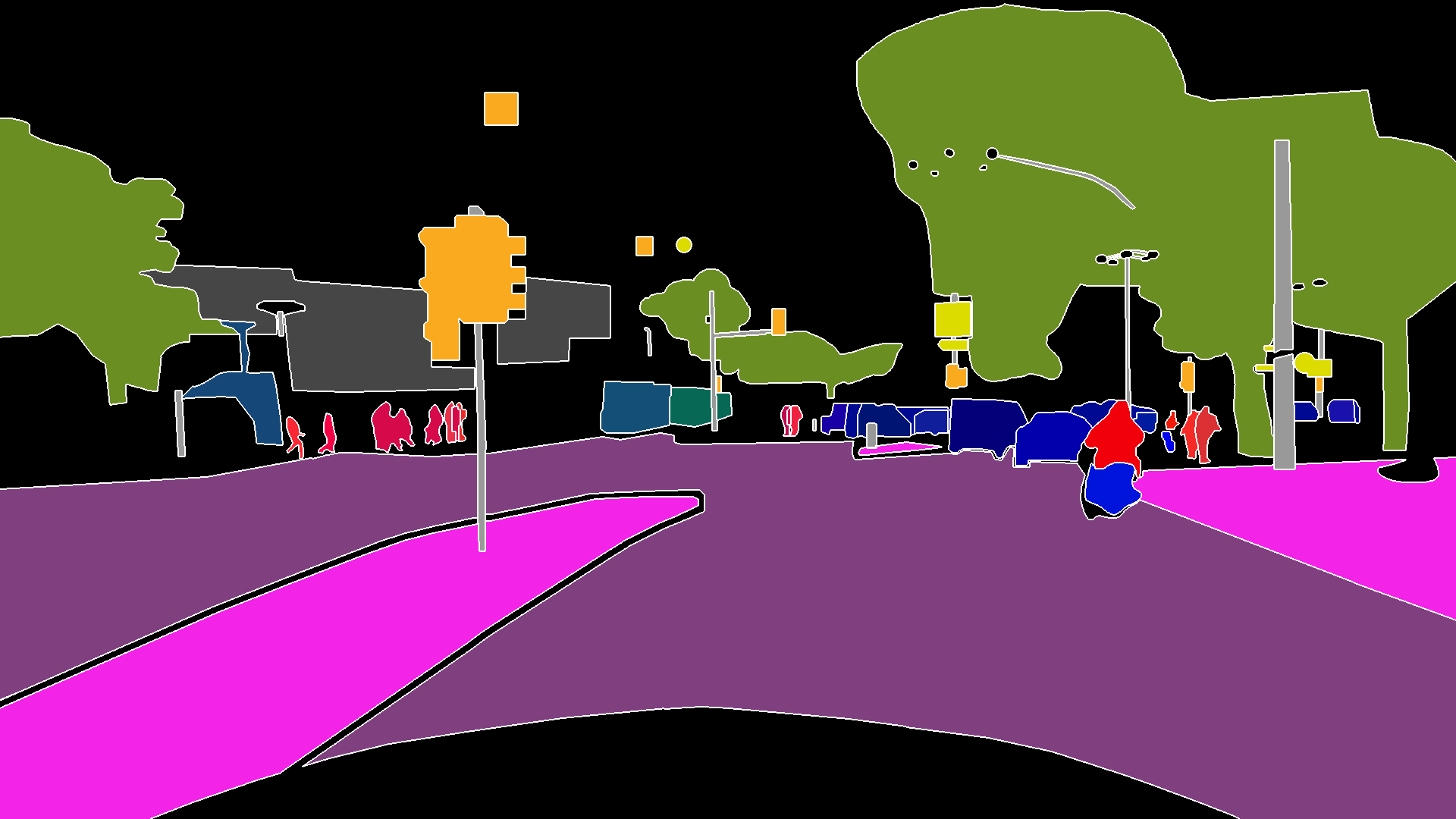} &
\includegraphics[width=0.14\textwidth]{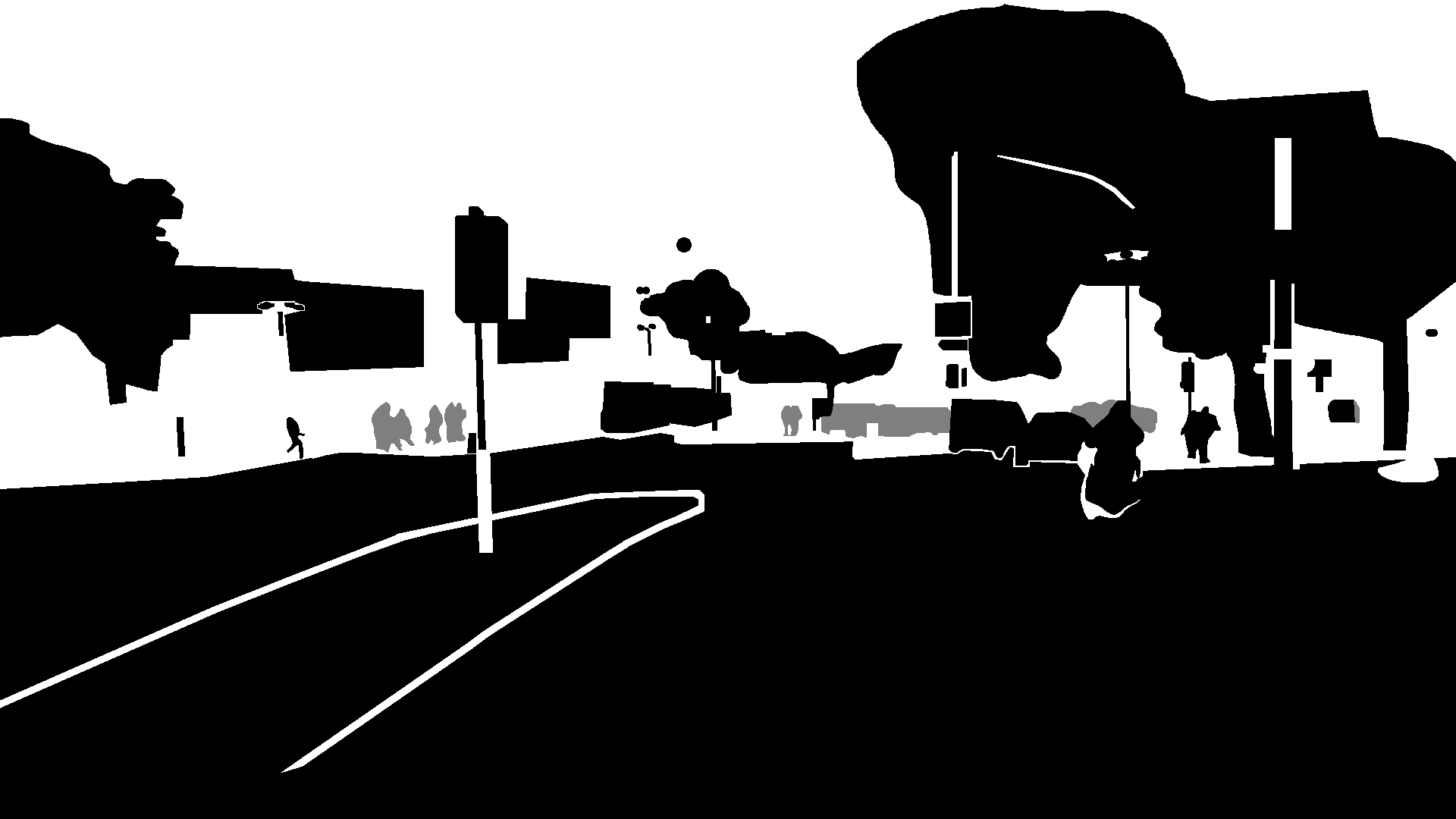}\\

% Rain Night 2
\includegraphics[width=0.14\textwidth]{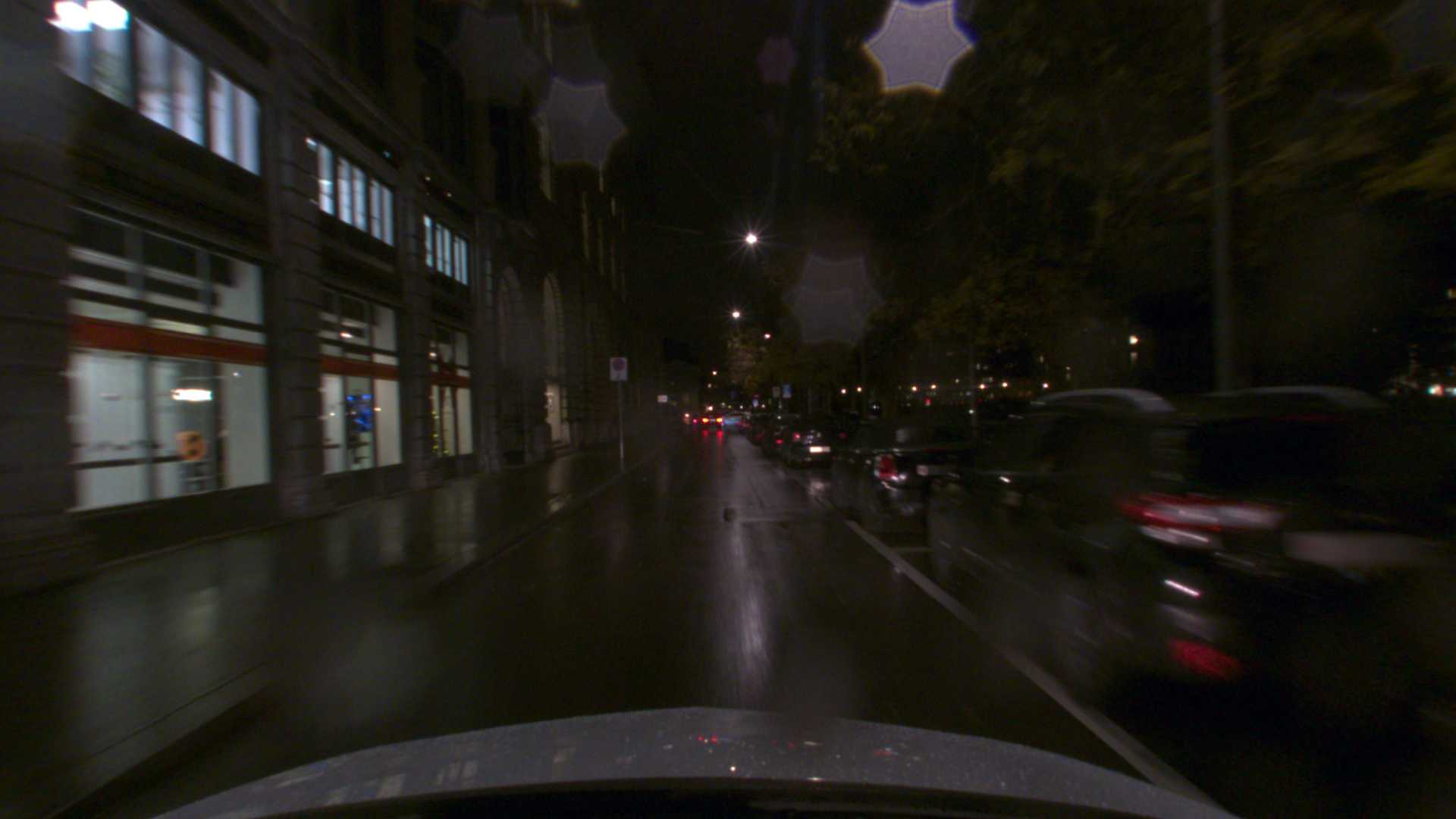} &
\includegraphics[width=0.14\textwidth]{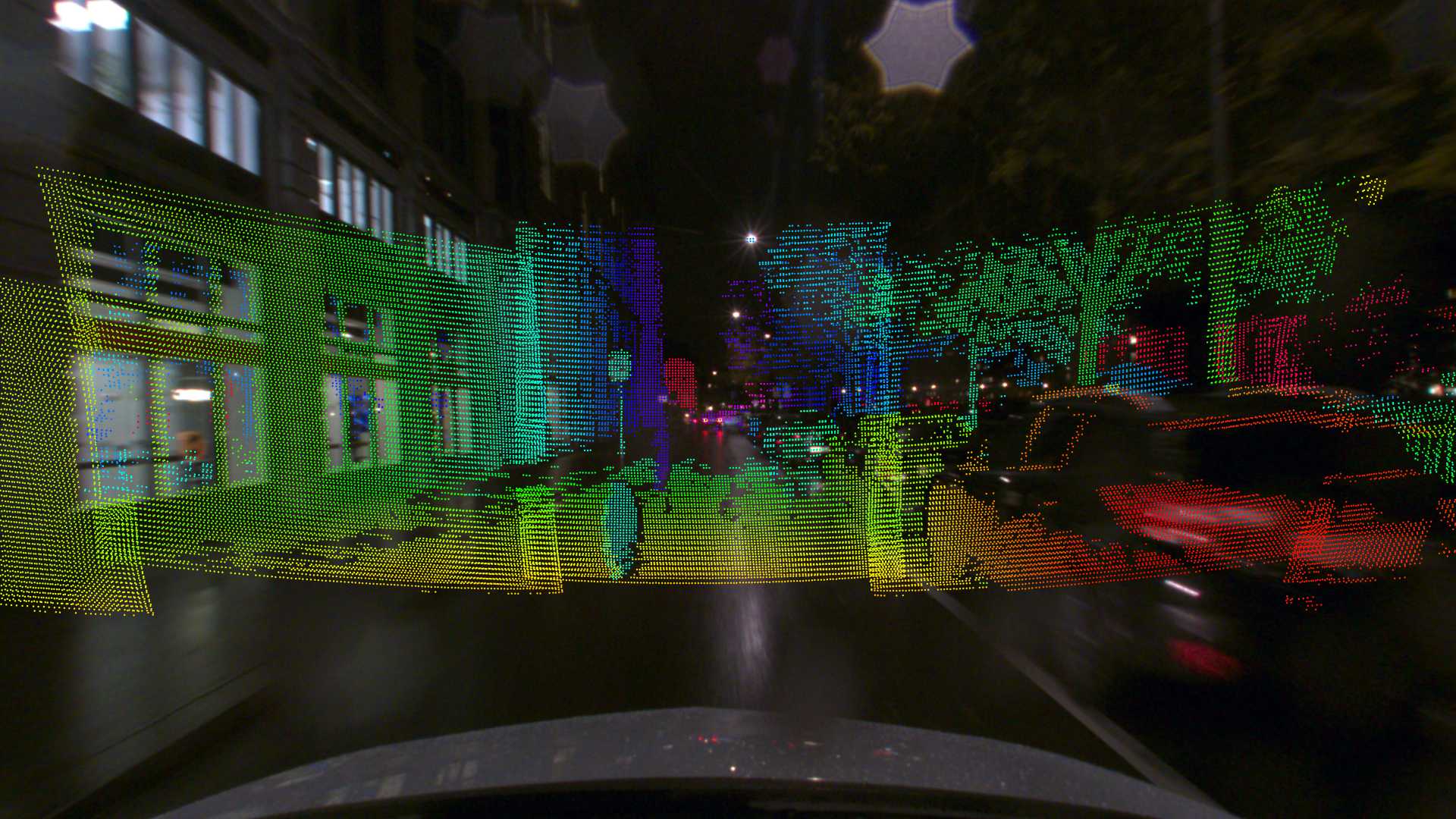} &
\includegraphics[width=0.14\textwidth]{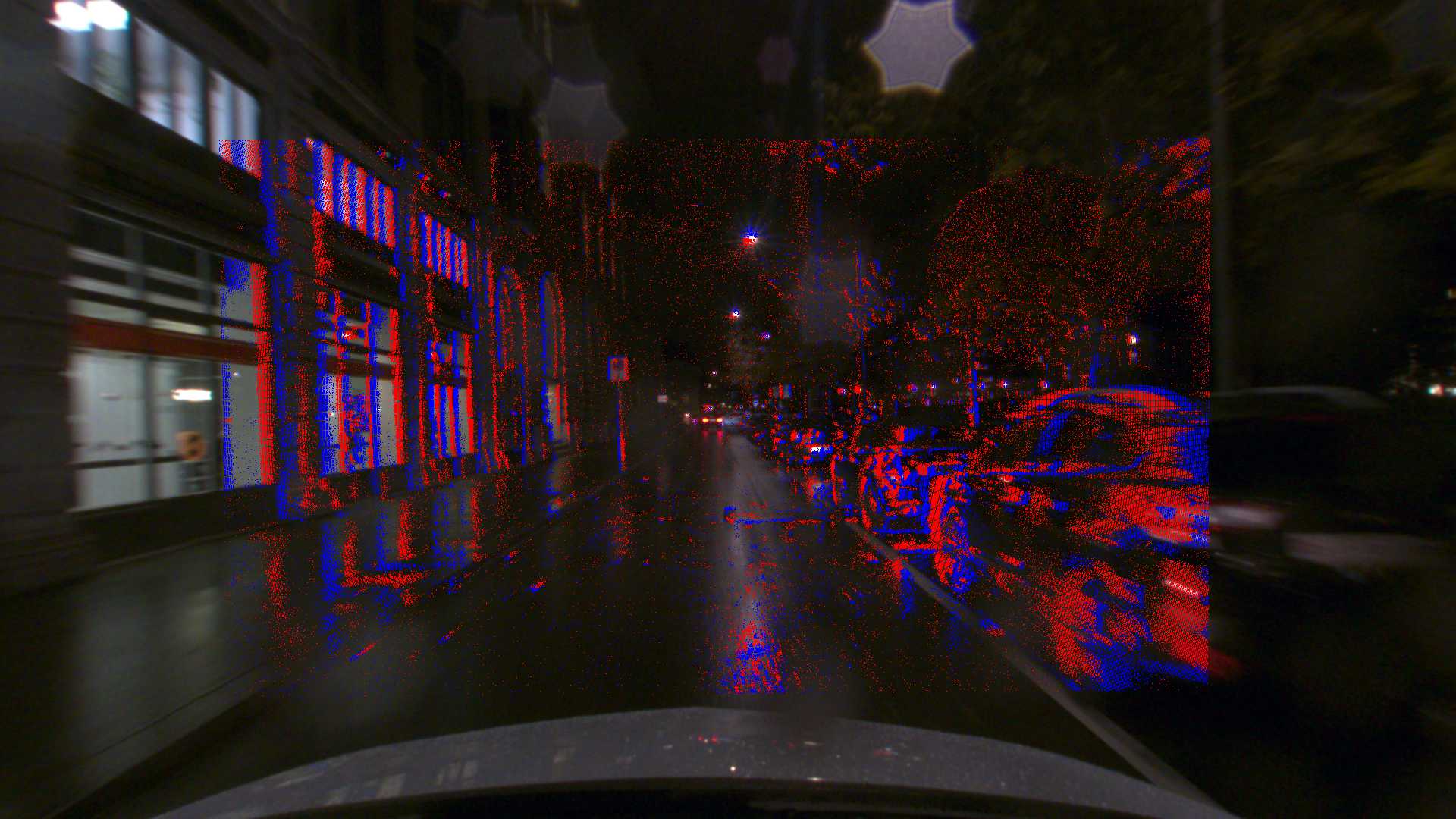} &
\includegraphics[angle=90, trim=0 6835 0 0, clip,width=0.14\textwidth]{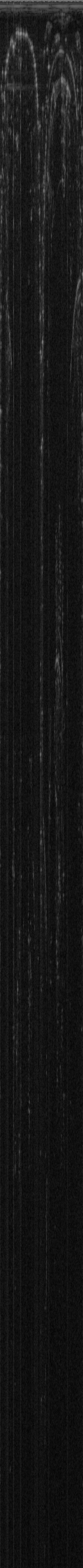}&
\includegraphics[width=0.14\textwidth]{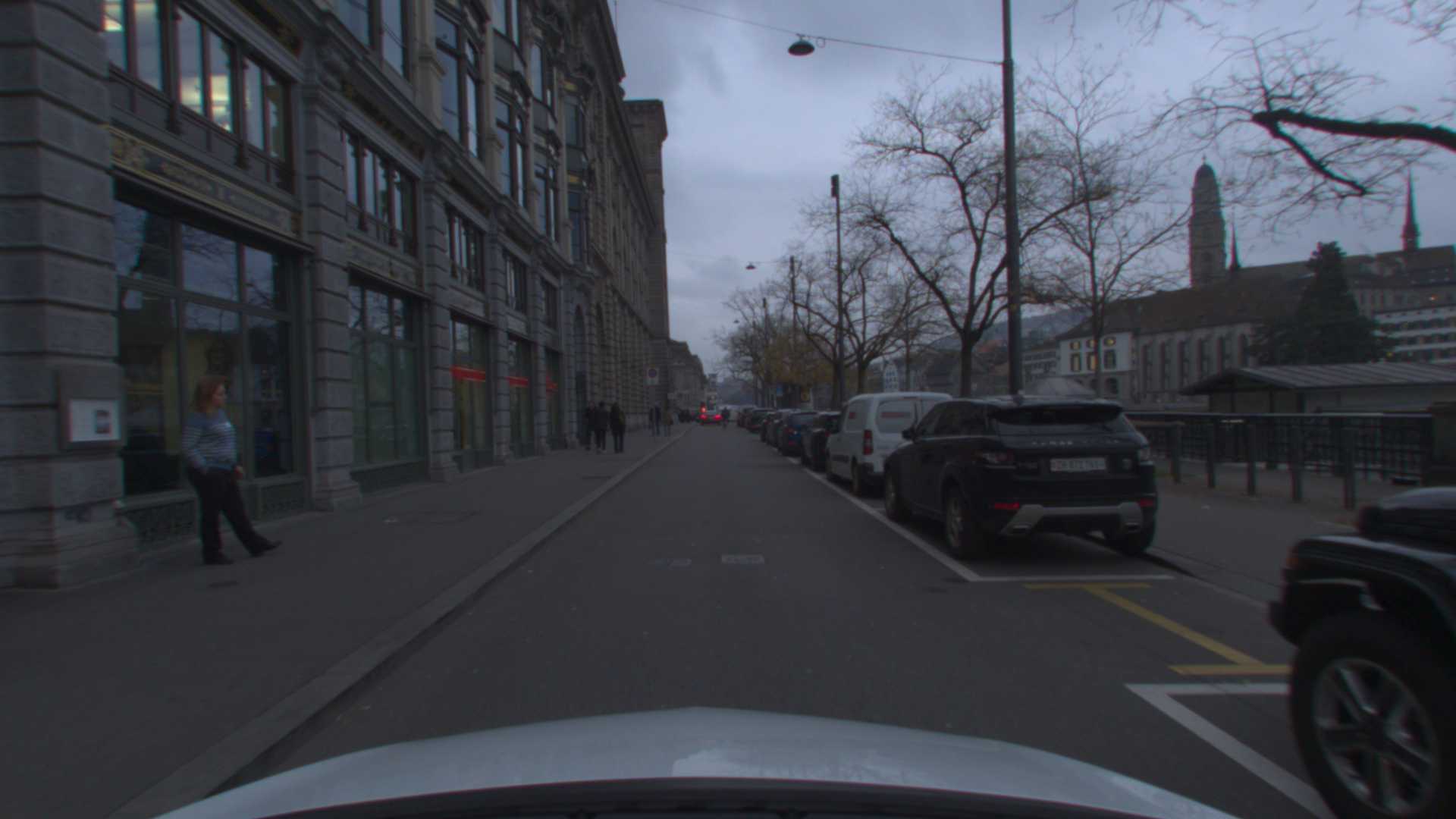} &
\includegraphics[width=0.14\textwidth]{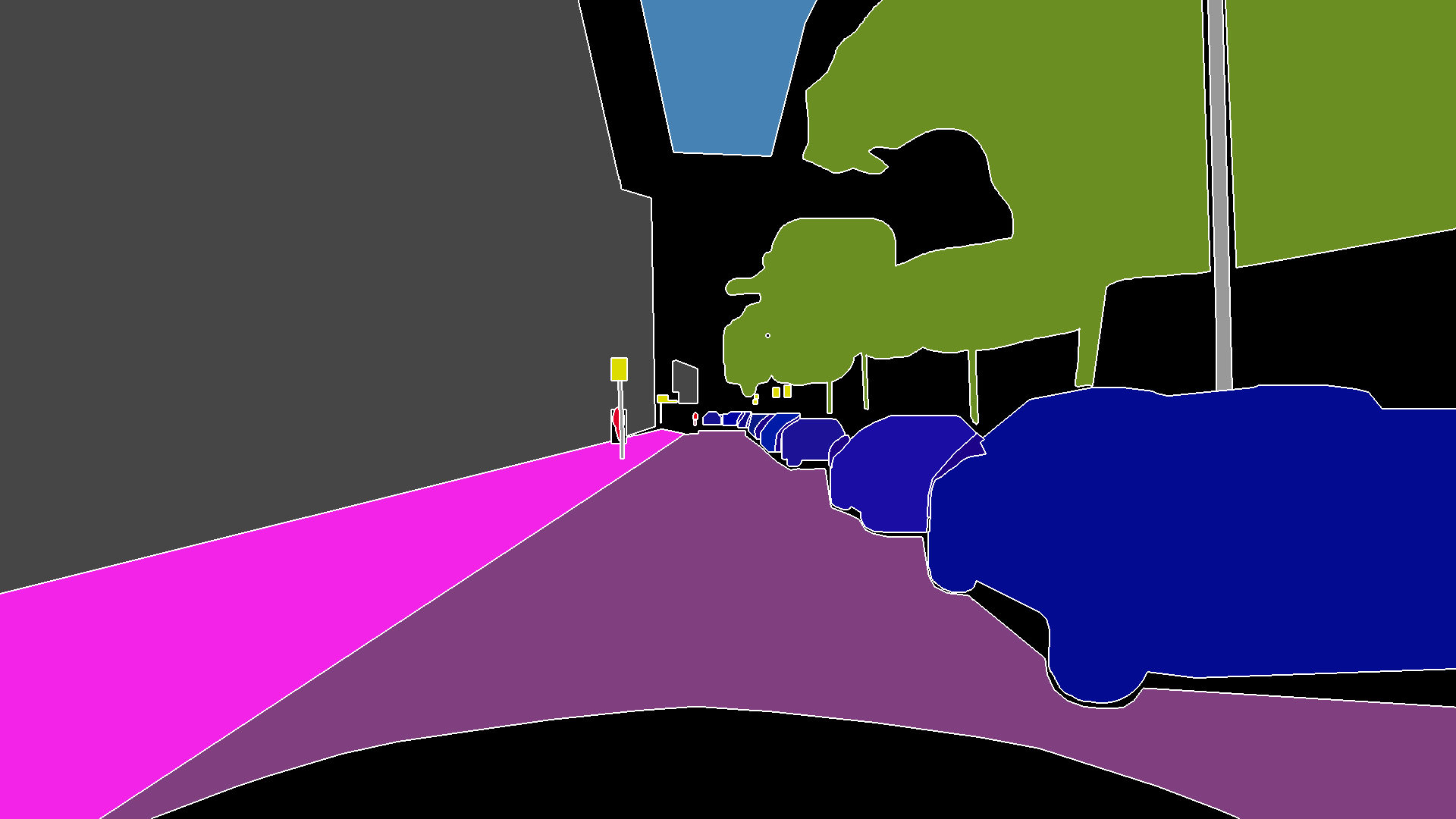} &
\includegraphics[width=0.14\textwidth]{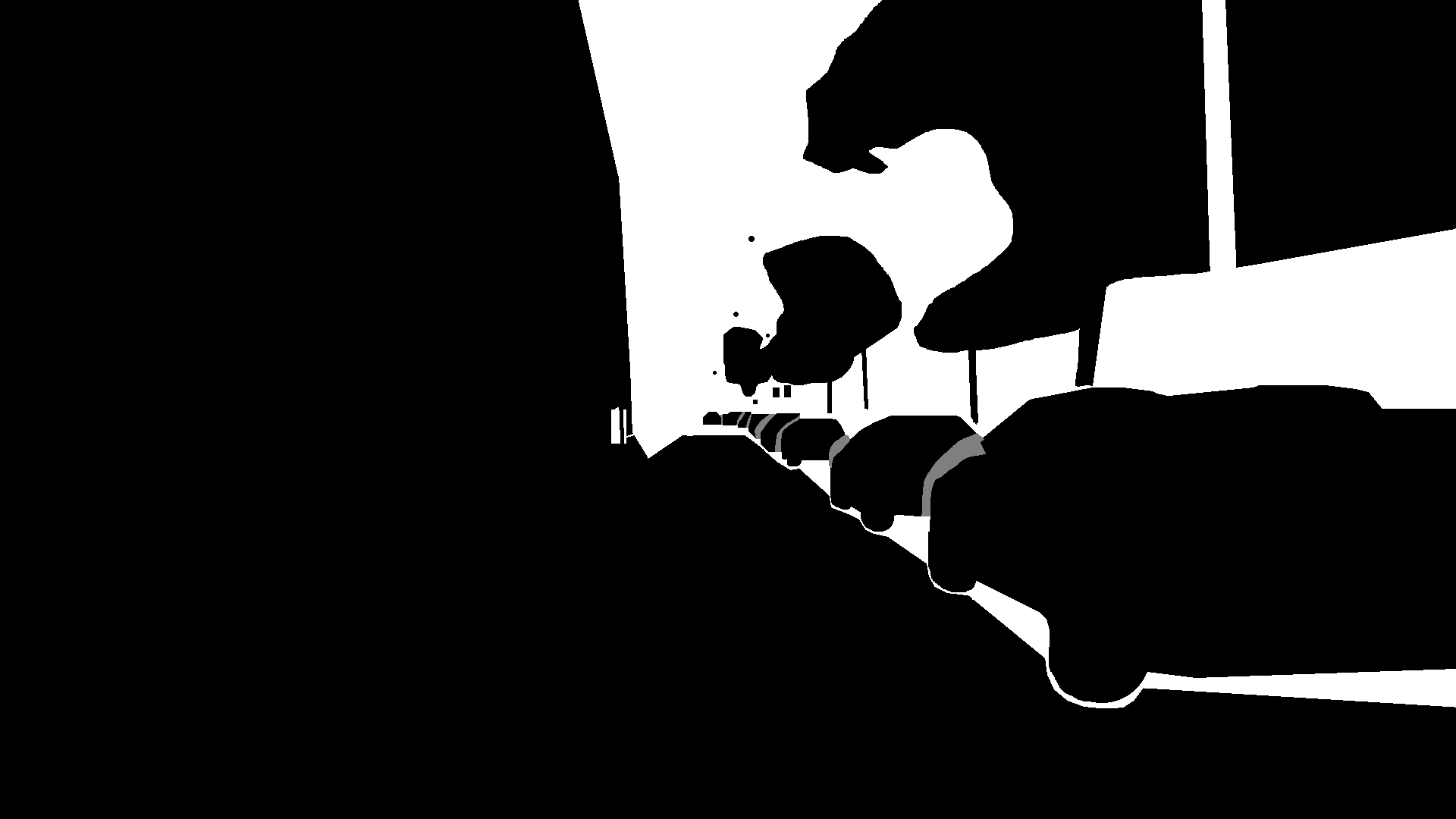}\\

% Snow Day 1
\includegraphics[width=0.14\textwidth]{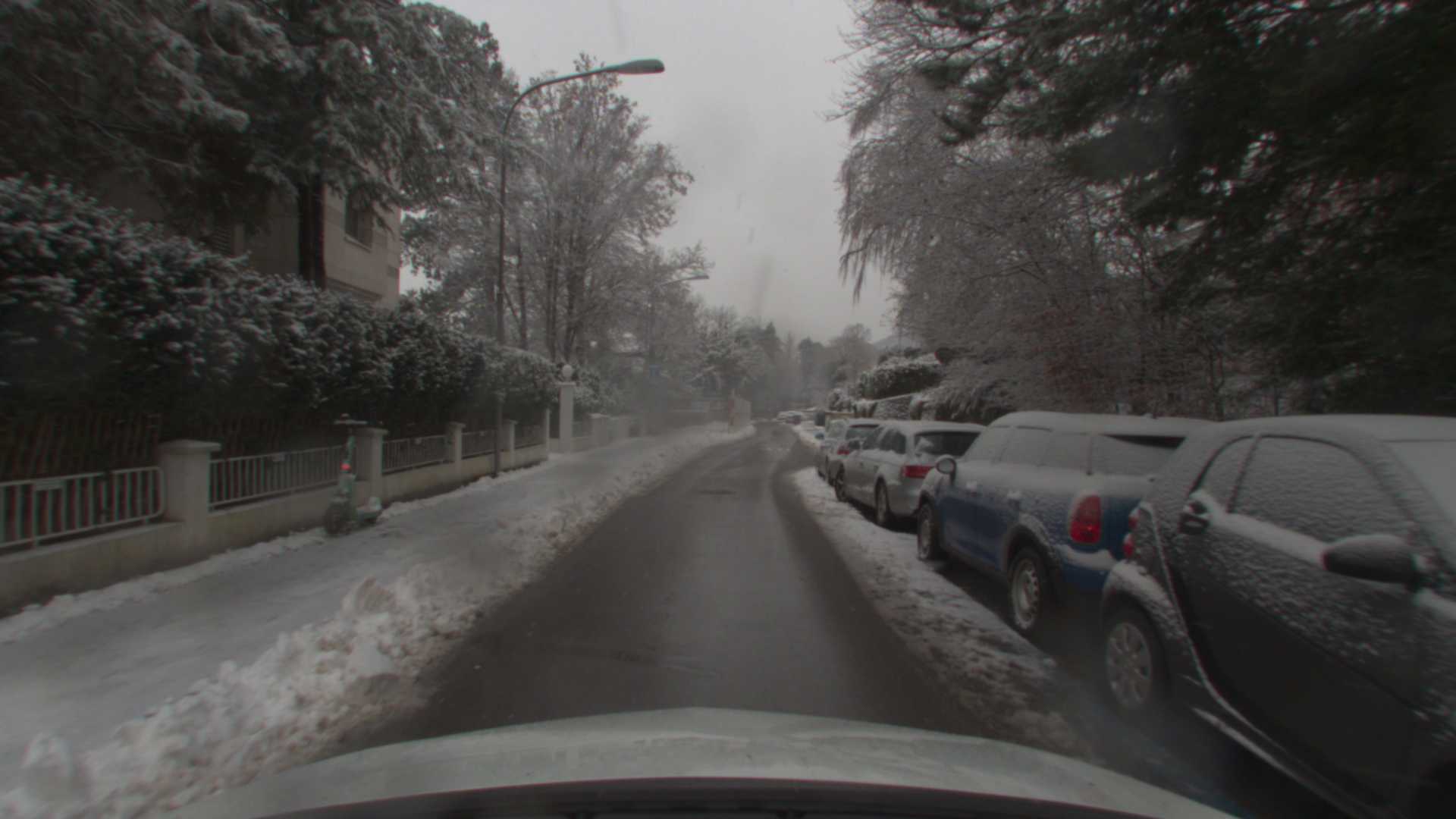} &
\includegraphics[width=0.14\textwidth]{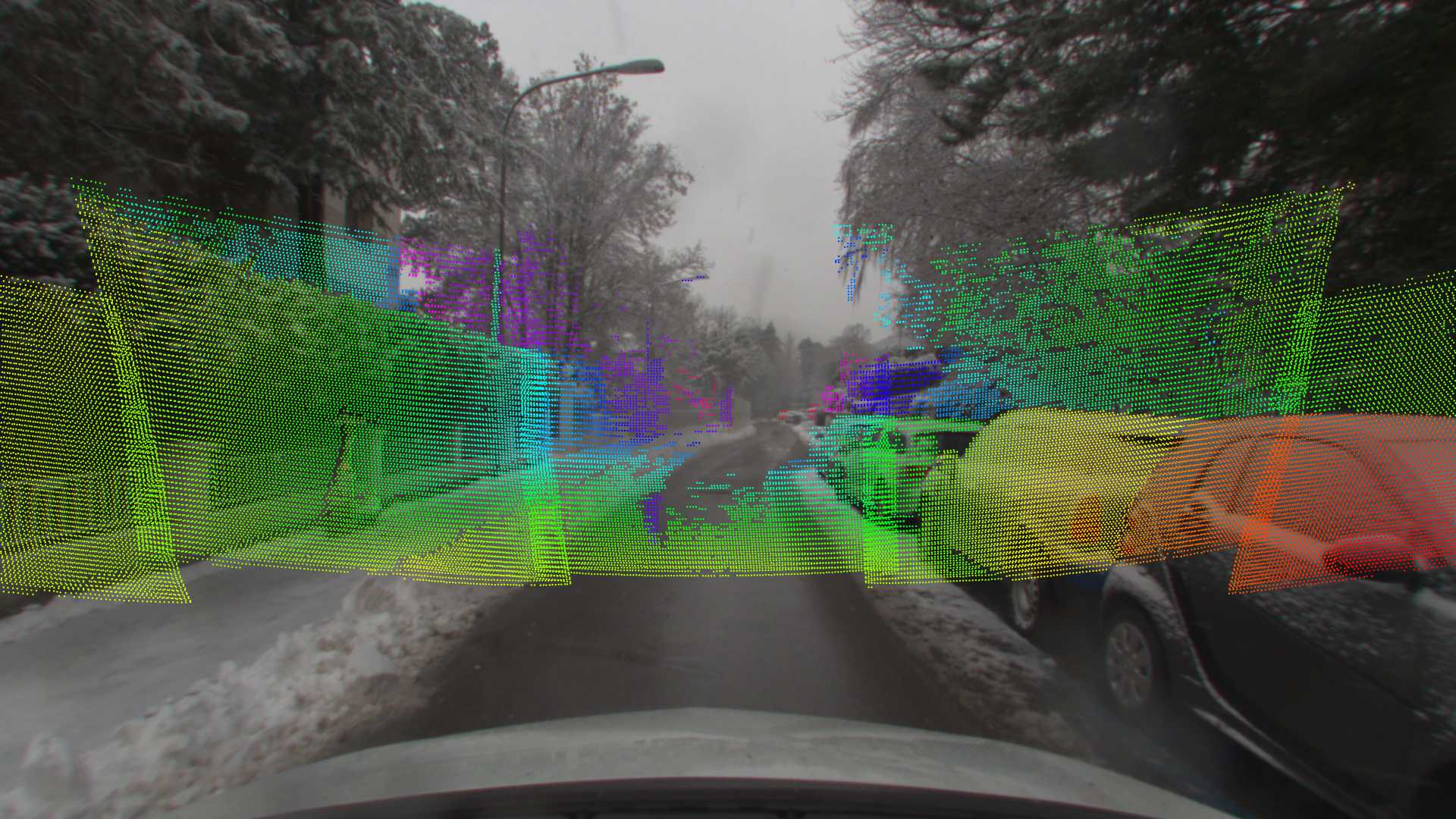} &
\includegraphics[width=0.14\textwidth]{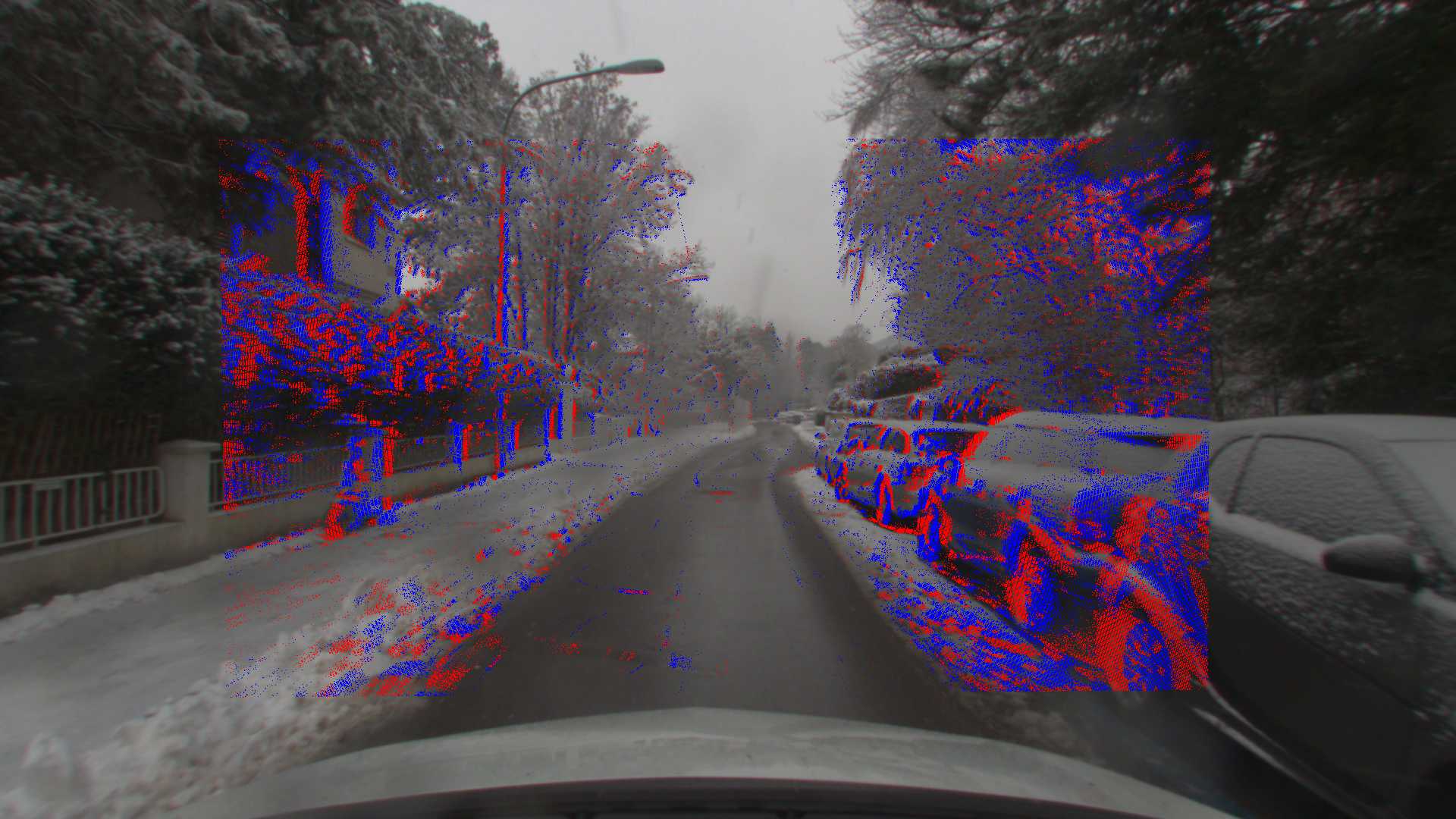} &
\includegraphics[angle=90, trim=0 6835 0 0, clip,width=0.14\textwidth]{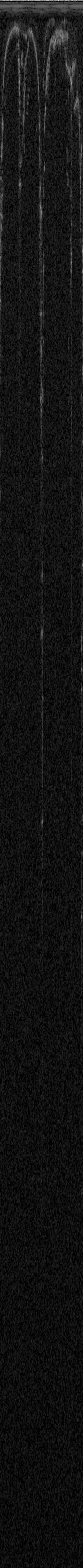}&
\includegraphics[width=0.14\textwidth]{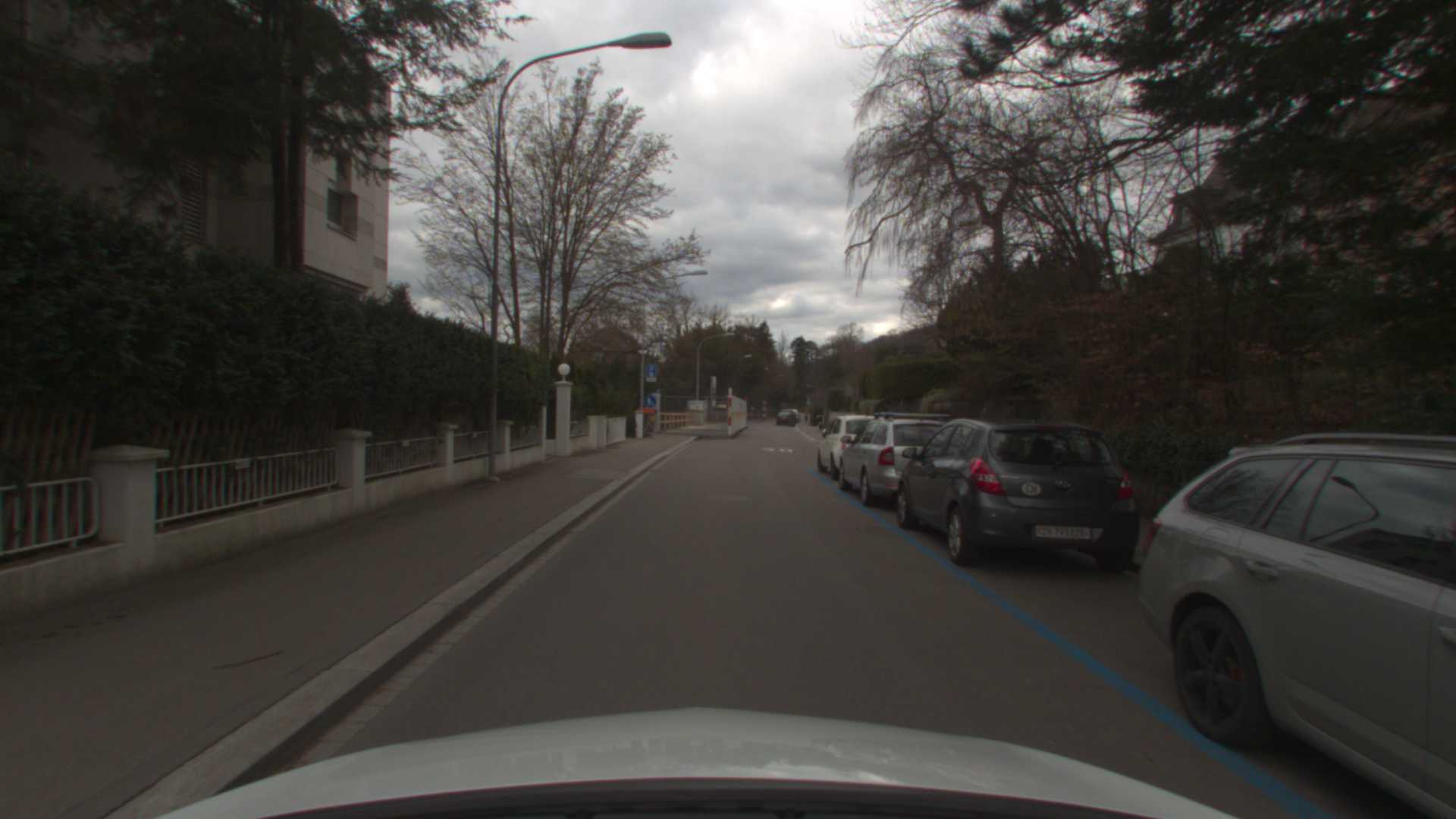} &
\includegraphics[width=0.14\textwidth]{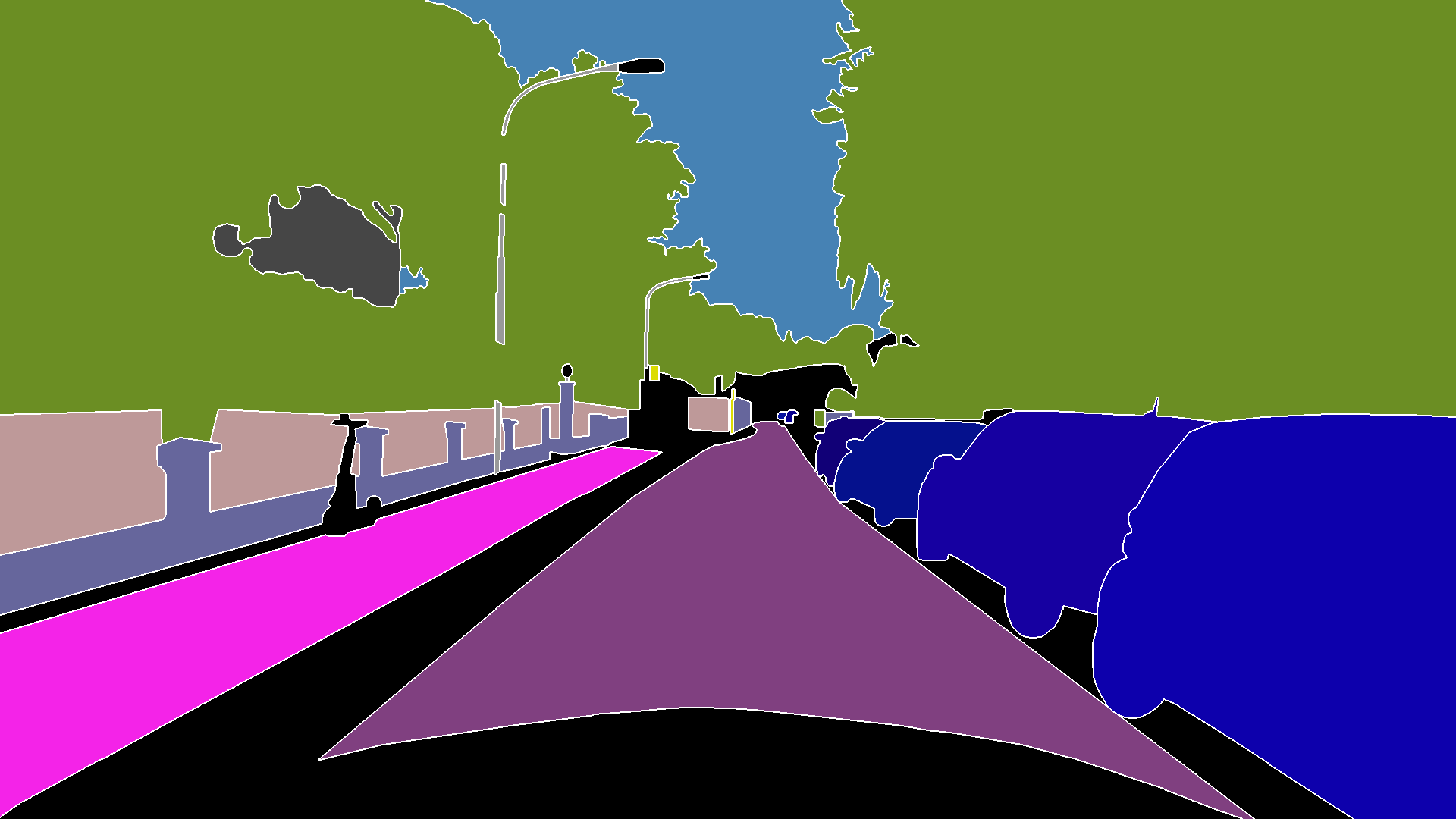} &
\includegraphics[width=0.14\textwidth]{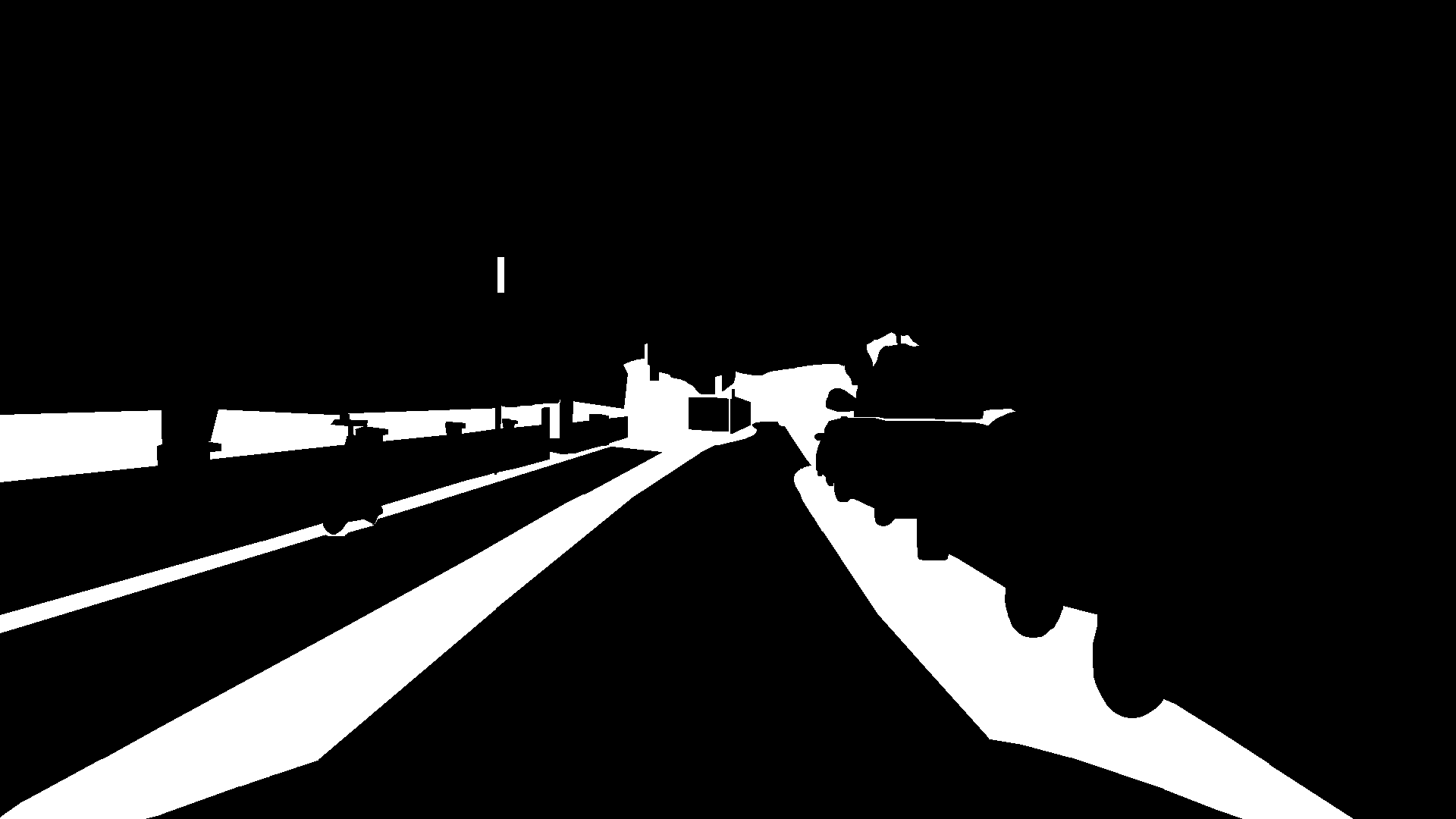}\\

% Snow Day 2
\includegraphics[width=0.14\textwidth]{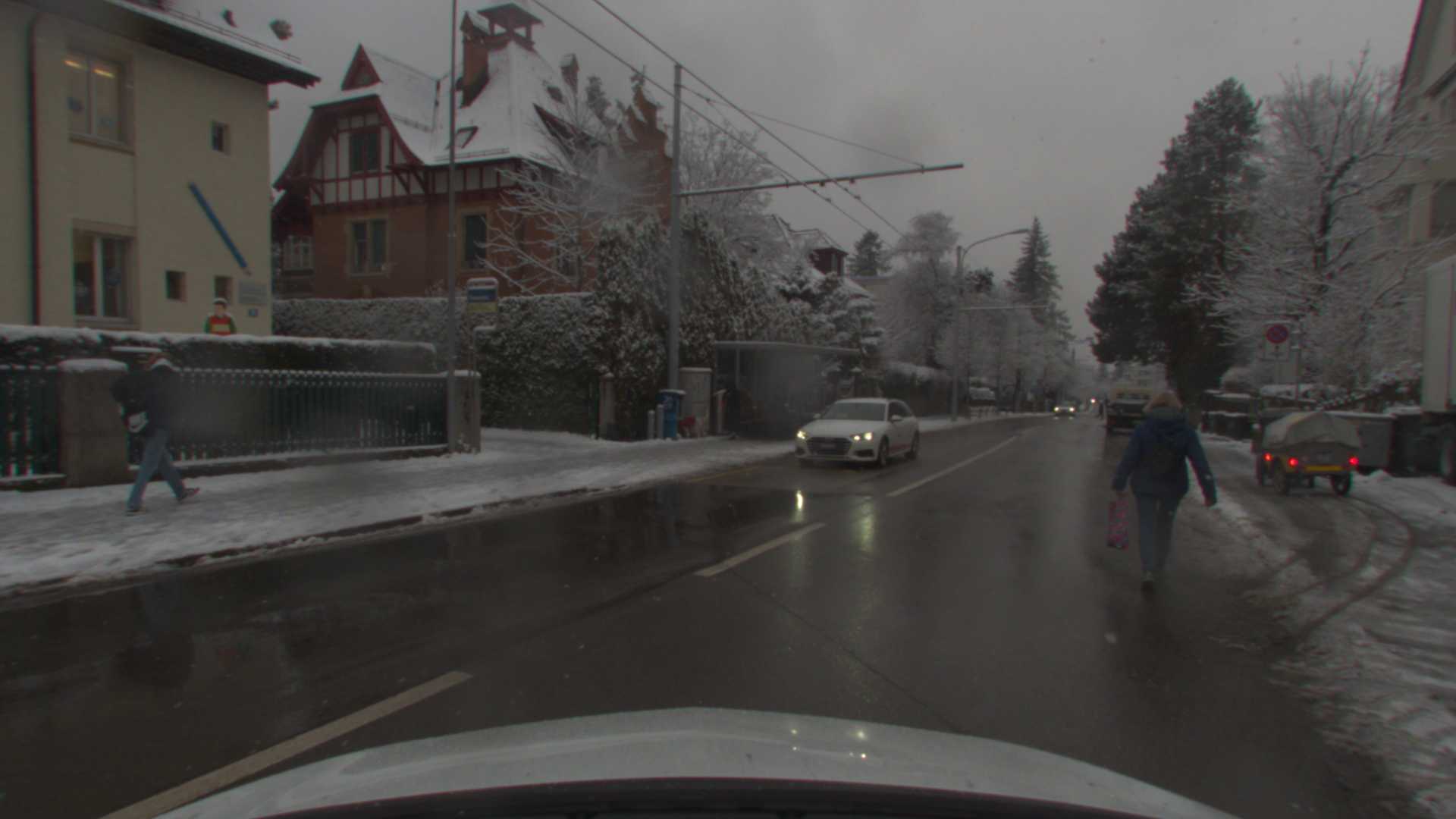} &
\includegraphics[width=0.14\textwidth]{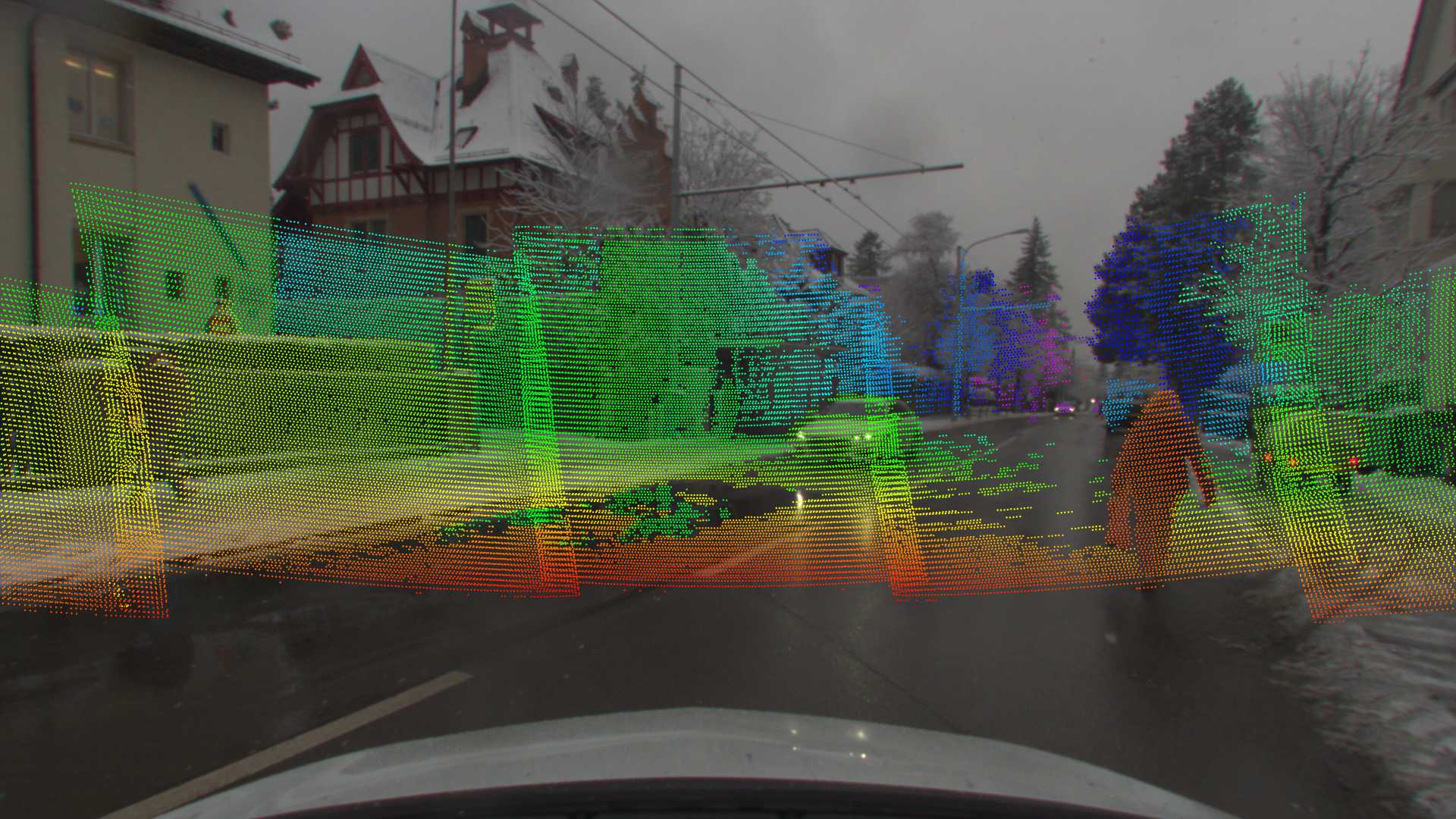} &
\includegraphics[width=0.14\textwidth]{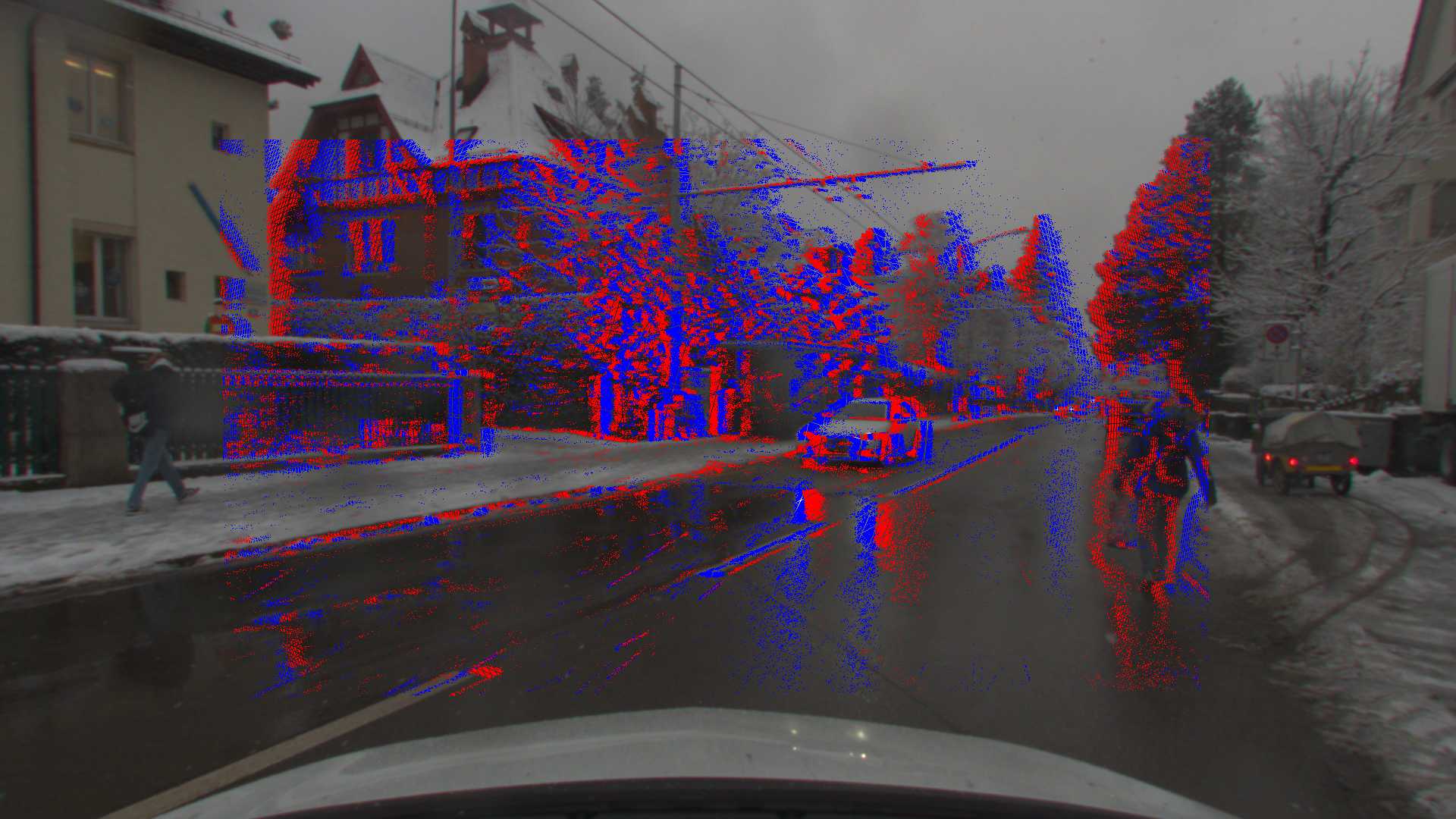} &
\includegraphics[angle=90, trim=0 6835 0 0, clip,width=0.14\textwidth]{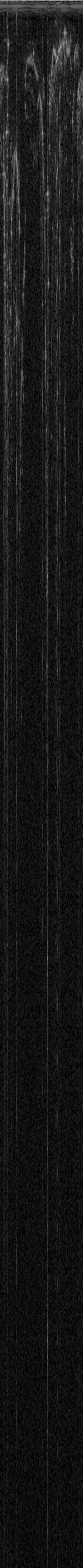}&
\includegraphics[width=0.14\textwidth]{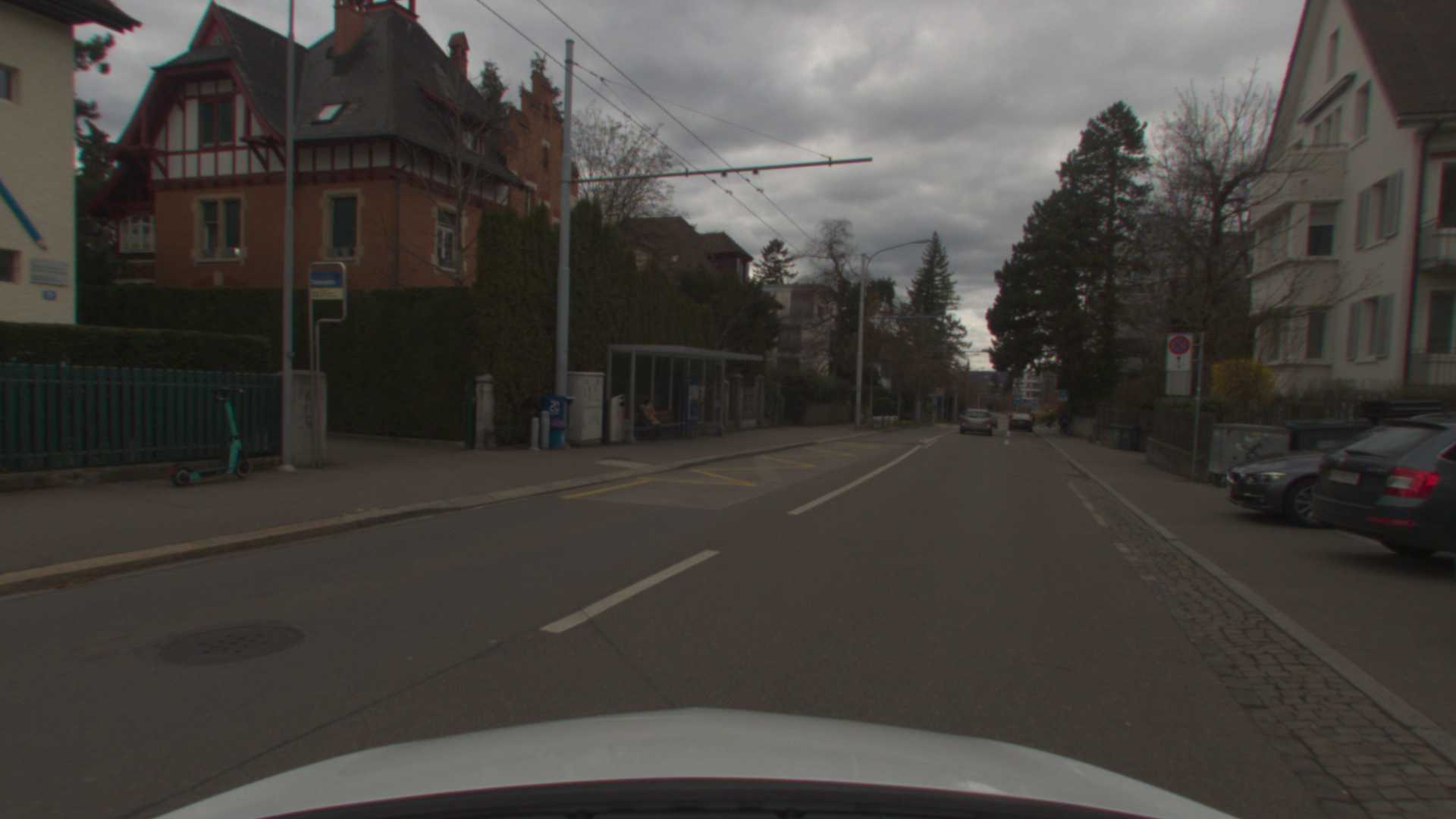} &
\includegraphics[width=0.14\textwidth]{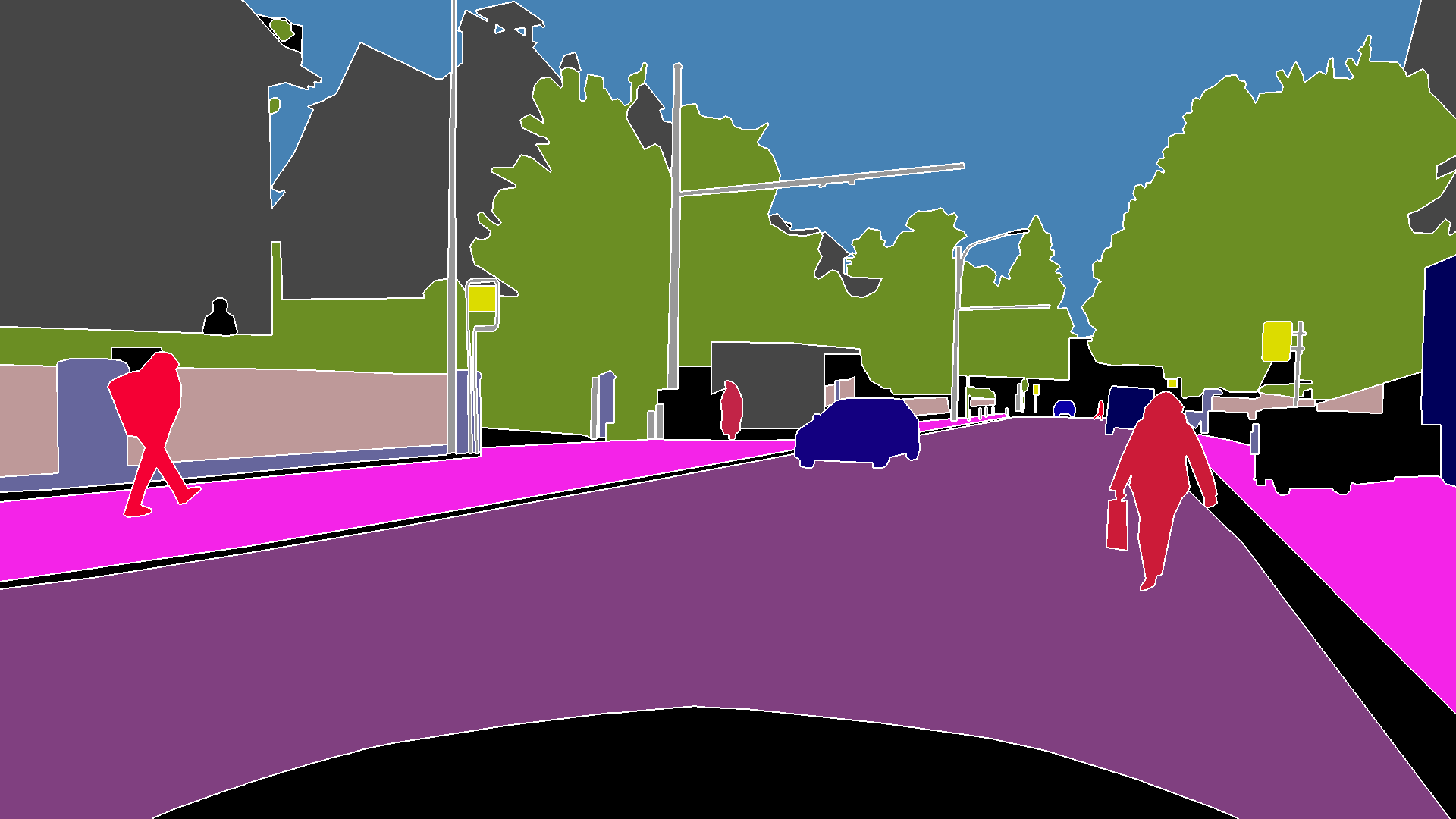} &
\includegraphics[width=0.14\textwidth]{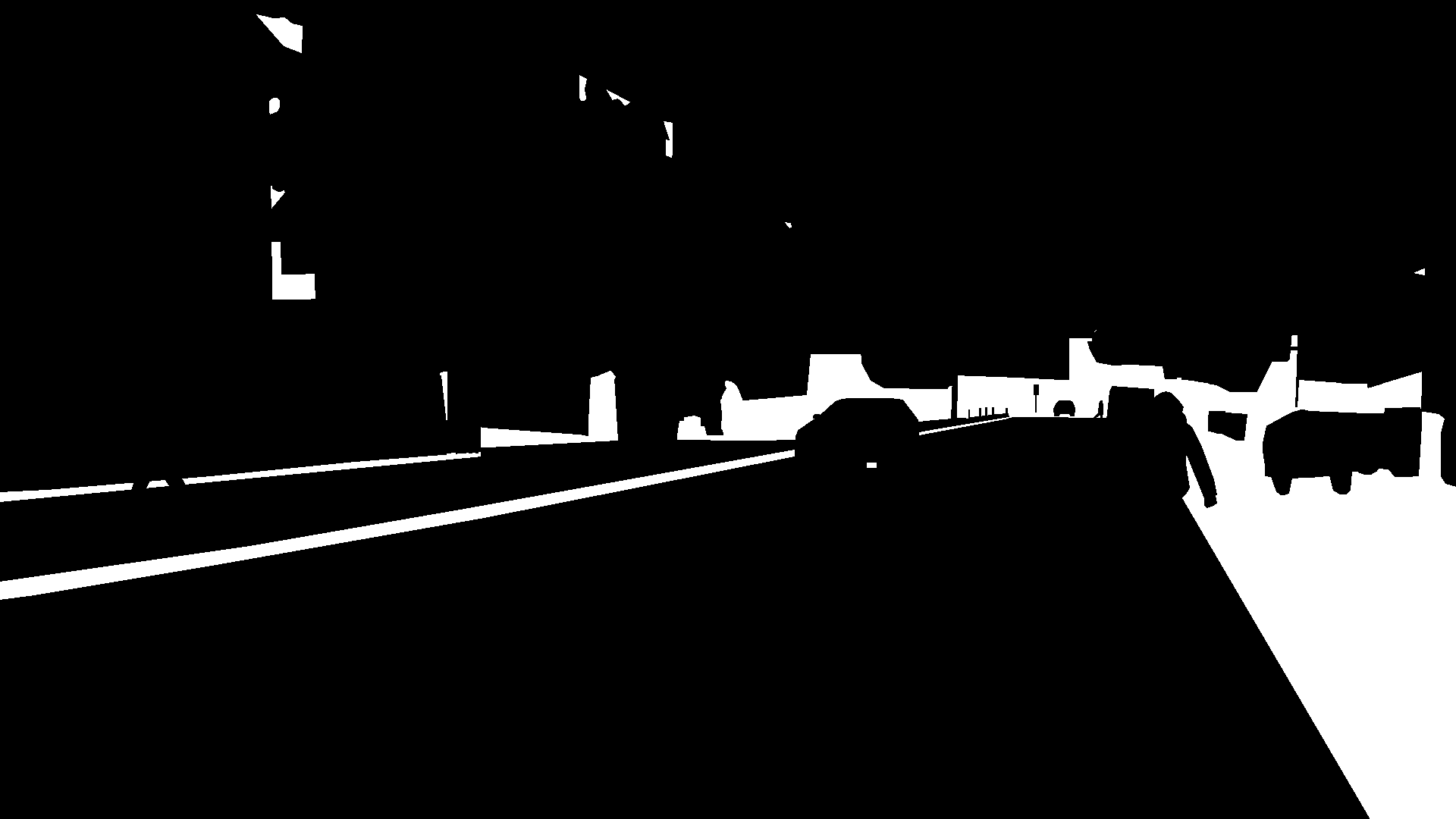}\\

% Snow Night 1
\includegraphics[width=0.14\textwidth]{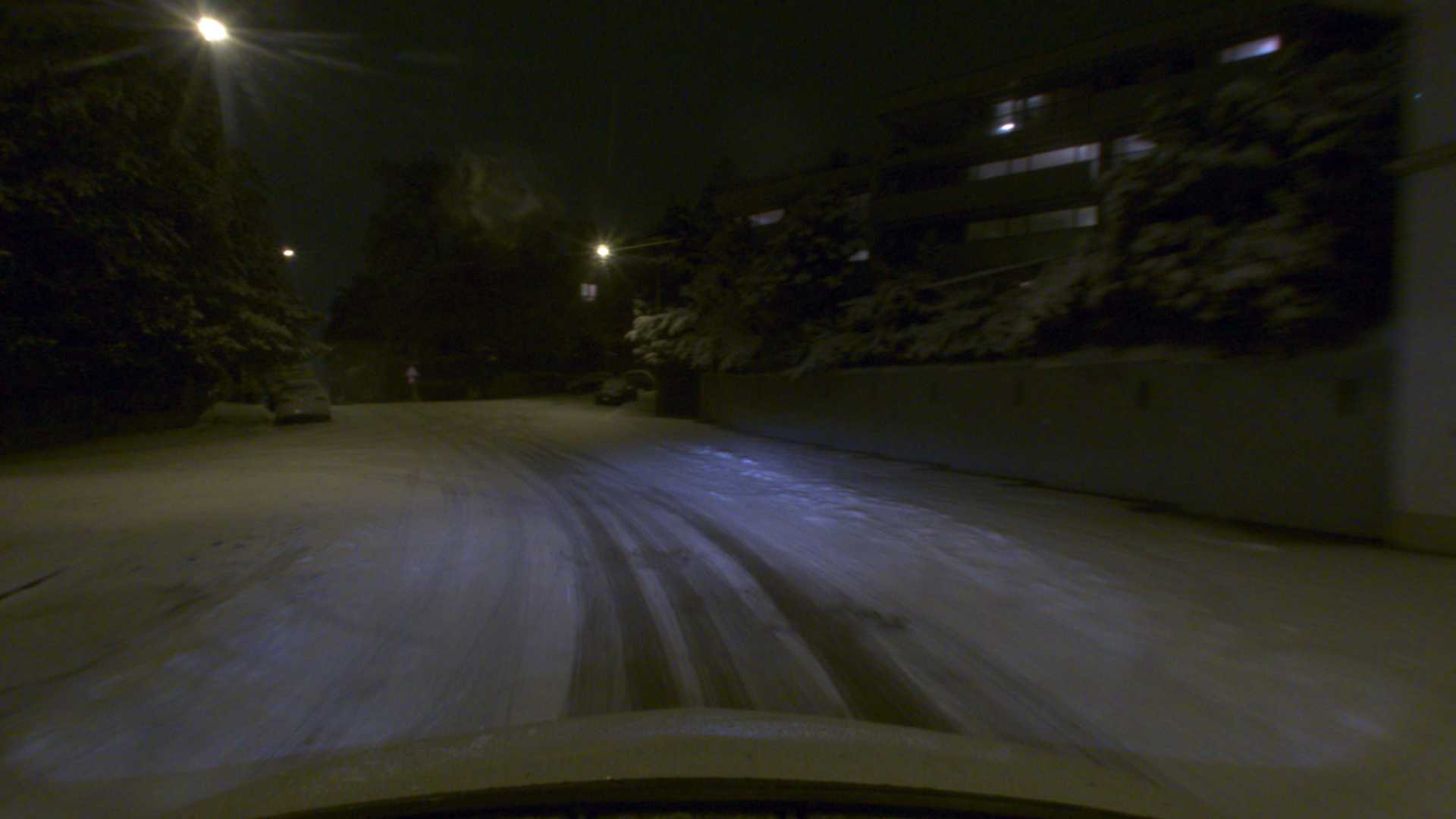} &
\includegraphics[width=0.14\textwidth]{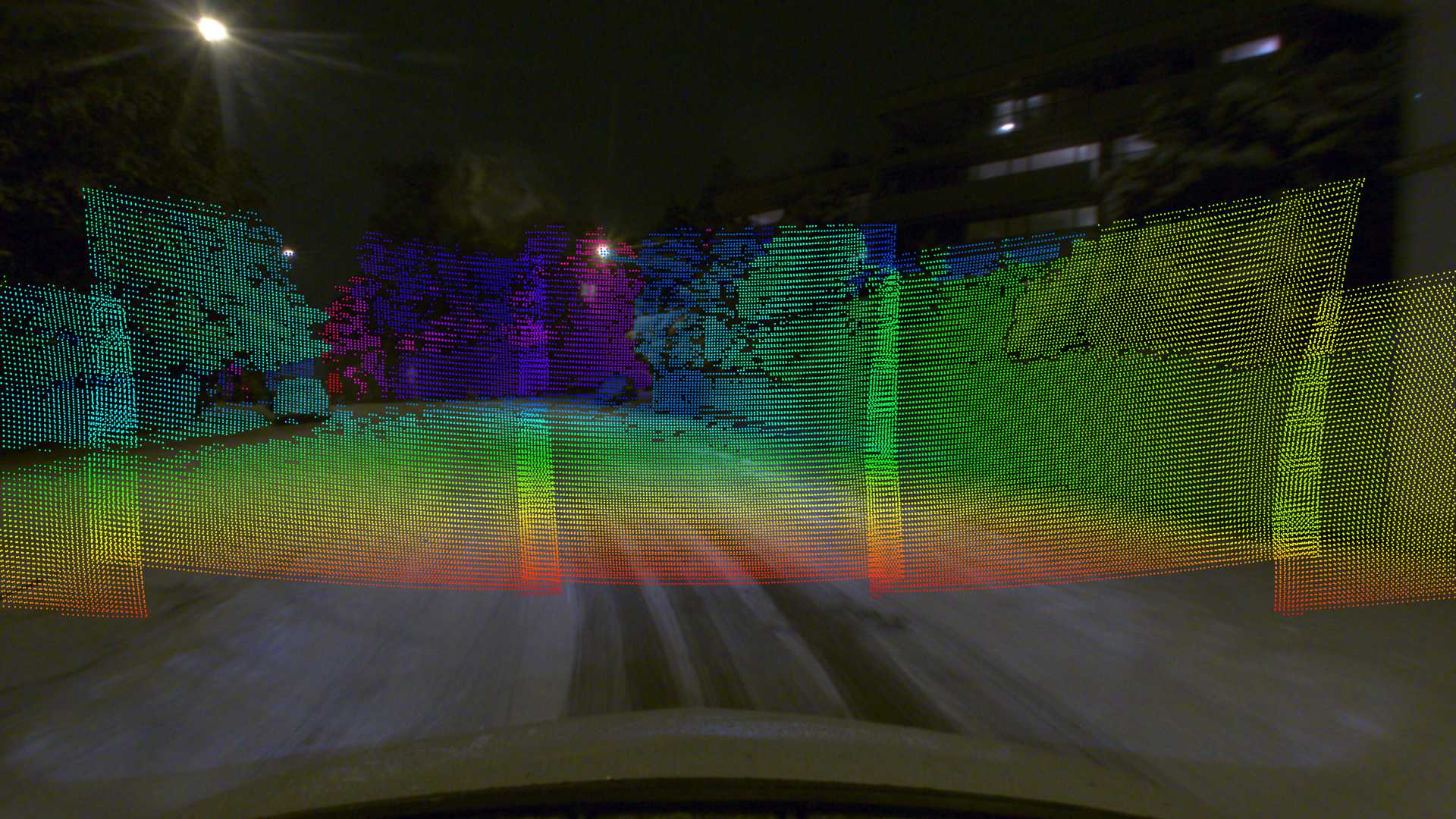} &
\includegraphics[width=0.14\textwidth]{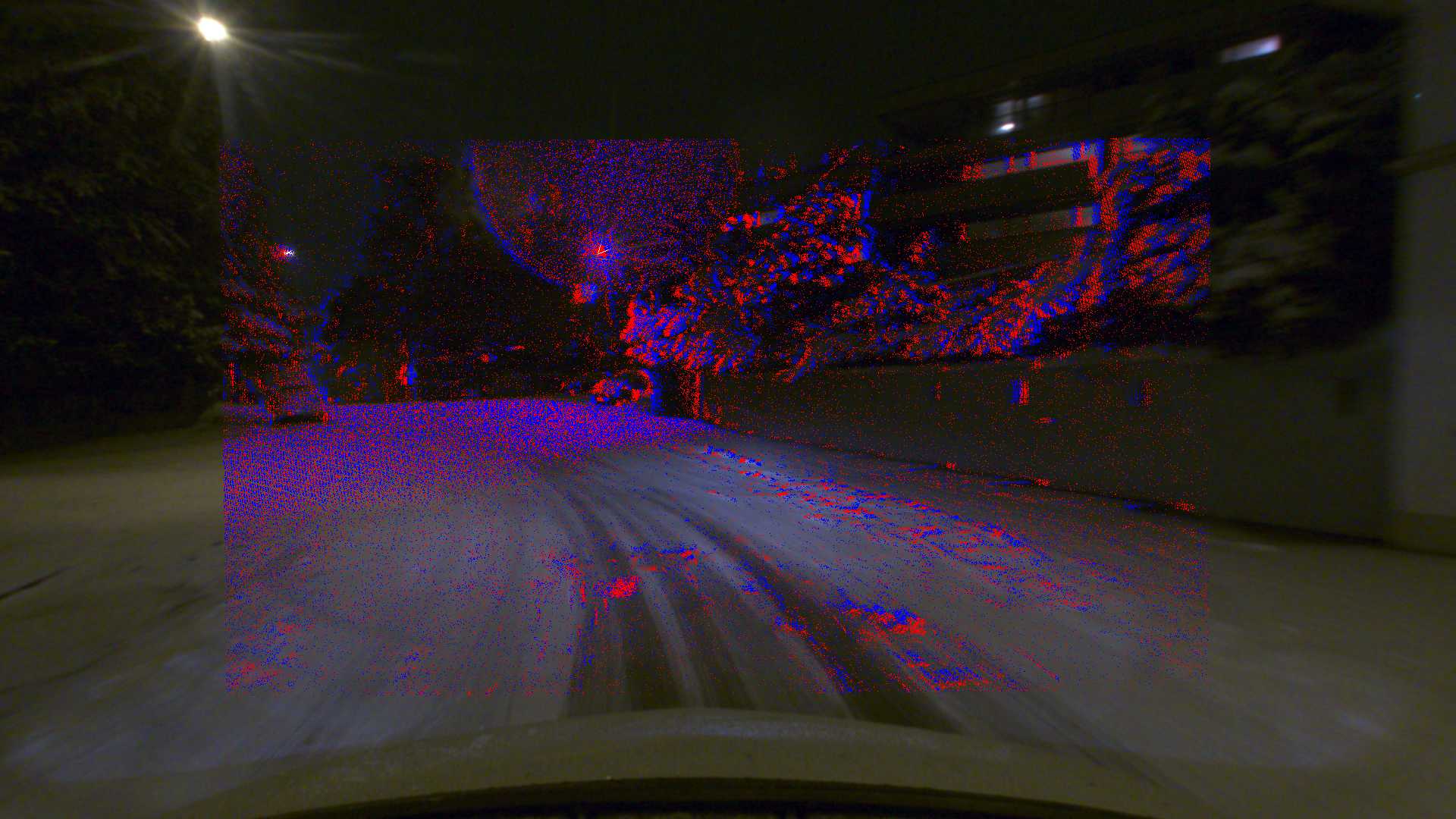} &
\includegraphics[angle=90, trim=0 6835 0 0, clip, width=0.14\textwidth]{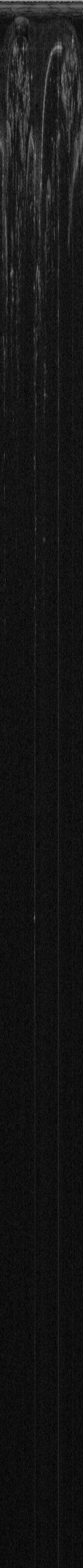}&
\includegraphics[width=0.14\textwidth]{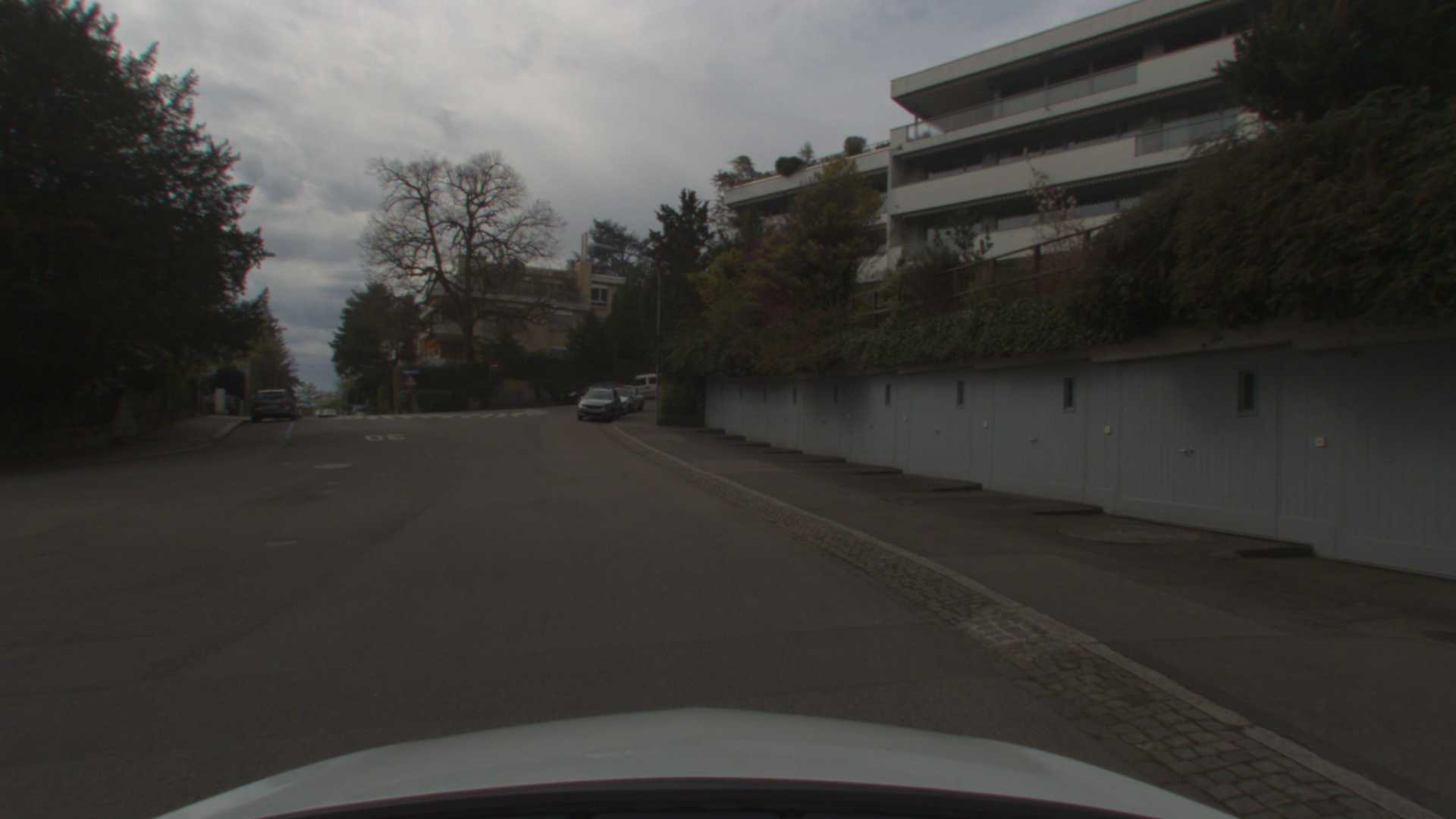} &
\includegraphics[width=0.14\textwidth]{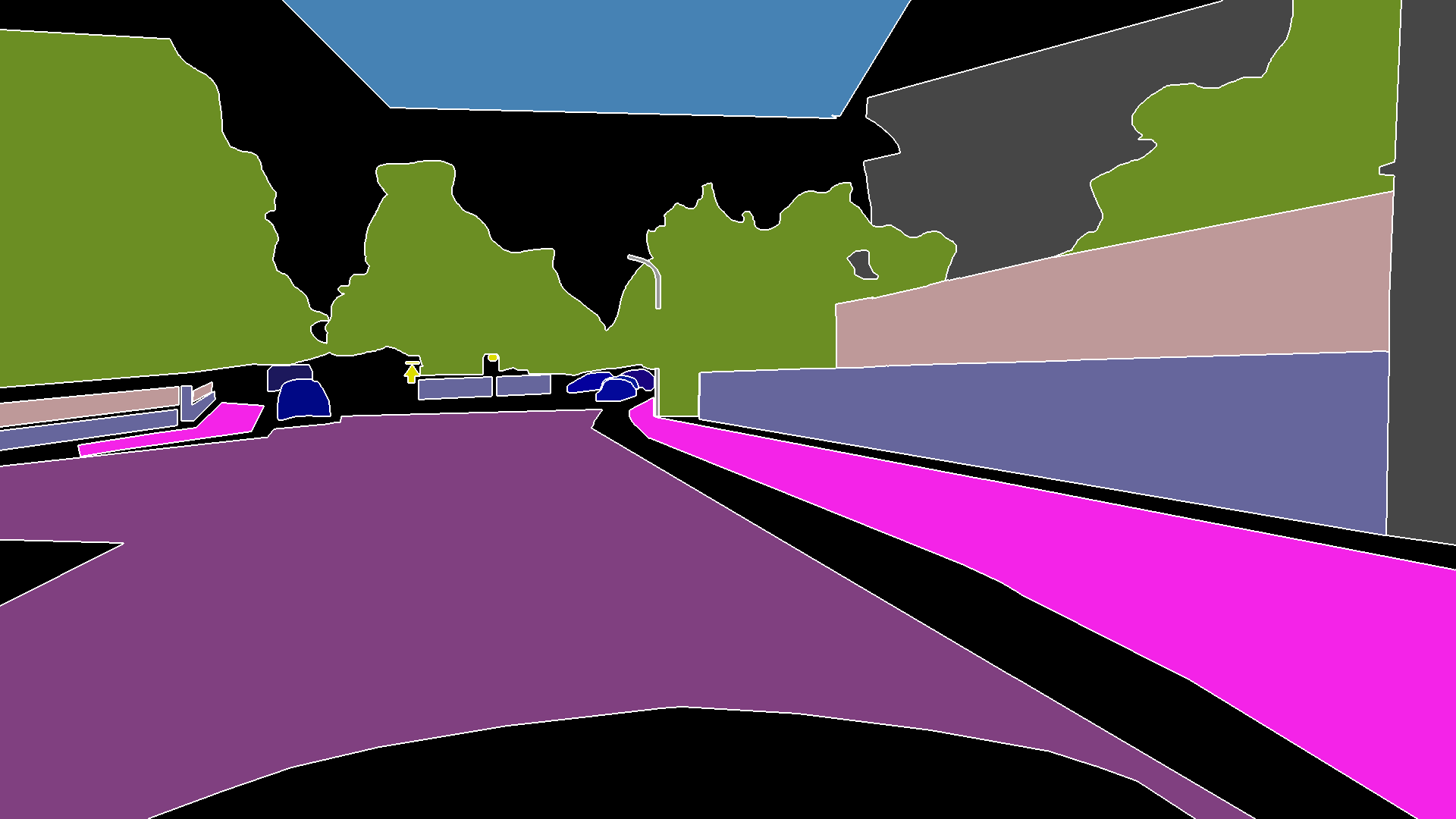} &
\includegraphics[width=0.14\textwidth]{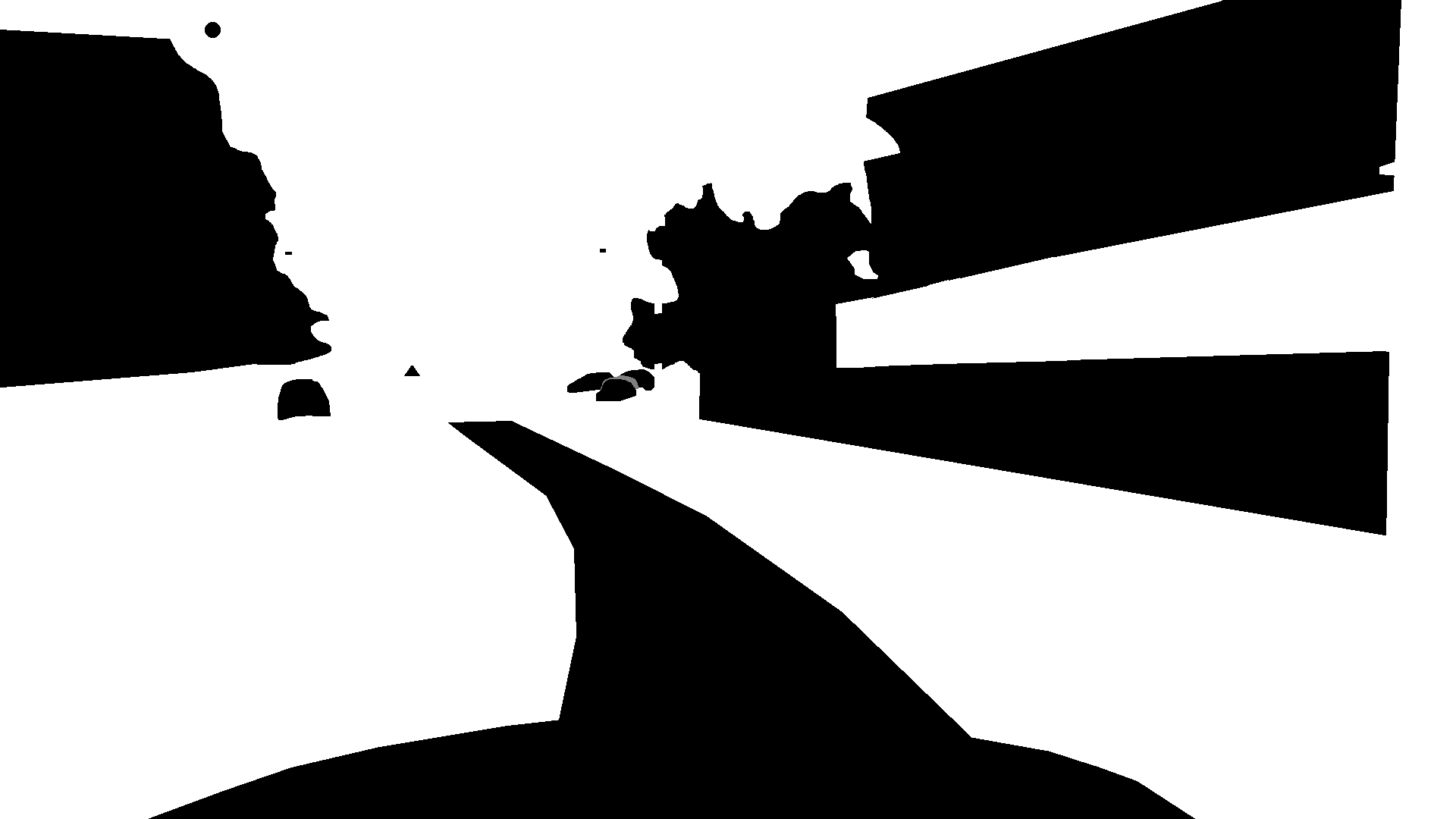}\\

% Snow Night 2
\includegraphics[width=0.14\textwidth]{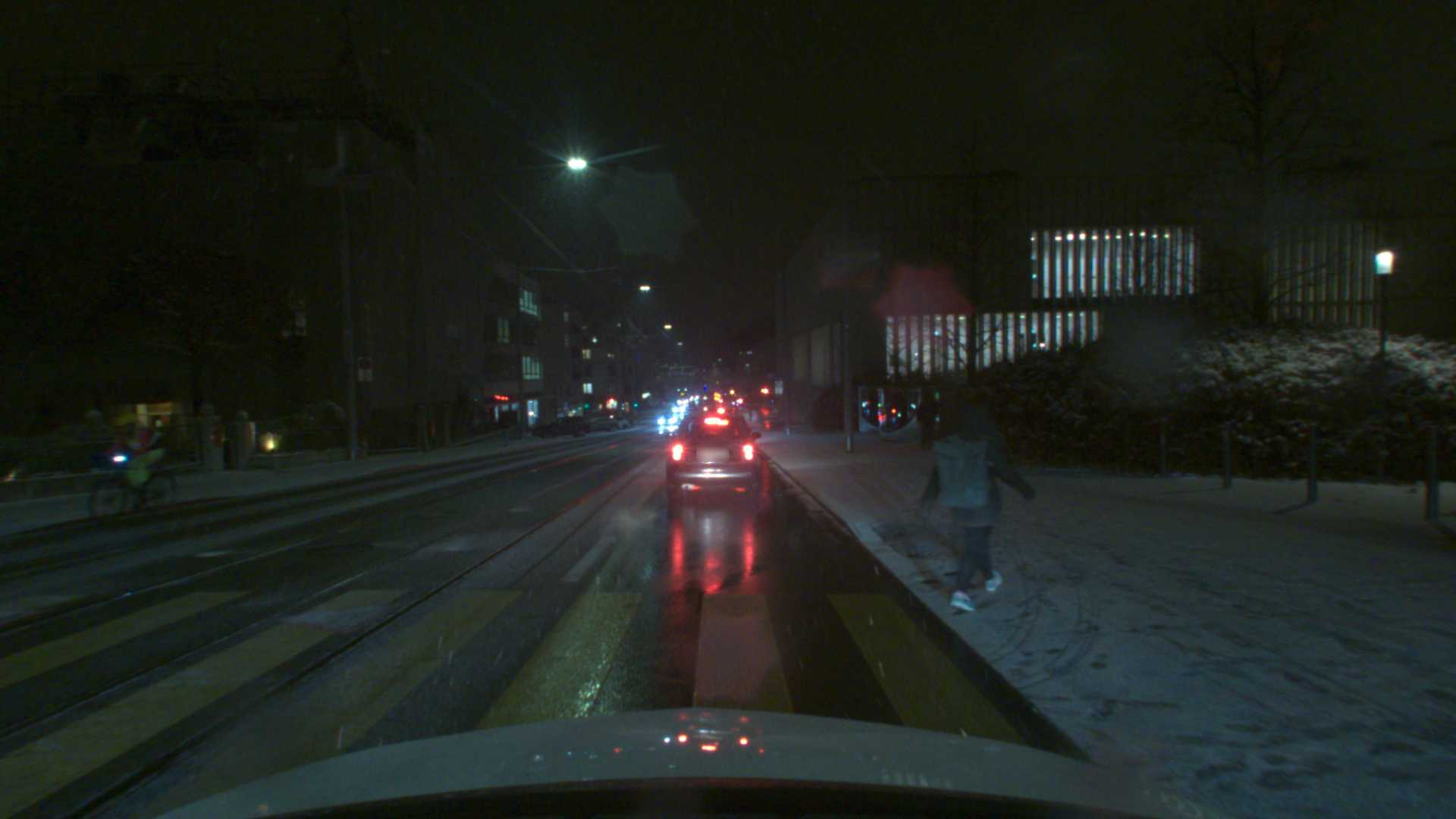} &
\includegraphics[width=0.14\textwidth]{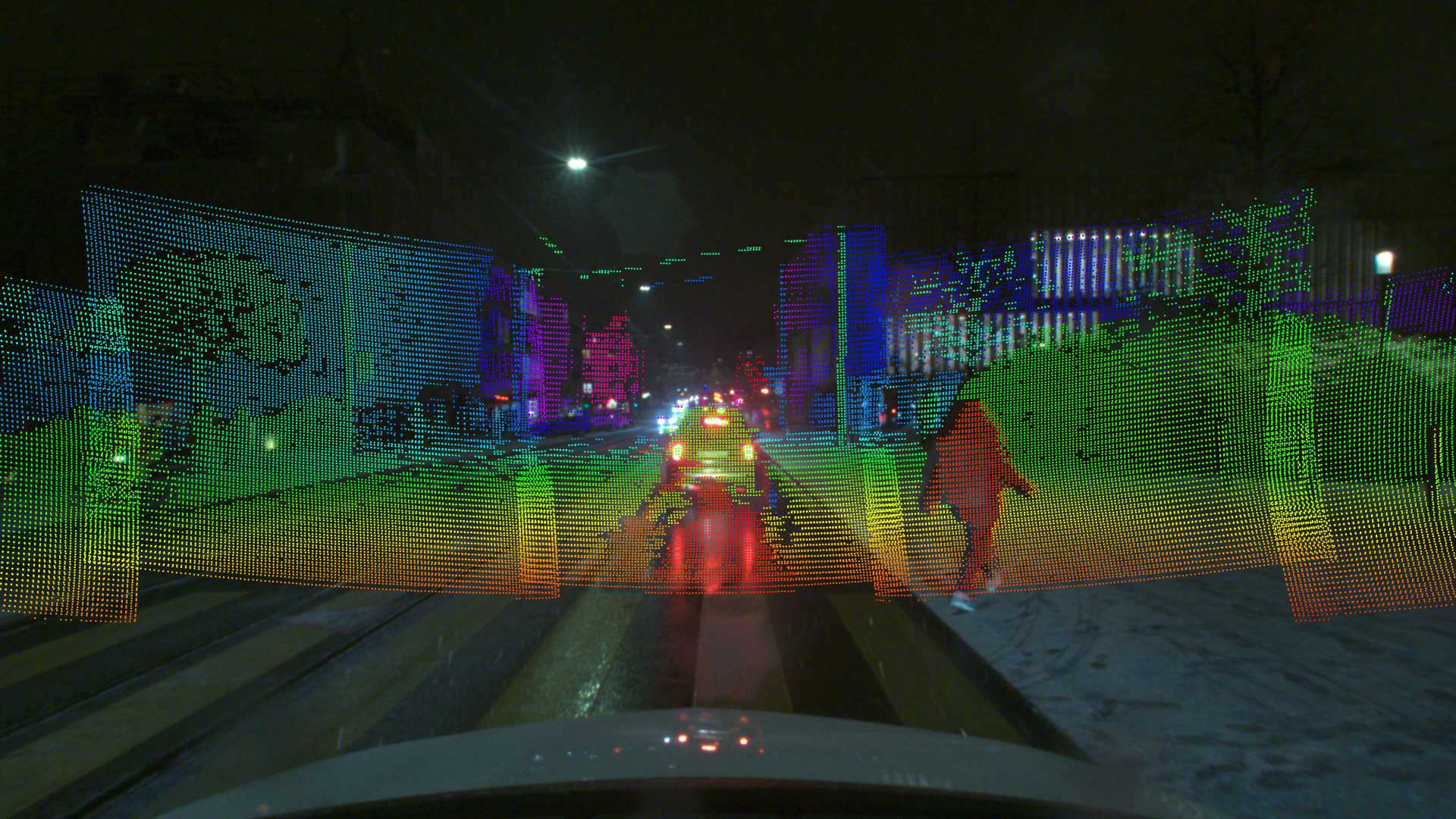} &
\includegraphics[width=0.14\textwidth]{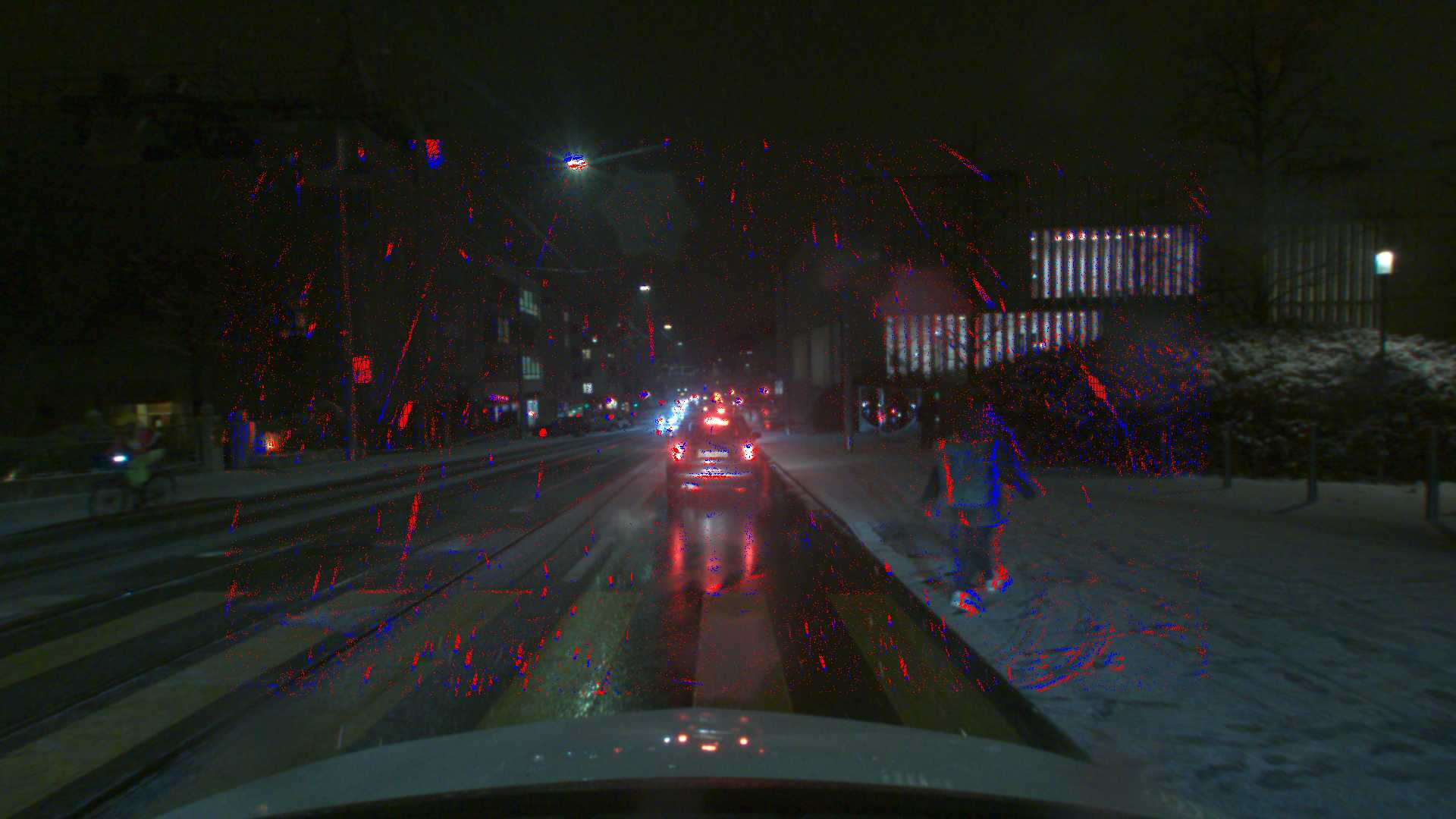} &
\includegraphics[angle=90, trim=0 6835 0 0, clip, width=0.14\textwidth]{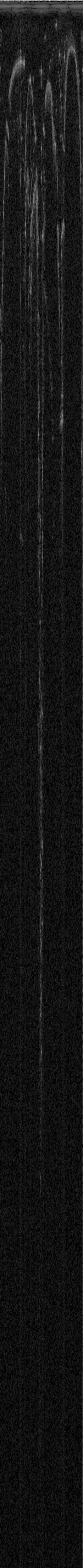}&
\includegraphics[width=0.14\textwidth]{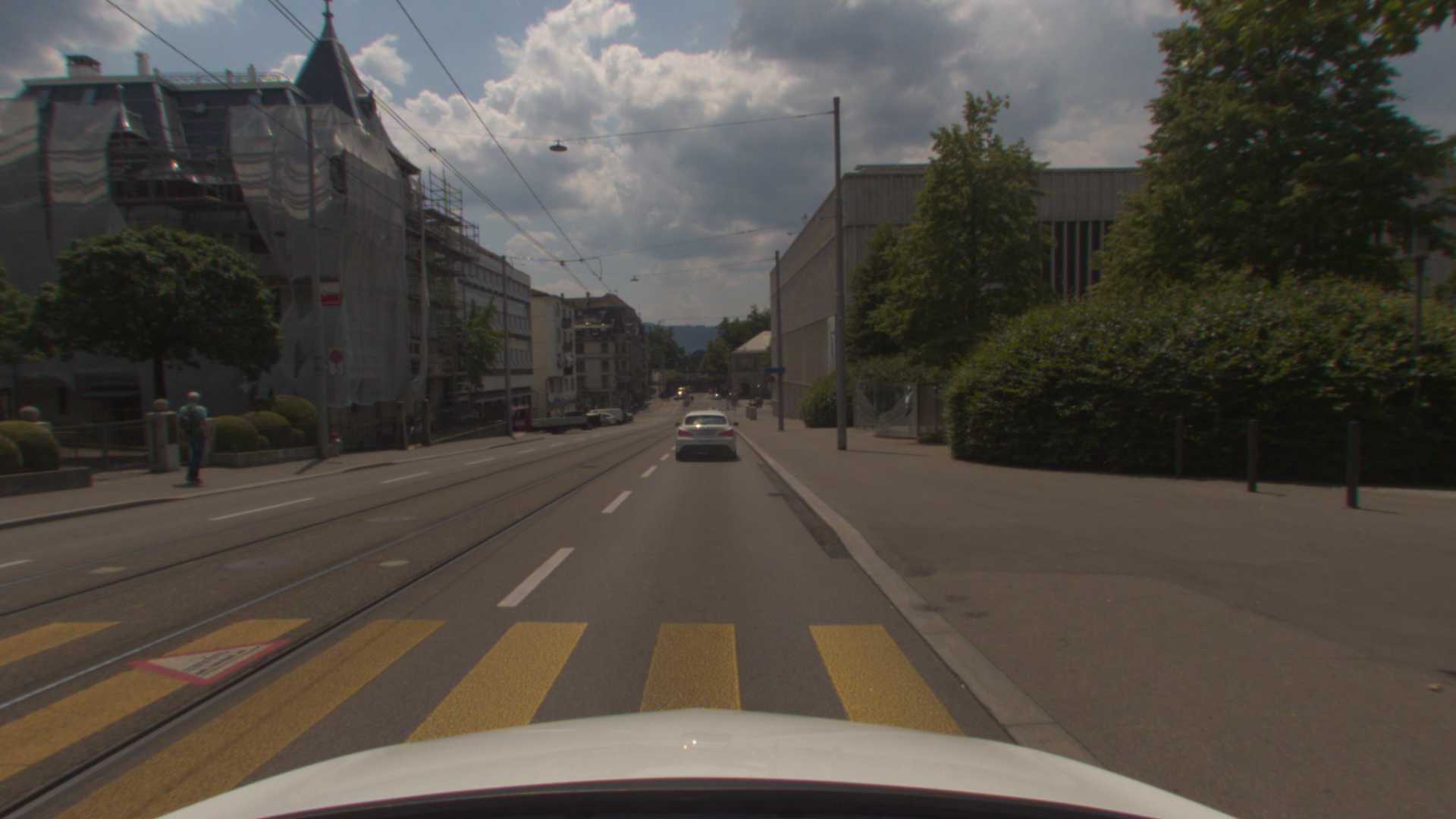} &
\includegraphics[width=0.14\textwidth]{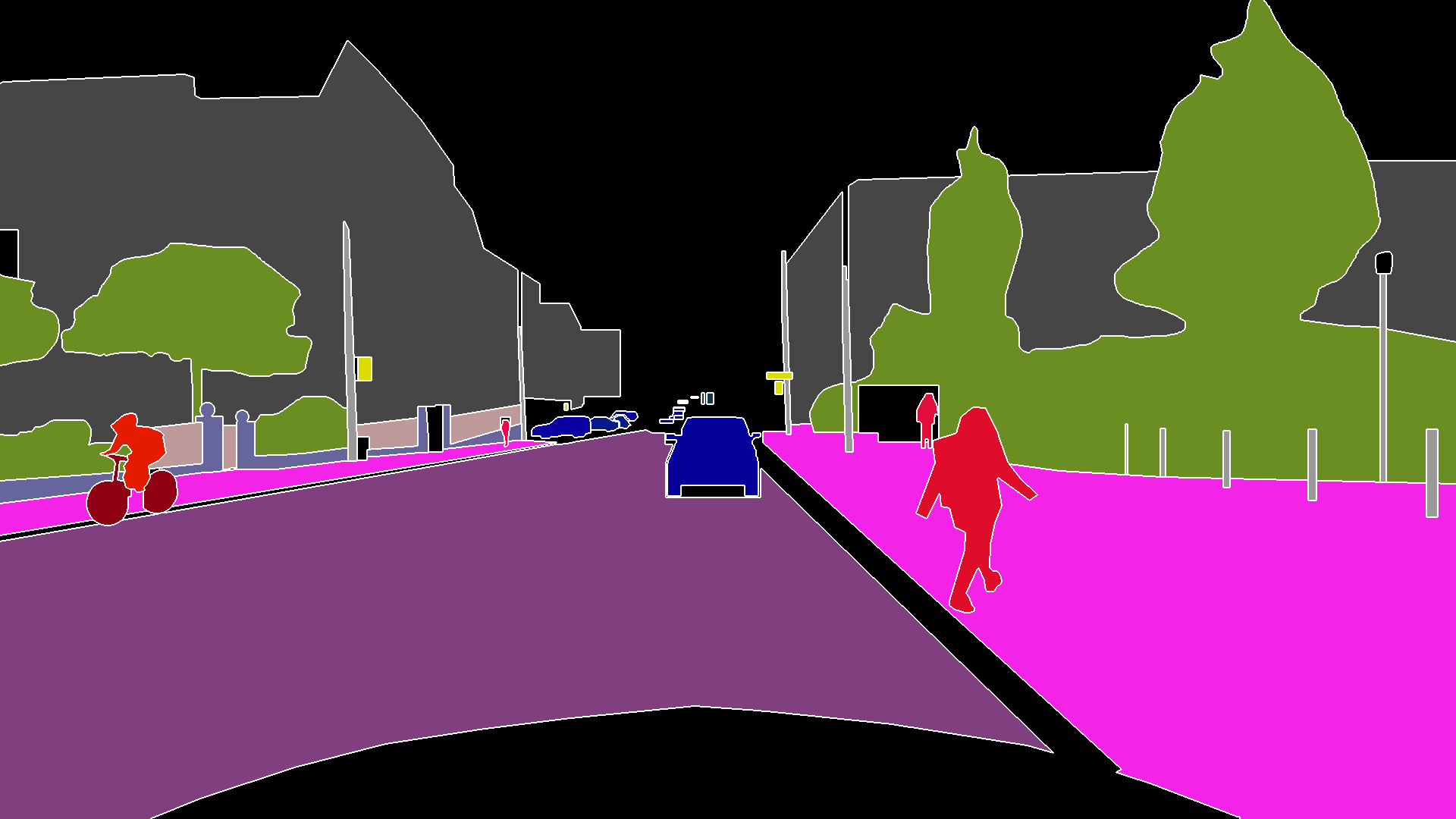} &
\includegraphics[width=0.14\textwidth]{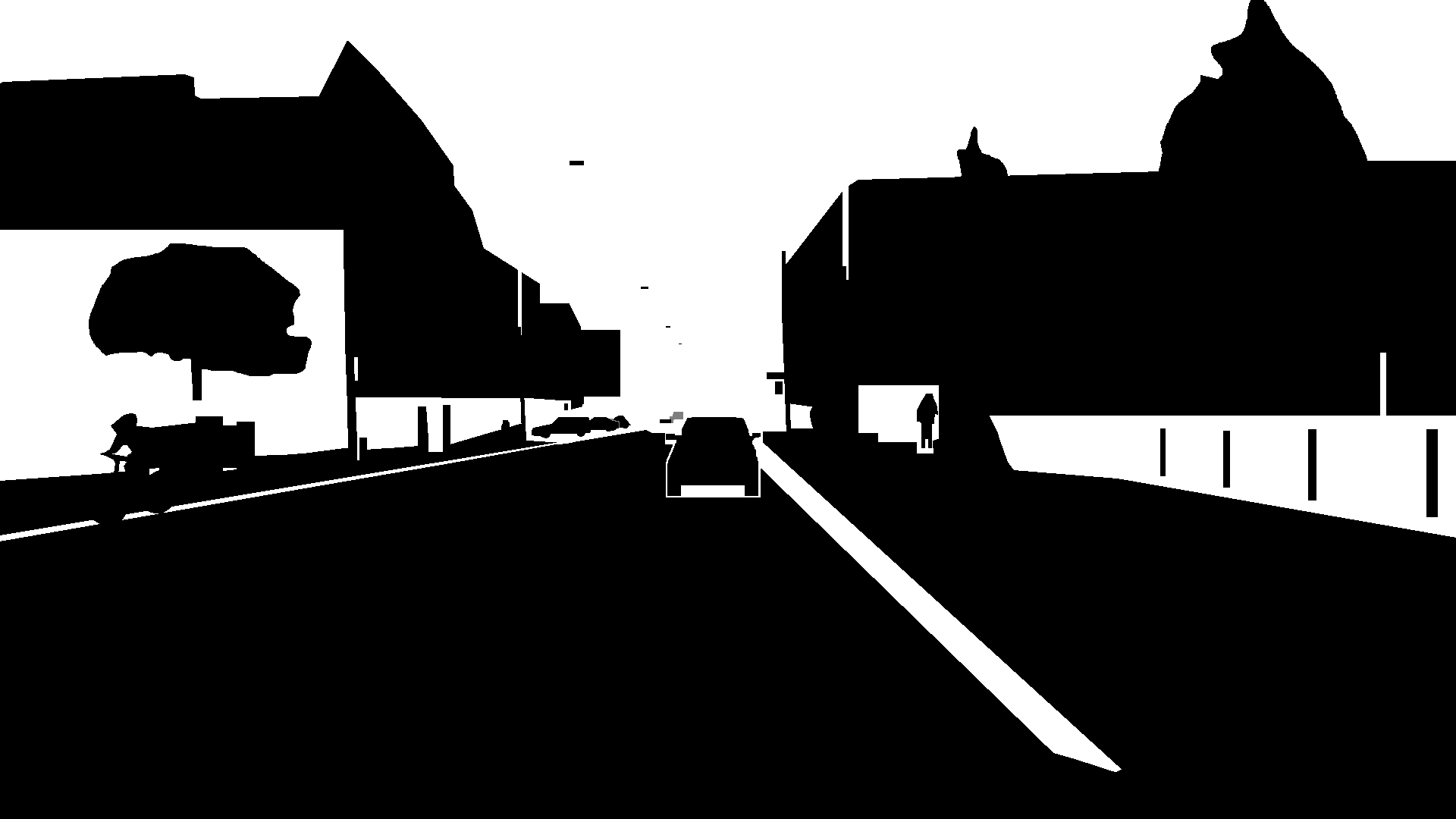}\\
\end{tabular}
\caption{\textbf{Visualization of \Ours{} samples (continued).} From left to right: RGB image; motion-compensated lidar points projected and overlaid with the image; events projected onto the image (assuming infinite distance); azimuth-range radar scan (with ranges above a threshold cropped out); corresponding normal-condition image; panoptic ground truth; difficulty map. Best viewed zoomed in.
}
\label{fig:suppl:annotation:aux:data:2}
\end{figure*}

\section{Qualitative Results}

\begin{figure*}
    \centering
    \begin{tabular}{@{}c@{\hspace{0.05cm}}c@{\hspace{0.05cm}}c@{}}
    % \begin{tabular}{@{}c@{}c@{}c@{}}
        \multicolumn{1}{c}{\scriptsize Ground Truth} &
        \multicolumn{1}{c}{\scriptsize Unimodal Mask2Former~\cite{cheng2022masked}} &
        \multicolumn{1}{c}{\scriptsize Quadrimodal Mask2Former~\cite{cheng2022masked}} \\
        \vspace{-0.07cm}        
        \includegraphics[width=0.328\textwidth]{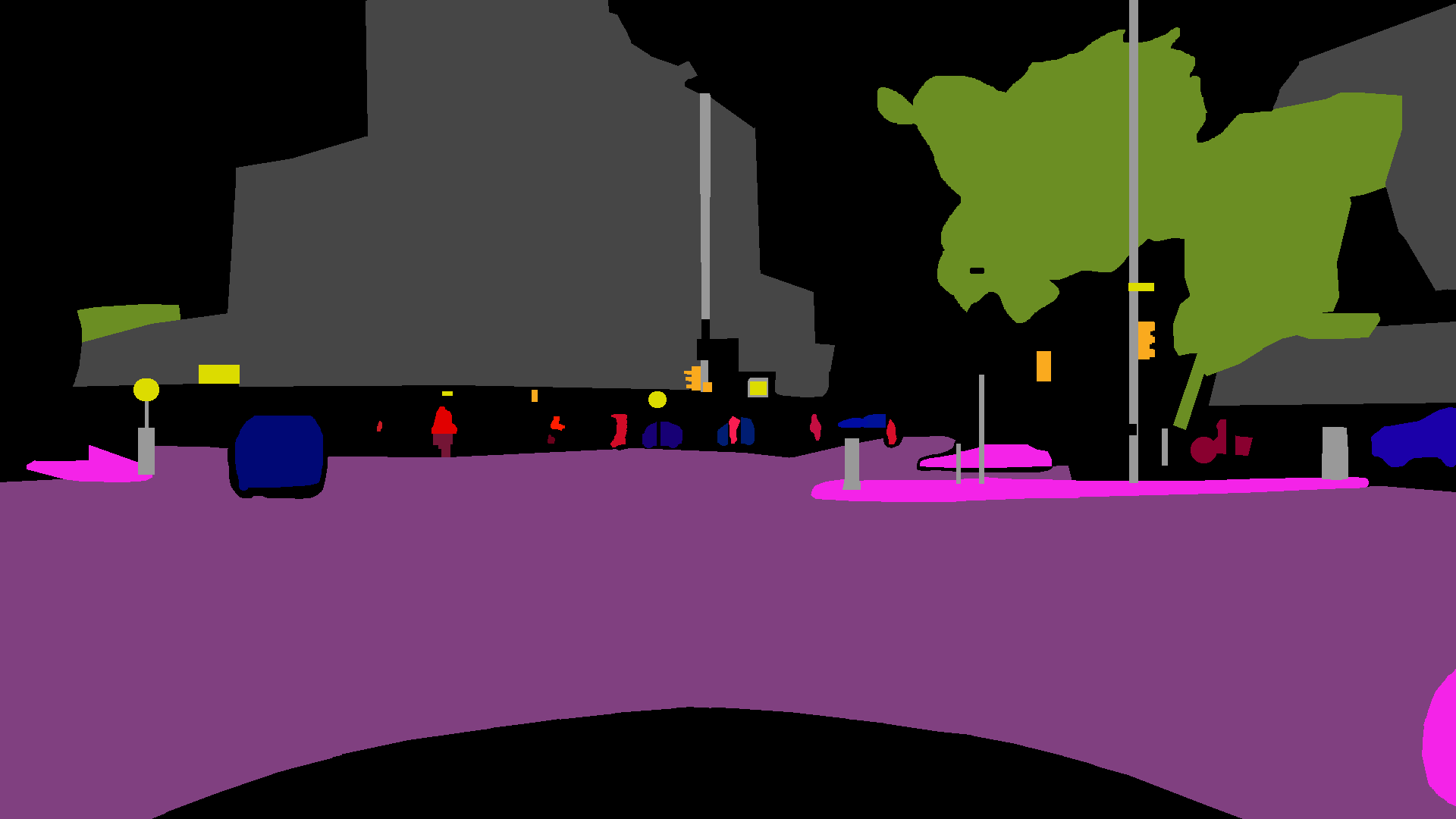} &
        \includegraphics[width=0.328\textwidth]{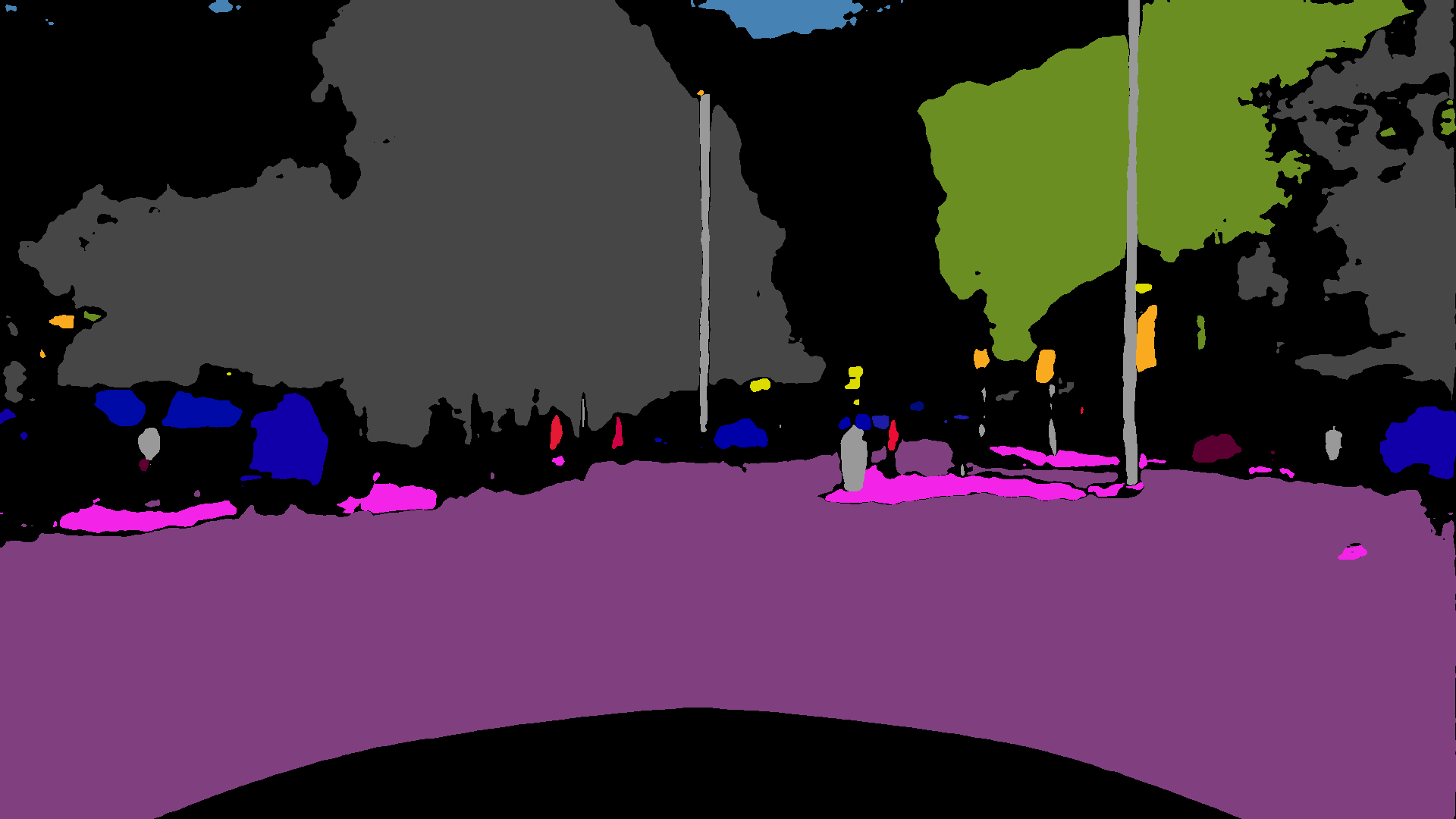} &
        \includegraphics[width=0.328\textwidth]{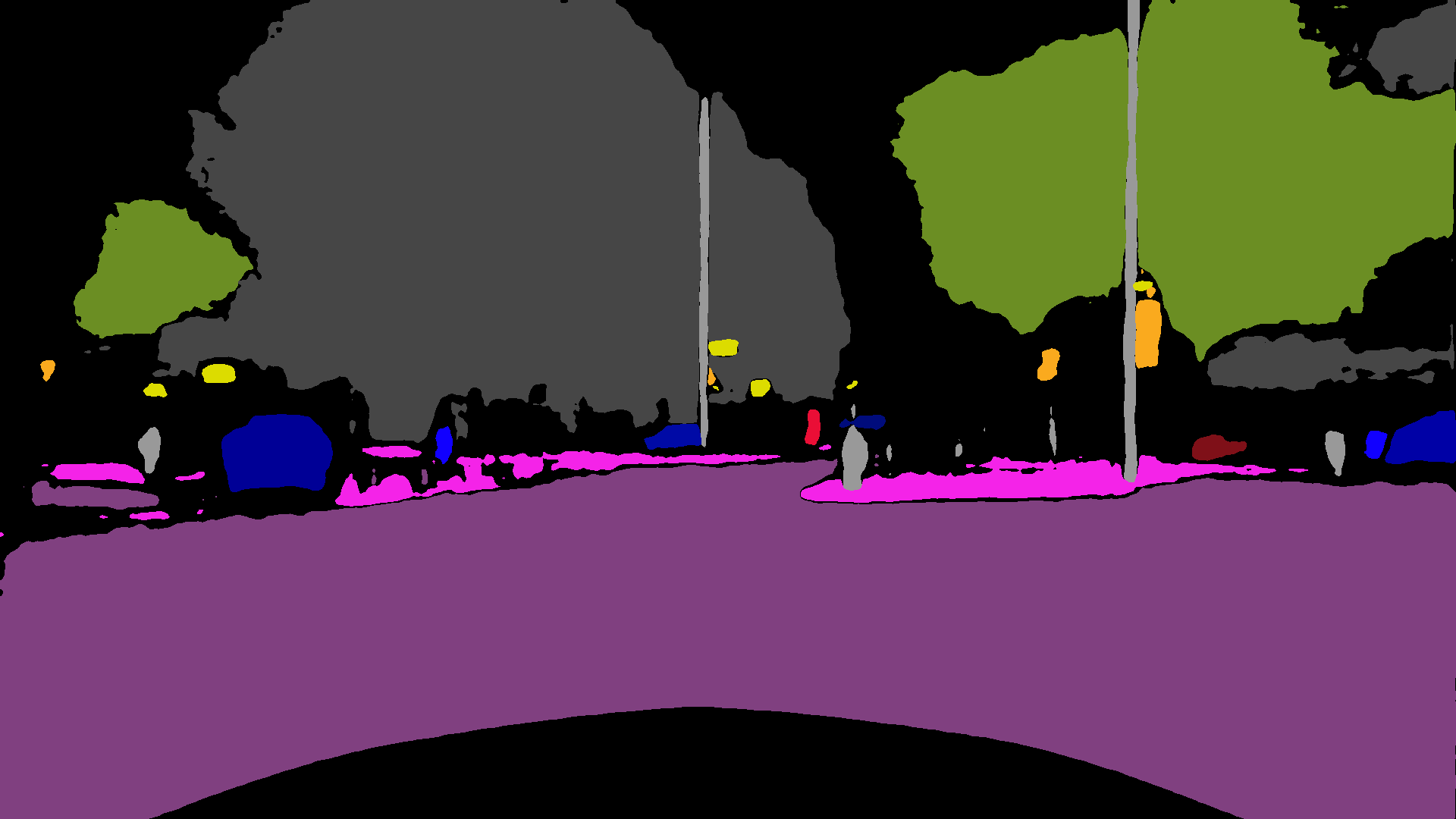} \\
        \vspace{-0.07cm}
        \includegraphics[width=0.328\textwidth]{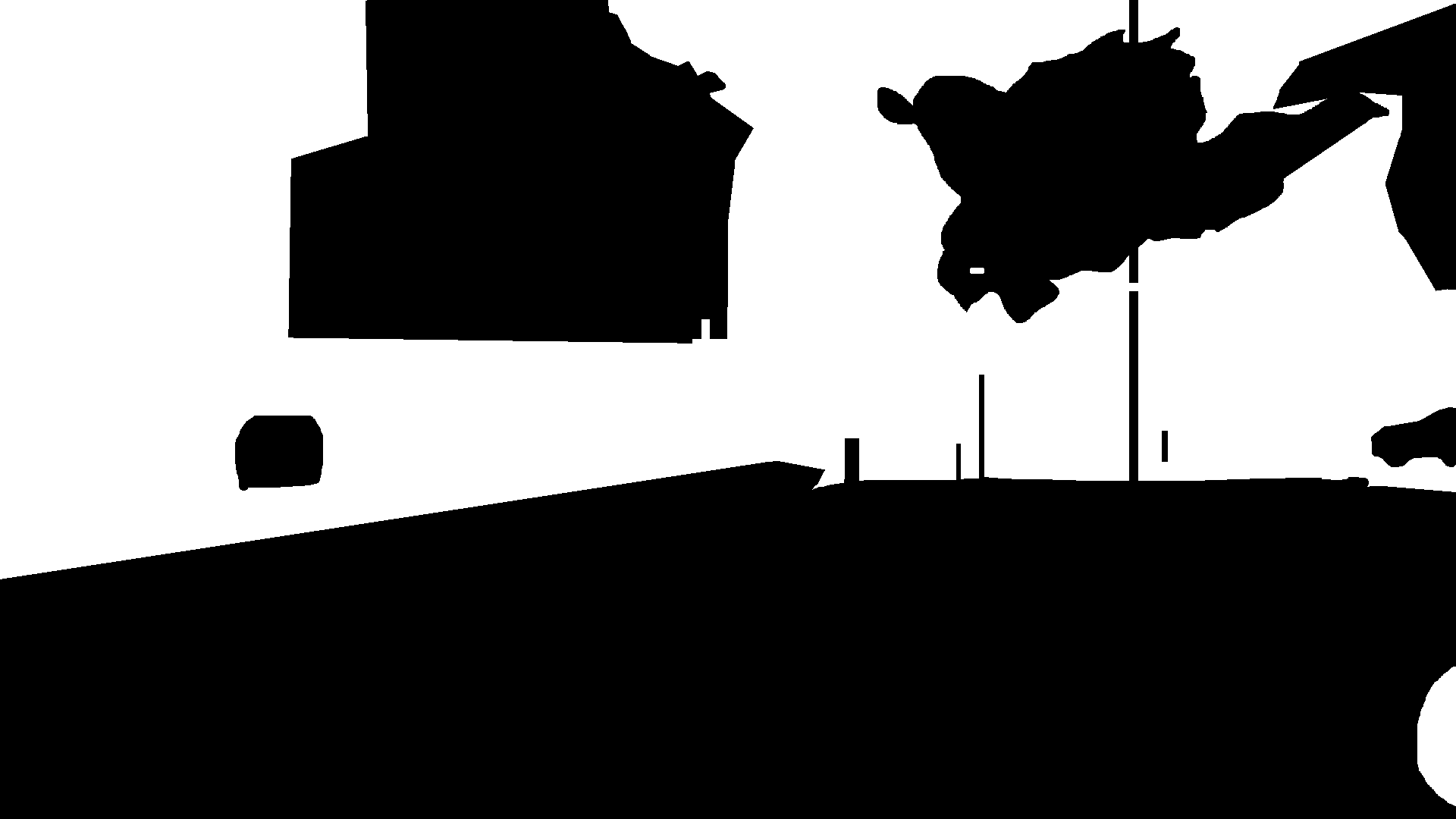} &
        \includegraphics[width=0.328\textwidth]{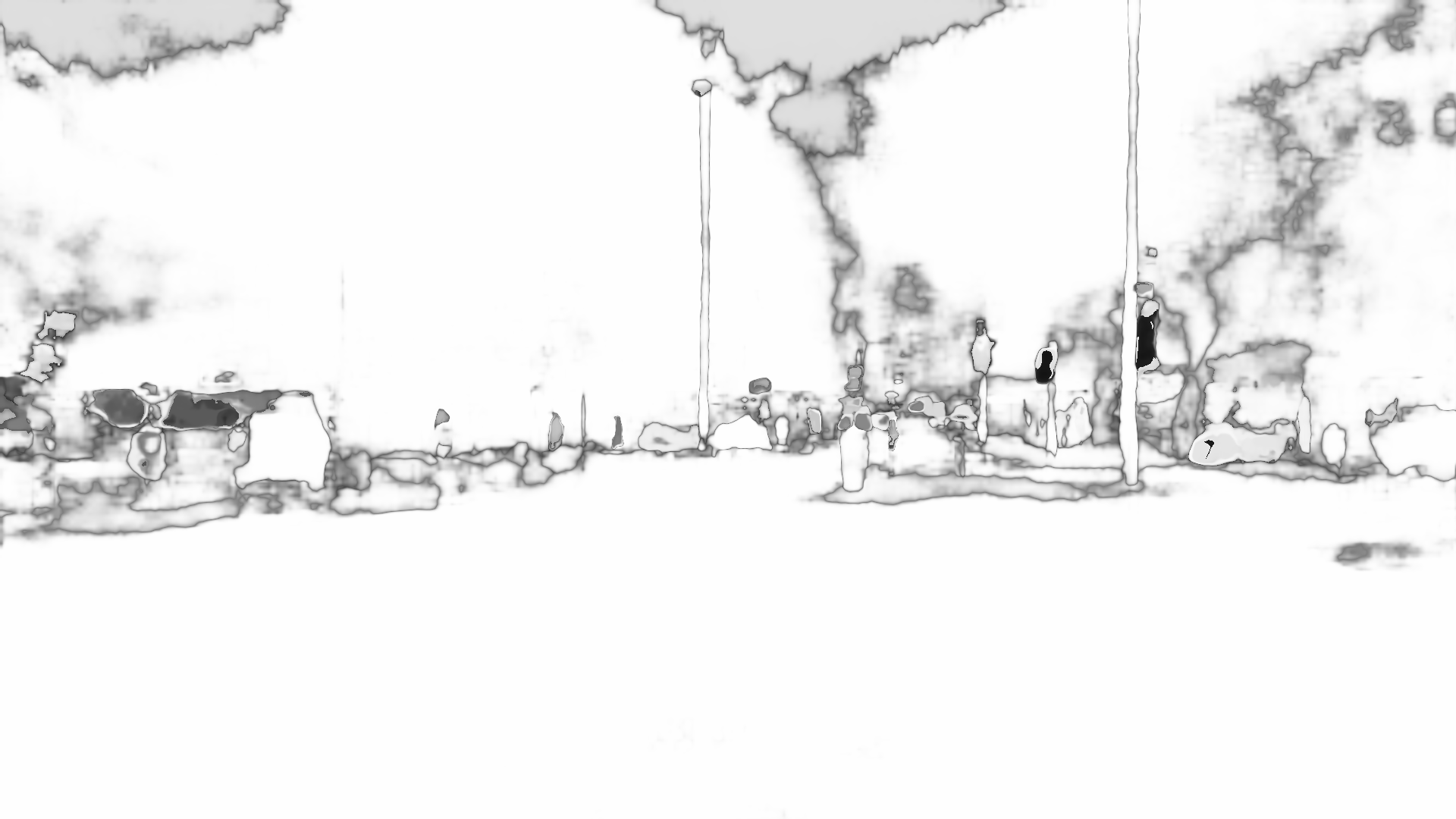} &
        \includegraphics[width=0.328\textwidth]{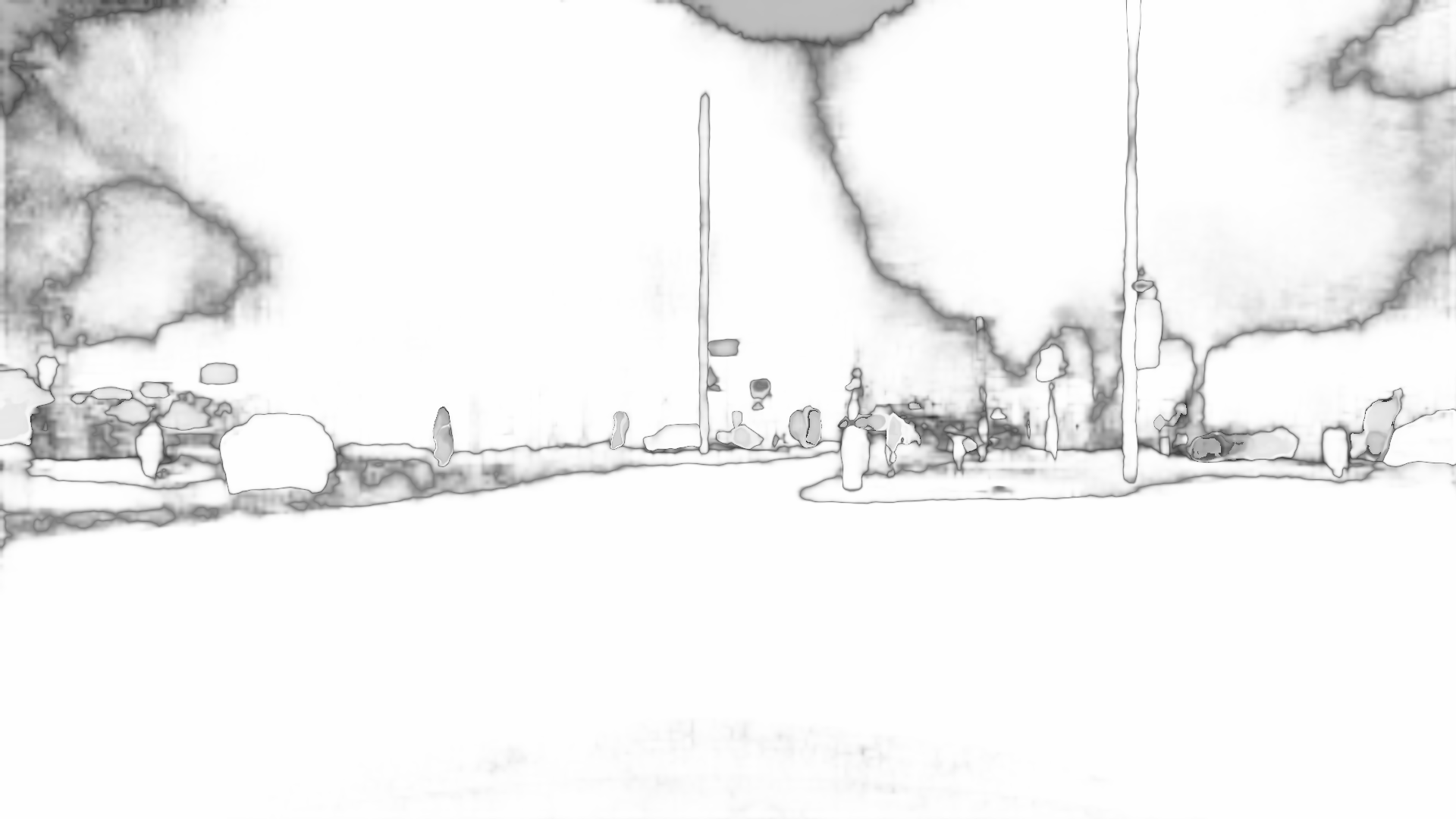} \\
        \vspace{-0.07cm}
        \includegraphics[width=0.328\textwidth]{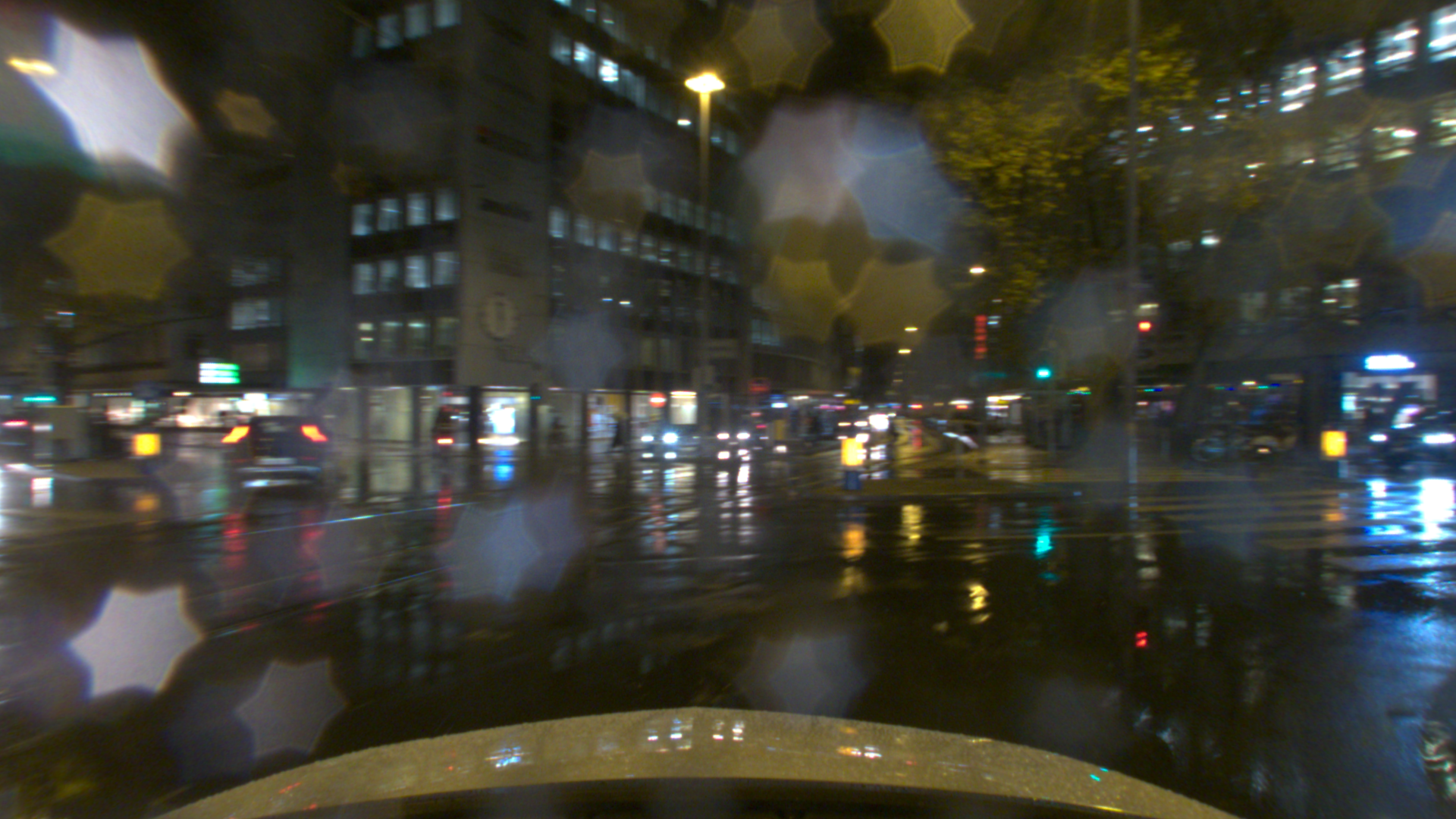} &
        \includegraphics[width=0.328\textwidth]{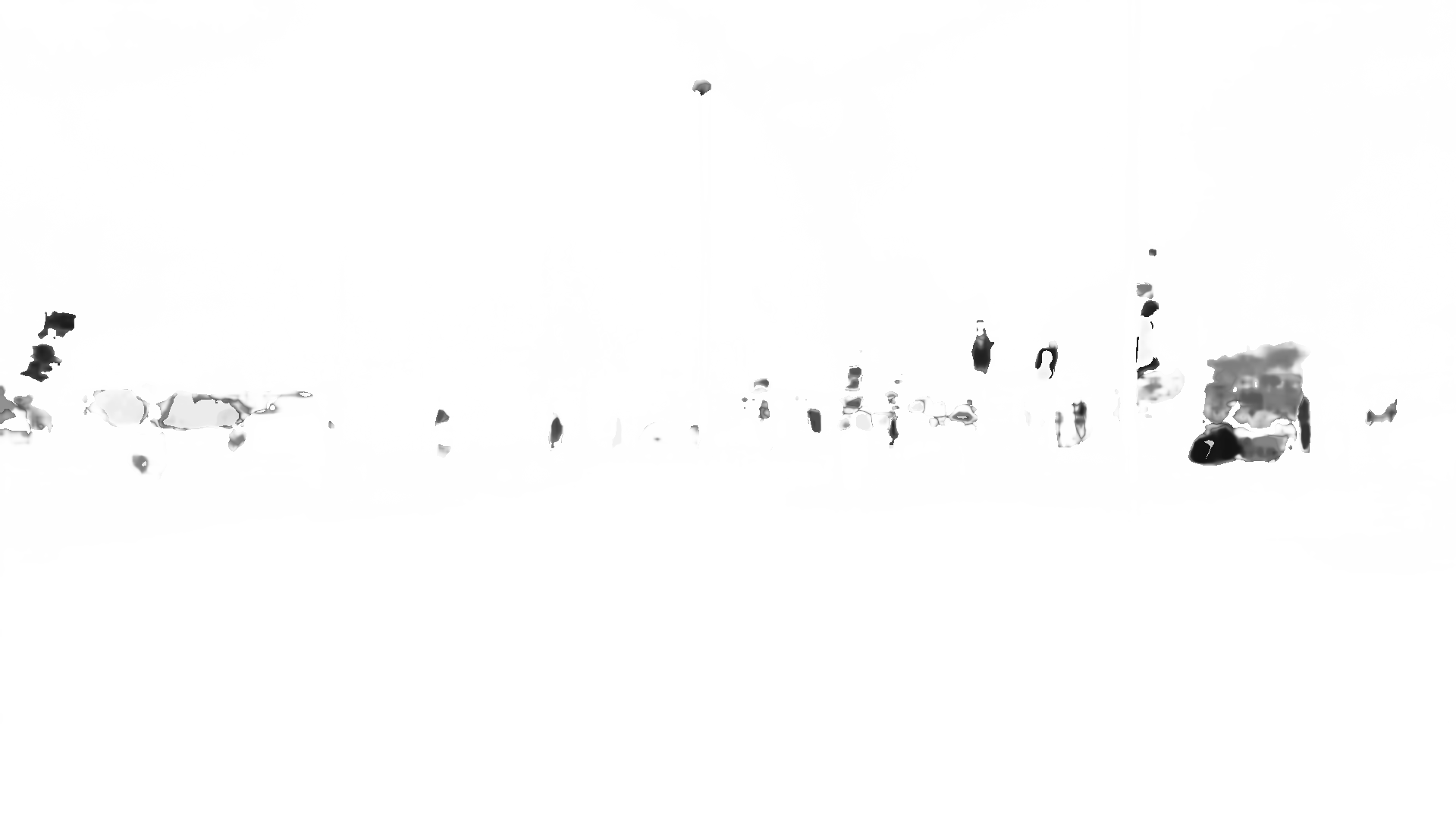} &
        \includegraphics[width=0.328\textwidth]{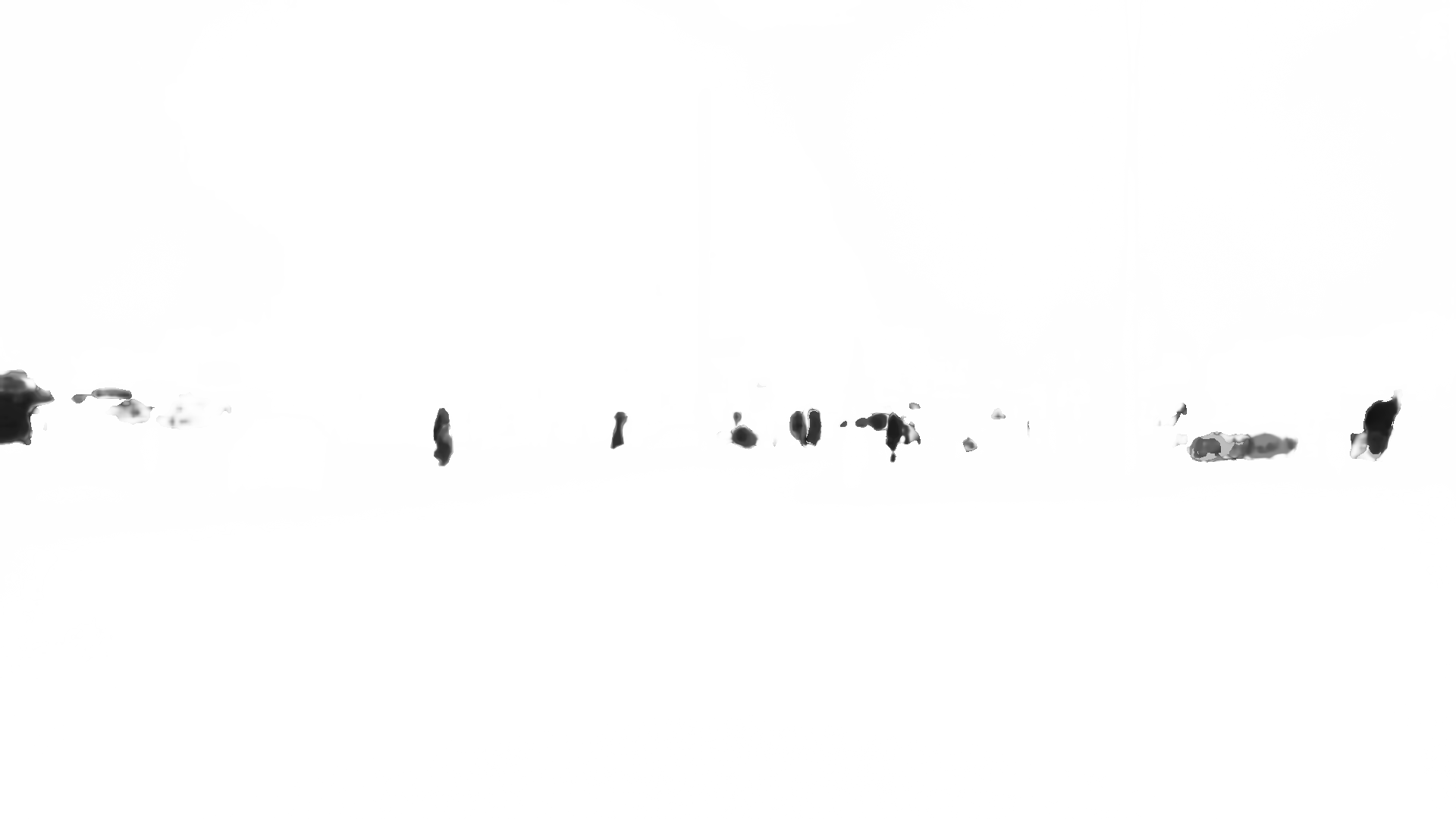} \\[8pt]
        
        \vspace{-0.07cm}
        \includegraphics[width=0.328\textwidth]{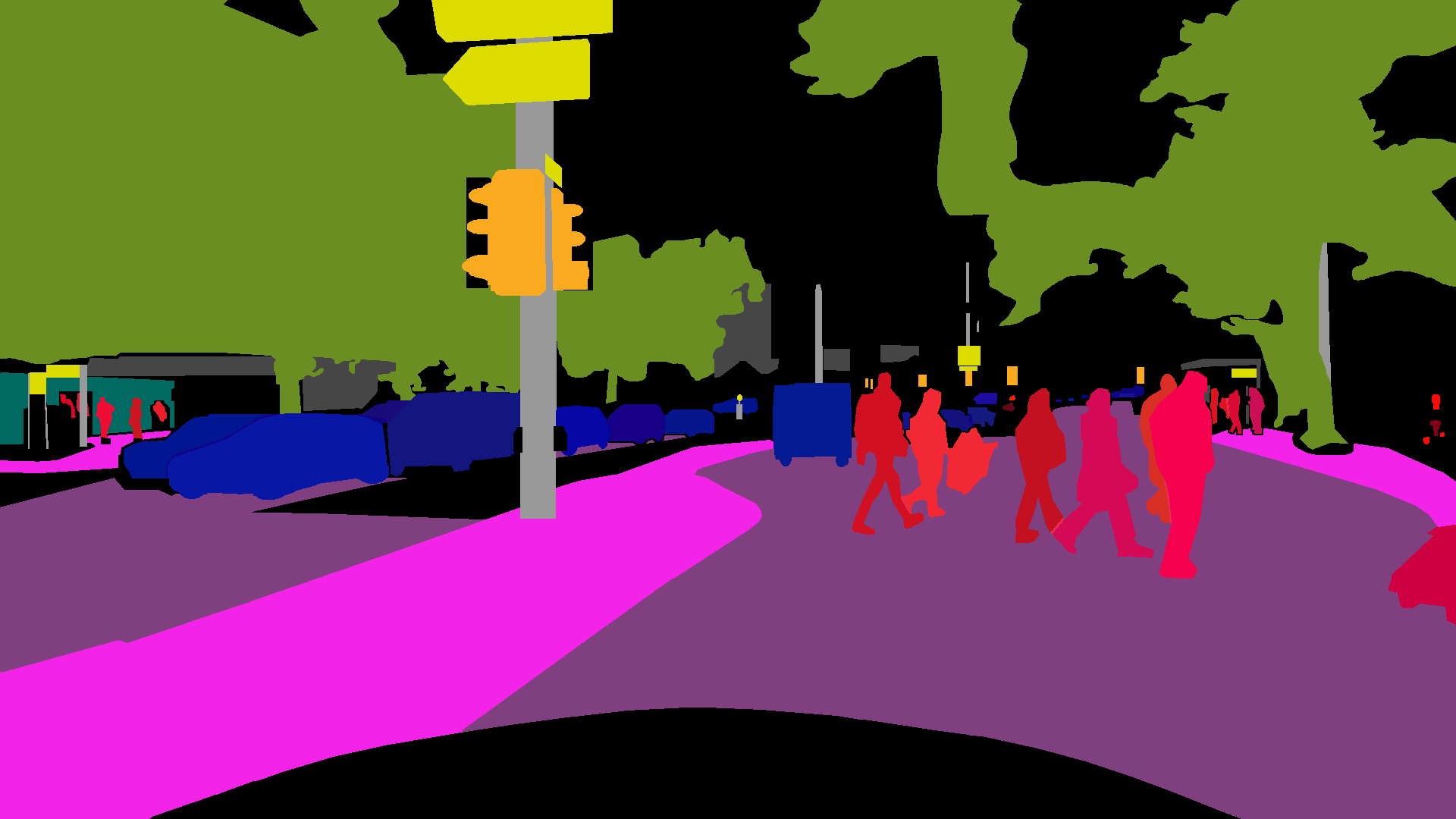} &
        \includegraphics[width=0.328\textwidth]{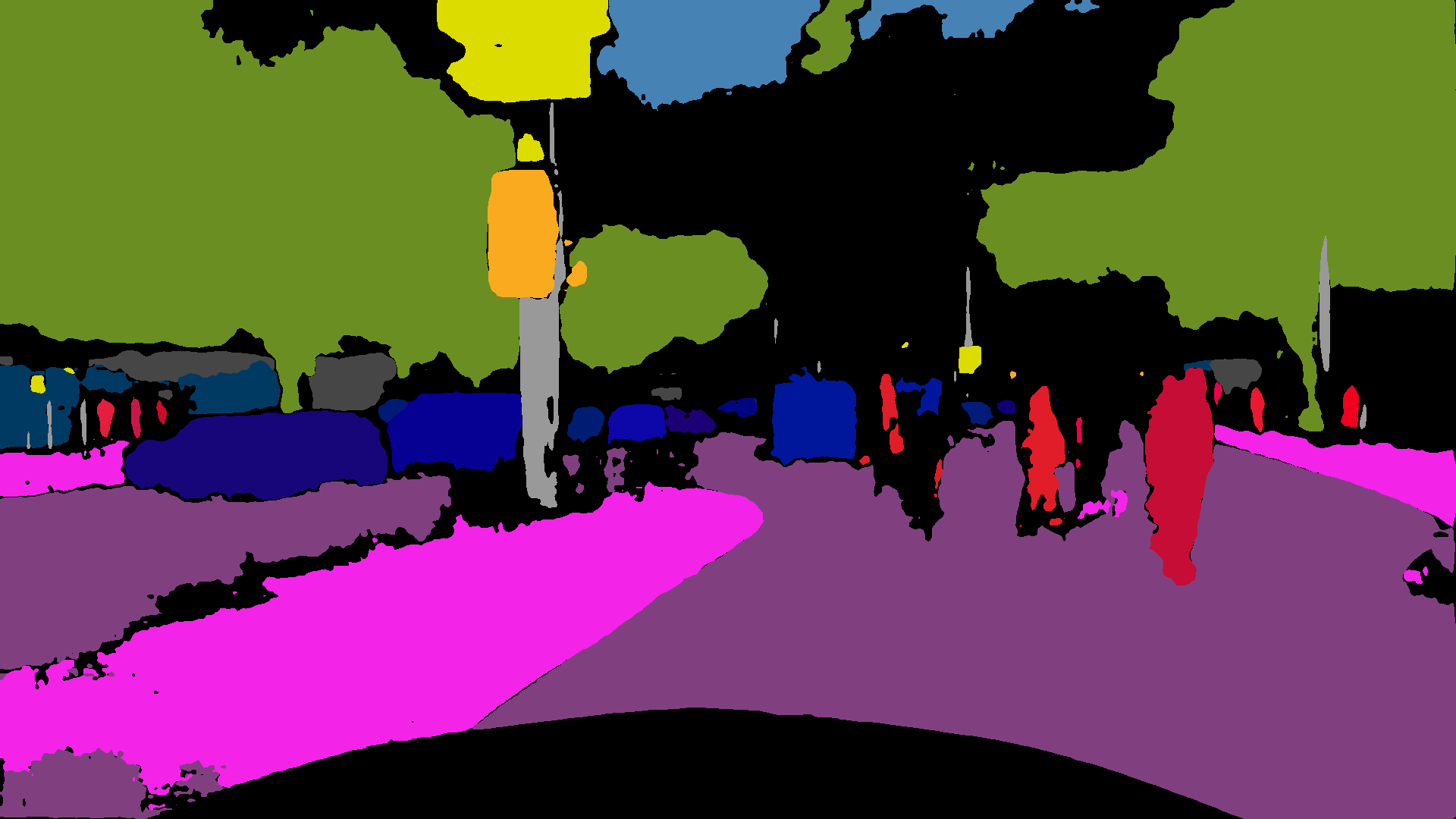} &
        \includegraphics[width=0.328\textwidth]{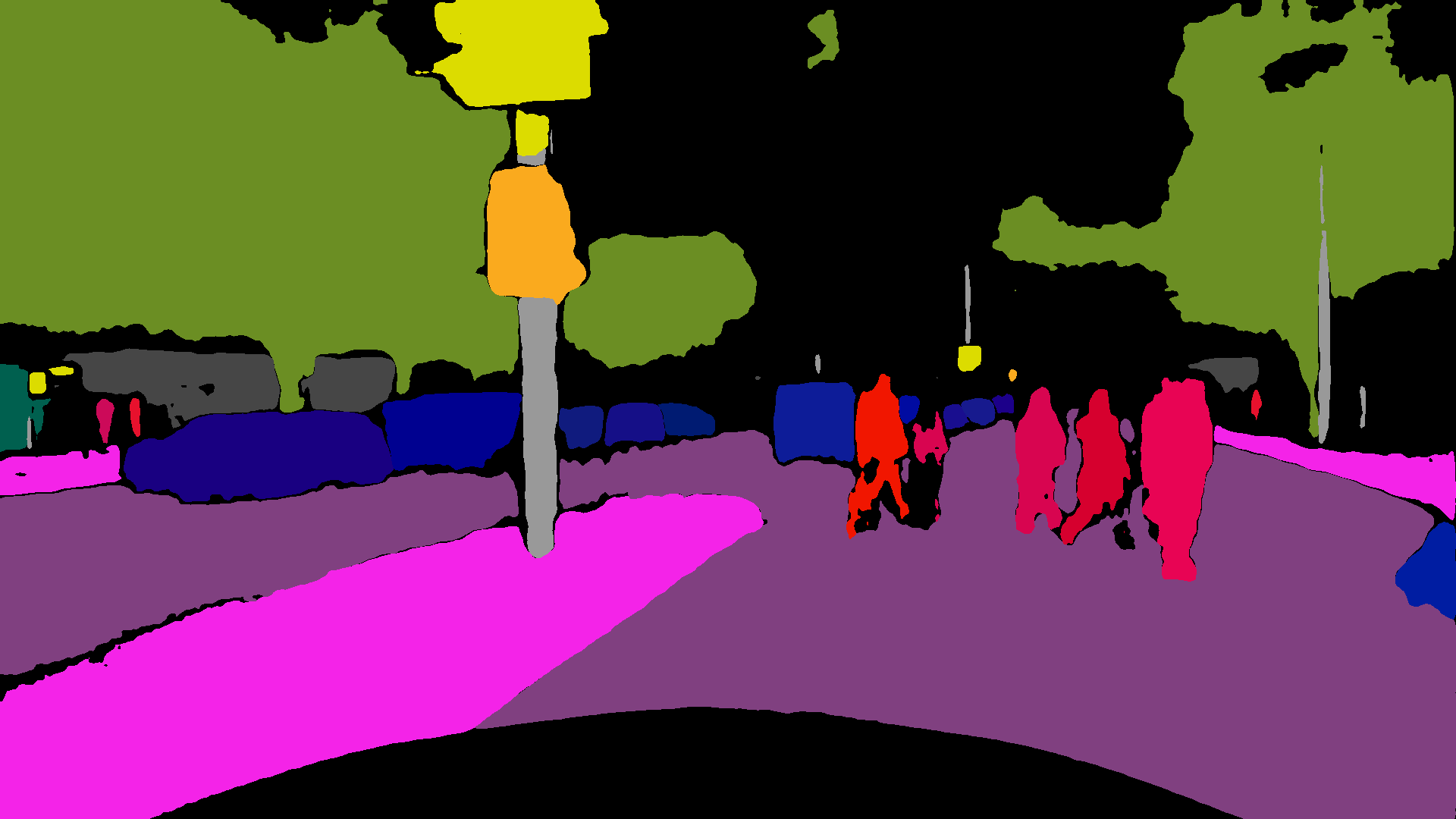} \\
        \vspace{-0.07cm}
        \includegraphics[width=0.328\textwidth]{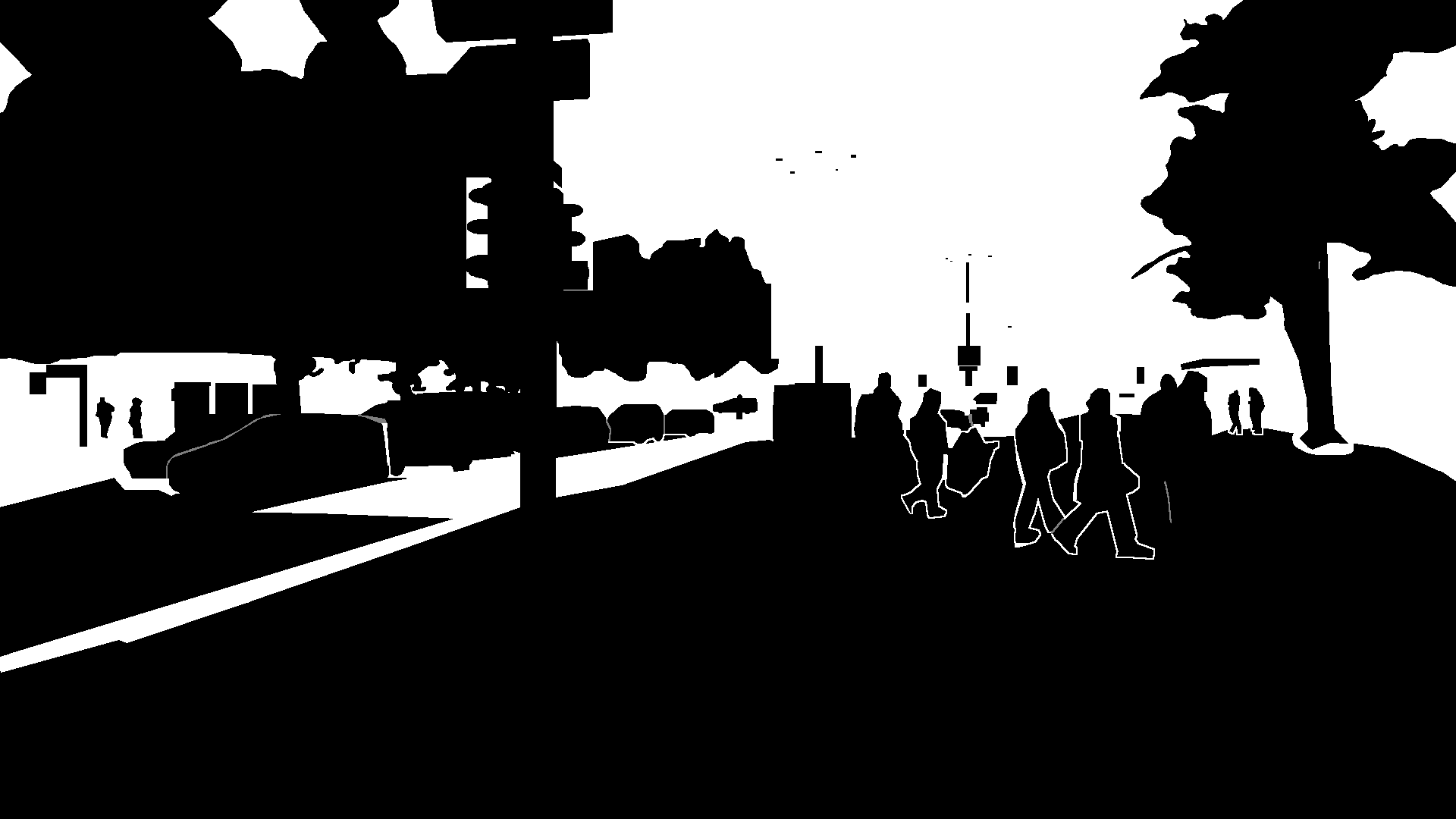} &
        \includegraphics[width=0.328\textwidth]{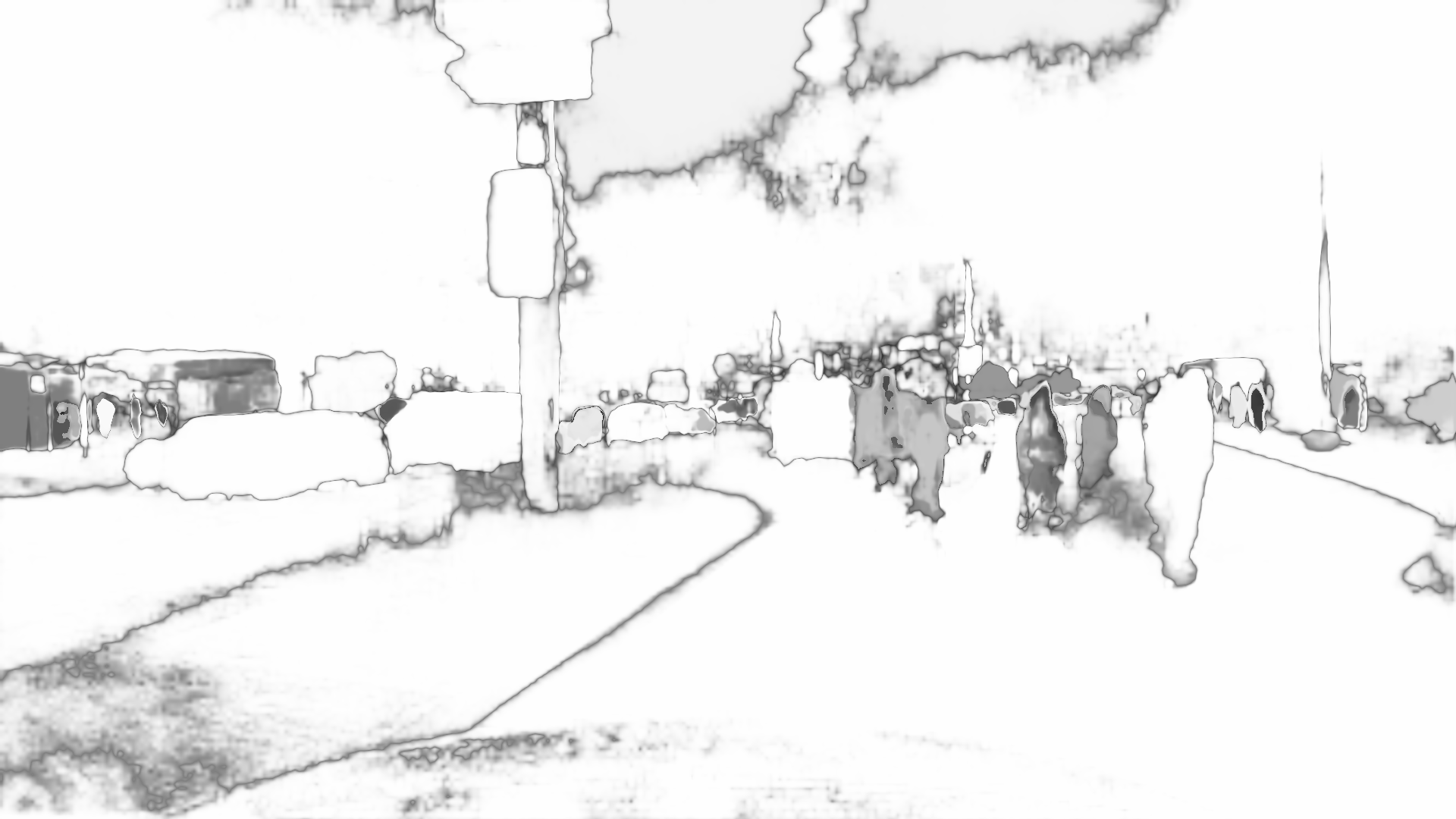} &
        \includegraphics[width=0.328\textwidth]{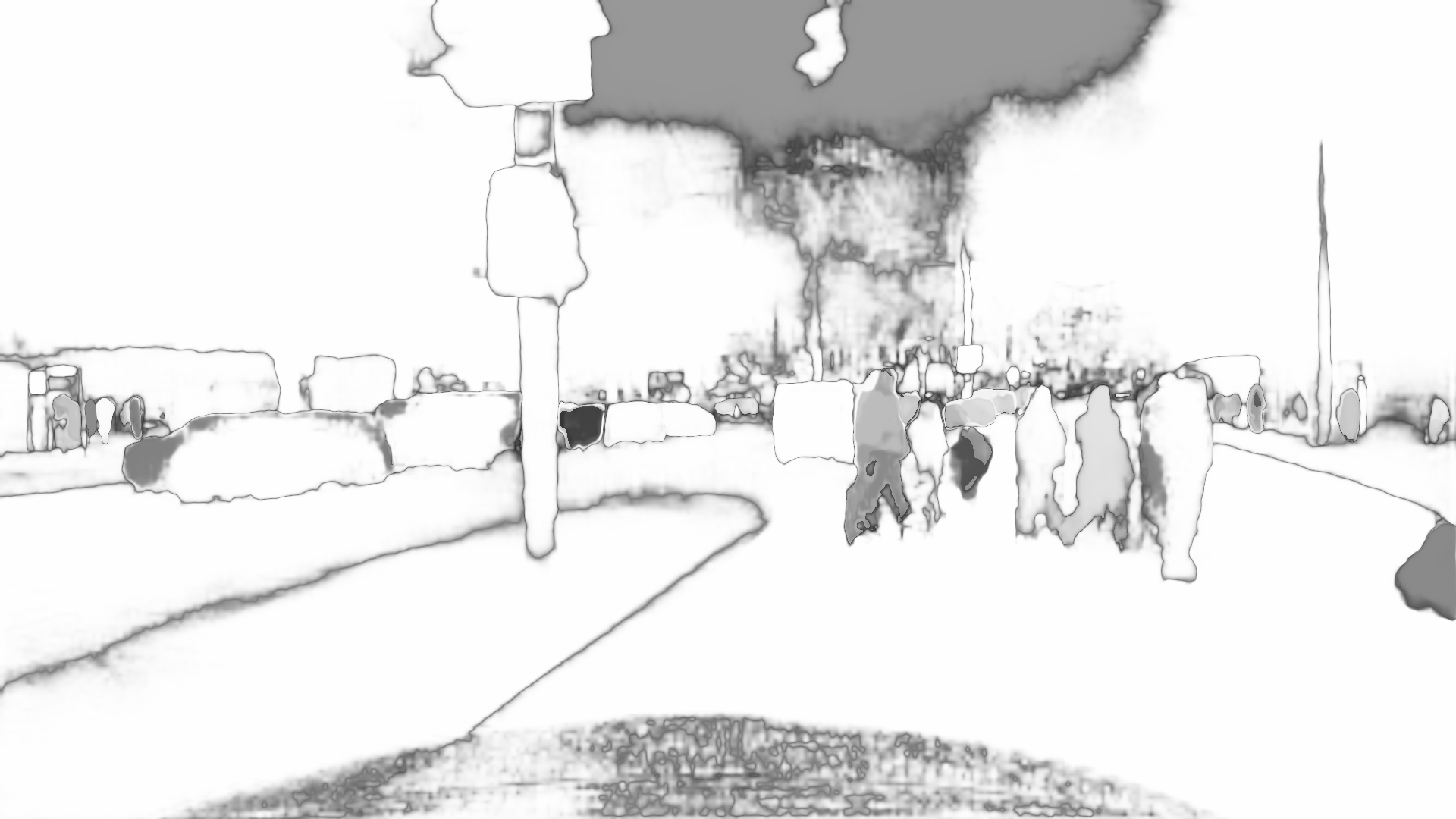} \\
        \vspace{-0.07cm}
        \includegraphics[width=0.328\textwidth]{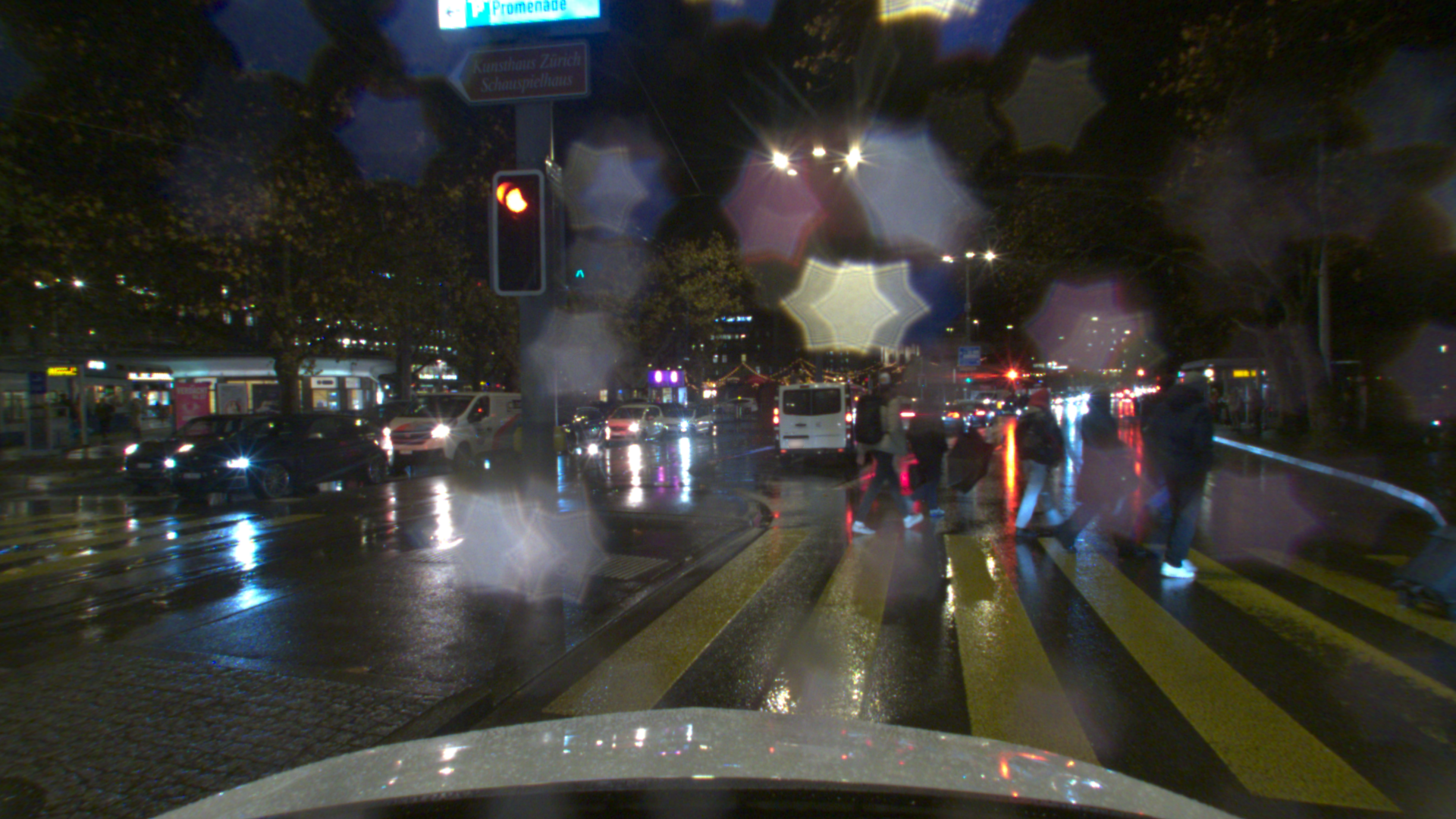} &
        \includegraphics[width=0.328\textwidth]{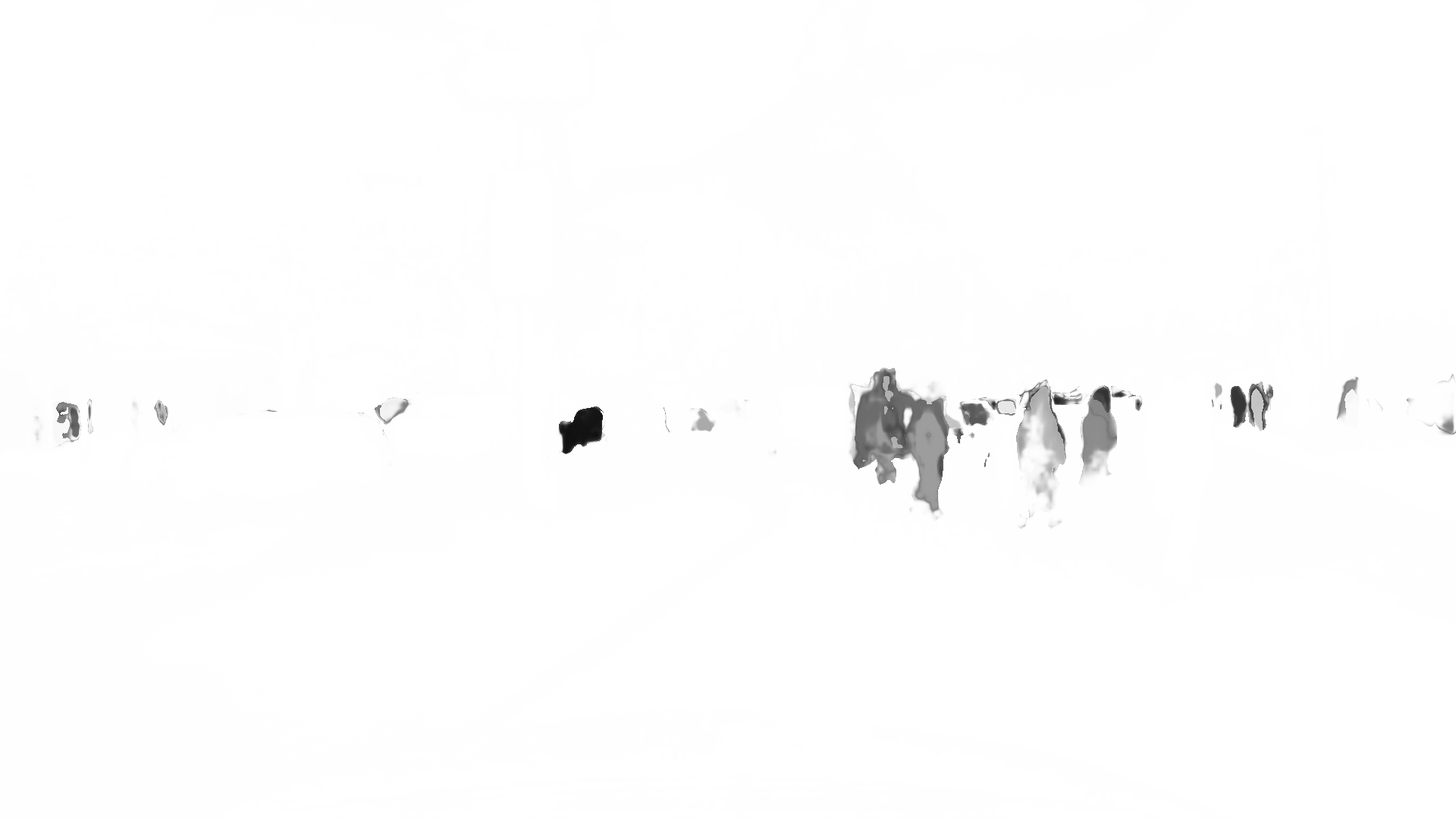} &
        \includegraphics[width=0.328\textwidth]{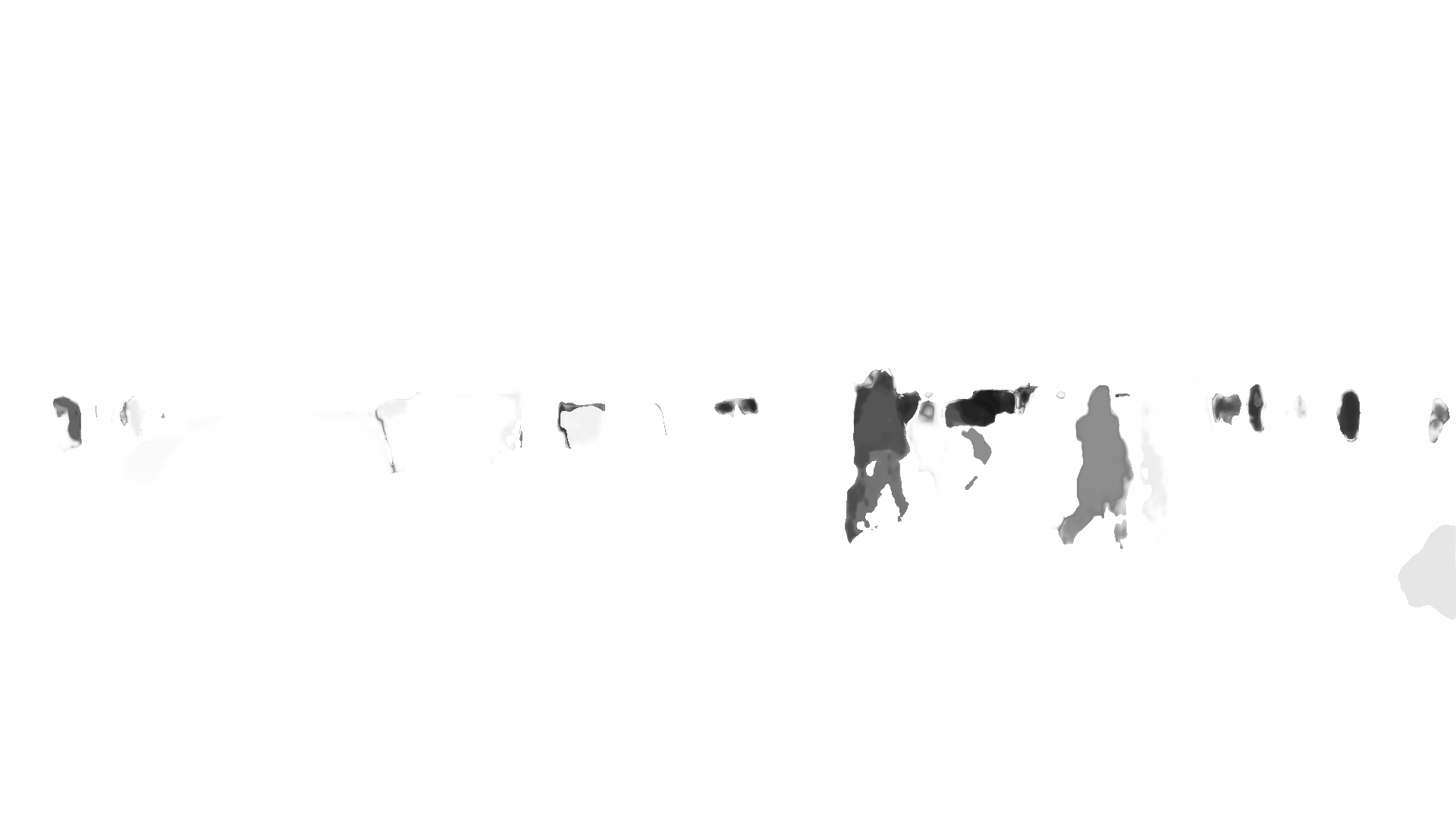} \\
    \end{tabular}
    \caption{\textbf{Qualitative panoptic results}. First column from top to bottom: panoptic segmentation ground-truth annotation, difficulty map, input frame camera image. Second and third columns from top to bottom: the panoptic predictions, class uncertainty scores, and instance uncertainty scores. Best viewed on a screen at full zoom.}
    \label{fig:suppl:qual:results:1}
\end{figure*}

\begin{figure*}
    \centering
    \begin{tabular}{@{}c@{\hspace{0.05cm}}c@{\hspace{0.05cm}}c@{}}
        \multicolumn{1}{c}{\scriptsize Ground Truth} &
        \multicolumn{1}{c}{\scriptsize Unimodal Mask2Former~\cite{cheng2022masked}} &
        \multicolumn{1}{c}{\scriptsize Quadrimodal Mask2Former~\cite{cheng2022masked}} \\
        \vspace{-0.07cm}
        \includegraphics[width=0.328\textwidth]{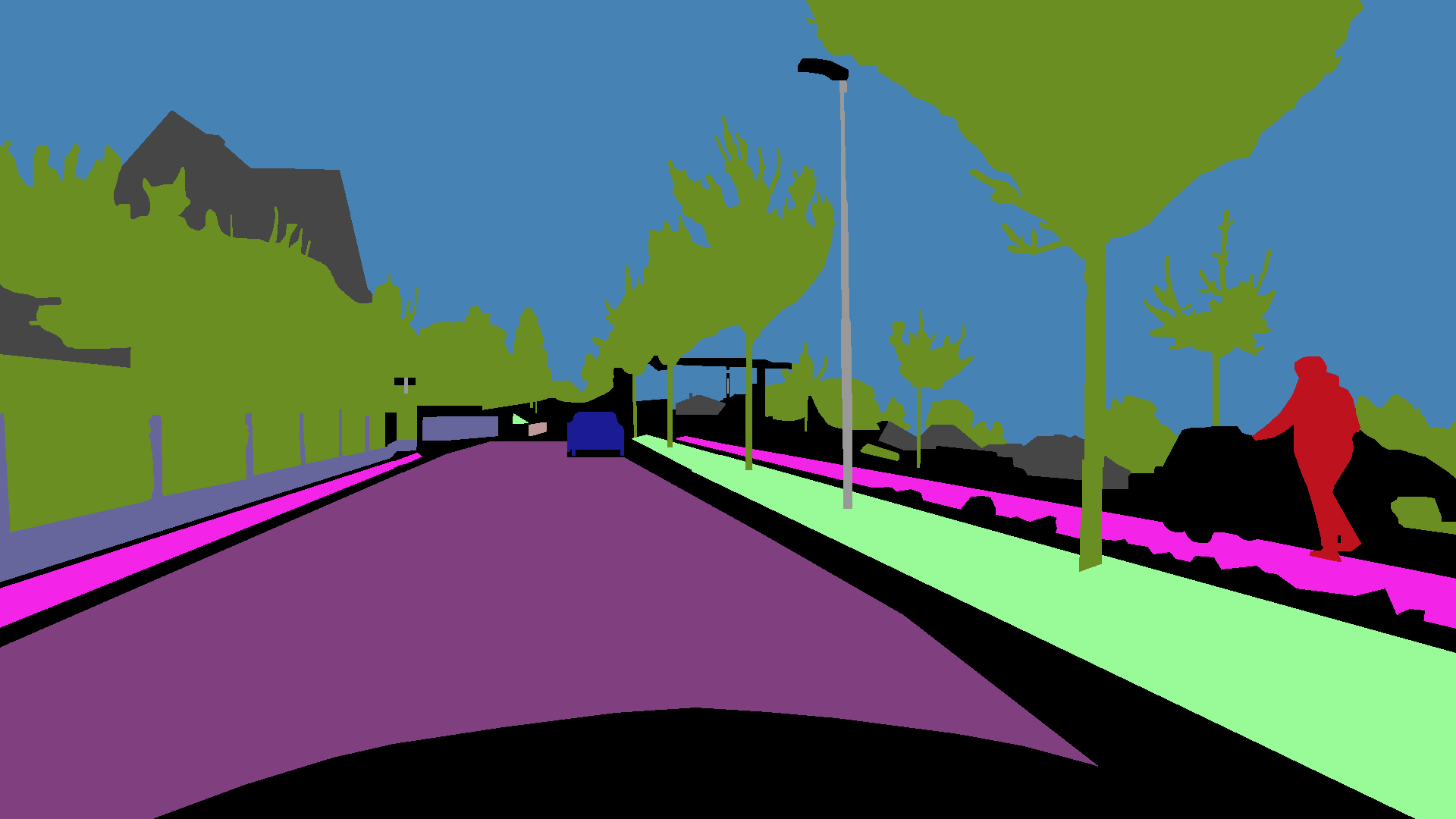} &
        \includegraphics[width=0.328\textwidth]{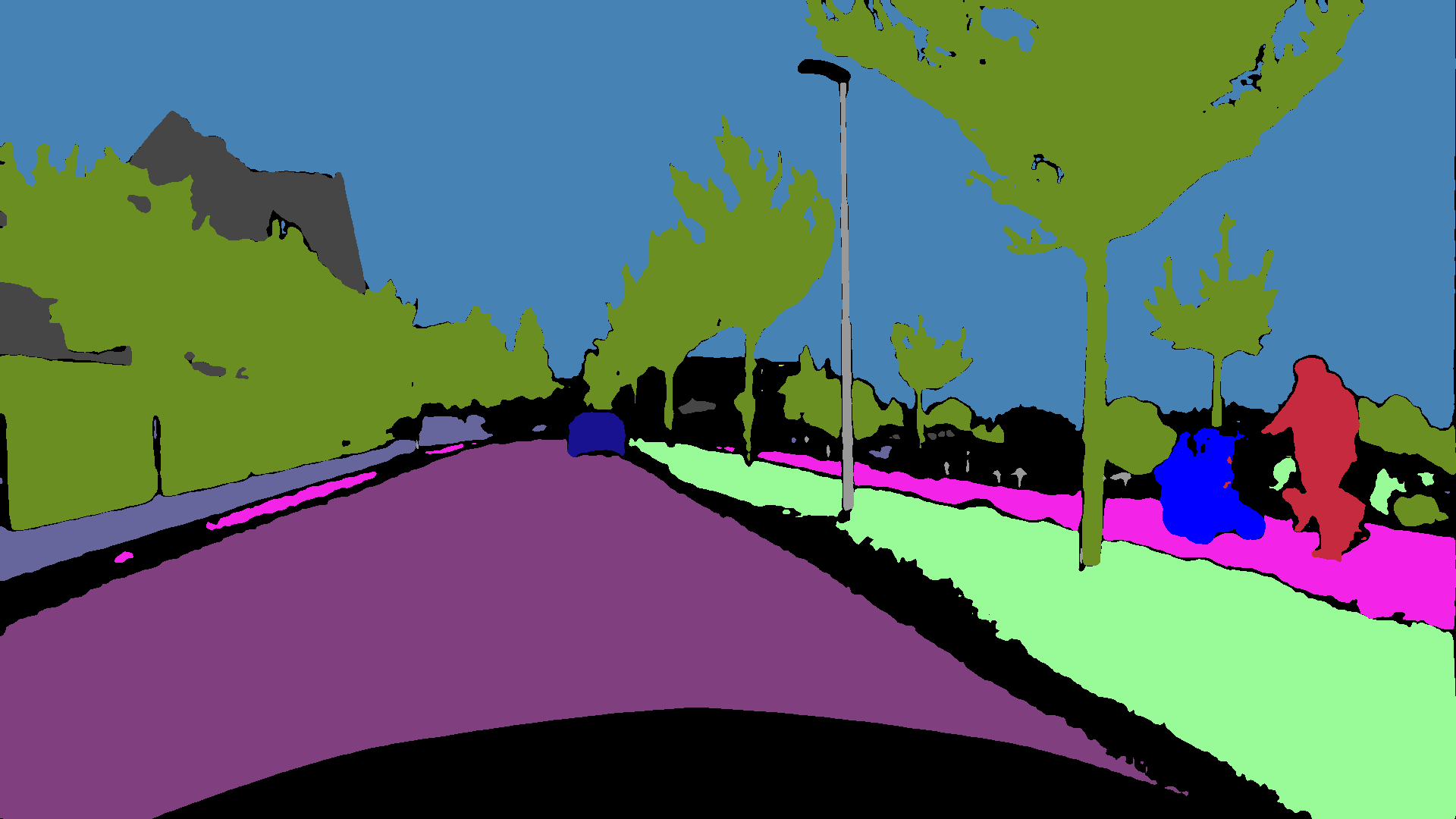} &
        \includegraphics[width=0.328\textwidth]{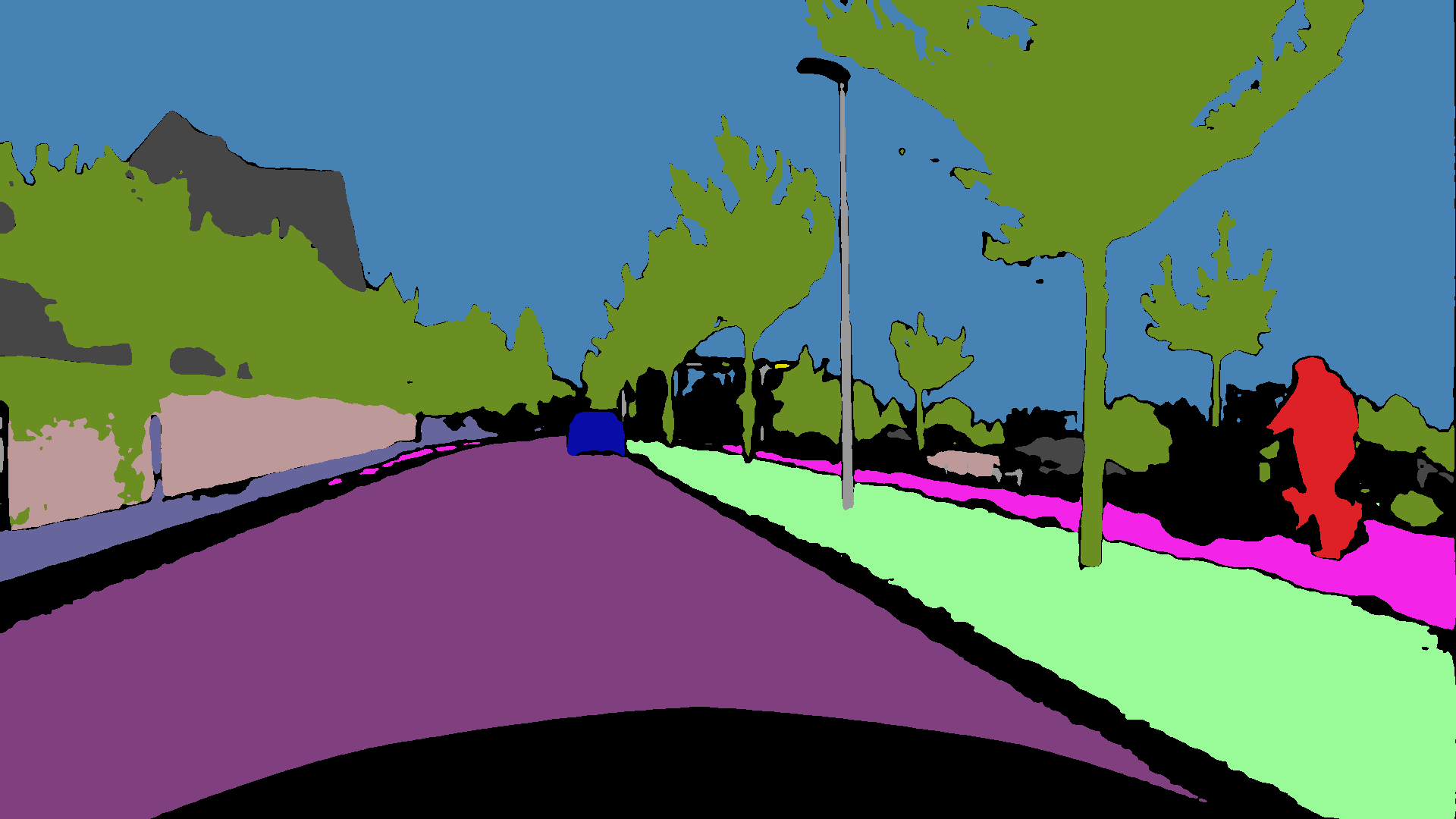} \\
        \vspace{-0.07cm}
        \includegraphics[width=0.328\textwidth]{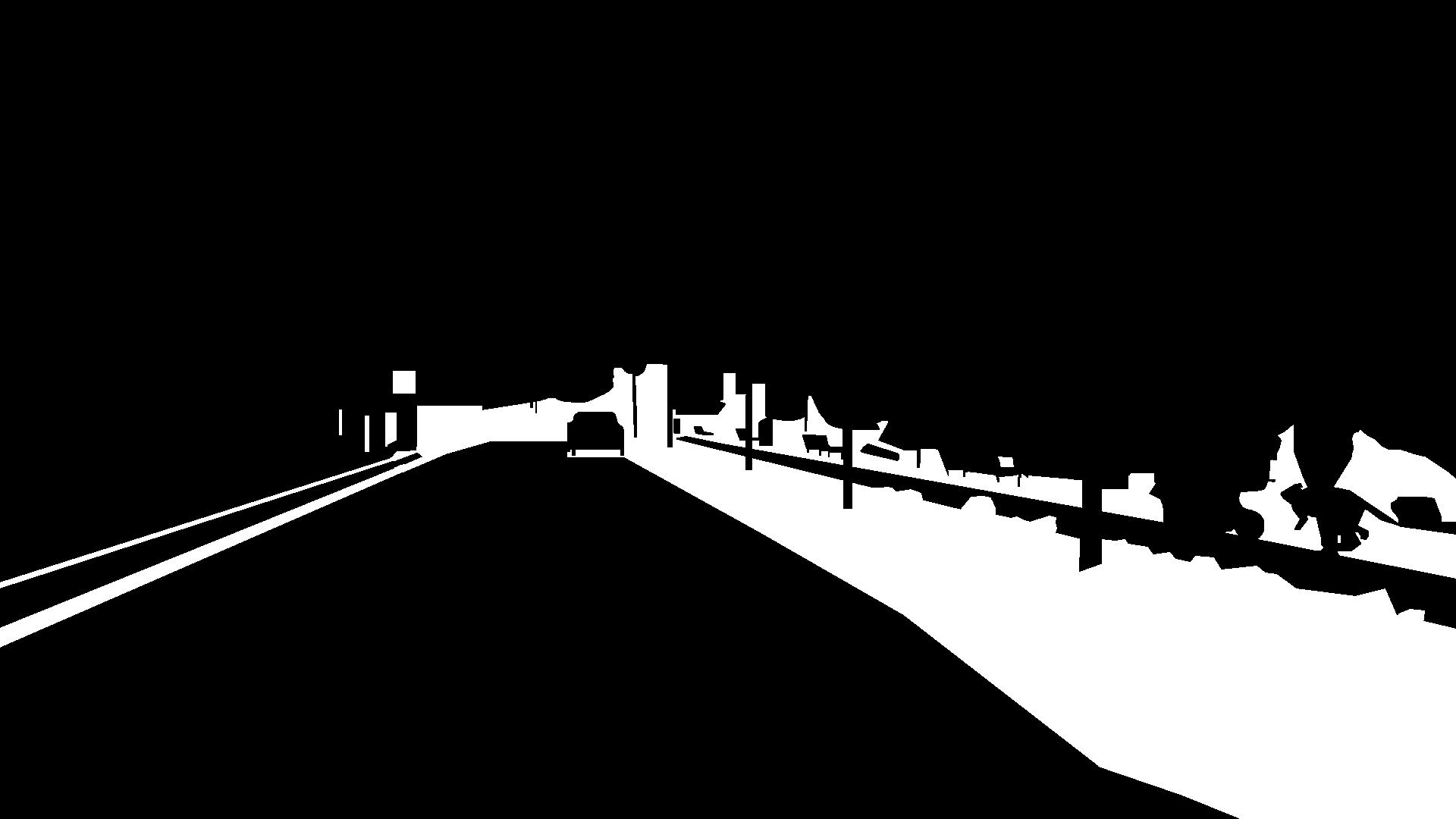} &
        \includegraphics[width=0.328\textwidth]{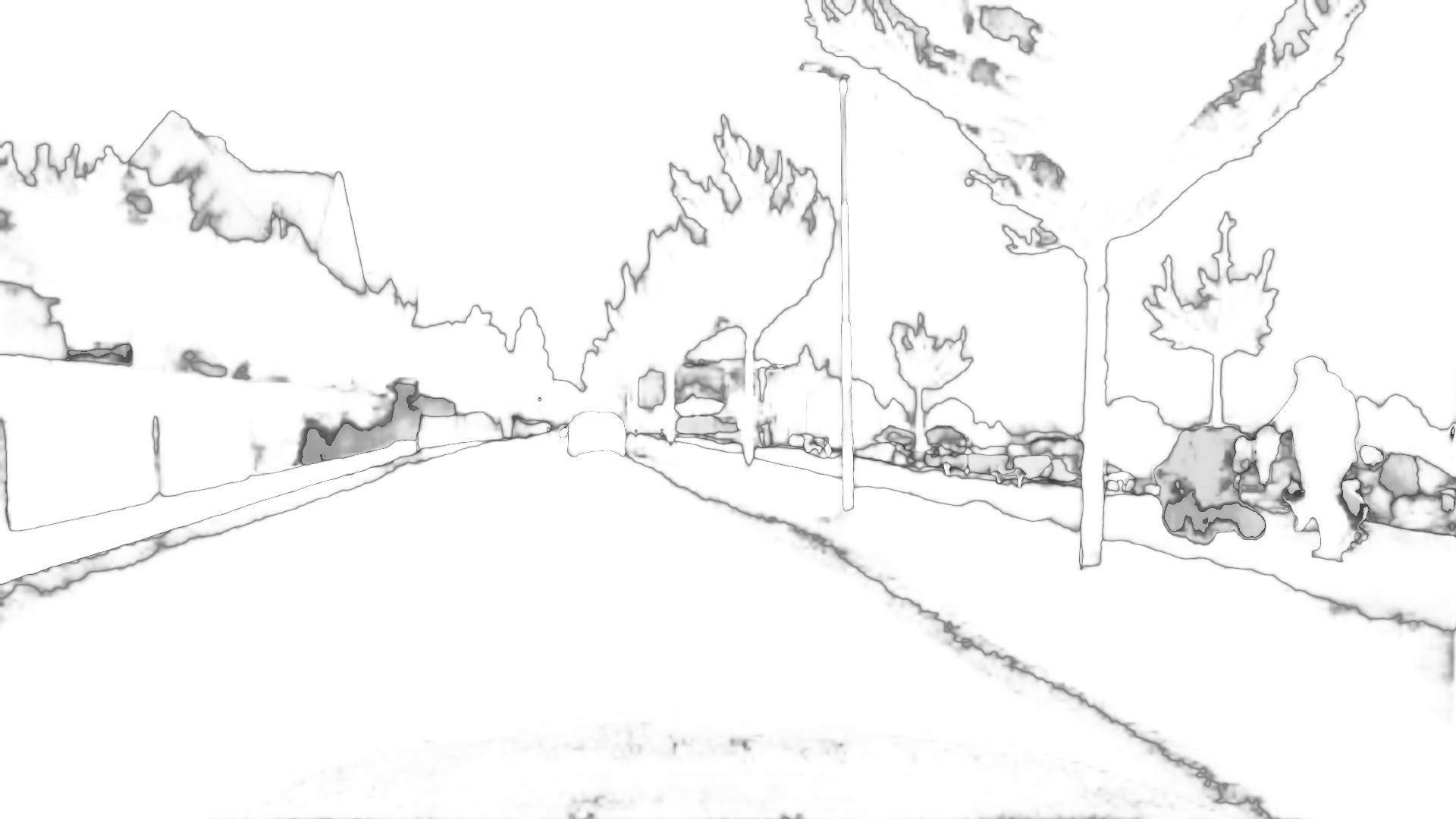} &
        \includegraphics[width=0.328\textwidth]{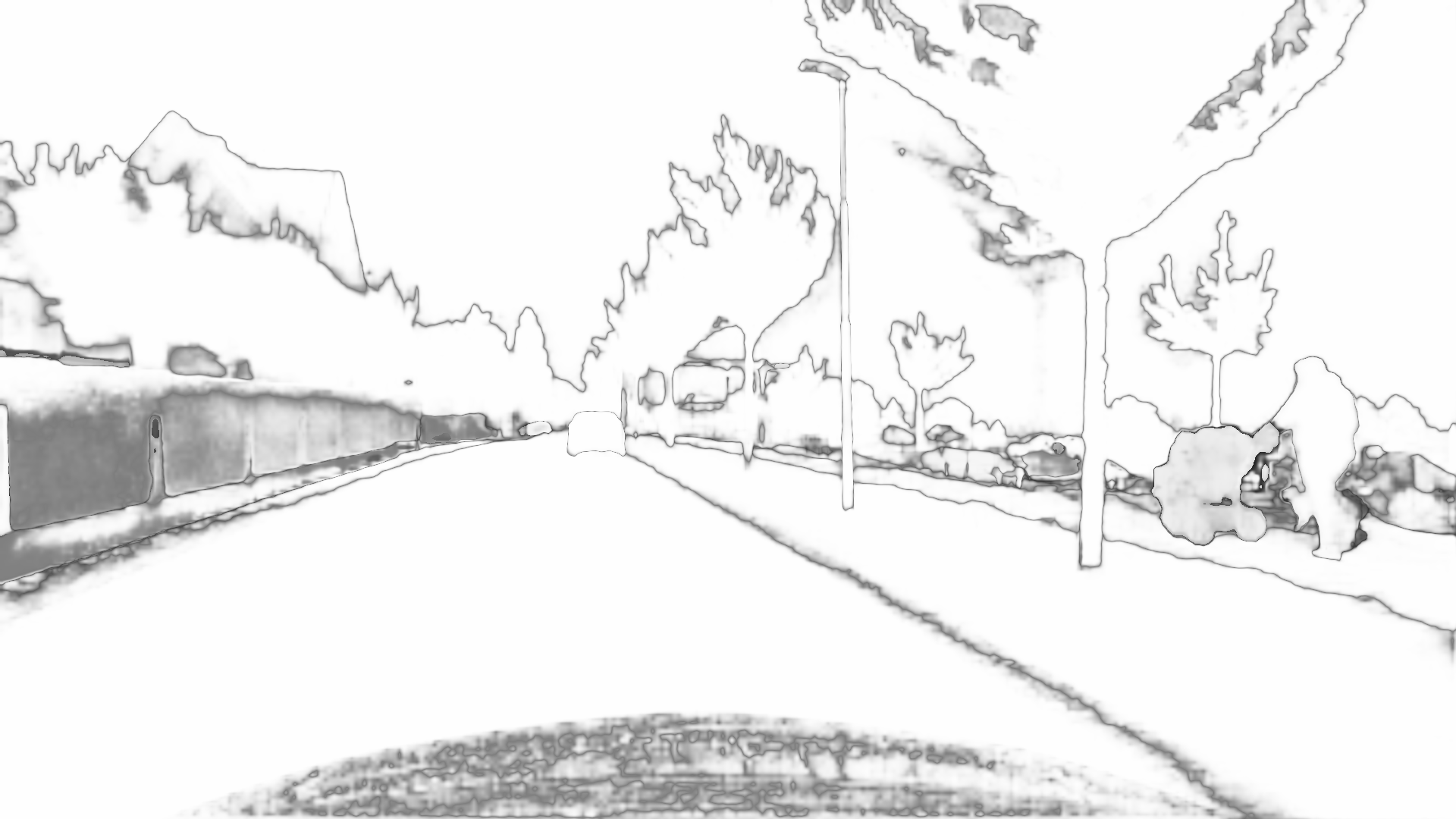} \\
        \vspace{-0.07cm}
        \includegraphics[width=0.328\textwidth]{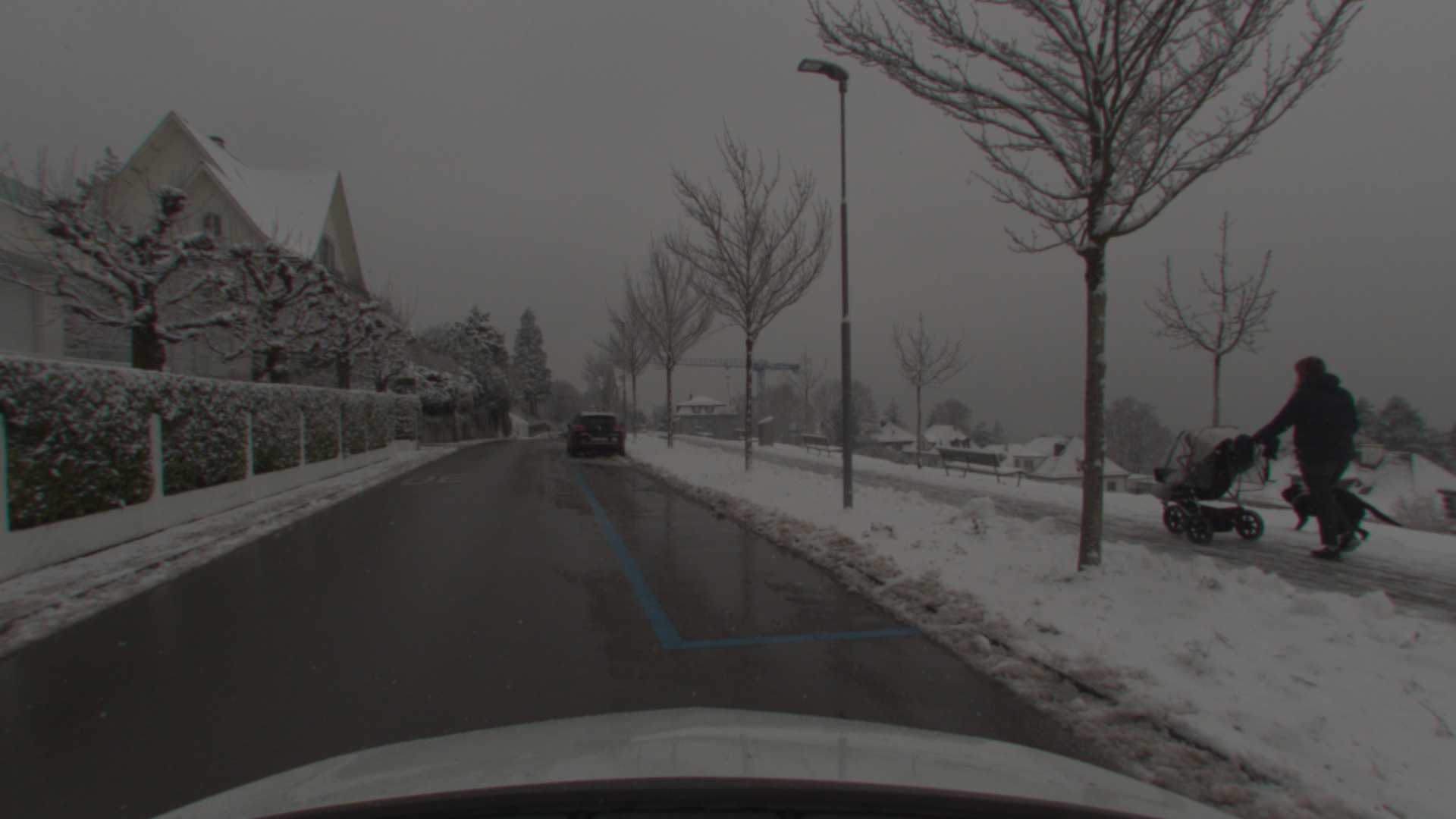} &
        \includegraphics[width=0.328\textwidth]{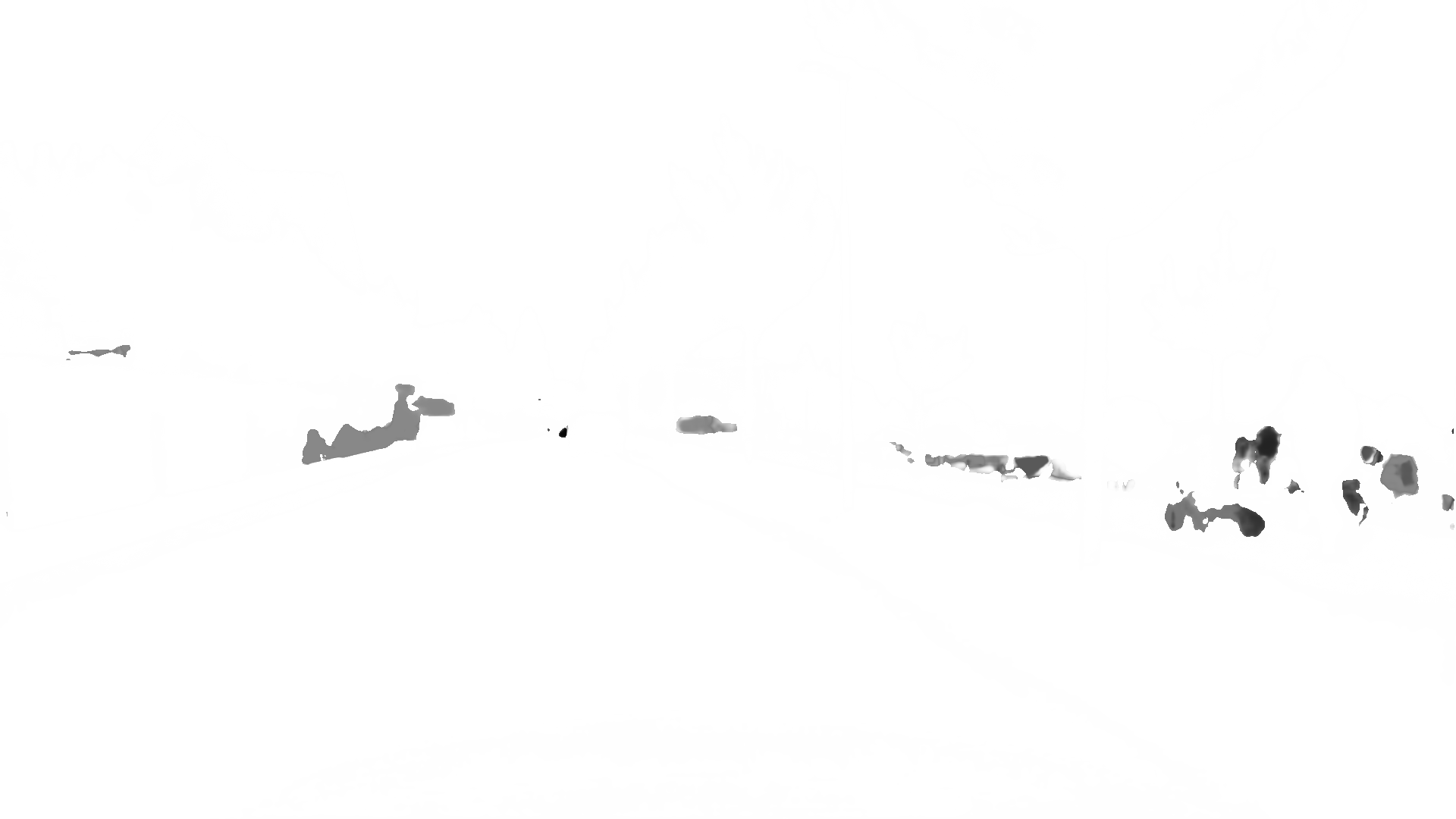} &
        \includegraphics[width=0.328\textwidth]{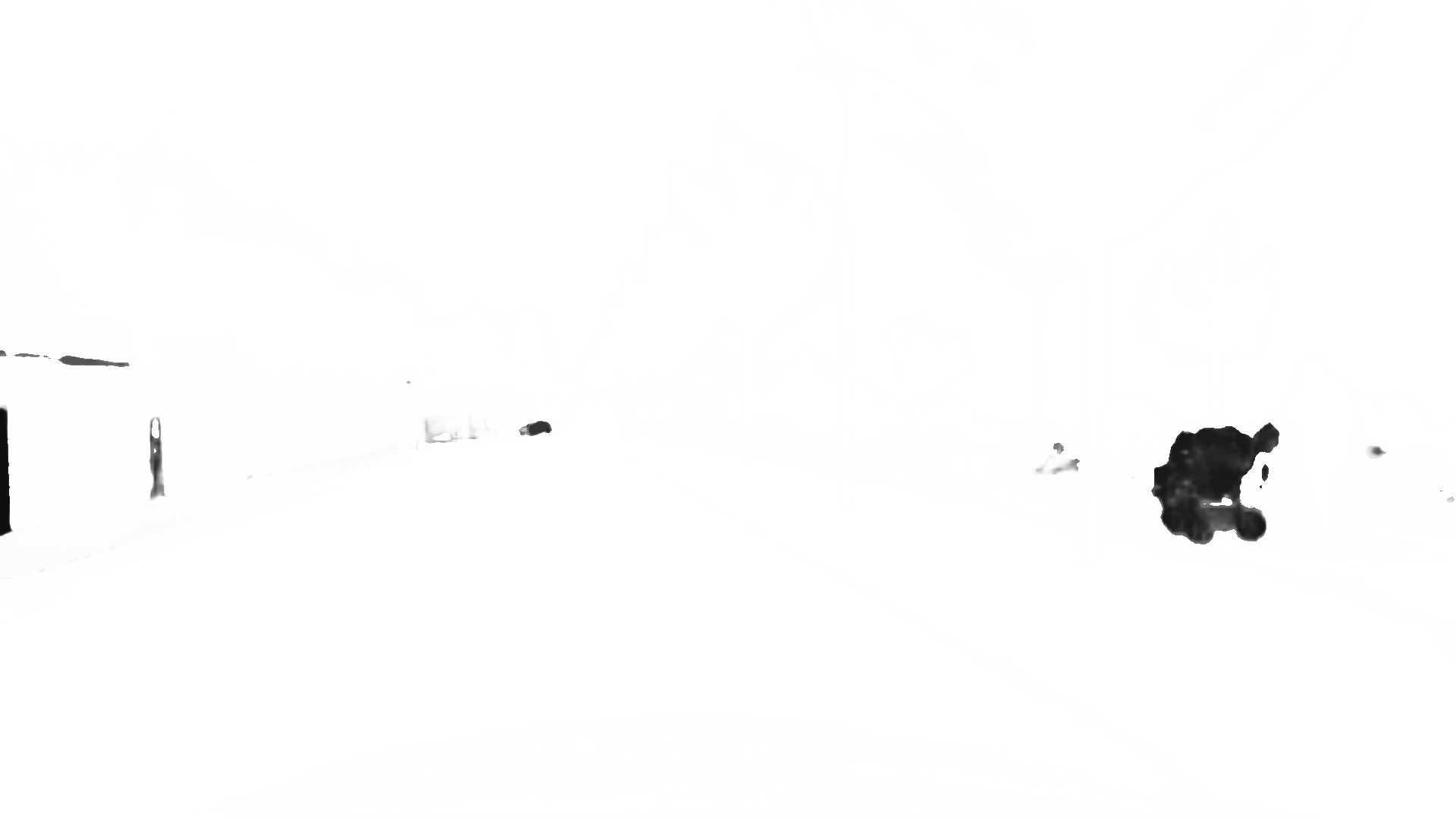} \\[8pt]
        
        \vspace{-0.07cm}
        \includegraphics[width=0.328\textwidth]{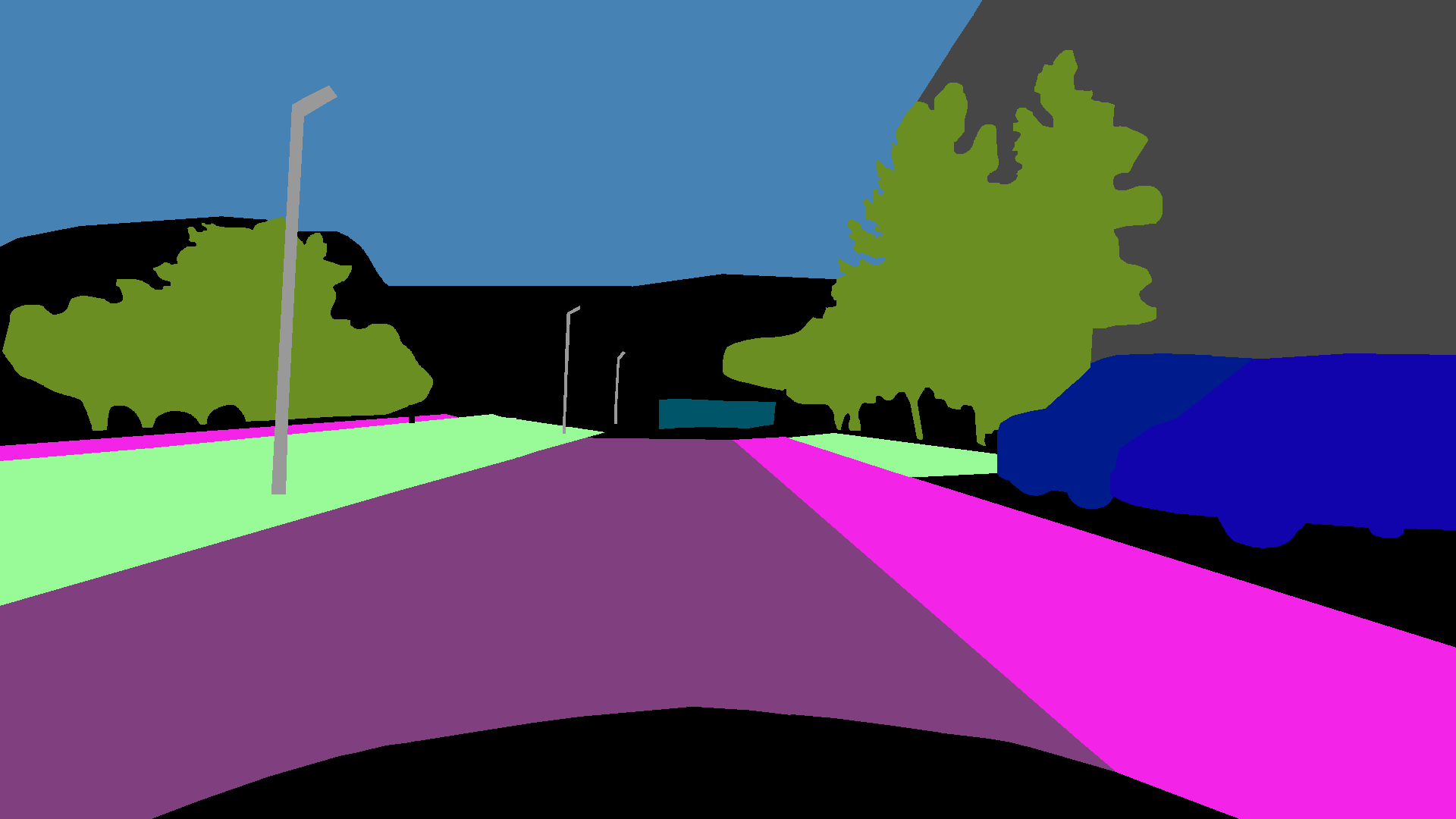} &
        \includegraphics[width=0.328\textwidth]{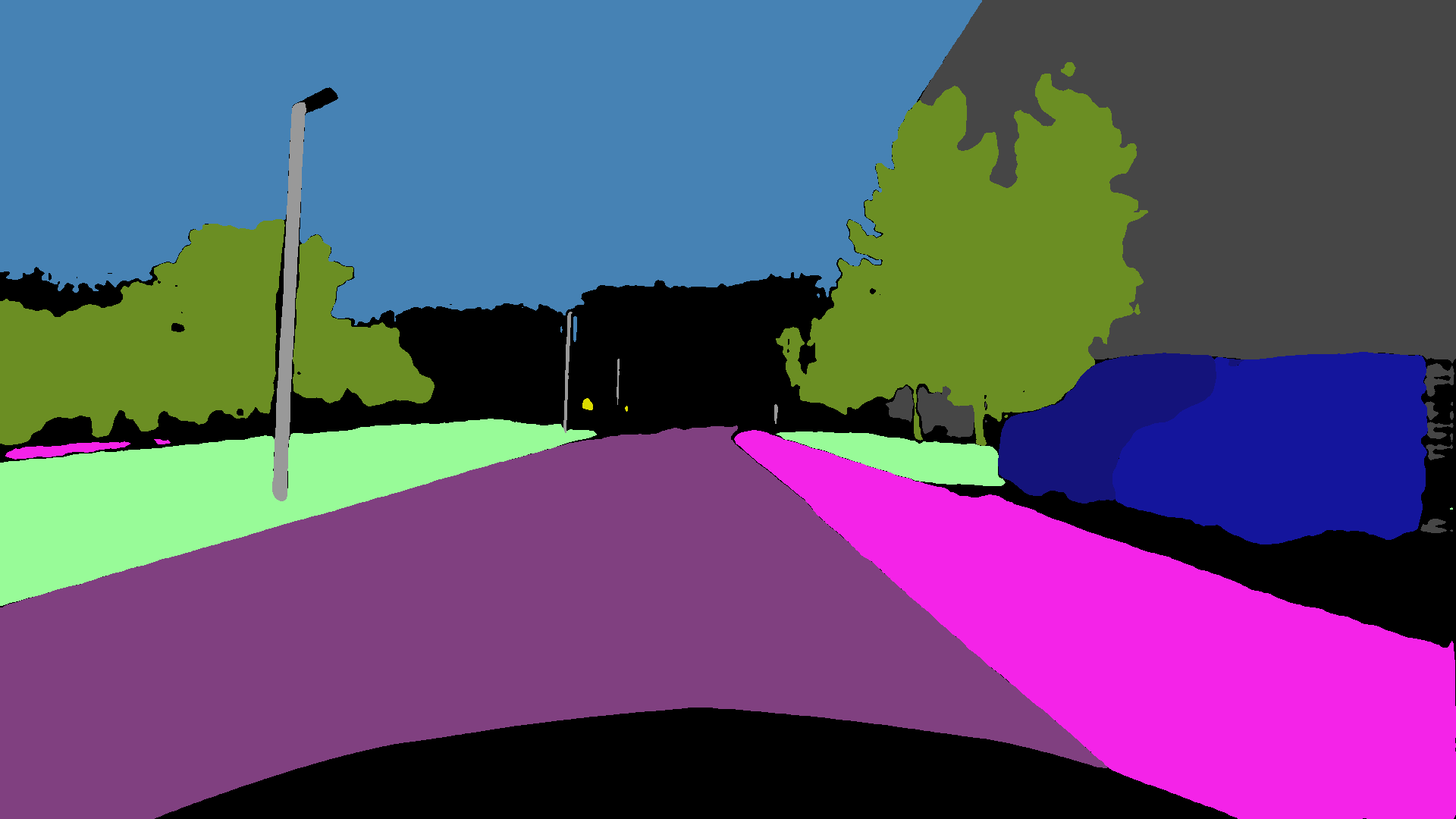} &
        \includegraphics[width=0.328\textwidth]{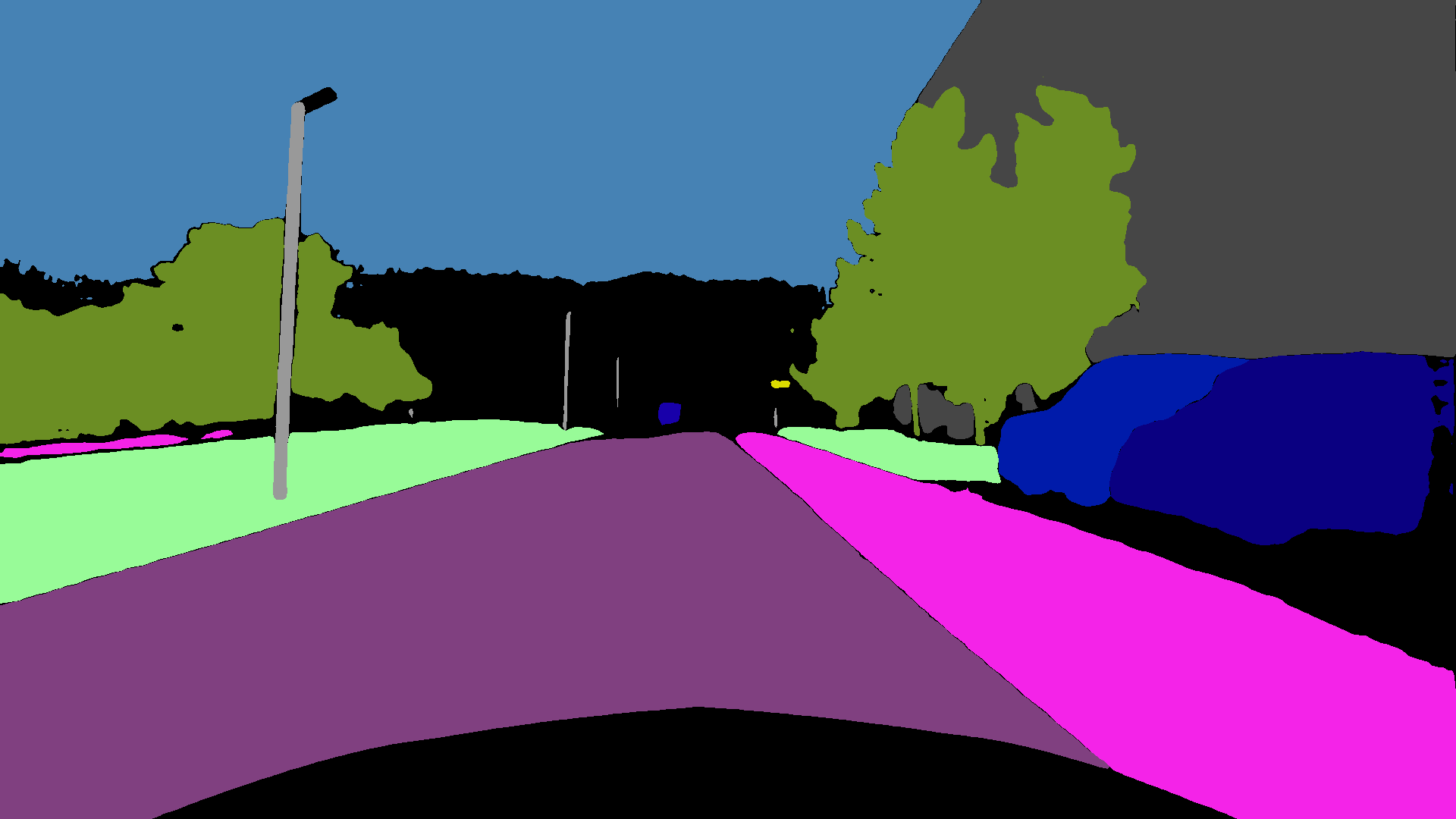} \\
        \vspace{-0.07cm}
        \includegraphics[width=0.328\textwidth]{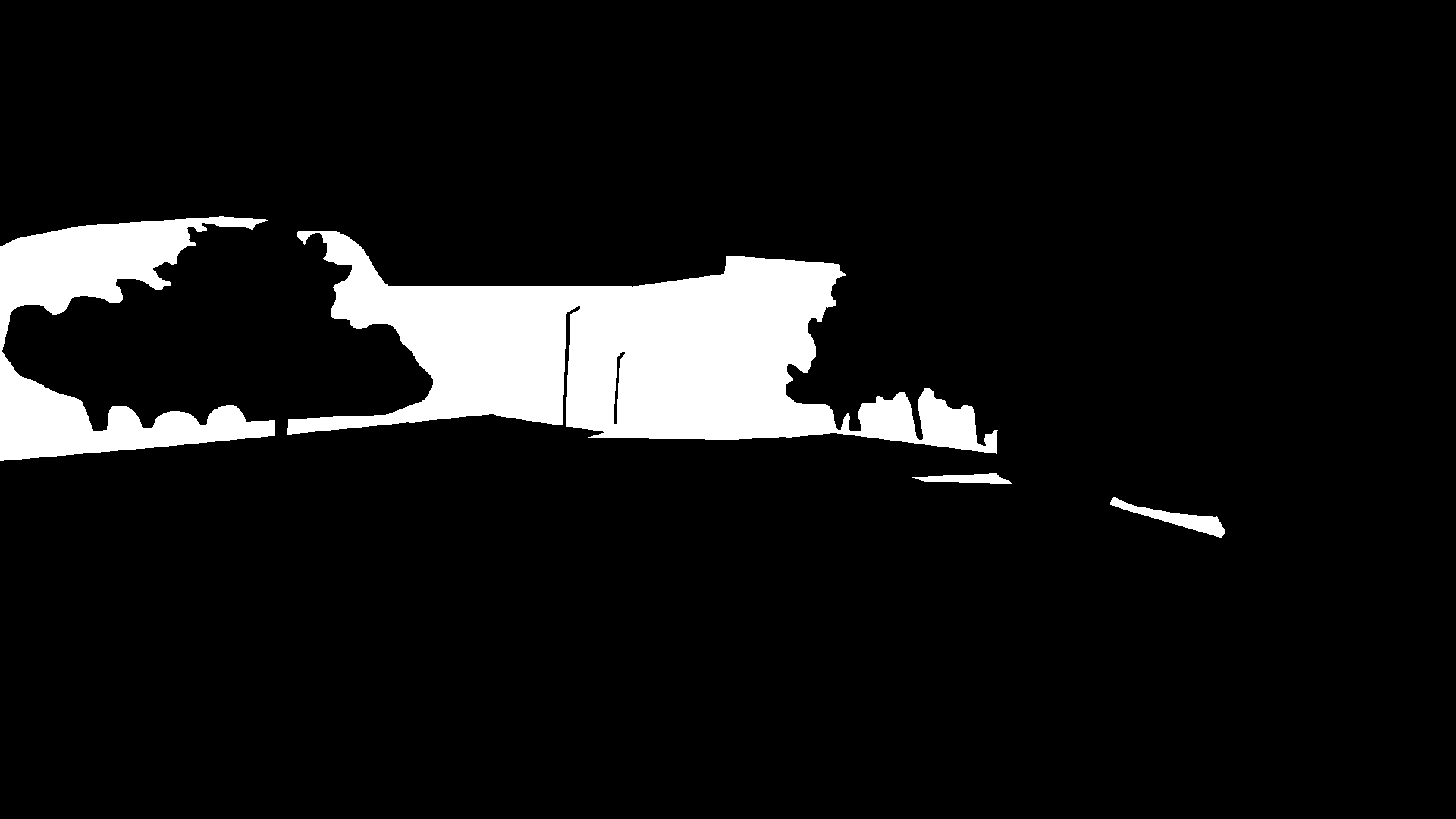} &
        \includegraphics[width=0.328\textwidth]{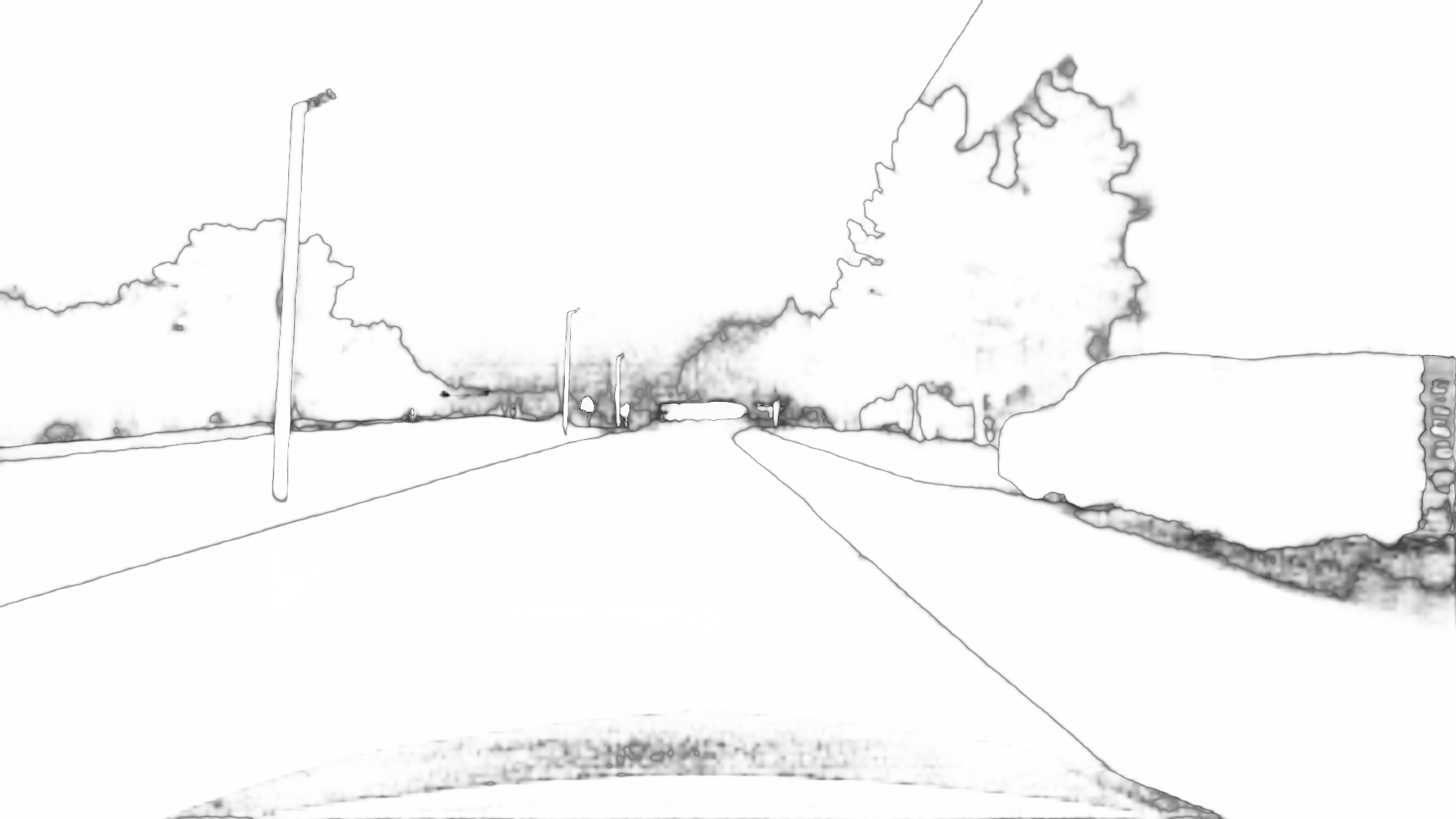} &
        \includegraphics[width=0.328\textwidth]{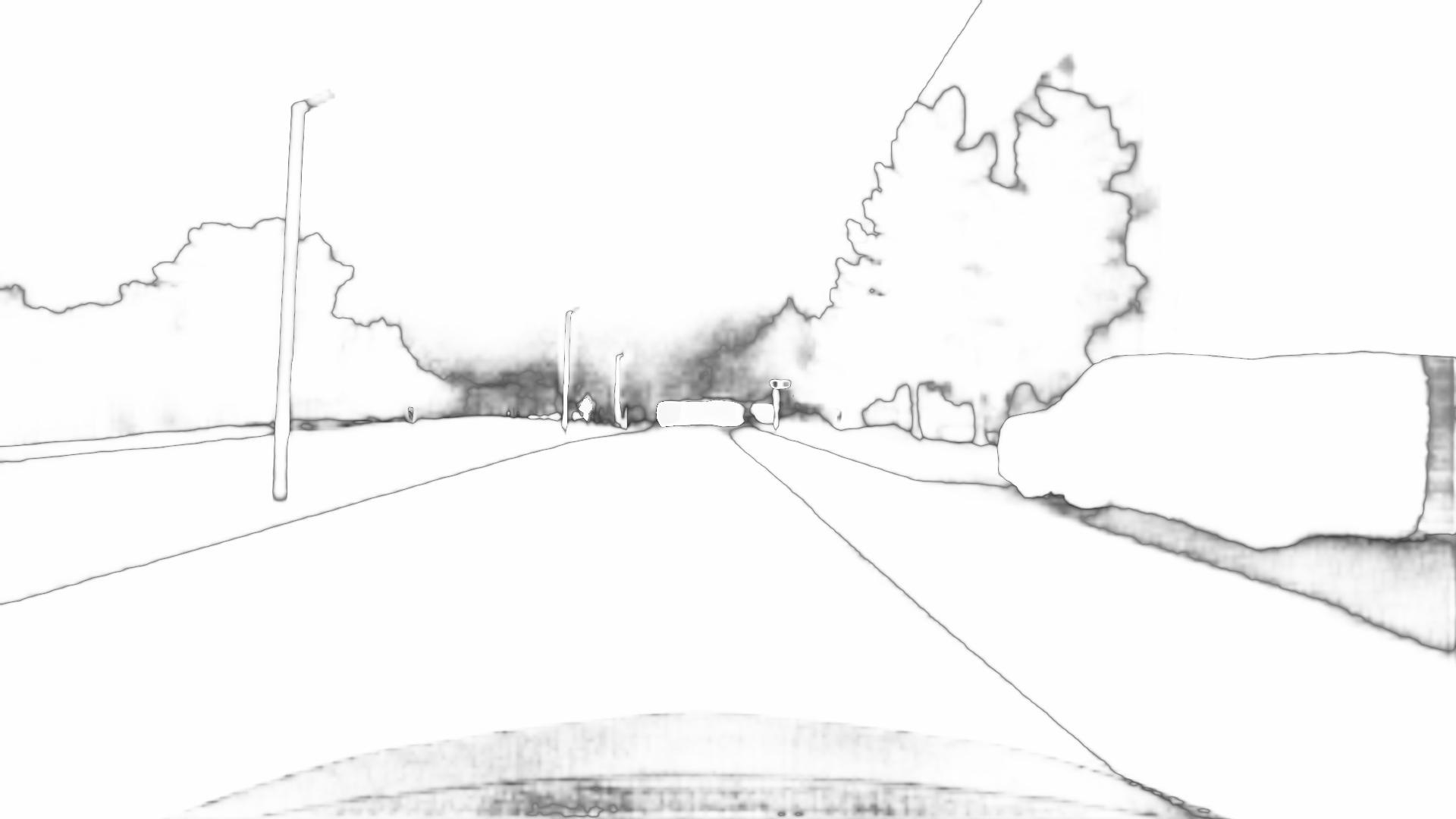} \\
        \vspace{-0.07cm}
        \includegraphics[width=0.328\textwidth]{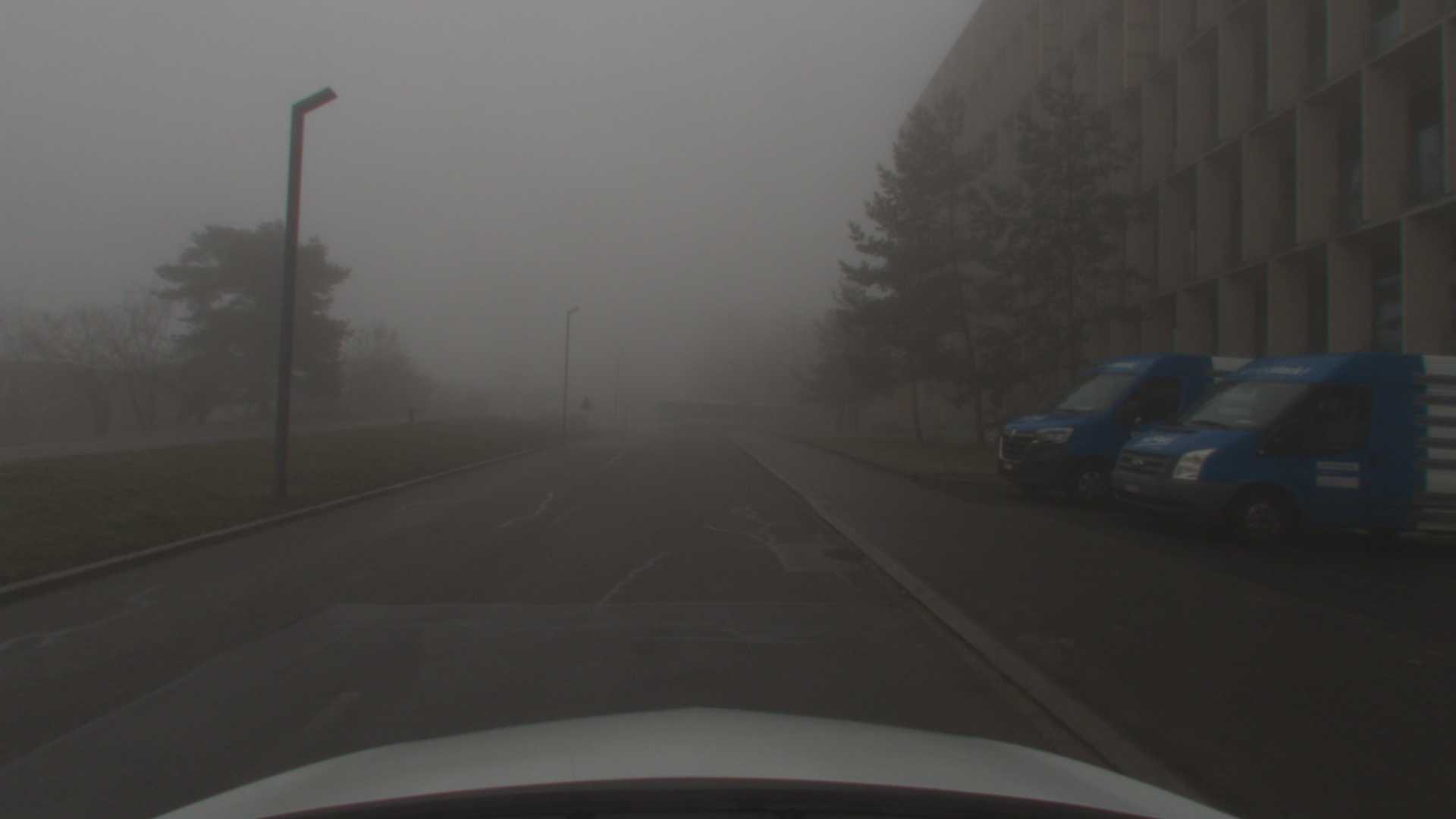} &
        \includegraphics[width=0.328\textwidth]{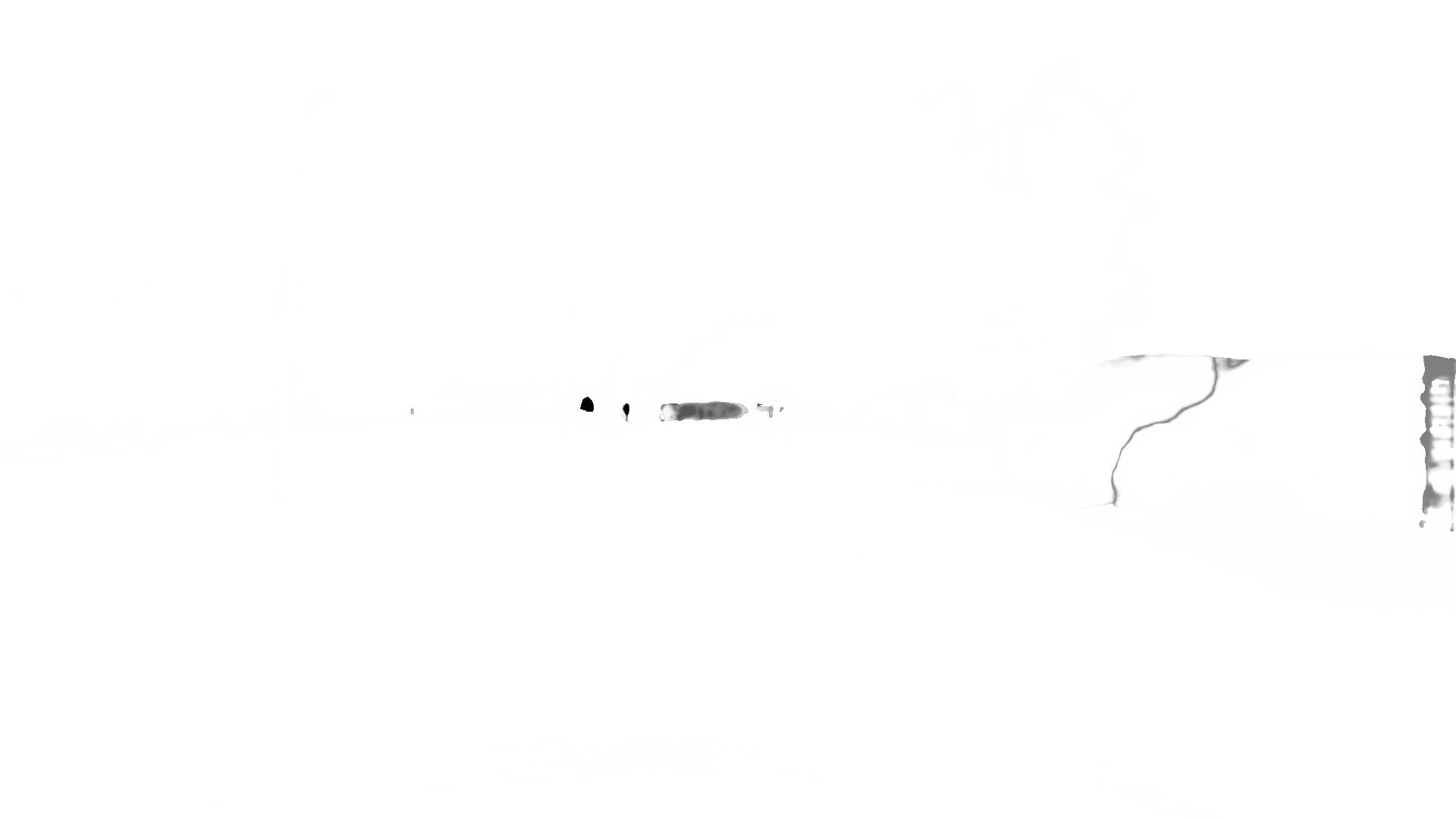} &
        \includegraphics[width=0.328\textwidth]{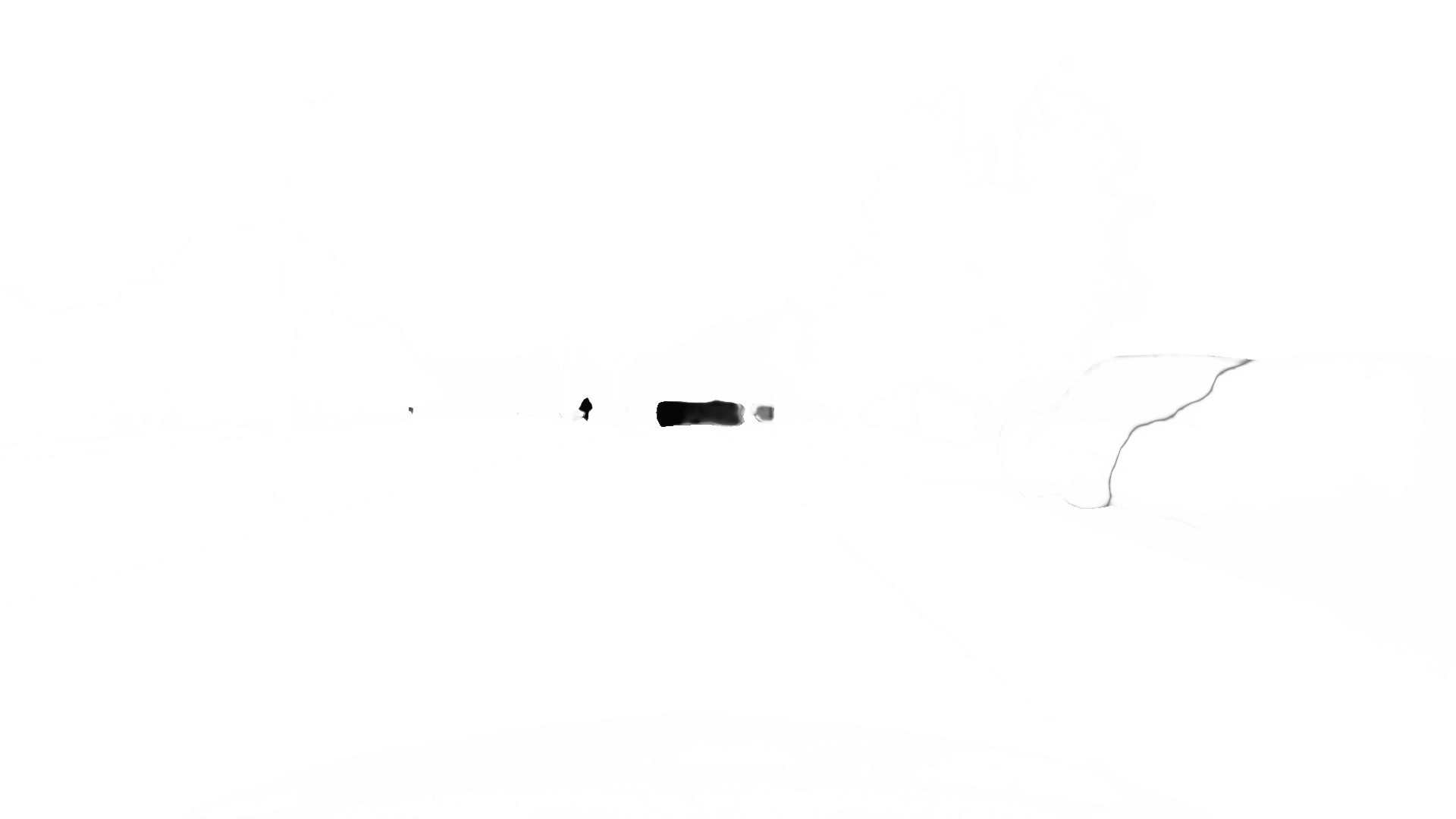} \\
    \end{tabular}
    \caption{\textbf{Qualitative panoptic results (continued)}. First column from top to bottom: panoptic segmentation ground-truth annotation, difficulty map, input frame camera image. Second and third columns from top to bottom: the panoptic predictions, class uncertainty scores, and instance uncertainty scores. Best viewed on a screen at full zoom.}
    \label{fig:suppl:qual:results:2}
\end{figure*}

We visualize some exemplary panoptic segmentation predictions results of the uni- and quadrimodal Mask2Former with their respective class and instance uncertainty maps in \cref{fig:suppl:qual:results:1,fig:suppl:qual:results:2}. The class and instance uncertainty scores are calculated according to \cref{subsec:sup:marg}.

\begin{table*}[!tb]
  \caption{\textbf{\textit{Clear}-test split class-wise PQ of Mask2Former~\cite{cheng2022masked} for different variations of input sensors}. A Swin-T~\cite{liu2021swin} backbone is used in all cases.}
  \label{table:supp:class:condition:clear}
  \centering
  \setlength\tabcolsep{4pt}
  \footnotesize
  \resizebox{\textwidth}{!}{
  \begin{tabular}{@{\extracolsep{\fill}} lccccccccccccccccccccccc@{}}
  \toprule
\rotatebox[origin=c]{90}{Frame camera} & \rotatebox[origin=c]{90}{Event camera} & \rotatebox[origin=c]{90}{Radar} & \rotatebox[origin=c]{90}{Lidar} && \rotatebox[origin=c]{90}{road} & \rotatebox[origin=c]{90}{sidewalk} & \rotatebox[origin=c]{90}{building} & \rotatebox[origin=c]{90}{wall} & \rotatebox[origin=c]{90}{fence} & \rotatebox[origin=c]{90}{pole} & \rotatebox[origin=c]{90}{light} & \rotatebox[origin=c]{90}{sign} & \rotatebox[origin=c]{90}{vegetation} & \rotatebox[origin=c]{90}{terrain} & \rotatebox[origin=c]{90}{sky} & \rotatebox[origin=c]{90}{person} & \rotatebox[origin=c]{90}{rider} & \rotatebox[origin=c]{90}{car} & \rotatebox[origin=c]{90}{truck} & \rotatebox[origin=c]{90}{bus} & \rotatebox[origin=c]{90}{train} & \rotatebox[origin=c]{90}{motorcycle} & \rotatebox[origin=c]{90}{bicycle} \\
  \midrule
  \yes & \no & \no & \no && 97.2 & 82.6 & 82.5 & 40.6 & 28.0 & 40.4 & 38.1 & 46.1 & 70.0 & 51.4 & 92.5 & 36.2 & 23.2 & 55.3 & 27.6 & 35.8 & 37.7 & 20.0 & 22.8 \\
  \yes & \yes & \no & \no &&97.6 & 83.7 & 83.8 & 43.9 & 28.7 & 43.2 & 38.4 & 49.5 & 70.7 & 53.8 & 92.2 & 41.6 & 27.1 & 59.6 & 38.2 & 44.7 & 34.3 & 31.4 & 27.1\\
  \yes & \no & \yes & \no  && 97.8 & 85.0 & 84.6 & 44.9 & 30.0 & 46.2 & 36.7 & 48.7 & 71.8 & 54.6 & 93.1 & 43.3 & 28.7 & 61.0 & 35.6 & 49.1 & 39.4 & 30.4 & 24.5\\
  \yes & \no & \no & \yes && 97.7 & 85.2 & 84.8 & 52.7 & 32.7 & 45.9 & 41.2 & 57.3 & 75.7 & 51.9 & 92.6 & 44.7 & 31.3 & 63.2 & 44.5 & 36.2 & 34.0 & 31.6 & 28.5\\
  \yes & \yes & \yes & \yes && 97.6 & 85.2 & 86.0 & 52.9 & 32.3 & 47.2 & 40.8 & 58.5 & 75.8 & 53.8 & 92.9 & 45.7 & 34.0 & 62.4 & 42.3 & 47.8 & 38.2 & 32.4 & 24.4\\
  \bottomrule
  \end{tabular}}
\end{table*}

\begin{table*}[!tb]
  \caption{\textbf{\textit{Fog}-test split class-wise PQ of Mask2Former~\cite{cheng2022masked} for different variations of input sensors}. A Swin-T~\cite{liu2021swin} backbone is used in all cases.}
  \label{table:supp:class:condition:fog}
  \centering
  \setlength\tabcolsep{4pt}
  \footnotesize
  \resizebox{\textwidth}{!}{
  \begin{tabular}{@{\extracolsep{\fill}} lccccccccccccccccccccccc@{}}
  \toprule
\rotatebox[origin=c]{90}{Frame camera} & \rotatebox[origin=c]{90}{Event camera} & \rotatebox[origin=c]{90}{Radar} & \rotatebox[origin=c]{90}{Lidar} && \rotatebox[origin=c]{90}{road} & \rotatebox[origin=c]{90}{sidewalk} & \rotatebox[origin=c]{90}{building} & \rotatebox[origin=c]{90}{wall} & \rotatebox[origin=c]{90}{fence} & \rotatebox[origin=c]{90}{pole} & \rotatebox[origin=c]{90}{light} & \rotatebox[origin=c]{90}{sign} & \rotatebox[origin=c]{90}{vegetation} & \rotatebox[origin=c]{90}{terrain} & \rotatebox[origin=c]{90}{sky}  & \rotatebox[origin=c]{90}{person} & \rotatebox[origin=c]{90}{rider} & \rotatebox[origin=c]{90}{car} & \rotatebox[origin=c]{90}{truck} & \rotatebox[origin=c]{90}{bus} & \rotatebox[origin=c]{90}{train} & \rotatebox[origin=c]{90}{motorcycle} & \rotatebox[origin=c]{90}{bicycle} \\
  \midrule
  \yes & \no & \no & \no && 94.8 & 57.8 & 69.9 & 25.1 & 32.2 & 45.7 & 71.9 & 39.1 & 67.7 & 63.8 & 79.5 & 44.6 & 0.0 & 61.1 & 43.6 & 35.5 & n/a & 0.0 & 4.4\\
  \yes & \yes & \no & \no && 94.2 & 58.1 & 76.7 & 12.8 & 31.2 & 50.9 & 72.2 & 42.9 & 70.7 & 65.5 & 74.2 & 48.3 & 0.0 & 61.6 & 39.5 & 32.5 & n/a & 46.5 & 10.8\\
  \yes & \no & \yes & \no && 95.5 & 62.2 & 74.8 & 23.0 & 30.4 & 51.6 & 65.5 & 41.0 & 73.5 & 68.2 & 77.8 & 48.8 & 0.0 & 66.5 & 45.2 & 46.6 & n/a & 0.0 & 20.7\\
  \yes & \no & \no & \yes && 94.0 & 57.3 & 76.0 & 29.1 & 38.0 & 50.5 & 71.0 & 53.6 & 77.4 & 67.4 & 76.9 & 43.4 & 0.0 & 64.8 & 48.3 & 41.2 & n/a & 0.0 & 15.3 \\
  \yes & \yes & \yes & \yes && 95.4 & 60.0 & 71.1 & 26.3 & 43.6 & 50.9 & 71.3 & 51.6 & 74.7 & 68.2 & 76.6 & 47.3 & 0.0 & 68.0 & 49.3 & 40.2 & n/a & 0.0 & 11.7\\
  \bottomrule
  \end{tabular}}
\end{table*}

\begin{table*}[!tb]
  \caption{\textbf{\textit{Rain}-test split class-wise PQ of Mask2Former~\cite{cheng2022masked} for different variations of input sensors}. A Swin-T~\cite{liu2021swin} backbone is used in all cases.}
  \label{table:supp:class:condition:rain}
  \centering
  \setlength\tabcolsep{4pt}
  \footnotesize  
  \resizebox{\textwidth}{!}{
  \begin{tabular}{@{\extracolsep{\fill}} lccccccccccccccccccccccc@{}}
  \toprule
\rotatebox[origin=c]{90}{Frame camera} & \rotatebox[origin=c]{90}{Event camera} & \rotatebox[origin=c]{90}{Radar} & \rotatebox[origin=c]{90}{Lidar} && \rotatebox[origin=c]{90}{road} & \rotatebox[origin=c]{90}{sidewalk} & \rotatebox[origin=c]{90}{building} & \rotatebox[origin=c]{90}{wall} & \rotatebox[origin=c]{90}{fence} & \rotatebox[origin=c]{90}{pole} & \rotatebox[origin=c]{90}{light} & \rotatebox[origin=c]{90}{sign} & \rotatebox[origin=c]{90}{vegetation} & \rotatebox[origin=c]{90}{terrain} & \rotatebox[origin=c]{90}{sky}  & \rotatebox[origin=c]{90}{person} & \rotatebox[origin=c]{90}{rider} & \rotatebox[origin=c]{90}{car} & \rotatebox[origin=c]{90}{truck} & \rotatebox[origin=c]{90}{bus} & \rotatebox[origin=c]{90}{train} & \rotatebox[origin=c]{90}{motorcycle} & \rotatebox[origin=c]{90}{bicycle} \\
  \midrule
  \yes & \no & \no & \no && 93.4 & 71.1 & 73.3 & 30.2 & 26.1 & 43.0 & 35.8 & 55.4 & 69.1 & 36.5 & 81.2 & 32.4 & 16.9 & 54.8 & 30.0 & 41.7 & 39.6 & 15.7 & 16.2\\
  \yes & \yes & \no & \no && 93.7 & 73.6 & 76.9 & 33.4 & 32.6 & 42.1 & 39.8 & 61.3 & 70.4 & 43.1 & 81.2 & 38.1 & 23.1 & 58.9 & 28.0 & 40.5 & 36.3 & 16.4 & 26.1\\
  \yes & \no & \yes & \no && 94.1 & 76.1 & 76.8 & 34.3 & 32.9 & 47.8 & 41.1 & 61.0 & 69.7 & 34.5 & 81.9 & 38.5 & 24.4 & 60.8 & 37.4 & 49.9 & 41.3 & 22.9 & 22.2 \\
  \yes & \no & \no & \yes && 94.6 & 79.0 & 78.3 & 39.6 & 37.4 & 50.4 & 45.6 & 67.8 & 73.6 & 44.4 & 83.3 & 42.7 & 32.2 & 63.1 & 38.9 & 50.7 & 50.4 & 28.3 & 25.1\\
  \yes & \yes & \yes & \yes && 94.9 & 78.8 & 79.9 & 42.8 & 44.5 & 49.0 & 44.2 & 66.9 & 74.0 & 40.0 & 84.0 & 42.1 & 35.3 & 64.4 & 40.7 & 40.2 & 45.8 & 27.3 & 26.6\\
  \bottomrule
  \end{tabular}}
\end{table*}

\begin{table*}[!tb]
  \caption{\textbf{\textit{Snow}-test split class-wise PQ of Mask2Former~\cite{cheng2022masked} for different variations of input sensors}. A Swin-T~\cite{liu2021swin} backbone is used in all cases.}
  \label{table:supp:class:condition:snow}
  \centering
  \setlength\tabcolsep{4pt}
  \footnotesize
   \resizebox{\textwidth}{!}{   
   \begin{tabular}{@{\extracolsep{\fill}} lccccccccccccccccccccccc@{}}
  \toprule
\rotatebox[origin=c]{90}{Frame camera} & \rotatebox[origin=c]{90}{Event camera} & \rotatebox[origin=c]{90}{Radar} & \rotatebox[origin=c]{90}{Lidar}  && \rotatebox[origin=c]{90}{road} & \rotatebox[origin=c]{90}{sidewalk} & \rotatebox[origin=c]{90}{building} & \rotatebox[origin=c]{90}{wall} & \rotatebox[origin=c]{90}{fence} & \rotatebox[origin=c]{90}{pole} & \rotatebox[origin=c]{90}{light} & \rotatebox[origin=c]{90}{sign} & \rotatebox[origin=c]{90}{vegetation} & \rotatebox[origin=c]{90}{terrain} & \rotatebox[origin=c]{90}{sky} & \rotatebox[origin=c]{90}{person} & \rotatebox[origin=c]{90}{rider} & \rotatebox[origin=c]{90}{car} & \rotatebox[origin=c]{90}{truck} & \rotatebox[origin=c]{90}{bus} & \rotatebox[origin=c]{90}{train} & \rotatebox[origin=c]{90}{motorcycle} & \rotatebox[origin=c]{90}{bicycle} \\
  \midrule
  \yes & \no & \no & \no && 94.5 & 68.4 & 76.4 & 43.6 & 30.8 & 30.3 & 35.3 & 50.9 & 71.7 & 34.8 & 81.1 & 31.8 & 18.1 & 56.7 & 15.0 & 64.3 & 32.4 & 6.2 & 14.7\\
  \yes & \yes & \no & \no && 94.8 & 70.9 & 78.1 & 44.9 & 33.8 & 31.1 & 38.9 & 54.0 & 73.0 & 33.8 & 80.1 & 35.8 & 27.1 & 58.6 & 23.8 & 51.2 & 37.5 & 16.4 & 21.6\\
  \yes & \no & \yes & \no && 94.9 & 69.0 & 78.2 & 46.7 & 36.6 & 34.0 & 36.6 & 56.3 & 72.4 & 37.7 & 81.2 & 37.2 & 30.0 & 61.2 & 24.0 & 68.0 & 57.5 & 16.6 & 19.5\\
  \yes & \no & \no & \yes && 94.3 & 72.4 & 80.4 & 50.6 & 33.0 & 36.6 & 38.7 & 59.3 & 75.1 & 39.6 & 80.5 & 36.7 & 19.1 & 64.2 & 28.9 & 60.1 & 39.9 & 17.7 & 22.0\\
  \yes & \yes & \yes & \yes && 95.0 & 72.7 & 79.3 & 50.3 & 33.6 & 34.8 & 40.2 & 60.7 & 75.2 & 41.3 & 79.9 & 36.2 & 35.7 & 63.9 & 23.4 & 52.8 & 38.7 & 24.5 & 21.7\\
  \bottomrule
  \end{tabular}}
\end{table*}

\begin{table*}[!tb]
  \caption{\textbf{\textit{Day}-test split class-wise PQ of Mask2Former~\cite{cheng2022masked} for different variations of input sensors}. A Swin-T~\cite{liu2021swin} backbone is used in all cases.}
  \label{table:supp:class:condition:day}
  \centering
  \setlength\tabcolsep{4pt}
  \footnotesize
   \resizebox{\textwidth}{!}{   
   \begin{tabular}{@{\extracolsep{\fill}} lccccccccccccccccccccccc@{}}
  \toprule
\rotatebox[origin=c]{90}{Frame camera} & \rotatebox[origin=c]{90}{Event camera} & \rotatebox[origin=c]{90}{Radar} & \rotatebox[origin=c]{90}{Lidar}  && \rotatebox[origin=c]{90}{road} & \rotatebox[origin=c]{90}{sidewalk} & \rotatebox[origin=c]{90}{building} & \rotatebox[origin=c]{90}{wall} & \rotatebox[origin=c]{90}{fence} & \rotatebox[origin=c]{90}{pole} & \rotatebox[origin=c]{90}{light} & \rotatebox[origin=c]{90}{sign} & \rotatebox[origin=c]{90}{vegetation} & \rotatebox[origin=c]{90}{terrain} & \rotatebox[origin=c]{90}{sky} & \rotatebox[origin=c]{90}{person} & \rotatebox[origin=c]{90}{rider} & \rotatebox[origin=c]{90}{car} & \rotatebox[origin=c]{90}{truck} & \rotatebox[origin=c]{90}{bus} & \rotatebox[origin=c]{90}{train} & \rotatebox[origin=c]{90}{motorcycle} & \rotatebox[origin=c]{90}{bicycle} \\
  \midrule
  \yes & \no & \no & \no && 95.7 & 72.9 & 79.6 & 38.8 & 32.1 & 39.7 & 43.9 & 46.2 & 84.0 & 53.4 & 95.7 & 34.8 & 19.2 & 56.2 & 30.0 & 46.0 & 34.9 & 17.6 & 16.9\\
  \yes & \yes & \no & \no && 96.1 & 74.7 & 81.6 & 40.1 & 33.9 & 42.3 & 43.5 & 49.4 & 84.8 & 55.1 & 96.1 & 40.2 & 23.9 & 59.6 & 35.3 & 45.0 & 28.8 & 27.8 & 24.2\\
  \yes & \no & \yes & \no && 96.2 & 74.9 & 81.4 & 40.3 & 35.8 & 45.4 & 44.4 & 49.8 & 85.2 & 54.6 & 96.0 & 41.0 & 25.0 & 61.4 & 38.5 & 54.1 & 30.6 & 26.6 & 24.2\\
  \yes & \no & \no & \yes && 95.8 & 75.6 & 81.6 & 46.0 & 36.5 & 45.8 & 44.6 & 55.7 & 84.8 & 54.8 & 95.3 & 41.5 & 26.2 & 63.1 & 42.6 & 48.6 & 38.3 & 27.1 & 24.7\\
  \yes & \yes & \yes & \yes && 96.0 & 75.4 & 81.1 & 44.7 & 39.5 & 44.7 & 44.6 & 55.6 & 84.6 & 55.8 & 95.8 & 42.1 & 28.6 & 63.0 & 41.7 & 45.0 & 32.4 & 28.9 & 27.5\\
  \bottomrule
  \end{tabular}}
\end{table*}

\begin{table*}[!tb]
  \caption{\textbf{\textit{Night}-test split class-wise PQ of Mask2Former~\cite{cheng2022masked} for different variations of input sensors}. A Swin-T~\cite{liu2021swin} backbone is used in all cases.}
  \label{table:supp:class:condition:night}
  \centering
  \setlength\tabcolsep{4pt}
  \footnotesize
   \resizebox{\textwidth}{!}{   
   \begin{tabular}{@{\extracolsep{\fill}} lccccccccccccccccccccccc@{}}
  \toprule
\rotatebox[origin=c]{90}{Frame camera} & \rotatebox[origin=c]{90}{Event camera} & \rotatebox[origin=c]{90}{Radar} & \rotatebox[origin=c]{90}{Lidar} && \rotatebox[origin=c]{90}{road} & \rotatebox[origin=c]{90}{sidewalk} & \rotatebox[origin=c]{90}{building} & \rotatebox[origin=c]{90}{wall} & \rotatebox[origin=c]{90}{fence} & \rotatebox[origin=c]{90}{pole} & \rotatebox[origin=c]{90}{light} & \rotatebox[origin=c]{90}{sign} & \rotatebox[origin=c]{90}{vegetation} & \rotatebox[origin=c]{90}{terrain} & \rotatebox[origin=c]{90}{sky}  & \rotatebox[origin=c]{90}{person} & \rotatebox[origin=c]{90}{rider} & \rotatebox[origin=c]{90}{car} & \rotatebox[origin=c]{90}{truck} & \rotatebox[origin=c]{90}{bus} & \rotatebox[origin=c]{90}{train} & \rotatebox[origin=c]{90}{motorcycle} & \rotatebox[origin=c]{90}{bicycle} \\
  \midrule
  \yes & \no & \no & \no && 94.2 & 70.6 & 70.9 & 37.5 & 22.3 & 40.1 & 29.1 & 52.9 & 43.4 & 39.6 & 49.4 & 32.6 & 19.4 & 55.0 & 0.0 & 22.5 & 40.4 & 11.0 & 17.3 \\
  \yes & \yes & \no & \no && 93.9 & 72.3 & 75.2 & 38.8 & 26.8 & 41.4 & 35.0 & 58.5 & 46.5 & 41.8 & 40.4 & 37.2 & 25.0 & 58.6 & 0.0 & 29.7 & 40.3 & 16.7 & 24.5 \\
  \yes & \no & \yes & \no && 95.1 & 75.2 & 75.6 & 43.3 & 26.5 & 44.3 & 32.4 & 57.3 & 47.3 & 46.6 & 48.6 & 38.7 & 26.8 & 60.9 & 0.0 & 41.5 & 50.3 & 20.9 & 19.8 \\
  \yes & \no & \no & \yes && 94.6 & 76.7 & 78.7 & 49.0 & 31.2 & 46.1 & 41.7 & 67.3 & 58.7 & 48.7 & 49.0 & 43.2 & 33.0 & 64.1 & 17.5 & 27.6 & 48.2 & 26.2 & 24.9 \\
  \yes & \yes & \yes & \yes && 95.6 & 78.2 & 79.0 & 51.8 & 33.3 & 46.9 & 40.5 & 67.9 & 57.8 & 48.6 & 46.4 & 42.3 & 37.9 & 64.6 & 12.3 & 45.7 & 49.2 & 28.0 & 18.9 \\
  \bottomrule
  \end{tabular}}
\end{table*}

